%% file: iclr2027_conference.tex
\pdfoutput=1
\PassOptionsToPackage{table}{xcolor}

\documentclass{article}%
\usepackage{iclr2027_conference,times}
\usepackage{xspace}
\usepackage{graphicx}
\newcommand{\negs}{\leavevmode\llap{\scalebox{0.5}[1]{$-$}}}%
\usepackage{booktabs}
\usepackage{makecell}
\usepackage{caption}
\newcolumntype{C}[1]{>{\centering\arraybackslash}p{#1}}
\newcommand{\hi}{{}^{\uparrow}}
\newcommand{\lo}{{}^{\downarrow}}
\usepackage{multirow}
\usepackage[table]{xcolor}
\usepackage{amsthm}
\usepackage{algorithm}
\usepackage{algorithmic}
\usepackage{bm}
\newcommand{\vect}[1]{\bm{#1}}
\newcommand{\mat}[1]{\mathbf{#1}}

\definecolor{papericeblue}{RGB}{234,242,251}%
\definecolor{actwalk}{HTML}{1E88E5}\definecolor{actrun}{HTML}{D81B60}\definecolor{actstand}{HTML}{6D4C41}
\definecolor{actlateral}{HTML}{43A047}\definecolor{actflexion}{HTML}{7B1FA2}\definecolor{actjump}{HTML}{FF9800}
\definecolor{meangray}{RGB}{245,245,246}%
\definecolor{imptint}{RGB}{252,231,231}\definecolor{imptintmean}{RGB}{251,221,221}%
\definecolor{dbetter}{HTML}{2E7D32}\definecolor{dworse}{HTML}{C62828}%
\newcommand{\tintbar}[1]{{\color{#1!6}\rule[-1pt]{2.5pt}{7pt}}{\color{#1!12}\rule[-1pt]{2.5pt}{7pt}}{\color{#1!18}\rule[-1pt]{2.5pt}{7pt}}{\color{#1!24}\rule[-1pt]{2.5pt}{7pt}}{\color{#1!30}\rule[-1pt]{2.5pt}{7pt}}{\color{#1!36}\rule[-1pt]{2.5pt}{7pt}}{\color{#1!42}\rule[-1pt]{2.5pt}{7pt}}{\color{#1!48}\rule[-1pt]{2.5pt}{7pt}}{\color{#1!54}\rule[-1pt]{2.5pt}{7pt}}{\color{#1!60}\rule[-1pt]{2.5pt}{7pt}}}%
\newlength{\rulepadabove}
\newlength{\rulepadbelow}
\newcommand{\padT}{\rule{0pt}{\dimexpr\ht\strutbox+\rulepadbelow\relax}}%
\newcommand{\padB}{\rule[-\dimexpr\dp\strutbox+\rulepadabove\relax]{0pt}{\dimexpr\dp\strutbox+\rulepadabove\relax}}%
\makeatletter
\newcommand{\fillL}{\gdef\CT@cell@color{\@tempdimb\z@\global\let\CT@cell@color\relax}}
\newcommand{\fillR}{\gdef\CT@cell@color{\@tempdimc\z@\global\let\CT@cell@color\relax}}
\newcommand{\cellcolorR}[1]{\gdef\CT@cell@color{\CT@color{#1}\@tempdimc\z@\global\let\CT@cell@color\relax}}%
\makeatother
\newcommand{\act}[1]{\textcolor{act#1}{\texttt{\textbf{#1}}}}
\newcommand{\actl}[2]{\textcolor{act#1}{#2}}%
\input{math_commands.tex}

\usepackage{hyperref}
\usepackage{url}
\providecommand{\doi}[1]{\href{https://doi.org/#1}{doi:#1}}%

\def\projname{ImpedanceMimic\xspace}

\usepackage[tracking=true,activate=false]{microtype}
\usepackage{float}%
\SetTracking{encoding=*,shape=sc}{0}%
\SetTracking{encoding=*,shape=sc,size=17.28}{-10}%

\title{ImpedanceMimic: Learned Impedance as Action \\ Space for Force-Accurate Motion Imitation}

\author{%
  \makebox[\textwidth][c]{%
    Matthias Jammot\textsuperscript{1}\setcounter{footnote}{1}\thanks{Corresponding author.} \qquad
    Pembe Gizem \"{O}zdil\textsuperscript{1,2}}\\[0.5em]
  {\textsuperscript{1}Harvard University},
  {\textsuperscript{2}Kempner Institute for the Study of Natural and Artificial Intelligence}\\[0.4em]
  \makebox[\textwidth][c]{\small\href{https://impedancemimic.github.io/ImpedanceMimic/}{\color{magenta}\texttt{https://impedancemimic.github.io/ImpedanceMimic/}}}
}

\newcommand{\fix}{\marginpar{FIX}}
\newcommand{\new}{\marginpar{NEW}}
\newcommand{\rev}[1]{{\color{red}#1}}%
\newlength{\toycolA}\newlength{\toycolB}%
\newtheorem{proposition}{Proposition}

\newtheorem{remark}{Remark}

\iclrfinalcopy
\ificlrfinal\newcommand{\preprintheader}{Preprint. Under review.}\fi%

\newlength{\defabovedisp}
\newlength{\defbelowdisp}
\newlength{\defabovedispshort}
\newlength{\defbelowdispshort}
\newlength{\deftextfloatsep}
\newlength{\deffloatsep}
\newlength{\defintextsep}
\newlength{\defparskip}
\makeatletter
\let\defsectionmacro\section \let\defsubsectionmacro\subsection \let\defparagraphmacro\paragraph
\makeatother
\makeatletter\def\thm@space@setup{\thm@preskip=2pt \thm@postskip=2pt}\makeatother

\ifdefined\pdfsuppressptexinfo\pdfsuppressptexinfo=-1\fi%
\begin{document}

\maketitle
\ifdefined\preprintheader\lhead{\preprintheader}\fi%
\ificlrfinal\else%
  \vspace{20.703pt}%
  \nointerlineskip\vbox to 0pt{\vss\hbox{\hskip\parindent\smash{\bf\begin{tabular}[t]{l}\makebox[\textwidth][c]{\small\href{https://impedancemimic.github.io/ImpedanceMimic/}{\color{magenta}\texttt{https://impedancemimic.github.io/ImpedanceMimic/}}}\end{tabular}}\hss}\kern25pt}%
  \global\let\anonfootnoterule\footnoterule\gdef\footnoterule{\ifnum\value{page}>1 \anonfootnoterule\fi}{\renewcommand\thefootnote{\protect\phantom{\ensuremath{\dagger}}}\footnotetext{\phantom{Corresponding author.}}}
\fi

\vspace{-2.5em}
\begin{abstract}
\vspace{-0.4em}
Recovering the forces behind a motion is ill-posed: different controllers can reproduce the same movement while generating different contact forces.
Physics-based motion imitation resolves this ambiguity through simulation, but relies on fixed-gain proportional--derivative (PD) control.
We show that fixed gains impose a trade-off between motion and force tracking, as contact and swing place conflicting demands on joint stiffness and damping.
We introduce \textbf{\projname{}}, a reinforcement-learning policy that modulates per-joint stiffness, damping, and position setpoints from motion and force measurements, but requires only motion at inference.
Across 129 participants and six activities, \projname{} outperforms matched fixed-gain PD under every integration scheme, direct-torque, muscle-actuated and existing imitation controllers in kinematic and force fidelity, approaching a dedicated motion-to-force regressor.
By expressing gains as multiples of inertia-scaled priors, a single policy controls subject-specific biomechanical models without explicit anthropometric input and transfers zero-shot to unseen humans and a humanoid robot.
The learned impedance forms a low-dimensional representation of motor state, shared across bodies, that generalizes to withheld activities: stiffness scales with load, and damping anticipates unloading.

\end{abstract}
\vspace{-0.8em}%

\section{Introduction}

Movement is governed by forces, which in locomotion arise from foot--ground contact and determine a sprinter's acceleration, a patient's loading asymmetry and a humanoid robot's balance.
Cameras have made motion datasets abundant~\citep{ionescu2013human3, mahmood2019amass, zhang2026romo}, but force data remains scarce. Force plates confine recordings to the laboratory, and wearable insoles resolve only the normal component and drift over time~\citep{koch2016validity}.
Recovering forces from motion alone would extend movement analysis into natural, unconstrained settings.

Recent corpora synchronize motion capture with force-plate measurements~\citep{werling2024addbiomechanics}, extending beyond steady gait to tasks like cutting, lunging and jumping~\citep{scherpereel2023human}.
Learning-based methods use such data to map kinematics to ground reaction forces (GRF), either through direct regression~\citep{mourot2022underpressure} or generative modeling~\citep{tan2026gaitdynamics}.
Their predictions can violate physics, for example by loading the ground during flight or placing the center of pressure (CoP) outside the foot, since a loss term at best encourages physical consistency.
The learned mapping is also opaque, going from motion to force directly without modeling the actuation and body dynamics that produce it, and therefore cannot drive a simulated body or robot.

Physics-based motion imitation addresses both limitations.
Tracking a recorded motion in a physics engine yields contact forces that are dynamically consistent by construction~\citep{peng2018deepmimic, luo2023perpetual}, and the controller producing these forces is explicitly defined.
However, dynamic consistency does not imply agreement with measured forces.
Most imitation policies output joint-position targets, which a proportional--derivative (PD) controller with hand-tuned gains converts into joint torques to drive the simulated body~\citep{peng2017learning}.
Different gains can track the same motion, yet produce different contact forces.
Recent work incorporates measured force as an observation~\citep{hori2026ground} or a reward~\citep{liu2026beyond}, but fixes the gains and reports accuracy either with force sensors available at inference or on a single participant.
Alternative actuation models, including learned gains~\citep{yuan2021simpoe}, muscle models~\citep{li2026towards} and direct torque control~\citep{cotton2025kintwin}, have not been systematically evaluated against measured forces.

\begin{figure}[t]
    \centering
    \includegraphics[width=1.0\linewidth]{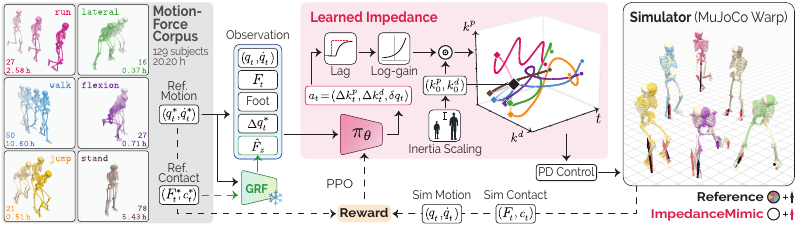}
    \vspace{-1em}
    \caption{\textbf{\projname{} learns joint impedance to reproduce measured contact forces.} From a reference motion and a frozen kinematics-to-GRF estimate, the policy modulates inertia-scaled nominal gains into time-varying stiffness and damping per joint. These gains set a PD law whose torques drive each participant's scaled model in its own MuJoCo Warp world, producing contact forces. PPO trains on kinematic and force rewards, and dashed lines mark training-only signals.}
    \label{fig:overview}
    \vspace{-1.5em}
\end{figure}

Here, we present \textbf{\projname{}}, a reinforcement-learning imitation policy that outputs per-joint stiffness, damping and position setpoints, supervised by reference kinematics, force-plate GRF and CoP.
On 129 participants, nine datasets and six activities, \projname{} outperforms fixed-PD, direct-torque, muscle and existing imitation controllers on kinematic and force accuracy.
To our knowledge, no prior controller reproduces measured per-foot contact forces across subjects and activities, including unseen ones, without force at inference.
We contribute:
\vspace{-0.8em}%
\begin{enumerate}
\item \textbf{Action-space bounds on force recovery.}
We derive a lower bound on stiffness, set by the torque the reference force demands, and an upper bound on damping, set by discrete-time stability, which oppose each other between contact and swing.
Empirically, a fixed-gain policy forfeits either motion or force tracking, whereas learned impedance satisfies both, extending the action-space analysis of \citet{peng2017learning} to force.

\item \textbf{A morphology-agnostic controller.}
Expressing impedance as a scaled inertia prior, our policy, trained on 107 subject-specific models without anthropometric input, transfers zero-shot to 22 subjects and a simulated humanoid robot (G1).
To our knowledge, this is the first physics-based imitator to transfer across unseen biomechanical and robotic embodiments.

\item \textbf{Learned impedance encodes motor state.}
The learned gains form a low-dimensional, interpretable representation in which stiffness scales with load and damping anticipates unloading. It generalizes to withheld activities better than fixed action spaces and identifies the activity from a single cycle of gains. This representation carries over to unseen subjects.

\item \textbf{An open benchmark.}
We release code, checkpoints and splits for a benchmark spanning 129 participants and six activities, with ablations, fixed-PD, direct-torque and muscle action spaces, and the existing methods PHC~\citep{luo2023perpetual}, SimPoE~\citep{yuan2021simpoe} and GaitDynamics~\citep{tan2026gaitdynamics}.
\end{enumerate}
\vspace{-0.55em}%

\vspace{-0.7em}%
\section{Related Work}
\vspace{-0.7em}%

\paragraph{Physics-based motion imitation.}
Executing a recorded motion in a simulator makes the resulting forces dynamically consistent. Reinforcement learning has made this routine at scale, from single-clip tracking~\citep{peng2018deepmimic} through adversarial objectives~\citep{peng2021amp} to controllers covering large motion corpora~\citep{luo2023perpetual, luo2024universal, tessler2024maskedmimic} and kinetic labeling~\citep{liu2025imdy}.
These methods pursue a kinematic objective because they originate in character animation, where simulation corrects generated motion~\citep{yuan2023physdiff} rather than recovering the forces behind it.
Comparing action spaces on motion-tracking tasks, \citet{peng2017learning} found fixed-gain PD control superior to torque, velocity and muscle parameterizations, and most later work adopted it.
In biomechanics, trajectory optimization over musculoskeletal models~\citep{dembia2020opensim} estimates individual muscle forces, but each trial requires its own slow optimization.
Recent work extends RL imitation to anatomically grounded models with torque or muscle actuation~\citep{cotton2025kintwin, simos2025kinesis, li2026towards, choi2026musculoskeletal}, with limited validation of the resulting forces against measured ones.
\citet{hori2026ground} provide measured force as an observation and therefore need inertial and insole sensors at inference.
\citet{liu2026beyond} add GRF and CoP to the reward, but validate on a single walking participant and rely on an external pelvis wrench~\citep{yuan2020residual}.
\projname{} is trained on measured per-foot force and CoP from 129 participants and six activities, and needs neither sensors nor an external wrench at inference.

\paragraph{Learned impedance control.}
Letting a policy set its own impedance originates in robotics, where variable stiffness and damping were learned to make contact-rich manipulation and legged hopping robust to contact uncertainty~\citep{buchli2011learning, martin2019variable, bogdanovic2020learning}, the contact force being a quantity to minimize rather than to reproduce.
The idea has since entered physics-based character control and motion capture.
SimPoE~\citep{yuan2021simpoe} scales nominal gains at every PD step and adds a residual root wrench to the action, ReGAIL~\citep{boursin2025imitation} puts servo gains and damping in the action so that a controller can alternate tension and relaxation, and Neural PhysCap~\citep{shimada2021neural} and OSDCap~\citep{le2024optimal} regress the gains from video, the latter together with an external force inside a neural Kalman filter.
In all four, the gains serve motion imitation or pose estimation, without being tied to the controlled body or evaluated against measured force.
\citet{le2026gravity} come closest, predicting the gains of a center-of-mass PD controller against force plates, though they distribute the force over body vertices instead of controlling joints.
No prior work has learned the impedance of an articulated human model against measured contact forces.
\projname{} learns joint impedance against measured force, with gains expressed as multiples of inertia-scaled priors that tie them to the body.
We first motivate our approach on a single joint, which we then scale to human motion imitation.

\vspace{-0.7em}%
\section{Method}
\vspace{-0.7em}%

\paragraph{Ill-posed contact forces.}
We model the body as a floating-base rigid-body system with generalized coordinates $\vect{q} \in \mathbb{R}^{n}$, $n = 31$, six of which describe the pelvis and $m = 15$ of which are actuated joints. Its dynamics obey the Newton--Euler equations of motion,
\begin{equation}
  \vspace{-0.1em}
  \mat{M}(\vect{q})\,\ddot{\vect{q}} \;+\; \vect{h}(\vect{q},\dot{\vect{q}})
    \;=\; \mat{S}^{\top}\vect{\tau}
    \;+\; \mat{J}_c(\vect{q})^{\top}\vect{\lambda},
  \label{eq:eom}
  \vspace{-0.1em}
\end{equation}
with mass matrix $\mat{M}$, bias forces $\vect{h}$ (Coriolis, centrifugal, gravitational), joint torques $\vect{\tau}$ mapped by the selection matrix $\mat{S}$, and contact forces $\vect{\lambda}$ through their Jacobian $\mat{J}_c$ (dimensions in App.~\ref{App:contact}).
Each GRF is the resultant of its contact forces applied at its CoP.
A recorded motion $\vect{q}^{*}(t)$ fixes the left-hand side of Eq.~\ref{eq:eom}, and its torque-free pelvis rows fix the net contact wrench to the rate of change of whole-body momentum, but not how that wrench is realized: in double support, its split between the feet and each CoP remains undetermined.
Simulation resolves this through a contact model and a joint-torque trajectory $\vect{\tau}(t)$, so the accuracy of the recovered forces is that of $\vect{\tau}(t)$.

\vspace{-0.5em}%
\subsection{Action-space Impedance Bounds}
\label{sec:bounds}
\vspace{-0.5em}%

A policy $\pi_\theta(\vect{a}_t \mid \vect{o}_t)$ acts every $\Delta t = 10$\,ms while the simulator steps every $h = 2$\,ms.
The action space determines how the joint torque $\hat{\vect{\tau}}$ in Eq.~\ref{eq:eom} evolves over the $N = 5$ substeps: a direct-torque policy holds $\hat{\vect{\tau}} = \tau_{\mathrm{scale}}\,\vect{a}_t$ fixed, so it cannot react between queries.
\citet{peng2017learning} wrap it in a proportional--derivative controller recomputed every substep, to smooth learning:
\begin{equation}
  \hat{\vect{\tau}} \;=\; \vect{k}^{p} \circ
      \bigl(\vect{q}^{*}_{t+1} + \alpha\,\delta\vect{q} - \vect{q}\bigr)
    \;+\; \vect{k}^{d} \circ
      \bigl(\dot{\vect{q}}^{*}_{t+1} - \dot{\vect{q}}\bigr),
  \label{eq:pd_torque}
\end{equation}
clipped to $\pm\vect{\tau}^{\mathrm{lim}}$, where $\delta\vect{q} \in [-1,1]^{m}$ are per-joint position offsets scaled by $\alpha$, and $\circ$ the element-wise product.
We extend this controller by emitting per-joint gain residuals $\Delta\vect{k}^{p}, \Delta\vect{k}^{d} \in [-1,1]^{m}$,
\begin{equation}
  \vect{k}^{p} = \vect{k}^{p}_{0} \circ \exp\!\bigl(s_p\,\Delta\vect{k}^{p}\bigr),
  \qquad
  \vect{k}^{d} = \vect{k}^{d}_{0} \circ \exp\!\bigl(s_d\,\Delta\vect{k}^{d}\bigr),
  \label{eq:learned_impedance}
\end{equation}
about base gains $\vect{k}^{p}_{0}, \vect{k}^{d}_{0}$, with log-scales $s_p, s_d > 0$ setting the gain range. The exponential keeps gains strictly positive, as stability at the joint requires.
We define two PD action spaces: \emph{Fixed-PD} ($\vect{a} = \delta\vect{q}$, gains held at $(\vect{k}^{p}_{0}, \vect{k}^{d}_{0})$), and \emph{Learned-PD} ($\vect{a} = (\Delta\vect{k}^{p}, \Delta\vect{k}^{d}, \delta\vect{q})$).
As observed empirically by \citet{bogdanovic2020learning}, in contact the environment constrains the joint, so $k^{d}$ has little effect and $k^{p}$ sets the force the PD can produce. Out of contact the joint drives only the light distal segment, so $k^{d}$ governs the closed-loop dynamics. We formalize this as two bounds.

\begin{proposition}[Force authority, Bound 1]\label{prop:authority}
At the reference state $(\vect{q}, \dot{\vect{q}}) = (\vect{q}^{*}, \dot{\vect{q}}^{*})$ the velocity-error term of Eq.~\ref{eq:pd_torque} vanishes and the displacement is $\mathcal{O}(\Delta t)$-small, so the PD torque reduces to $\vect{\tau} \approx \alpha\,\vect{k}^{p} \circ \delta\vect{q}$: the offset $\delta\vect{q}$ is the policy's sole lever for torque production at the reference pose, giving an achievable interval of half-width $\alpha\vect{k}^{p}$.
Producing the reference contact force therefore requires the inverse-dynamics torque $\vect{\tau}^{*}$ of Eq.~\ref{eq:eom}, which is attainable at the reference pose only if
\begin{equation}
  k^{p} \;\ge\; \frac{\max_t |\tau^{*}(t)|}{\alpha}.
  \label{eq:bound1}
\end{equation}
\end{proposition}
Below this bound, the lever $\alpha\vect{k}^{p}$ is too small to supply the torque required by the reference force.
The offset $\delta\vect{q}$ saturates, leaving the tracking term $\vect{k}^{p} \circ (\vect{q}^{*} - \vect{q})$ as the only source of torque.
A fixed-gain policy below the bound must therefore choose between forfeiting kinematic or force tracking.

\begin{proposition}[Feedback stability, Bound 2]\label{prop:stability}
In continuous time any positive $k^{d}$ is stabilizing.
In simulation the PD of Eq.~\ref{eq:pd_torque} is applied as an explicit actuator force at every physics step $h$~\citep{peng2017learning}.
On a free joint of inertia $I$, discretizing the error dynamics yields a $2 \times 2$ linear map whose eigenvalues must lie inside the unit circle.
The Jury criterion gives
(App.~\ref{App:single_joint_theory})
\addtolength{\abovedisplayskip}{3.6pt}\addtolength{\belowdisplayskip}{-3.6pt}%
\begin{equation}
  0 < k^{d} < \frac{2I}{h}
  \qquad\text{and}\qquad
  0 < k^{p} < \frac{4I}{h^{2}} - \frac{2 k^{d}}{h}.
  \label{eq:bound2}
\end{equation}
\end{proposition}
The damping condition is the more restrictive case, and depends on $I$, the changing inertia that the joint drives at a given moment.
During contact the joint acts on the kinematic chain of the whole body above it, whereas in swing it drives only the distal segment, whose inertia is orders of magnitude smaller.
A damping gain that is stable in contact can therefore be unstable in flight, where the feedback amplifies the error between policy steps and the torque alternates at $\pm\vect{\tau}^{\mathrm{lim}}$.
This chatter is the classical limit on the impedance a sampled-data controller can render~\citep{colgate1994factors}.
Under the explicit PD of Eq.~\ref{eq:pd_torque}, each joint thus inherits a damping ceiling set by its distal inertia.
Stable PD avoids it by evaluating the controller at the next state~\citep{tan2011stable}, and implicit damping by moving the damping term into the integrator~\citep{todorov2012mujoco}.
\vspace{0.2em}%
\begin{remark}[Phase-dependent feasibility]\label{cor:sensitivity}
A stiff joint, as Bound~1 demands in contact, needs higher damping against oscillation, whereas Bound~2 needs a lightly damped joint in swing.
Fixed gains must therefore favor one phase over the other, whereas state-dependent gains can match each bound.
\end{remark}
\begin{figure}
    \centering
    \includegraphics[width=\linewidth]{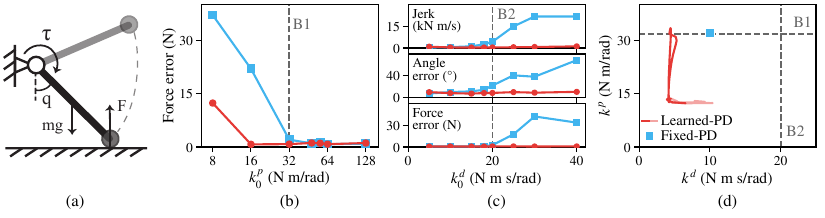}
    \vspace{-1.5em}%
    \caption{\textbf{Action-space bounds on a single joint.} \textbf{(a)}~A hinged rod tracks a trajectory into a compliant floor, drawn from its swing apex (light) to contact (dark). \textbf{(b)}~Force error of Learned-PD (red) and Fixed-PD (blue) versus base stiffness, with Bound~1 (B1) at $32$\,Nm/rad. \textbf{(c)}~Jerk, angle and force error versus base damping, with Bound~2 (B2) at $20$\,Nm\,s/rad. \textbf{(d)}~Both bounds, the Fixed-PD gains and the Learned-PD gains over one cycle, solid in contact and translucent in swing.}
    \label{fig:action_space_bounds}
    \vspace{-1.5em}
\end{figure}

\vspace{-0.85em}%
\paragraph{Validation.}
We test both bounds on the rod of Fig.~\ref{fig:action_space_bounds}a, where they evaluate to $k^{p} \ge 32$\,Nm/rad and $k^{d} < 20$\,Nm\,s/rad (App.~\ref{App:action_space_bounds}).
Trained with identical base gains, reward and PPO~\citep{schulman2017proximal}, Fixed-PD forfeits force tracking when its stiffness is below Bound~1 (Fig.~\ref{fig:action_space_bounds}b) and loses kinematic and force tracking for damping above Bound~2 (Fig.~\ref{fig:action_space_bounds}c).
Learned-PD instead reaches the same force accuracy by raising $k^{p}$ to Bound~1 in contact and holding $k^{d}$ under Bound~2 throughout (Fig.~\ref{fig:action_space_bounds}d).
Section~\ref{sec:seen} verifies these bounds on the human model (Fig.~\ref{fig:humanoid_bounds}, Table~\ref{tab:humanoid_bounds}).

\vspace{-0.5em}%
\subsection{\projname{}}
\label{sec:impedancemimic}
\vspace{-0.5em}%

\paragraph{Action space.}
The policy outputs $\vect{a}_t = (\Delta\vect{k}^p_t, \Delta\vect{k}^d_t, \delta\vect{q}_t) \in [-1,1]^{3m}$ with
log-scales $s_p = 1.25$ and $s_d = 1.5$.
A first-order lag mimics activation dynamics by smoothing gain changes,
\begin{equation}
  \tilde{\Delta \vect{k}}_t = \tilde{\Delta \vect{k}}_{t-1}
    + \beta\bigl(\Delta \vect{k}_t - \tilde{\Delta \vect{k}}_{t-1}\bigr),
  \qquad
  \beta = 1 - e^{-\Delta t/T_{\mathrm{act}}},\quad
  T_{\mathrm{act}} = 50~\mathrm{ms},
  \label{eq:gain_lag}
\end{equation}
 that would otherwise appear as force transients, and requires the policy to stiffen ahead of touchdown.
Base gains scale with each subject $i$'s diagonal joint-space inertia $\vect{I}_{jj}(i)$ at the rest pose,
\begin{equation}
  \vect{k}^{p}_{0} = \vect{I}_{jj} \circ \vect{\omega}_{n}^{2},
  \qquad
  \vect{k}^{d}_{0} = 2\,\vect{\zeta} \circ \vect{I}_{jj} \circ \vect{\omega}_{n},
  \label{eq:inertia_scaling}
\end{equation}
where the natural frequency $\vect{\omega}_n$ and damping ratio $\vect{\zeta}$ are set per joint by nominal gains (Table~\ref{tab:hyperparams}), such that a multiple of the nominal results in the same closed-loop behavior across bodies.

\vspace{-0.3em}%
\paragraph{Kinematics-to-GRF estimator.}
Setting the impedance before touchdown requires anticipating the contact force, which the policy cannot read from a force plate at inference.
The actor therefore receives an estimate $\smash{\hat{\vect{F}}}_t$ computed from the reference kinematics.
A center-of-mass physics prior, corrected by a residual network ensemble, is fitted once per split on the plate-valid frames of the training clips and then frozen (App.~\ref{App:estimator}).
It reaches a GRF correlation of $0.979$ on seen and $0.943$ on unseen subjects, on par with the dedicated GaitDynamics regressor~\citep{tan2026gaitdynamics}.

\paragraph{Observations.}
The actor observes $\vect{o}_t \in \mathbb{R}^{215}$ (Table~\ref{tab:obs}): proprioception and contact state from the simulator, and three lookaheads at $\{10, 50, 250\}$~ms, namely the kinematic offset $\vect{q}^*_{t+k} - \vect{q}_t$, the estimated GRF $\smash{\hat{\vect{F}}_{t+k}}$, and foot positioning.
Every entry is a simulator state or a function of the reference kinematics, so no force-plate signal reaches the actor and the policy runs plate-free at inference (Fig.~\ref{fig:overview}).
The critic receives the current-frame GRF to lower advantage variance~\citep{pinto2017asymmetric}, and
a linear head on the actor, discarded at inference, regresses the measured GRF at three lookahead offsets through an auxiliary Huber loss in the PPO objective (App.~\ref{App:ImpedanceMimic}).%
\vspace{-0.21em}%
\paragraph{Reward.}
The reward combines kinematic imitation with force-plate supervision,
\begin{equation}
  r_t = r^{\mathrm{kin}}_t + w_{\mathrm{grf}}\, r^{\mathrm{grf}}_t
      + 0.25\, r^{\mathrm{shear}}_t + 0.15\, r^{\mathrm{ee}}_t,
  \label{eq:reward}
\end{equation}
with $w_{\mathrm{grf}}$ annealed linearly from $0.25$ to $0.5$, such that the policy first learns to track and then to produce faithful forces.
The kinematic term combines pose, velocity and pelvis tracking (App.~\ref{App:reward}).
Each joint's pose error is normalized by its amplitude over the clip, and the root term tracks pelvis velocity, independent of the subject's leg length.
With $\Delta\vect{F}_f = \vect{F}_f - \vect{F}^*_f$ and $\Delta\vect{c}_f = \vect{c}_f - \vect{c}^*_f$ the force and CoP errors of foot $f$, and $\bar{F}_f = \max(F^*_{z,f},\,0.5)$ the load that scales the force error,
\begin{equation}
  r^{\mathrm{grf}} = \exp\Bigl(-\sum_f g_f \bigl[\alpha_{\mathrm{grf}} \tfrac{\|\Delta\vect{F}_f\|^2}{\bar{F}_f^2}
    + 5\,\|\Delta\vect{c}_f\|^2\bigr]\Bigr),
  \quad
  r^{\mathrm{shear}} = \exp\Bigl(-5\sum_f g_f\,\|\Delta\vect{F}_{f,\parallel}\|^2\Bigr),
  \label{eq:rgrf}
\end{equation}
with $\alpha_{\mathrm{grf}} = 4$.
The contact gate $g_f$ is active when either the simulated or the reference vertical force exceeds $0.1$~BW, to penalize mistimed contacts.
$r^{\mathrm{ee}}$ tracks heel and toe positions.
\vspace{-0.21em}%
\paragraph{Datasets.}
We assemble 20\,h of motion from 129 participants across 9 public datasets with synchronized force plates (Fig.~\ref{fig:overview}, App.~\ref{App:dataset_descriptions}), distributed through AddBiomechanics~\citep{werling2024addbiomechanics} with joint kinematics, per-foot GRF and centers of pressure.
Recordings cover six activity classes, walking, running, standing, lateral movements, flexion and jumping, with body masses ranging from 44 to 126~kg and heights from 1.46 to 1.95~m.
We resample to 100~Hz, low-pass filter, cut event-aligned clips of 1--10~s and normalize forces to body weight (App.~\ref{App:Datasets_Preprocessing}).
\vspace{-0.21em}%
\paragraph{Simulation model.}
Every subject is simulated with the no-arm Rajagopal model, scaled by \mbox{AddBiomechanics} and converted to MuJoCo~\citep{todorov2012mujoco} with MyoConverter~\citep{wang2022myosim, ikkala2022converting}, with torque-driven joints.
All subjects share 14 bodies and the $n = 31$ coordinates of Eq.~\ref{eq:eom}, while masses, inertias, segment lengths, contact geometry and joint ranges are subject-specific.
Each foot carries eight contact spheres building on \citet{falisse2019rapid} (App.~\ref{App:contact}).
The policy runs at 100~Hz, the rate of the reference data, and the physics at $h = 2$~ms.
\vspace{-1.3em}%
\paragraph{Training and evaluation.}
We train the actor and critic, each an MLP ($[512,256,128]$, SiLU), with PPO for 2 billion steps on 8{,}192 parallel worlds in one batched MuJoCo Warp environment, each world carrying its own subject-scaled model (App.~\ref{App:ImpedanceMimic}).
Episodes start at a uniformly sampled frame with noise on pose and velocity and terminate when the clip ends or the pelvis drops below $0.4$~m~\citep{peng2018deepmimic}.
Our policy receives no subject identifier or anthropometric descriptor: morphology enters only through the priors of Eq.~\ref{eq:inertia_scaling}.
We report three benchmarks for every method (App.~\ref{App:splits}): \emph{seen} (train on all 129 subjects, evaluate on the held-out $20\,\%$ of each subject's clips), \emph{unseen} (train on 107 subjects, evaluate zero-shot on 22 held out) and \emph{leave-one-activity-out} (train on five activities, evaluate on the sixth).
Evaluation rollouts report survival, RMSE and Pearson correlation for joint angles and per-foot forces, center-of-pressure error and torque consistency (App.~\ref{App:results}).
\vspace{-1.3em}%
\paragraph{Baselines.}
We compare \projname{} to other action spaces, extending the analysis of \citet{peng2017learning} to contact forces.
The baselines match \projname{} in everything but actuation, as far as training allowed.
\textit{(i)} \emph{Fixed-PD} outputs the position offset $\delta\vect{q}_t$ with fixed inertia-scaled base gains (App.~\ref{App:humanoid_bounds}).
\textit{(ii)} \emph{Torque} outputs $\hat{\vect{\tau}}_t = \vect{\tau}_{\mathrm{scale}} \circ \vect{a}_t$ with per-joint, per-subject scale $\vect{\tau}_{\mathrm{scale}} = \alpha\,\vect{k}^{p}_0$, its authority inertia-normalized like the PD baselines (App.~\ref{App:reward}).
\textit{(iii)} \emph{Muscle} outputs excitations for the 80 Hill-type muscles of each participant's scaled model (App.~\ref{App:muscle}).
We also compare against existing methods.
\textit{(iv)} \emph{GaitDynamics}~\citep{tan2026gaitdynamics} regresses bilateral GRF and CoP from joint kinematics (App.~\ref{App:baselines}).
With no physics and no kinematics to imitate, it serves as a reference for force accuracy.
\textit{(v)} \emph{PHC}~\citep{luo2023perpetual} and \textit{(vi)} \emph{SimPoE}~\citep{yuan2021simpoe} are physics-based imitation controllers, the latter with learned PD gains, re-implemented as faithfully as paper and code allow, on the same data, physics and evaluation (App.~\ref{App:baselines}).%
We retrain Fixed-PD and \projname{} under stable PD and implicit damping, repeat every action space without the force reward, and run ablations of \projname{} and robustness tests (\S\ref{sec:benchmark}, App.~\ref{App:ablation}).

\vspace{-0.6em}%
\section{Results}
\vspace{-0.9em}%

\subsection{Action Space Governs Force Fidelity}
\label{sec:seen}
\vspace{-0.9em}%

\begin{table*}[!b]%
\vspace{-1.6em}%
\centering
\caption{
  \textbf{ImpedanceMimic produces force-accurate motion imitation and transfers zero-shot to unseen morphologies and activities.}
  Each baseline trains one policy on all six activities across 1)
  \textit{Seen subjects}: held-out clips, 2)
\textit{Unseen subjects}: all clips of a disjoint subject set $H$, and one policy per fold for 3) \textit{Withheld activity}: trained on five activities and evaluated zero-shot on the sixth.
  Standard deviations over 5 seeds in Tables~\ref{tab:golden_full}, \ref{tab:golden_unseen} and~\ref{tab:loao_full}.
  \textbf{Bold} / \underline{underline} = best / second best.
  }
  \vspace{-0.4em}%
\label{tab:golden}
{\scriptsize
\setlength{\tabcolsep}{0.7pt}%
\newcommand{\gd}[1]{\textit{#1}}
\setlength{\aboverulesep}{0pt}\setlength{\belowrulesep}{0pt}%
\begin{tabular}{@{}p{52.5pt}>{\hspace{0.3em}}lll>{\hspace{0.37em}}lll>{\hspace{0.37em}}lll>{\hspace{0.37em}}lll>{\hspace{0.37em}}lll>{\hspace{0.37em}}lll|>{\columncolor{meangray}\hspace{0.5em}}l>{\columncolor{meangray}}l>{\columncolor{meangray}}l>{\columncolor{meangray}\fillR\hspace{2.88pt}}l@{}}\toprule
\padT\padB\texttt{\textbf{Activity}} & \multicolumn{3}{c}{\act{walk}} & \multicolumn{3}{c}{\act{run}} & \multicolumn{3}{c}{\act{stand}} & \multicolumn{3}{c}{\act{lateral}} & \multicolumn{3}{c}{\act{flexion}} & \multicolumn{3}{c|}{\act{jump}} & \multicolumn{4}{>{\columncolor{meangray}}c}{\textbf{Mean}} \\
\cmidrule(l{0.5em}r{0.5em}){2-4}\cmidrule(l{0.5em}r{0.5em}){5-7}\cmidrule(l{0.5em}r{0.5em}){8-10}\cmidrule(l{0.5em}r{0.5em}){11-13}\cmidrule(l{0.5em}r{0.5em}){14-16}\cmidrule(l{0.5em}r{0.85em}){17-19}\cmidrule(l{0.85em}r{0.2em}){20-23}
Method\padT\padB
 & Ep.$\hi$ & $q\lo$ & $F^{\uparrow}_{\mathrm{Z}_r}$
 & Ep.$\hi$ & $q\lo$ & $F^{\uparrow}_{\mathrm{Z}_r}$
 & Ep.$\hi$ & $q\lo$ & $F^{\uparrow}_{\mathrm{Z}_r}$
 & Ep.$\hi$ & $q\lo$ & $F^{\uparrow}_{\mathrm{Z}_r}$
 & Ep.$\hi$ & $q\lo$ & $F^{\uparrow}_{\mathrm{Z}_r}$
 & Ep.$\hi$ & $q\lo$ & $F^{\uparrow}_{\mathrm{Z}_r}$
 & Ep.$\hi$ & $q\lo$ & \phantom{.00}\llap{$F^{\uparrow}_{\mathrm{Z}_r}$} & $F^{\uparrow}_{\mathrm{ML}_r}$ \\
\midrule
\rowcolor{papericeblue}
\multicolumn{19}{@{}>{\fillL}l|}{\padT\textbf{Seen subjects, held-out clips}} & \multicolumn{4}{l}{} \\[1pt]
\gd{GaitDynamics}$^\ast$ & -- & -- & \gd{1.00} & -- & -- & \gd{1.00} & -- & -- & \gd{.75} & -- & -- & \gd{.90} & -- & -- & \gd{.92} & -- & -- & \gd{.90} & -- & -- & \gd{.91} & \gd{.92} \\
Fixed-PD       & \underline{99.9} & 2.48 & .95 & 99.7 & \underline{3.36} & .96 & 97.8 & \underline{1.34} & .36 & 95.9 & \underline{3.77} & .85 & 91.3 & \underline{2.76} & .63 & 73.6 & \underline{3.76} & .64 & 93.0 & \underline{2.91} & .73 & .53 \\
Torque         & 99.7 & \underline{2.33} & \textbf{.98} & \underline{99.8} & 3.51 & .96 & \underline{99.6} & 2.33 & \textbf{.64} & \underline{97.6} & 6.49 & \underline{.89} & \underline{96.8} & 4.62 & \textbf{.73} & \underline{78.9} & 7.61 & \textbf{.74} & \underline{95.4} & 4.48 & \textbf{.82} & \underline{.62} \\
Muscle         & 98.6 & 2.72 & .97 & 99.7 & 3.84 & \underline{.98} & 93.2 & 2.14 & .52 & 92.4 & 8.91 & \underline{.89} & 68.2 & 12.3 & .62 & 56.4 & 13.2 & .62 & 84.8 & 7.18 & .77 & \underline{.62} \\
\rowcolor{imptint}\fillL\textcolor[RGB]{229,57,53}{\textbf{ImpedanceMimic}}\padB & \textbf{100} & \textbf{1.66} & \textbf{.98} & \textbf{100} & \textbf{2.24} & \textbf{.99} & \textbf{99.7} & \textbf{1.07} & \underline{.56} & \textbf{99.9} & \textbf{3.11} & \textbf{.91} & \textbf{97.7} & \textbf{2.40} & \underline{.70} & \textbf{83.8} & \textbf{3.22} & \underline{.69} & \cellcolor{imptintmean}\textbf{96.9} & \cellcolor{imptintmean}\textbf{2.28} & \cellcolor{imptintmean}\underline{.80} & \cellcolorR{imptintmean}\textbf{.70} \\
\midrule
\rowcolor{papericeblue}
\multicolumn{19}{@{}>{\fillL}l|}{\padT\textbf{Unseen subjects, zero-shot}} & \multicolumn{4}{l}{} \\[1pt]
\gd{GaitDynamics}$^\ast$ & -- & -- & \gd{.99} & -- & -- & \gd{1.00} & -- & -- & \gd{.65} & -- & -- & \gd{.89} & -- & -- & \gd{.91} & -- & -- & \gd{.88} & -- & -- & \gd{.89} & \gd{.92} \\
Fixed-PD       & \underline{99.8} & 2.75 & .94 & \underline{98.6} & \underline{3.31} & .96 & 99.1 & \underline{1.35} & .49 & \underline{93.7} & \underline{4.02} & .78 & \underline{86.6} & \underline{3.19} & .52 & \underline{71.5} & \underline{4.18} & .48 & \underline{91.5} & \underline{3.13} & .70 & .45 \\
Torque         & \underline{99.8} & \underline{2.69} & \textbf{.98} & 94.2 & 3.61 & .96 & \textbf{99.8} & 2.24 & \textbf{.64} & 71.4 & 8.77 & .79 & 64.3 & 5.77 & \underline{.62} & 45.7 & 6.94 & \textbf{.72} & 79.2 & 5.01 & \textbf{.79} & .53 \\
Muscle         & 90.9 & 3.98 & .95 & 97.0 & 4.19 & \textbf{.98} & 92.0 & 2.30 & .52 & 79.3 & 12.9 & \underline{.83} & 49.1 & 14.4 & .58 & 43.7 & 14.2 & .57 & 75.3 & 8.65 & .74 & \underline{.62} \\
\rowcolor{imptint}\fillL\textcolor[RGB]{229,57,53}{\textbf{ImpedanceMimic}}\padB & \textbf{100} & \textbf{1.95} & \underline{.97} & \textbf{100} & \textbf{2.24} & \textbf{.98} & \underline{99.6} & \textbf{.98} & \underline{.54} & \textbf{99.9} & \textbf{3.62} & \textbf{.84} & \textbf{99.7} & \textbf{2.53} & \textbf{.64} & \textbf{89.3} & \textbf{3.41} & \underline{.62} & \cellcolor{imptintmean}\textbf{98.1} & \cellcolor{imptintmean}\textbf{2.45} & \cellcolor{imptintmean}\underline{.76} & \cellcolorR{imptintmean}\textbf{.65} \\
\midrule
\rowcolor{papericeblue}
\multicolumn{19}{@{}>{\fillL}l|}{\padT\textbf{Withheld activity, zero-shot}} & \multicolumn{4}{l}{} \\[1pt]
\gd{GaitDynamics}$^\ast$ & -- & -- & \gd{.96} & -- & -- & \gd{.99} & -- & -- & \gd{.85} & -- & -- & \gd{.86} & -- & -- & \gd{.86} & -- & -- & \gd{.87} & -- & -- & \gd{.90} & \gd{.89} \\
Fixed-PD       & 40.0 & \underline{4.52} & .43 & 73.4 & \underline{5.10} & .75 & 68.6 & \underline{3.08} & .39 & \underline{93.7} & \underline{4.10} & .77 & \underline{71.7} & \underline{3.52} & .61 & 71.2 & \underline{3.98} & .58 & 69.8 & \underline{4.05} & .59 & .35 \\
Torque         & \underline{79.5} & 6.53 & \textbf{.76} & 73.8 & 9.29 & \underline{.85} & \textbf{93.2} & 4.92 & .49 & 86.9 & 7.23 & \underline{.82} & 67.9 & 8.71 & \underline{.64} & \underline{74.5} & 8.72 & \textbf{.67} & \underline{79.3} & 7.57 & \underline{.71} & .47 \\
Muscle         & 32.8 & 12.2 & \underline{.74} & \underline{80.6} & 8.39 & \underline{.85} & 65.0 & 7.45 & \underline{.53} & 83.3 & 10.1 & \textbf{.84} & 64.0 & 11.7 & .60 & 57.5 & 13.3 & .53 & 63.9 & 10.5 & .68 & \textbf{.61} \\
\rowcolor{imptint}\fillL\textcolor[RGB]{229,57,53}{\textbf{ImpedanceMimic}}\padB & \textbf{87.1} & \textbf{3.43} & .66 & \textbf{95.6} & \textbf{3.55} & \textbf{.92} & \underline{91.9} & \textbf{2.34} & \textbf{.64} & \textbf{98.9} & \textbf{3.41} & \underline{.82} & \textbf{80.5} & \textbf{3.06} & \textbf{.65} & \textbf{83.8} & \textbf{3.48} & \underline{.65} & \cellcolor{imptintmean}\textbf{89.6} & \cellcolor{imptintmean}\textbf{3.21} & \cellcolor{imptintmean}\textbf{.72} & \cellcolorR{imptintmean}\underline{.57} \\
\bottomrule
\end{tabular}%
}
\vspace{0pt}
\parbox{\textwidth}{\scriptsize\raggedright
$^\ast$GaitDynamics is shown in italics for reference and excluded from the ranking, since it is not a controller.
  Ep. = episode survival before termination (\%). $q$ = joint-angle RMSE ($^\circ$). $F_{\mathrm{Z}_r}$, $F_{\mathrm{ML}_r}$ = Pearson correlation w.r.t. reference vertical and lateral~GRF.}
\vspace{-0.7em}%
\end{table*}
\begin{figure}[t]
    \centering
    \includegraphics[width=\linewidth]{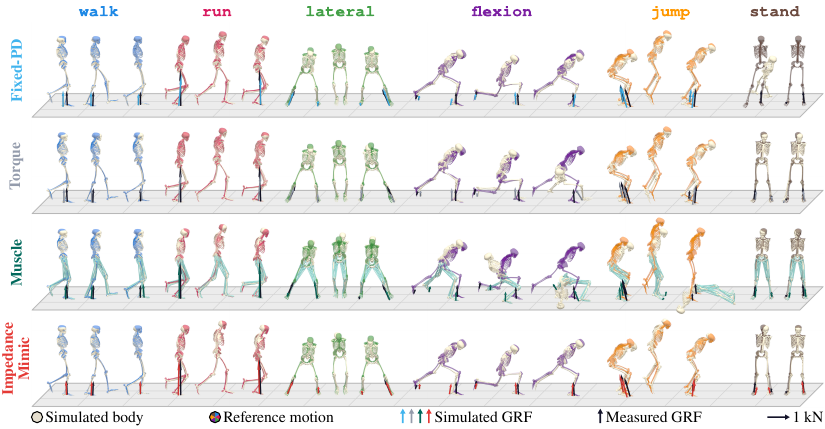}
    \vspace{-1.3em}%
    \caption{\textbf{Action space drives force recovery.} Fixed-PD, Torque, Muscle and \projname{} on six held-out clips, one per activity. Three sequential frames per clip, overlaid on the reference motion and measured ground reaction forces at matched time points. Table~\ref{tab:golden} reports results.}
    \label{fig:actionspace}
    \vspace{-1.6em}
\end{figure}

Across six activities and 129 human morphologies, \projname{} outperforms the other action spaces in imitating both a motion and its contact forces (Table~\ref{tab:golden}, Fig.~\ref{fig:actionspace}).
On seen subjects, it has the lowest joint-angle error and the highest survival in every activity, and it leads on nearly every additional kinematic, force and center-of-pressure metric (Table~\ref{tab:golden_full}, Fig.~\ref{fig:waveforms}).
Its force accuracy approaches that of GaitDynamics, a dedicated kinematics-to-force regressor that serves as a reference, even though \projname{} must also produce the motion itself by driving each participant's body model under full physics, whereas the regressor reads the forces directly off the recorded kinematics.
Fixed-PD has the second-lowest joint-angle error but the largest vertical-force RMSE of the three action-space baselines in standing, flexion and jumping (Table~\ref{tab:golden_full}), activities spent largely in double support, where the motion does not fix how the force is split between the feet.
Intuitively, the legs share the load like springs in parallel, and Fixed-PD can shift the load only through position offsets that the policy needs to hold the pose, whereas \projname{} stiffens the loaded joints without compromising kinematic tracking.
The action-space bounds constrain fixed gains further in other activities (App.~\ref{App:humanoid_bounds}).
In running, the nominal stiffness misses Bound~1 on up to half of the knee and ankle contact frames, against under a tenth for the learned stiffness (Fig.~\ref{fig:humanoid_bounds}), and in swing the nominal ankle damping violates Bound~2 on every frame, and capping it raises the vertical-force correlation (Table~\ref{tab:existing_full}).
Torque has a vertical-force accuracy similar to \projname{}'s, its slim lead in correlation ($0.82$ against $0.80$) coming mostly from standing, where the force stays near body weight.
Overall, however, it has twice the joint-angle error and is less accurate on horizontal forces (Table~\ref{tab:golden_full}), as a torque held over the $10$\,ms policy step cannot correct deviations the way a PD law does at every $2$\,ms physics step.
Acting at the physics rate leaves this error unchanged ($4.55^\circ$ against $4.48^\circ$, Table~\ref{tab:torque_variants}), since a torque policy must learn the feedback the PD law computes from the joint state~\citep{peng2017learning}.%
Muscle produces the smoothest torques and near-best forces in walking and running, but the lowest survival, particularly in flexion and jumping, likely because its activation dynamics (App.~\ref{App:muscle}) leave too little margin to recover from rapid load changes.

\vspace{-0.225em}%
Learning the gains triples the action dimension, from 15 to 45, without changing the network size or the simulation throughput.
The wider action space does not slow learning, in line with reports that variable impedance raises the learning rate in contact-rich manipulation~\citep{martin2019variable, bogdanovic2020learning}.
\projname{}, Fixed-PD and Torque each take about $1.3$ GPU-hours per run on an NVIDIA RTX PRO 6000, against $1.8$ for Muscle (Table~\ref{tab:compute}), and \projname{} converges fastest, approaching its plateau within $0.5 \times 10^9$ steps (Fig.~\ref{fig:training_curves}).

\vspace{-0.225em}%
These results agree with the kinematic comparison of \citet{peng2017learning}, who found PD targets comparable to the best action space on every motion, torques among the worst and muscle activations the slowest to learn, and attributed the advantage of PD to its local feedback.
Scoring the contact forces as well shows that fixed gains keep the pose at the cost of the force, and that learned impedance, which keeps the local feedback of PD while adapting it to the state, leads on both.
\label{sec:plausibility}
\vspace{-0.8em}\vspace{0pt}%

\subsection{Zero-Shot Transfer Across Human and Robot Embodiments}
\label{sec:morphology}
\vspace{-0.6em}%

\begin{figure}[!b]
    \vspace{-1.6em}%
    \centering
    \includegraphics[width=\linewidth]{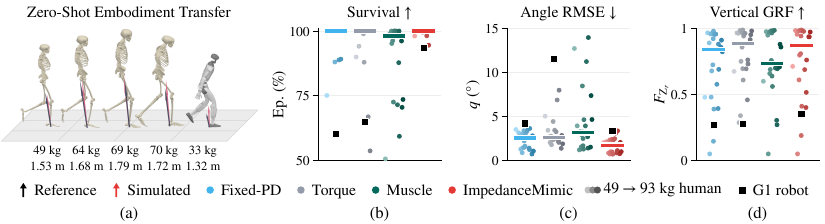}
    \vspace{-1.3em}%
    \caption{\textbf{Zero-shot transfer across humans and robots.} \textbf{(a)}~A single \projname{} policy without anthropometric input drives unseen human and robot embodiments. \textbf{(b--d)}~Survival, joint-angle RMSE and vertical-GRF correlation on the 22 held-out subjects. Each point denotes one subject (square for the G1), placed and shaded by body mass, with medians as horizontal lines.}
    \label{fig:morphology}
    \vspace{-1.7em}
\end{figure}

\projname{} transfers zero-shot to new human and robot bodies without any anthropometric input (Table~\ref{tab:golden}, Fig.~\ref{fig:morphology}).
On the 22 held-out participants, who span 49--93\,kg and 1.50--1.82\,m, it retains the survival and joint-angle error of its seen-split counterpart with little loss of force accuracy.
Fixed-PD keeps its survival and joint-angle error, and with them its force deficit.
Torque keeps its vertical-force accuracy but loses survival, mostly in the non-cyclic activities, as does Muscle.

\vspace{-0.225em}%
The inertia prior expresses the controller in the units of each body.
With base gains proportional to the joint-space inertia (Eq.~\ref{eq:inertia_scaling}), a given gain multiple yields the same closed-loop natural frequency and damping ratio on every body, and the PD feedback corrects the remaining differences at every physics step, which Torque, normalized by the same inertia, cannot.
Removing the scaling leaves survival intact but multiplies torque jerk by $2.7$ on unseen subjects (Table~\ref{tab:ablation_unseen}), since the damping limit of Bound~2 moves with the inertia of each body while unscaled gains do not.

\vspace{-0.225em}%
The same policies drive a simulated Unitree G1 humanoid without retraining, on held-out clips retargeted to the robot, with gain priors from its own inertia (Fig.~\ref{fig:morphology}a, App.~\ref{App:embodiments}).
It is lighter and shorter (33\,kg, 1.32\,m) than all 107 training participants.
Without force plates, its reference forces come from the estimator's physics prior on its center-of-mass acceleration.
\projname{} is best on every activity-mean metric among the action spaces (Table~\ref{tab:embodiments}, Fig.~\ref{fig:morphology}b--d), with its survival falling from $98.1$ to $82.1\,\%$ against losses of about 35 points for Fixed-PD ($56.9\,\%$) and Torque ($44.0\,\%$).

\subsection{Learned Impedance Encodes Motor State}
\label{sec:loao}
\vspace{-0.2em}%
\begin{figure}[t]
    \centering
    \includegraphics[width=\linewidth]{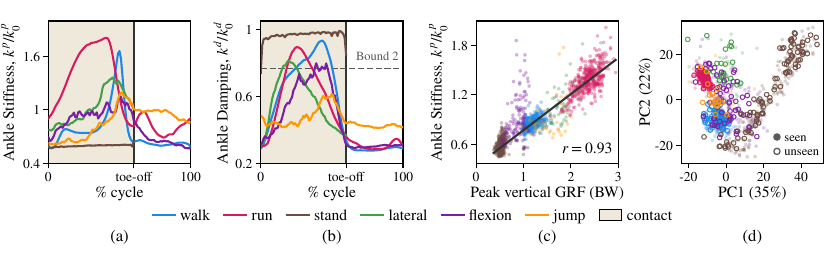}
    \vspace{-1.35em}%
    \caption{\textbf{The learned impedance encodes motor state.} \textbf{(a)}~Ankle stiffness and \textbf{(b)}~ankle damping over the normalized cycle, as medians over one task per activity. \textbf{(c)}~Contact-mean ankle stiffness scales with peak vertical GRF, one point per cycle. \textbf{(d)}~Principal components of the seen subjects' impedance gains (filled circles), with unseen subjects (open circles) on the same activity clusters.}
    \vspace{-1.6em}%
    \label{fig:encoder}
\end{figure}

The learned impedance follows motor state, stiffening loaded joints and releasing the ankle before the foot unloads.
We segment held-out clips of seen subjects into movement cycles and characterize each cycle by hip, knee and ankle stiffness and damping, normalized by the subject's nominal gains (Figs.~\ref{fig:waveforms} and~\ref{fig:encoder}a,b, App.~\ref{App:learned_impedance}).
Two patterns recur across activities, one for each action-space bound.
Stiffness follows the torque demanded by the contact force, as Bound~1 requires.
Within a walking stride, ankle stiffness rises from $0.6\times$ nominal at heel strike to $1.5\times$ at push-off, where ankle torque peaks, and across all six activities, its contact mean increases linearly with the cycle's peak vertical force ($r = 0.93$, Fig.~\ref{fig:encoder}c).
Ankle damping is released before unloading, peaking $128$--$196$\,ms before toe-off or take-off in walking, running and jumping, and falling to about a third of nominal by toe-off (Fig.~\ref{fig:encoder}b).
At lift-off, the ankle drives only the foot, which lowers Bound~2 below the peak contact damping (App.~\ref{App:humanoid_bounds}).
Because the gains lag by $50$\,ms (Eq.~\ref{eq:gain_lag}), damping must fall beforehand and then stays below Bound~2 on $99.99\,\%$ of swing frames (App.~\ref{App:learned_impedance}).

\vspace{-0.1em}%
The learned gains partly match human ankle impedance (App.~\ref{App:identifiability}).
Ankle stiffness rises over contact, though less steeply than measured by \citet{rouse2014estimation}, and holds $0.4$ of the toppling stiffness in quiet standing, against $0.9$ in humans~\citep{loram2002direct}.
The damping release matches human data in timing~\citep{lee2015timevarying} but not in magnitude, which Bound~2 caps.

\vspace{-0.1em}%
Across activities, these schedules form a low-dimensional code of motor state.
In a principal component analysis of the standardized, cycle-normalized gain trajectories, two components carry $57\,\%$ of the variance (Fig.~\ref{fig:pca_scree}a).
The first encodes activity, which accounts for $89\,\%$ of its variance, and the second encodes activity and subject (Figs.~\ref{fig:encoder}d and~\ref{fig:pca_scree}b).
Projected onto these components, cycles of the 22 unseen subjects fall on the same activity clusters (App.~\ref{App:learned_impedance}), so new bodies reuse the code learned in training.
A logistic regression on the first ten principal components, cross-validated across subjects, recovers the activity of a single cycle with a macro recall of $92.5\,\%$ (Fig.~\ref{fig:pca_scree}c).
The activity can thus be read from the gains of one cycle alone, without kinematics or forces, making learned impedance a decodable latent of motor state.
A fixed waveform per activity would form the same clusters, but replacing the learned gains with their subject-mean, phase-indexed schedule and retraining the offsets lowers survival below that of fixed gains ($88.5$ against $93.0\,\%$, App.~\ref{App:controller}), meaning the policy also adapts the gains to the state within each cycle.

\vspace{-0.1em}%
This shared code carries over to activities withheld from training, where \projname{} generalizes best (Table~\ref{tab:golden}, bottom block).
Trained on five activities and evaluated zero-shot on the sixth, it has the lowest joint-angle error in every fold and the highest survival in all but standing, losing 7 points of survival relative to training on all six, against 16 to 23 for the baselines.

\vspace{-0.1em}%
\projname{} loses force accuracy mainly when walking is withheld, keeping the pose but dropping to a vertical-force correlation of $0.66$, against $0.98$ when walking is trained (Tables~\ref{tab:golden} and~\ref{tab:loao_full}).
On these clips, the damping waveform of the policy trained without walking correlates with that of the policy trained on walking ($r = 0.99$), including the release before toe-off and the swing level, but reaches only $66\,\%$ of the peak damping and $46\,\%$ of the push-off stiffening (App.~\ref{App:learned_impedance}).
This split between timing and magnitude is consistent with the two bounds, since the release is set by Bound~2, which depends only on foot inertia, the same in every activity, whereas the push-off stiffening is set by Bound~1, which depends on the torque each activity demands.
\label{sec:encoder}

\subsection{Open Benchmark: Ablations, Sensitivity and Robustness}
\label{sec:benchmark}
\vspace{-0.6em}%

\begin{table}[t]%
\centering
\caption{\textbf{ImpedanceMimic recovers the most accurate forces under every integration scheme.} Activity means over 5 seeds, per benchmark (full results in Tables~\ref{tab:existing_full} and~\ref{tab:existing_transfer}). Fixed-PD holds the nominal gains, its damping capped at $I_{jj}/h$ under explicit PD only. $2\times$ doubles the stiffness priors, and under stable and implicit PD also the damping priors. \textbf{Bold} / \underline{underline} = best / second best.}
\label{tab:existing}
\vspace{-0.3em}%
\scriptsize%
\setlength{\tabcolsep}{3.15pt}
\setlength{\aboverulesep}{0pt}%
\begin{tabular}{@{}l<{\hspace{13.70pt}}l>{\hspace{23.25pt}}cccl>{\hspace{23.25pt}}cccl>{\hspace{23.25pt}}ccc>{\fillR}l<{\hspace{10.95pt}}@{}}
\toprule
& & \multicolumn{4}{>{\hspace{23.25pt}}c}{\textbf{Seen subjects}} & \multicolumn{4}{>{\hspace{23.25pt}}c}{\textbf{Unseen subjects}} & \multicolumn{4}{>{\hspace{23.25pt}}c<{\hspace{7.80pt}}}{\textbf{Withheld activity}} \\
\cmidrule(l{25.90pt}r{2.65pt}){3-6}\cmidrule(l{25.90pt}r{2.65pt}){7-10}\cmidrule(l{25.90pt}r{10.20pt}){11-14}
\textbf{PD} & \textbf{Method} & Ep.$\hi$ & $q\lo$ & \makebox[0pt]{$F^{\uparrow}_{\mathrm{Z}_r}$} & \makebox[0pt][l]{$F^{\uparrow}_{\mathrm{ML}_r}$} & Ep.$\hi$ & $q\lo$ & \makebox[0pt]{$F^{\uparrow}_{\mathrm{Z}_r}$} & \makebox[0pt][l]{$F^{\uparrow}_{\mathrm{ML}_r}$} & Ep.$\hi$ & $q\lo$ & \makebox[0pt]{$F^{\uparrow}_{\mathrm{Z}_r}$} & \makebox[0pt][l]{$F^{\uparrow}_{\mathrm{ML}_r}$} \\
\midrule
 & Fixed-PD & 93.0 & 2.91 & .73 & .53 & 91.5 & 3.13 & .70 & .45 & 69.8 & 4.05 & .59 & .35 \\
 & Fixed-PD, $2\times$ & 96.2 & 2.53 & .72 & .50 & 95.9 & 2.80 & .68 & .44 & 83.1 & 3.53 & .59 & .35 \\
\rowcolor{imptint}\cellcolor{white} & \textcolor[RGB]{229,57,53}{\textbf{\projname{}}} & \underline{96.9} & 2.28 & \textbf{.80} & \textbf{.70} & \textbf{98.1} & 2.45 & \textbf{.76} & \textbf{.65} & 89.6 & 3.21 & \textbf{.72} & \textbf{.57} \\
\rowcolor{imptint}\cellcolor{white}\multirow{-4}{*}{Explicit} & \textcolor[RGB]{229,57,53}{\textbf{\projname{}, $2\times$}} & \underline{96.9} & 2.30 & \underline{.79} & .68 & \underline{97.8} & 2.47 & \underline{.75} & .63 & 90.3 & 3.22 & .69 & .49 \\
\midrule
 & Fixed-PD & 93.5 & 2.44 & .75 & .65 & 92.2 & 2.70 & .72 & .59 & 78.9 & 3.32 & .69 & .51 \\
 & Fixed-PD, $2\times$ & 96.6 & \underline{2.10} & .75 & .67 & 96.2 & 2.35 & .71 & .61 & 85.8 & \underline{2.94} & .68 & .51 \\
 & \textit{SimPoE}$^\ast$ & \textit{96.4} & \textit{2.68} & \textit{.76} & \textit{.62} & \textit{96.2} & \textit{2.99} & \textit{.71} & \textit{.59} & \textit{93.0} & \textit{3.87} & \textit{.67} & \textit{.54} \\
\rowcolor{imptint}\cellcolor{white} & \textcolor[RGB]{229,57,53}{\textbf{\projname{}}} & 93.8 & 2.22 & \underline{.79} & \textbf{.70} & 94.9 & 2.45 & .74 & \underline{.64} & 83.5 & 3.16 & .70 & .52 \\
\rowcolor{imptint}\cellcolor{white}\multirow{-5}{*}{Stable} & \textcolor[RGB]{229,57,53}{\textbf{\projname{}, $2\times$}} & 96.3 & \textbf{2.07} & .77 & .68 & 97.4 & \textbf{2.27} & .73 & .63 & 88.0 & \textbf{2.92} & .69 & .52 \\
\midrule
 & Fixed-PD & 95.7 & 2.52 & .72 & .61 & 94.6 & 2.72 & .70 & .55 & 80.8 & 3.46 & .67 & .50 \\
 & Fixed-PD, $2\times$ & 95.8 & 2.21 & .67 & .56 & 96.5 & 2.47 & .65 & .50 & 83.4 & 3.13 & .60 & .48 \\
 & PHC & 93.5 & 8.66 & .71 & .59 & 88.0 & 10.8 & .66 & .57 & 70.2 & 13.3 & .61 & .52 \\
\rowcolor{imptint}\cellcolor{white} & \textcolor[RGB]{229,57,53}{\textbf{\projname{}}} & \textbf{97.2} & 2.22 & .77 & .68 & \underline{97.8} & 2.38 & \underline{.75} & .62 & \textbf{92.6} & 3.07 & \underline{.71} & \textbf{.57} \\
\rowcolor{imptint}\cellcolor{white}\multirow{-5}{*}{Implicit} & \textcolor[RGB]{229,57,53}{\textbf{\projname{}, $2\times$}} & 96.8 & 2.16 & .72 & .62 & 96.8 & \underline{2.33} & .70 & .56 & \underline{90.4} & 3.00 & .65 & .50 \\
\bottomrule
\end{tabular}
\vspace{0pt}
\parbox{\textwidth}{\scriptsize\raggedright $^\ast$SimPoE is shown in italics for reference and excluded from the ranking, since its residual wrench acts directly on the unactuated pelvis.\\ Ep. = survival before termination (\%). $q$ = joint-angle RMSE ($^\circ$). $F_{\mathrm{Z}_r}$, $F_{\mathrm{ML}_r}$ = Pearson correlation w.r.t. reference vertical and lateral~GRF.}
\vspace{-2.4em}%
\end{table}

We compare \projname{} with the existing imitation controllers PHC~\citep{luo2023perpetual} and SimPoE~\citep{yuan2021simpoe}, each under the PD integration of its original implementation (Table~\ref{tab:existing}, App.~\ref{App:existing}).
PHC, with our full reward in place of its own, still reaches $8.7$ to $13.3^\circ$ of joint-angle error and trails nominal Fixed-PD and \projname{} under implicit damping in vertical-force correlation.
SimPoE survives nearly as often as the best controllers but tracks joint angles and vertical forces less accurately than \projname{}. Its survival, however, depends on a residual root wrench that no joint or contact produces and that balances it instead of its feet (Table~\ref{tab:simpoe_wrench}); without it, its survival drops to $85$, $80$ and $56\,\%$ (Tables~\ref{tab:existing_full} and~\ref{tab:existing_transfer}).
Retraining Fixed-PD and \projname{} under stable PD~\citep{tan2011stable} and implicit damping~\citep{todorov2012mujoco} separates the integration scheme from the learned gains.
With the same gain priors, \projname{} leads Fixed-PD in joint-angle error and every force correlation on every benchmark (Table~\ref{tab:existing}), survives more often in all but one cell ($96.3$ against $96.6\,\%$ on seen subjects at doubled stable priors), and its explicit-PD force correlations exceed those of Fixed-PD under any scheme.
The lead carries to the Unitree G1, where the learned gains keep the highest survival under every scheme and gain level, by $1.4$ to $25.2$ points (Table~\ref{tab:embodiments_integrators}).
Doubled priors lower the joint-angle error of Fixed-PD under every scheme and reach their lowest values for \projname{} under stable PD, yet never raise the vertical-force correlation of either method, as the integrator lifts Bound~2 but leaves Bound~1 and the split of the load in double support to the gains.
At nominal priors under stable PD, \projname{} trails doubled fixed gains in survival, a gap concentrated in flexion ($85.1$ against $99.0\,\%$ seen) that doubled priors close ($98.4\,\%$) and that the policy closes by itself under explicit PD ($97.7$ against $91.3\,\%$). The learned gains modulate around the prior and do not replace the choice of its scale.

\begin{figure}[t]%
    \centering
    \includegraphics[width=\linewidth]{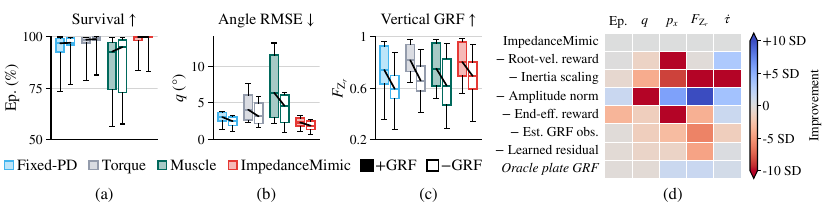}
    \vspace{-1.5em}%
    \caption{\textbf{Ablations.} \textbf{(a)}~Survival, \textbf{(b)}~joint-angle RMSE and \textbf{(c)}~vertical-GRF correlation of each action space with (solid) and without (hollow) the GRF reward, over activity means (Table~\ref{tab:reward_factorial}). \textbf{(d)}~Change in (a--c), root drift $p_x$ and torque jerk $\dot\tau$, in seed standard deviations of \projname{}, when a component is removed or the measured GRF replaces its estimate (oracle, Table~\ref{tab:ablation_seen}).}
    \label{fig:benchmark}
    \vspace{-1.5em}%
\end{figure}

\vspace{-0.1em}
Beyond the three benchmarks, we examine what drives the fidelity and robustness of \projname{} (App.~\ref{App:ablation}).
To isolate the contribution of force supervision, we retrain every action space without the GRF reward (Fig.~\ref{fig:benchmark}a--c, Table~\ref{tab:reward_factorial}).
In every action space, joint-angle error decreases while vertical-force correlation drops, so closer kinematic tracking does not recover the forces.
Without this reward, \projname{} surpasses Torque in force correlation, recovering more force from kinematic tracking alone than any other action space.

\vspace{-0.1em}
We perform ablations of \projname{}, from the inertia-scaled gains and reward terms to the force estimate and training-only signals (Fig.~\ref{fig:benchmark}d, Table~\ref{tab:ablation_seen}).
Removing the inertia scaling has the largest effect, quadrupling torque jerk and lowering every force correlation.
The estimated GRF improves force correlation, and the measured GRF adds little beyond it (Table~\ref{tab:ablation_unseen}).
Force fidelity requires no privileged information, as removing the auxiliary GRF head or the asymmetric critic, which use measured forces only in training, leaves all metrics within seed variation.
The log-ranges $s_p$ and $s_d$ of Eq.~\ref{eq:learned_impedance} bound how far stiffness and damping depart from their nominal values, and the activation lag $T_{\mathrm{act}}$ of Eq.~\ref{eq:gain_lag} limits how fast they change.
Varying each has little effect on survival, but fidelity depends far more on the damping than on the stiffness range (Table~\ref{tab:sensitivity}), as Bound~2 binds at the ankle in every swing phase and Bound~1 mainly in running and lateral cutting (App.~\ref{App:humanoid_bounds}).
Restricting damping to within $10\,\%$ of nominal multiplies torque jerk sixfold and lowers the vertical-force correlation four times as much as the same restriction on stiffness.
An activation lag of up to $50$\,ms is neutral, whereas $200$\,ms triples torque jerk, as the gains trail the contact events.
Finally, we evaluate the four action spaces without retraining under eleven test-time perturbations, where \projname{} keeps the lowest joint-angle error throughout and the highest survival under all but the largest contact spheres (Table~\ref{tab:robustness}).
Under a $20$\,ms observation delay, \projname{} retains $93.5\,\%$ survival, as its PD law acts on the current state at every simulation step, whereas Torque, whose corrections pass through the delayed policy, falls to $3.3\,\%$.
Fixed-PD closes the same loop with fixed gains yet falls to $65.6\,\%$, a collapse largely of explicit PD ($89.1$ and $84.4\,\%$ under stable PD and implicit damping, against $91.6$ and $91.4\,\%$ for \projname{}, Table~\ref{tab:robustness_integrators}).

\vspace{-0.8em}\vspace{0pt}%
\section{Discussion and Conclusion}
\label{sec:discussion}
\vspace{-0.6em}%
We introduced \projname{}, a motion-imitation policy that outputs per-joint stiffness, damping and setpoint offsets, supervised by force measurements withheld at inference.
Across 129 participants and six activities, \projname{} is the best force-accurate motion imitation controller, at no additional training cost.
It outperforms the Torque and Muscle action spaces, as well as the existing imitation controllers PHC and SimPoE, and its advantage over fixed gains holds under every PD integration scheme.
Its gains, learned as multiples of inertia-scaled priors, transfer zero-shot to unseen humans and a simulated humanoid robot.
The learned impedance forms a decodable representation of motor state, stiffening with load and releasing damping before the foot unloads, an anticipatory behavior the policy reproduces even in activities withheld from training.

\vspace{-0.225em}%
Our results further highlight the role of physical simulation in recovering physically realizable forces.
GaitDynamics, a regressor that leads in vertical-force correlation, loads the ground on $30\,\%$ of lateral-cutting and $39\,\%$ of jump flight frames, against $11\,\%$ and $17\,\%$ for \projname{} (a tracking error in contact timing, not a physical inconsistency), and places the center of pressure outside the foot on $6\,\%$ and $17\,\%$ of their loaded frames, against at most $0.2\,\%$ for \projname{} (App.~\ref{App:physics_realisticness}).
Yet simulation alone does not guarantee consistency with the dynamics of the body, as is the case for SimPoE, where a residual external wrench supplies more than half of the mediolateral force and with it much of the balance (Table~\ref{tab:simpoe_wrench}).
Without such a wrench, balance rests on the foot-contact model, as our robustness tests show (App.~\ref{App:robustness}).
Together with the subject models provided by the datasets, which lack the arms humans use to recover balance~\citep{bruijn2010effects}, and the low coverage of non-cyclic activities ($16\,\%$ of the training clips, Table~\ref{tab:splits}), this likely explains why every method performs worse in lateral cutting, flexion and jumping.
For a controller that acts only through its joints, the action-space bounds explain why fixed gains cannot reconcile force authority in contact with stability in swing.
With stiff gains, a stable integrator recovers the kinematic accuracy of learned impedance but not its force accuracy, as only a load-dependent stiffness redistributes the load between the feet in double support.
Learned impedance sets damping independently of stiffness, releasing the ankle damping ahead of toe-off while the stiffness peaks at push-off (Fig.~\ref{fig:encoder}).
On hardware, joint-level motor drivers evaluate the PD law at kilohertz rates without the mass matrix that stable PD requires, and their shorter step $h$ raises Bound~2, leaving the learned damping conservative.
They would, however, expose the policy to effects that our simulation omits, including actuator bandwidth, stiction and torque limits (App.~\ref{App:embodiments}).
Powered lower-limb prostheses already modulate joint impedance with gait phase, using gains tuned by hand for each user~\citep{sup2008design}, and a learned, state-dependent impedance could replace this tuning once its force estimate is causal, as the estimator and the reference lookahead currently draw on up to $400$\,ms of future motion.

\newpage

\subsection*{Acknowledgments}

\ificlrfinal
This work has been made possible in part by a gift from the Chan Zuckerberg Initiative Foundation to establish the Kempner Institute for the Study of Natural and Artificial Intelligence at Harvard University.
\else
Omitted for double-blind review.\\\mbox{}\\\mbox{}%
\fi

\subsection*{AI use statement}

In this work, we used generative AI tools for literature research, for polishing author-written text, and as a coding assistant, including for LaTeX table formatting. All research ideas, the experimental design, and the analysis are our own. We therefore did not use generative AI for ideation or for the generation of research artifacts such as figures, data, or proofs.

We have reviewed all AI-assisted work: we read every paper we cite in its original published form, and we reviewed every line of AI-assisted code. We take responsibility for the final content of this work, including text, claims, and artifacts produced with the aid of generative AI.

\subsection*{Reproducibility statement}

Code is released openly at {\color{magenta}\texttt{https://impedancemimic.github.io/ImpedanceMimic/}}.
Experimental data used in this work is open-sourced at {\color{magenta}\texttt{https://addbiomechanics.org/}}.

{\small%
\bibliography{iclr2027_conference}
\bibliographystyle{iclr2027_conference}
}

\newpage
\appendix
\raggedbottom%
\captionsetup{skip=4pt}
\setlength{\abovedisplayskip}{\defabovedisp}\setlength{\belowdisplayskip}{\defbelowdisp}
\setlength{\abovedisplayshortskip}{\defabovedispshort}\setlength{\belowdisplayshortskip}{\defbelowdispshort}
\setlength{\textfloatsep}{\deftextfloatsep}\setlength{\floatsep}{\deffloatsep}\setlength{\intextsep}{\defintextsep}\setlength{\parskip}{\defparskip}
\let\section\defsectionmacro \let\subsection\defsubsectionmacro \let\paragraph\defparagraphmacro
\section{Appendix}

\subsection{Datasets and Preprocessing}
\label{App:Datasets_Preprocessing}

\subsubsection{Dataset Descriptions}
\label{App:dataset_descriptions}

Our corpus pools nine public datasets that record joint kinematics together with force-plate GRF.
Eight come from the AddBiomechanics release~\citep{werling2024addbiomechanics}, which provides a no-arm Rajagopal model~\citep{rajagopal2016full} scaled to each subject and, for every trial, joint angles and force-plate data in a unified format.
We add \citet{scherpereel2023human} to cover non-cyclic activities, processing its raw motion-capture files with the AddBiomechanics tool~\citep{werling2023addbiomechanics}.
After curation, the corpus contains 129 participants, 972 trials, 8,493 clips and 20.2 hours of motion at 100~Hz (Table~\ref{tab:dataset_summary}).
We exclude the two largest paired motion--force datasets of AddBiomechanics, as \citet{carter2024full} lacks CoP data and \citet{camargo2021comprehensive} has excessive marker error.

{\setlength{\intextsep}{6.5pt plus 2pt minus 2pt}%
\begin{table}[h]
\centering
\caption{\textbf{Dataset summary.} Subj.~= retained / recorded participants. $H$~= participants held out in the unseen split (App.~\ref{App:splits}). Rate~= source sampling rate, resampled to 100\,Hz. TM~= instrumented treadmill, OG~= overground force plates. Activities: \texttt{\textbf{\actl{walk}{W}}}alk, \texttt{\textbf{\actl{run}{R}}}un, \texttt{\textbf{\actl{stand}{S}}}tand, \texttt{\textbf{\actl{flexion}{F}}}lexion, \texttt{\textbf{\actl{lateral}{L}}}ateral, \texttt{\textbf{\actl{jump}{J}}}ump.}
\vspace{-0.4em}
\label{tab:dataset_summary}
\small
\setlength{\tabcolsep}{3.1pt}%
\begin{tabular}{@{}lccrrrrccl@{}}
\toprule
\textbf{Dataset} & \textbf{Subj.} & \textbf{\boldmath$H$} & \textbf{Trials} & \textbf{Clips} & \textbf{Frames} & \textbf{Hours} & \textbf{Rate (Hz)} & \textbf{Setting} & \texttt{\textbf{Activities}} \\
\midrule
\citet{dos2017data}           & 49/49 & 9 & 294 & 1,764 & 1,762,236 &  4.90 & 100      & OG    & \texttt{\textbf{\actl{stand}{S}}} \\
\citet{fregly2012grand}           & 6/6   & 1 & 113 &  115 &  19,061 & 0.05 & 120--200 & OG    & \texttt{\textbf{\actl{walk}{W}}} \\
\citet{hamner2013muscle}           & 10/10 & 2 &  36 &   36 &  5,226 & 0.01 & 100--120 & TM    & \texttt{\textbf{\actl{run}{R}}} \\
\citet{han2023groundlink}                 & 7/7   & 1 & 117 &  144 & 90,479 & 0.25 & 250      & OG    & \texttt{\textbf{\actl{walk}{W} \actl{stand}{S} \actl{flexion}{F} \actl{lateral}{L} \actl{jump}{J}}} \\
\citet{moore2015elaborate}             & 12/15 & 2 &  35 & 2,047 & 1,967,364 & 5.46 & 100      & TM    & \texttt{\textbf{\actl{walk}{W} \actl{stand}{S}}} \\
\citet{scherpereel2023human} & 11/12 & 2 & 116 & 1,001 & 464,459 & 1.29 & 200      & OG    & \texttt{\textbf{\actl{stand}{S} \actl{flexion}{F} \actl{lateral}{L} \actl{jump}{J}}} \\
\citet{tan2022strike}                 & 8/16  & 1 &  79 &  818 & 771,555 & 2.14 & 200      & TM    & \texttt{\textbf{\actl{run}{R}}} \\
\citet{tan2022imu}                 & 17/17 & 3 &  68 & 1,648 & 1,577,293 & 4.38 & 100      & TM    & \texttt{\textbf{\actl{walk}{W}}} \\
\citet{wang2023wearable}               & 9/12  & 1 & 114 &  920 &  614,376 & 1.71 & 100      & TM/OG & \texttt{\textbf{\actl{walk}{W} \actl{run}{R} \actl{flexion}{F} \actl{jump}{J}}} \\
\midrule[\cmidrulewidth]
\textbf{Total} & 129 & 22 & 972 & 8,493 & 7,272,049 & 20.20 & & & \\
\bottomrule
\end{tabular}
\end{table}}

Quality checks exclude whole subjects, trials and frames.
Subjects are dropped when AddBiomechanics processing fails or yields an inconsistent scaled model (Table~\ref{tab:dataset_summary}), and trials when they involve unmeasured external forces, such as walking poles, incomplete force-plate coverage, or geometry we do not simulate, such as stairs, ramps and chairs.
Frames are marked invalid where the plates miss part of the force, detected from the per-foot vertical force at a 0.02\,BW threshold, such as double contact in running or both feet unloaded in walking, standing and lateral cutting.

\vspace{-0.35em}
\subsubsection{Data Preprocessing}
\label{App:preprocessing}
\vspace{-0.25em}

\paragraph{Subject models and trial extraction.}
Each AddBiomechanics subject file (\texttt{.b3d}) holds a scaled OpenSim model and its trials.
We read the final dynamics pass and export one HDF5 file per trial with the 23 generalized coordinates of the no-arm Rajagopal model~\citep{rajagopal2016full}, their velocities and the per-foot GRF and CoP, rejoining recordings that AddBiomechanics splits into 20\,s segments.
Each scaled \texttt{.osim} model is converted to MuJoCo with MyoConverter~\citep{wang2022myosim}, keeping the skeleton, inertias and coupling constraints (App.~\ref{App:ImpedanceMimic}).
Only the Muscle baseline retains the 80 converted muscle--tendon units (App.~\ref{App:muscle}).
The MuJoCo model adds eight auxiliary knee coordinates coupled to the knee angles, giving the $n = 31$ coordinates of Eq.~\ref{eq:eom}, of which six describe the pelvis, 15 are actuated and ten are passive.
Joint torques are recomputed with MuJoCo inverse dynamics on each subject's model.
GRF are normalized to body weight, estimated from standing trials or, where a subject has none, from the pelvis residual force.

\paragraph{Signal processing.}
Joint angles and GRF are resampled to 100\,Hz with a polyphase FIR filter.
Isolated GRF dropouts of up to three frames are interpolated, and two frames are trimmed from each trial end.
GRF, CoP and pelvis velocity are rotated into a heading-local frame, invariant to laboratory orientation.
For the five treadmill datasets (Table~\ref{tab:dataset_summary}), we convert the belt motion into equivalent overground motion.
A loaded foot cannot slide, so its forward-kinematic displacement measures the belt travel, which we spline-interpolate across strides and add to the pelvis translation, velocity and CoP, leaving a residual foot slide below 2\,cm in walking.
Loaded frames whose CoP falls outside an oriented bounding box of the foot are interpolated from neighboring frames or zeroed.

\paragraph{Clip segmentation.}
Valid trials are cut into clips of 1--10\,s at activity-specific boundaries, namely double-support onsets in walking, foot strikes in running, 10\,s windows in standing and the midpoints of bilateral standing phases in non-cyclic activities.
Cyclic clips span whole strides and have median lengths of 9.7--10\,s, against 2.1, 3.4 and 4.1\,s for lateral, flexion and jump clips.

\vspace{-0.35em}%
\subsubsection{Subject Characteristics}
\label{App:subj_characteristics}
\vspace{-0.25em}

The 129 participants of Table~\ref{tab:participants}, 45 female and 84 male, weigh $69.5 \pm 12.5$\,kg (44--126\,kg) and measure $1.71 \pm 0.10$\,m (1.46--1.95\,m).
Age, reported for 102 participants, is $34 \pm 18$ years (19--84) and unreported for \citet{fregly2012grand}, \citet{hamner2013muscle} and \citet{scherpereel2023human}.

\begin{table}[h]
  \centering
  \caption{\textbf{Subject characteristics.} Activities: \texttt{\textbf{\actl{walk}{W}}}alk, \texttt{\textbf{\actl{run}{R}}}un, \texttt{\textbf{\actl{stand}{S}}}tand, \texttt{\textbf{\actl{flexion}{F}}}lexion, \texttt{\textbf{\actl{lateral}{L}}}ateral, \texttt{\textbf{\actl{jump}{J}}}ump, each with (\#trials,\,\#clips). A trial yields one or more clips and counts toward each of its activities. Mass in kg, Ht.~=~height in m. H-O = held-out clips (seen split). \textbf{Bold} = held-out participants (unseen split).}
  \label{tab:participants}
  \vspace{-0.4em}
  \scriptsize
  \setlength{\tabcolsep}{1.5pt}
  \renewcommand{\arraystretch}{0.85}
  \begin{tabular*}{\dimexpr0.5\textwidth-3pt\relax}[t]{@{}lcccc l@{\extracolsep{\fill}}c@{}}
  \toprule
  \textbf{Subj.} & \textbf{Sex} & \textbf{Age} & \textbf{Mass} & \textbf{Ht.}\ & \texttt{\textbf{Activities}} & \textbf{H-O} \\
  \midrule
  \rowcolor{papericeblue} \multicolumn{7}{@{}>{\fillL}l}{\emph{\citet{dos2017data}}} \\
  s0 & M & 25 & 74.3 & 1.72 & \texttt{\textbf{\actl{stand}{S}}}(6,36) & 7 \\
  s1 & M & 26 & 126.3 & 1.92 & \texttt{\textbf{\actl{stand}{S}}}(6,36) & 7 \\
  s2 & F & 22 & 52.9 & 1.67 & \texttt{\textbf{\actl{stand}{S}}}(6,36) & 7 \\
  s3 & M & 33 & 75.8 & 1.79 & \texttt{\textbf{\actl{stand}{S}}}(6,36) & 7 \\
  s4 & M & 24 & 61.0 & 1.84 & \texttt{\textbf{\actl{stand}{S}}}(6,36) & 7 \\
  \textbf{s5} & \textbf{M} & \textbf{25} & \textbf{83.2} & \textbf{1.74} & \textbf{\texttt{\textbf{\actl{stand}{S}}}(6,36)} & \textbf{7} \\
  \textbf{s6} & \textbf{F} & \textbf{24} & \textbf{71.8} & \textbf{1.58} & \textbf{\texttt{\textbf{\actl{stand}{S}}}(6,36)} & \textbf{7} \\
  s7 & M & 36 & 64.0 & 1.82 & \texttt{\textbf{\actl{stand}{S}}}(6,36) & 7 \\
  \textbf{s8} & \textbf{F} & \textbf{25} & \textbf{61.2} & \textbf{1.69} & \textbf{\texttt{\textbf{\actl{stand}{S}}}(6,36)} & \textbf{7} \\
  s9 & F & 31 & 61.7 & 1.62 & \texttt{\textbf{\actl{stand}{S}}}(6,36) & 7 \\
  s10 & M & 32 & 77.6 & 1.92 & \texttt{\textbf{\actl{stand}{S}}}(6,36) & 7 \\
  s11 & F & 24 & 48.8 & 1.58 & \texttt{\textbf{\actl{stand}{S}}}(6,36) & 7 \\
  s12 & M & 70 & 65.3 & 1.75 & \texttt{\textbf{\actl{stand}{S}}}(6,36) & 7 \\
  s13 & M & 30 & 95.4 & 1.71 & \texttt{\textbf{\actl{stand}{S}}}(6,36) & 7 \\
  \textbf{s14} & \textbf{F} & \textbf{31} & \textbf{64.9} & \textbf{1.53} & \textbf{\texttt{\textbf{\actl{stand}{S}}}(6,36)} & \textbf{7} \\
  \textbf{s15} & \textbf{M} & \textbf{23} & \textbf{89.3} & \textbf{1.81} & \textbf{\texttt{\textbf{\actl{stand}{S}}}(6,36)} & \textbf{7} \\
  s16 & M & 31 & 77.9 & 1.73 & \texttt{\textbf{\actl{stand}{S}}}(6,36) & 7 \\
  \textbf{s17} & \textbf{F} & \textbf{61} & \textbf{70.6} & \textbf{1.50} & \textbf{\texttt{\textbf{\actl{stand}{S}}}(6,36)} & \textbf{7} \\
  s18 & M & 63 & 72.2 & 1.72 & \texttt{\textbf{\actl{stand}{S}}}(6,36) & 7 \\
  s19 & M & 28 & 79.1 & 1.86 & \texttt{\textbf{\actl{stand}{S}}}(6,36) & 7 \\
  s20 & F & 28 & 59.3 & 1.68 & \texttt{\textbf{\actl{stand}{S}}}(6,36) & 7 \\
  s21 & M & 62 & 70.5 & 1.64 & \texttt{\textbf{\actl{stand}{S}}}(6,36) & 7 \\
  s22 & M & 29 & 68.2 & 1.76 & \texttt{\textbf{\actl{stand}{S}}}(6,36) & 7 \\
  s23 & M & 68 & 70.2 & 1.67 & \texttt{\textbf{\actl{stand}{S}}}(6,36) & 7 \\
  s24 & F & 21 & 62.5 & 1.70 & \texttt{\textbf{\actl{stand}{S}}}(6,36) & 7 \\
  s25 & M & 22 & 61.5 & 1.81 & \texttt{\textbf{\actl{stand}{S}}}(6,36) & 7 \\
  s26 & F & 25 & 61.2 & 1.70 & \texttt{\textbf{\actl{stand}{S}}}(6,36) & 7 \\
  s27 & M & 62 & 73.1 & 1.74 & \texttt{\textbf{\actl{stand}{S}}}(6,36) & 7 \\
  s28 & M & 84 & 66.3 & 1.56 & \texttt{\textbf{\actl{stand}{S}}}(6,36) & 7 \\
  s29 & F & 68 & 49.2 & 1.47 & \texttt{\textbf{\actl{stand}{S}}}(6,36) & 7 \\
  s30 & M & 61 & 74.2 & 1.62 & \texttt{\textbf{\actl{stand}{S}}}(6,36) & 7 \\
  s31 & F & 72 & 65.7 & 1.46 & \texttt{\textbf{\actl{stand}{S}}}(6,36) & 7 \\
  s32 & F & 72 & 62.0 & 1.54 & \texttt{\textbf{\actl{stand}{S}}}(6,36) & 7 \\
  s33 & F & 63 & 59.8 & 1.61 & \texttt{\textbf{\actl{stand}{S}}}(6,36) & 7 \\
  \textbf{s34} & \textbf{F} & \textbf{67} & \textbf{66.3} & \textbf{1.60} & \textbf{\texttt{\textbf{\actl{stand}{S}}}(6,36)} & \textbf{7} \\
  s35 & M & 69 & 73.2 & 1.60 & \texttt{\textbf{\actl{stand}{S}}}(6,36) & 7 \\
  s36 & F & 80 & 52.7 & 1.51 & \texttt{\textbf{\actl{stand}{S}}}(6,36) & 7 \\
  \textbf{s37} & \textbf{M} & \textbf{72} & \textbf{92.9} & \textbf{1.72} & \textbf{\texttt{\textbf{\actl{stand}{S}}}(6,36)} & \textbf{7} \\
  s38 & F & 69 & 66.1 & 1.60 & \texttt{\textbf{\actl{stand}{S}}}(6,36) & 7 \\
  s39 & F & 28 & 44.3 & 1.54 & \texttt{\textbf{\actl{stand}{S}}}(6,36) & 7 \\
  s40 & F & 68 & 65.3 & 1.51 & \texttt{\textbf{\actl{stand}{S}}}(6,36) & 7 \\
  s41 & F & 63 & 55.6 & 1.61 & \texttt{\textbf{\actl{stand}{S}}}(6,36) & 7 \\
  s42 & M & 63 & 96.0 & 1.78 & \texttt{\textbf{\actl{stand}{S}}}(6,36) & 7 \\
  \textbf{s43} & \textbf{M} & \textbf{61} & \textbf{82.2} & \textbf{1.70} & \textbf{\texttt{\textbf{\actl{stand}{S}}}(6,36)} & \textbf{7} \\
  s44 & F & 24 & 51.7 & 1.61 & \texttt{\textbf{\actl{stand}{S}}}(6,36) & 7 \\
  s45 & M & 27 & 101.8 & 1.72 & \texttt{\textbf{\actl{stand}{S}}}(6,36) & 7 \\
  s46 & F & 34 & 53.5 & 1.56 & \texttt{\textbf{\actl{stand}{S}}}(6,36) & 7 \\
  s47 & M & 37 & 69.6 & 1.55 & \texttt{\textbf{\actl{stand}{S}}}(6,36) & 7 \\
  s48 & F & 64 & 60.8 & 1.58 & \texttt{\textbf{\actl{stand}{S}}}(6,36) & 7 \\
  \addlinespace[1pt]

  \rowcolor{papericeblue} \multicolumn{7}{@{}>{\fillL}l}{\emph{\citet{fregly2012grand}}} \\
  1GC & M &  & 64.6 & 1.66 & \texttt{\textbf{\actl{walk}{W}}}(14,14) & 3 \\
  2GC & M &  & 67.0 & 1.72 & \texttt{\textbf{\actl{walk}{W}}}(15,15) & 3 \\
  3GC & F &  & 78.4 & 1.67 & \texttt{\textbf{\actl{walk}{W}}}(31,31) & 6 \\
  4GC & M &  & 66.7 & 1.68 & \texttt{\textbf{\actl{walk}{W}}}(31,31) & 6 \\
  5GC & M &  & 75.0 & 1.80 & \texttt{\textbf{\actl{walk}{W}}}(1,3) & 1 \\
  \textbf{6GC} & \textbf{M} &  & \textbf{70.0} & \textbf{1.72} & \textbf{\texttt{\textbf{\actl{walk}{W}}}(21,21)} & \textbf{4} \\
  \addlinespace[1pt]
  \rowcolor{papericeblue} \multicolumn{7}{@{}>{\fillL}l}{\emph{\citet{hamner2013muscle}}} \\
  \textbf{s01} & \textbf{M} &  & \textbf{72.8} & \textbf{1.77} & \textbf{\texttt{\textbf{\actl{run}{R}}}(4,4)} & \textbf{1} \\
  \textbf{s02} & \textbf{M} &  & \textbf{76.5} & \textbf{1.82} & \textbf{\texttt{\textbf{\actl{run}{R}}}(3,3)} & \textbf{1} \\
  s03 & M &  & 64.0 & 1.78 & \texttt{\textbf{\actl{run}{R}}}(4,4) & 1 \\
  s04 & M &  & 80.5 & 1.77 & \texttt{\textbf{\actl{run}{R}}}(4,4) & 1 \\
  s08 & M &  & 82.4 & 1.79 & \texttt{\textbf{\actl{run}{R}}}(4,4) & 1 \\
  s10 & M &  & 69.3 & 1.70 & \texttt{\textbf{\actl{run}{R}}}(3,3) & 1 \\
  s11 & M &  & 69.3 & 1.74 & \texttt{\textbf{\actl{run}{R}}}(4,4) & 1 \\
  s17 & M &  & 68.4 & 1.68 & \texttt{\textbf{\actl{run}{R}}}(3,3) & 1 \\
  s19 & M &  & 65.1 & 1.72 & \texttt{\textbf{\actl{run}{R}}}(4,4) & 1 \\
  s20 & M &  & 67.2 & 1.67 & \texttt{\textbf{\actl{run}{R}}}(3,3) & 1 \\

  \bottomrule
    \end{tabular*}\hfill
  \begin{tabular*}{\dimexpr0.5\textwidth-3pt\relax}[t]{@{}lcccc l@{\extracolsep{\fill}}c@{}}
  \toprule
  \textbf{Subj.} & \textbf{Sex} & \textbf{Age} & \textbf{Mass} & \textbf{Ht.}\ & \texttt{\textbf{Activities}} & \textbf{H-O} \\
  \midrule
  \rowcolor{papericeblue} \multicolumn{7}{@{}>{\fillL}l}{\emph{\citet{han2023groundlink}}} \\
  s001 & F & 22 & 69.9 & 1.68 & \texttt{\textbf{\actl{walk}{W}}}(3,3),\texttt{\textbf{\actl{stand}{S}}}(11,14),\texttt{\textbf{\actl{flexion}{F}}}(8,13),\texttt{\textbf{\actl{lateral}{L}}}(3,3) & 8 \\
  s002 & M & 21 & 66.7 & 1.70 & \texttt{\textbf{\actl{walk}{W}}}(2,2),\texttt{\textbf{\actl{stand}{S}}}(8,9),\texttt{\textbf{\actl{flexion}{F}}}(9,11) & 4 \\
  s003 & F & 23 & 53.1 & 1.59 & \texttt{\textbf{\actl{walk}{W}}}(3,3),\texttt{\textbf{\actl{flexion}{F}}}(10,14),\texttt{\textbf{\actl{lateral}{L}}}(2,2) & 4 \\
  s004 & M & 20 & 71.7 & 1.82 & \texttt{\textbf{\actl{walk}{W}}}(2,2),\texttt{\textbf{\actl{stand}{S}}}(4,4),\texttt{\textbf{\actl{flexion}{F}}}(6,10),\texttt{\textbf{\actl{lateral}{L}}}(1,1) & 3 \\
  s005 & M & 24 & 90.7 & 1.95 & \texttt{\textbf{\actl{walk}{W}}}(3,3),\texttt{\textbf{\actl{stand}{S}}}(5,8),\texttt{\textbf{\actl{flexion}{F}}}(8,11),\texttt{\textbf{\actl{lateral}{L}}}(1,1) & 5 \\
  \textbf{s006} & \textbf{F} & \textbf{35} & \textbf{49.0} & \textbf{1.53} & \textbf{\texttt{\textbf{\actl{walk}{W}}}(3,3),\texttt{\textbf{\actl{stand}{S}}}(4,4),\texttt{\textbf{\actl{flexion}{F}}}(9,11),\texttt{\textbf{\actl{lateral}{L}}}(3,3),\texttt{\textbf{\actl{jump}{J}}}(2,2)} & \textbf{5} \\
  s007 & F & 22 & 64.0 & 1.75 & \texttt{\textbf{\actl{stand}{S}}}(4,4),\texttt{\textbf{\actl{flexion}{F}}}(3,3) & 2 \\
  \addlinespace[1pt]
  \rowcolor{papericeblue} \multicolumn{7}{@{}>{\fillL}l}{\emph{\citet{moore2015elaborate}}} \\
  \textbf{s3} & \textbf{F} & \textbf{32} & \textbf{60.0} & \textbf{1.62} & \textbf{\texttt{\textbf{\actl{walk}{W}}}(3,129),\texttt{\textbf{\actl{stand}{S}}}(3,6)} & \textbf{27} \\
  s5 & M & 23 & 65.0 & 1.73 & \texttt{\textbf{\actl{walk}{W}}}(3,186),\texttt{\textbf{\actl{stand}{S}}}(3,6) & 38 \\
  s6 & M & 26 & 80.0 & 1.77 & \texttt{\textbf{\actl{walk}{W}}}(2,122),\texttt{\textbf{\actl{stand}{S}}}(2,4) & 25 \\
  s7 & F & 29 & 63.0 & 1.72 & \texttt{\textbf{\actl{walk}{W}}}(3,185),\texttt{\textbf{\actl{stand}{S}}}(3,6) & 38 \\
  s8 & M & 20 & 70.0 & 1.57 & \texttt{\textbf{\actl{walk}{W}}}(3,134),\texttt{\textbf{\actl{stand}{S}}}(3,6) & 28 \\
  s9 & M & 20 & 64.0 & 1.69 & \texttt{\textbf{\actl{walk}{W}}}(3,138),\texttt{\textbf{\actl{stand}{S}}}(3,6) & 29 \\
  \textbf{s10} & \textbf{M} & \textbf{19} & \textbf{90.7} & \textbf{1.77} & \textbf{\texttt{\textbf{\actl{walk}{W}}}(3,186),\texttt{\textbf{\actl{stand}{S}}}(3,6)} & \textbf{38} \\
  s12 & M & 22 & 81.0 & 1.85 & \texttt{\textbf{\actl{walk}{W}}}(3,179),\texttt{\textbf{\actl{stand}{S}}}(3,6) & 37 \\
  s13 & F & 21 & 64.0 & 1.70 & \texttt{\textbf{\actl{walk}{W}}}(3,186),\texttt{\textbf{\actl{stand}{S}}}(3,6) & 38 \\
  s15 & M & 22 & 79.4 & 1.83 & \texttt{\textbf{\actl{walk}{W}}}(3,183),\texttt{\textbf{\actl{stand}{S}}}(2,4) & 38 \\
  s16 & F & 28 & 52.2 & 1.69 & \texttt{\textbf{\actl{walk}{W}}}(3,180),\texttt{\textbf{\actl{stand}{S}}}(3,6) & 37 \\
  s17 & M & 23 & 87.0 & 1.86 & \texttt{\textbf{\actl{walk}{W}}}(3,171),\texttt{\textbf{\actl{stand}{S}}}(3,6) & 35 \\
  \addlinespace[1pt]
  \rowcolor{papericeblue} \multicolumn{7}{@{}>{\fillL}l}{\emph{\citet{scherpereel2023human}}} \\
  \textbf{AB01} & \textbf{M} &  & \textbf{79.1} & \textbf{1.75} & \textbf{\texttt{\textbf{\actl{stand}{S}}}(1,11),\texttt{\textbf{\actl{flexion}{F}}}(2,20),\texttt{\textbf{\actl{lateral}{L}}}(4,38),\texttt{\textbf{\actl{jump}{J}}}(2,15)} & \textbf{17} \\
  AB03 & M &  & 113.9 & 1.79 & \texttt{\textbf{\actl{stand}{S}}}(1,11),\texttt{\textbf{\actl{flexion}{F}}}(2,18),\texttt{\textbf{\actl{lateral}{L}}}(5,36),\texttt{\textbf{\actl{jump}{J}}}(2,11) & 14 \\
  AB05 & F &  & 71.8 & 1.69 & \texttt{\textbf{\actl{stand}{S}}}(1,6),\texttt{\textbf{\actl{flexion}{F}}}(2,20),\texttt{\textbf{\actl{lateral}{L}}}(5,43),\texttt{\textbf{\actl{jump}{J}}}(2,10) & 18 \\
  AB06 & M &  & 80.7 & 1.83 & \texttt{\textbf{\actl{stand}{S}}}(1,8),\texttt{\textbf{\actl{flexion}{F}}}(2,26),\texttt{\textbf{\actl{lateral}{L}}}(5,44),\texttt{\textbf{\actl{jump}{J}}}(2,18) & 19 \\
  AB07 & F &  & 62.0 & 1.65 & \texttt{\textbf{\actl{stand}{S}}}(1,13),\texttt{\textbf{\actl{flexion}{F}}}(2,24),\texttt{\textbf{\actl{lateral}{L}}}(5,42),\texttt{\textbf{\actl{jump}{J}}}(3,14) & 17 \\
  AB08 & M &  & 88.1 & 1.71 & \texttt{\textbf{\actl{stand}{S}}}(1,9),\texttt{\textbf{\actl{flexion}{F}}}(2,28),\texttt{\textbf{\actl{lateral}{L}}}(5,41),\texttt{\textbf{\actl{jump}{J}}}(3,13) & 19 \\
  \textbf{AB09} & \textbf{M} &  & \textbf{83.9} & \textbf{1.77} & \textbf{\texttt{\textbf{\actl{stand}{S}}}(1,9),\texttt{\textbf{\actl{flexion}{F}}}(2,18),\texttt{\textbf{\actl{lateral}{L}}}(5,39),\texttt{\textbf{\actl{jump}{J}}}(3,13)} & \textbf{18} \\
  AB10 & M &  & 67.5 & 1.73 & \texttt{\textbf{\actl{stand}{S}}}(1,10),\texttt{\textbf{\actl{flexion}{F}}}(2,25),\texttt{\textbf{\actl{lateral}{L}}}(5,41),\texttt{\textbf{\actl{jump}{J}}}(3,10) & 19 \\
  AB11 & F &  & 64.2 & 1.57 & \texttt{\textbf{\actl{stand}{S}}}(1,8),\texttt{\textbf{\actl{flexion}{F}}}(2,36),\texttt{\textbf{\actl{lateral}{L}}}(5,39),\texttt{\textbf{\actl{jump}{J}}}(3,18) & 23 \\
  AB12 & F &  & 63.3 & 1.65 & \texttt{\textbf{\actl{stand}{S}}}(1,9),\texttt{\textbf{\actl{flexion}{F}}}(2,31),\texttt{\textbf{\actl{lateral}{L}}}(5,48),\texttt{\textbf{\actl{jump}{J}}}(3,21) & 22 \\
  AB13 & F &  & 67.7 & 1.65 & \texttt{\textbf{\actl{stand}{S}}}(1,9),\texttt{\textbf{\actl{flexion}{F}}}(2,37),\texttt{\textbf{\actl{lateral}{L}}}(5,41),\texttt{\textbf{\actl{jump}{J}}}(3,20) & 24 \\
  \addlinespace[1pt]
  \rowcolor{papericeblue} \multicolumn{7}{@{}>{\fillL}l}{\emph{\citet{tan2022strike}}} \\
  \textbf{s1} & \textbf{F} & \textbf{25} & \textbf{56.1} & \textbf{1.58} & \textbf{\texttt{\textbf{\actl{run}{R}}}(11,109)} & \textbf{22} \\
  s3 & F & 24 & 57.4 & 1.69 & \texttt{\textbf{\actl{run}{R}}}(5,48) & 10 \\
  s4 & M & 25 & 69.0 & 1.78 & \texttt{\textbf{\actl{run}{R}}}(7,82) & 16 \\
  s5 & F & 22 & 52.1 & 1.58 & \texttt{\textbf{\actl{run}{R}}}(12,122) & 24 \\
  s6 & M & 24 & 68.0 & 1.75 & \texttt{\textbf{\actl{run}{R}}}(11,114) & 23 \\
  s7 & M & 23 & 72.3 & 1.78 & \texttt{\textbf{\actl{run}{R}}}(11,113) & 23 \\
  s8 & F & 24 & 53.0 & 1.59 & \texttt{\textbf{\actl{run}{R}}}(10,104) & 21 \\
  s9 & M & 23 & 66.0 & 1.75 & \texttt{\textbf{\actl{run}{R}}}(12,126) & 25 \\
  \addlinespace[1pt]
  \rowcolor{papericeblue} \multicolumn{7}{@{}>{\fillL}l}{\emph{\citet{tan2022imu}}} \\
  02 & M & 25 & 71.4 & 1.83 & \texttt{\textbf{\actl{walk}{W}}}(4,97) & 19 \\
  04 & M & 23 & 57.0 & 1.62 & \texttt{\textbf{\actl{walk}{W}}}(4,97) & 19 \\
  \textbf{05} & \textbf{M} & \textbf{23} & \textbf{64.0} & \textbf{1.68} & \textbf{\texttt{\textbf{\actl{walk}{W}}}(4,97)} & \textbf{19} \\
  \textbf{06} & \textbf{M} & \textbf{24} & \textbf{67.2} & \textbf{1.75} & \textbf{\texttt{\textbf{\actl{walk}{W}}}(4,97)} & \textbf{19} \\
  07 & M & 22 & 74.4 & 1.71 & \texttt{\textbf{\actl{walk}{W}}}(4,97) & 19 \\
  08 & M & 22 & 59.0 & 1.79 & \texttt{\textbf{\actl{walk}{W}}}(4,97) & 19 \\
  \textbf{09} & \textbf{M} & \textbf{24} & \textbf{70.7} & \textbf{1.77} & \textbf{\texttt{\textbf{\actl{walk}{W}}}(4,96)} & \textbf{19} \\
  10 & M & 22 & 58.1 & 1.76 & \texttt{\textbf{\actl{walk}{W}}}(4,97) & 19 \\
  11 & M & 23 & 66.9 & 1.80 & \texttt{\textbf{\actl{walk}{W}}}(4,97) & 19 \\
  12 & M & 24 & 63.8 & 1.73 & \texttt{\textbf{\actl{walk}{W}}}(4,97) & 19 \\
  13 & M & 22 & 73.2 & 1.76 & \texttt{\textbf{\actl{walk}{W}}}(4,97) & 19 \\
  14 & M & 22 & 61.8 & 1.81 & \texttt{\textbf{\actl{walk}{W}}}(4,97) & 19 \\
  15 & M & 24 & 64.6 & 1.76 & \texttt{\textbf{\actl{walk}{W}}}(4,97) & 19 \\
  17 & M & 26 & 74.8 & 1.87 & \texttt{\textbf{\actl{walk}{W}}}(4,97) & 19 \\
  18 & M & 23 & 68.9 & 1.82 & \texttt{\textbf{\actl{walk}{W}}}(4,97) & 19 \\
  19 & M & 23 & 91.4 & 1.85 & \texttt{\textbf{\actl{walk}{W}}}(4,97) & 19 \\
  20 & M & 22 & 57.4 & 1.70 & \texttt{\textbf{\actl{walk}{W}}}(4,97) & 19 \\
  \addlinespace[1pt]
  \rowcolor{papericeblue} \multicolumn{7}{@{}>{\fillL}l}{\emph{\citet{wang2023wearable}}} \\
  s04 & M & 23 & 75.2 & 1.80 & \texttt{\textbf{\actl{walk}{W}}}(6,42),\texttt{\textbf{\actl{run}{R}}}(3,21),\texttt{\textbf{\actl{flexion}{F}}}(2,21),\texttt{\textbf{\actl{jump}{J}}}(2,18) & 20 \\
  s05 & F & 23 & 65.1 & 1.71 & \texttt{\textbf{\actl{walk}{W}}}(6,44),\texttt{\textbf{\actl{run}{R}}}(3,19),\texttt{\textbf{\actl{flexion}{F}}}(2,28),\texttt{\textbf{\actl{jump}{J}}}(2,19) & 23 \\
  s06 & F & 23 & 56.0 & 1.68 & \texttt{\textbf{\actl{walk}{W}}}(6,42),\texttt{\textbf{\actl{run}{R}}}(3,14),\texttt{\textbf{\actl{flexion}{F}}}(2,34),\texttt{\textbf{\actl{jump}{J}}}(2,21) & 22 \\
  s07 & M & 23 & 80.8 & 1.82 & \texttt{\textbf{\actl{walk}{W}}}(6,40),\texttt{\textbf{\actl{run}{R}}}(3,19),\texttt{\textbf{\actl{flexion}{F}}}(2,18),\texttt{\textbf{\actl{jump}{J}}}(2,21) & 20 \\
  s08 & F & 28 & 61.5 & 1.62 & \texttt{\textbf{\actl{walk}{W}}}(6,40),\texttt{\textbf{\actl{run}{R}}}(3,20),\texttt{\textbf{\actl{flexion}{F}}}(2,30),\texttt{\textbf{\actl{jump}{J}}}(2,19) & 22 \\
  s09 & M & 26 & 81.0 & 1.82 & \texttt{\textbf{\actl{walk}{W}}}(6,40),\texttt{\textbf{\actl{run}{R}}}(3,18),\texttt{\textbf{\actl{flexion}{F}}}(2,26),\texttt{\textbf{\actl{jump}{J}}}(2,17) & 20 \\
  s10 & M & 28 & 71.6 & 1.66 & \texttt{\textbf{\actl{walk}{W}}}(6,41),\texttt{\textbf{\actl{run}{R}}}(3,18),\texttt{\textbf{\actl{flexion}{F}}}(2,22),\texttt{\textbf{\actl{jump}{J}}}(2,19) & 20 \\
  s11 & F & 25 & 62.4 & 1.68 & \texttt{\textbf{\actl{walk}{W}}}(4,26),\texttt{\textbf{\actl{run}{R}}}(3,18),\texttt{\textbf{\actl{flexion}{F}}}(2,24),\texttt{\textbf{\actl{jump}{J}}}(2,20) & 18 \\
  \textbf{s12} & \textbf{M} & \textbf{30} & \textbf{69.0} & \textbf{1.79} & \textbf{\texttt{\textbf{\actl{walk}{W}}}(6,41),\texttt{\textbf{\actl{run}{R}}}(2,12),\texttt{\textbf{\actl{flexion}{F}}}(2,30),\texttt{\textbf{\actl{jump}{J}}}(2,18)} & \textbf{20} \\
  \bottomrule
  \end{tabular*}
\end{table}

\subsubsection{Data Splits and Evaluation Protocol}
\label{App:splits}

\paragraph{Splits.}
All splits are drawn once with seed 42 and shared by every experiment (Tables~\ref{tab:participants} and~\ref{tab:splits}).
\emph{Seen subjects, held-out clips} withholds $\mathrm{round}(0.2\,n_{\mathrm{clips}})$ randomly sampled clips of each participant per activity and trains on the remaining 6,796 clips of all 129 participants.
\emph{Unseen subjects, zero-shot} withholds $\lfloor 0.2 N_d \rfloor$ randomly sampled participants of each dataset $d$ and trains on the 7,128 clips of the remaining 107.
Since every dataset has $N_d \ge 6$, each withholds at least one participant, giving 22 unseen participants in total and 3--14 per activity.
\emph{Leave-one-activity-out} (LOAO) trains on five activities and evaluates zero-shot on the sixth.
Datasets that record only the withheld activity are excluded from its fold, since their participants would also be unseen, confounding transfer to a new activity with transfer to a new body.
This removes \citet{dos2017data} from the standing fold, \citet{fregly2012grand} and \citet{tan2022imu} from the walking fold, and \citet{hamner2013muscle} and \citet{tan2022strike} from the running fold, leaving 4,112 of the 8,493 clips for evaluation.

\begin{table}[h]
\centering
\caption{\textbf{Training and evaluation data per split.} \emph{Seen}~= Seen subjects, held-out clips. \mbox{\emph{Unseen}~=} Unseen subjects, zero-shot. \emph{LOAO}~= Withheld activity, with the participants available for training and the clips evaluated in each fold, excluding datasets that record only the withheld activity.}
\vspace{-0.5em}
\label{tab:splits}
\small
\begin{tabular*}{\linewidth}{@{\extracolsep{\fill}}lrrrrrr@{}}
\toprule
& \multicolumn{2}{c}{\textbf{\emph{Seen}}} & \multicolumn{2}{c}{\textbf{\emph{Unseen}}} & \multicolumn{2}{c}{\textbf{\emph{LOAO}}} \\
\cmidrule(lr){2-3}\cmidrule(lr){4-5}\cmidrule(l){6-7}
\texttt{\textbf{Activity}} & Train clips & Eval clips & Train clips & Eval clips & Train subj. & Eval clips \\
\midrule
\act{walk}    & 3,298 & 816 & 3,444 & 670 & 106 & 2,351 \\
\act{run}     &   806 & 207 &   885 & 128 & 111 &   159 \\
\act{stand}   & 1,591 & 387 & 1,618 & 360 &  80 &   214 \\
\act{lateral} &   369 &  93 &   382 &  80 & 129 &   462 \\
\act{flexion} &   464 & 125 &   510 &  79 & 129 &   589 \\
\act{jump}    &   268 &  69 &   289 &  48 & 129 &   337 \\
\midrule
\textbf{Total} & 6,796 & 1,697 & 7,128 & 1,365 & & 4,112 \\
\bottomrule
\end{tabular*}
\vspace{-0.5em}
\end{table}

\paragraph{Evaluation protocol.}
Each evaluation clip is split into valid segments of at least 25 frames, each rolled out as one episode from its first reference frame with the policy's mean action until the segment ends or terminates (pelvis below 0.4\,m or numerically invalid state).
Survival is the percentage of a clip's reference frames completed before termination.
Fidelity metrics exclude the first 0.1\,s, since the reset places the model in the reference pose without settling.
Rollouts that diverge (peak vertical GRF above 50\,BW or joint-velocity RMSE above $10^4\,^\circ$/s) keep their survival but are excluded from the other metrics, which affects six rollouts of the main evaluations, all of Torque or PHC on a withheld activity, and three Torque rollouts of the robustness tests.
Metrics are computed per clip (Tables~\ref{tab:golden_full} and~\ref{tab:loao_full}) and averaged within each activity, and the overall score is the mean of the six activity means.
All scores are reported as mean and standard deviation over five training seeds.

\subsection{Method}
\label{App:method}

\subsubsection{Action-Space Impedance Bounds}
\label{App:action_space_bounds}

\paragraph{Proof of Proposition~\ref{prop:stability}.}
\label{App:single_joint_theory}
\begin{proof}
Between policy queries the command $q^* + \alpha\,\delta q$ and gains $(k^p, k^d)$ are held fixed. Define $e = q - (q^* + \alpha\,\delta q)$ and $v = \dot q - \dot q^*$. The PD torque (Eq.~\ref{eq:pd_torque}) reduces to $\hat\tau = -k^p e - k^d v$. On a free joint of inertia $I$, Newton's law gives $I\dot v = -k^p e - k^d v$. Discretizing with semi-implicit Euler ($v_+ = v + h\dot v$, $e_+ = e + h v_+$):
\begin{equation}
\begin{pmatrix} e_+ \\ v_+ \end{pmatrix}
= A \begin{pmatrix} e \\ v \end{pmatrix},
\qquad
A = \begin{pmatrix} 1 - h^2 k^p / I & h(1 - h k^d / I) \\ -h k^p / I & 1 - h k^d / I \end{pmatrix}.
\label{eq:error_map}
\end{equation}
Asymptotic stability requires all eigenvalues of $A$ inside the unit circle. For a real $2\times 2$ matrix the Jury criterion \citep{aastrom2013computer} gives necessary and sufficient conditions: $|\det A| < 1$ and $|\operatorname{tr} A| < 1 + \det A$. Computing:
\begin{equation}
\det A = 1 - \frac{h k^d}{I}, \qquad \operatorname{tr} A = 2 - \frac{h k^d}{I} - \frac{h^2 k^p}{I}.
\end{equation}
The condition $|\det A| < 1$ gives $0 < k^d < 2I/h$. Substituting into $|\operatorname{tr} A| < 1 + \det A$ and simplifying gives $0 < k^p < 4I/h^2 - 2k^d/h$.
\end{proof}

\paragraph{Single-joint model.}
The model is a uniform rod on a hinge whose tip presses on a compliant floor, so that one joint alternates between a contact phase, which demands force, and a free swing within each cycle (Table~\ref{tab:toy_setup}).
The gain priors and both bounds derive from the rod's inertia and geometry, and all results are averaged over three seeds.

\begin{table}[h]
\centering
\caption{\textbf{Single-joint model, controller and protocol.}}
\vspace{-0.5em}
\label{tab:toy_setup}
\small
\settowidth{\toycolA}{Randomization}%
\begin{tabular}{@{}l>{\raggedright\arraybackslash}p{\dimexpr\linewidth-\toycolA-2\tabcolsep\relax}@{}}%
\toprule
Rod & $L = 0.20$\,m, $m = 1.5$\,kg, $I_{jj} = mL^2/3 = 0.020$\,kg\,m$^2$, hinge height $z = 60$\,mm\\
Contact & sphere $r = 20$\,mm, contact at $q_c = 11.5^\circ$, solref $= (0.01, 1)$ \\
Rates & physics $h = 2$\,ms, policy $100$\,Hz ($N = 5$ substeps) \\
Actuator & $|\hat\tau| \le \tau^{\mathrm{lim}} = 40$\,Nm \\
Gain priors & $k^p_0 = I_{jj}\omega_n^2 = 32$\,Nm/rad ($\omega_n = 40$\,rad/s), $k^d_0 = I_{jj}/h = 10$\,Nm\,s/rad $= \tfrac12 \cdot 2I_{jj}/h$ \\
Learned-PD & $k = k_0\,e^{\overline{\Delta k}}$, $\overline{\Delta k} \in [-1, 1]$ (log-scale $s = 1$), $\alpha = 0.5$\,rad \\
Reference & $1$\,s cycle: a $\sin^2$ contact-force bell $F^*$ peaking at $mg = 14.7$\,N ($6mg$ for the stiffness sweep), with $q^*$ the angle producing it on the floor, then a $\sin^2$ swing $20^\circ$ off the floor \\
Reward & $r^{\mathrm{kin}} + 0.25\,r^{\mathrm{force}}$, with $r^{\mathrm{kin}} = \exp(-2\,q_e^2) + \exp(-0.1\,\dot q_e^2)$ and $r^{\mathrm{force}} = \exp\bigl(-2\,\1[F^* > 0.1\,F^*_{\max}]\,F_e^2\bigr)$, $F_e = (F - F^*)/F^*_{\max}$ \\
Randomization & per episode, unobserved: floor time constant $\times\,e^{\mathcal U(-\ln 1.4,\, \ln 1.4)}$, rod inertia $\times\,\mathcal U(0.85, 1.15)$, reference-state initialization, termination at $|q - q^*| > 20^\circ$ \\
Gain sweeps & $k^p_0 \times \{0.25, \ldots, 4\}$ at $6mg$, or $k^d_0 = \{0.25, \ldots, 2\} \times 2I_{jj}/h$ at $mg$ \\
PPO & $128$ envs $\times$ $32$ steps $\times$ $2000$ updates ($8.2$\,M steps per policy), network $(64, 64)$ \\
Evaluation & angle error $\bar e_q = \mathrm{mean}_t|q - q^*|$, force error $e_F = \mathrm{RMS}_{\mathrm{contact}}(F - F^*)$ \\
\bottomrule
\end{tabular}
\end{table}

\paragraph{Bound values for this model.}
Bound~1 is the net joint torque demanded by the contact, divided by the offset authority $\alpha$.
With the contact force acting at the rod tip and the rod's weight at its center,
\begin{equation}
  \frac{\max_t|\tau^*(t)|}{\alpha}
  = \frac{L\cos q_c}{\alpha}\Bigl(F^*_{\max} - \tfrac12 mg\Bigr),
  \label{eq:bound1_rod}
\end{equation}
which evaluates to $2.9$\,Nm/rad at the nominal reference ($F^*_{\max} = mg$), where lifting the rod in swing binds instead ($4.5$\,Nm/rad), and to $31.7$\,Nm/rad at the sixfold contact force ($F^*_{\max} = 6mg$).
Bound~2 is $2I_{jj}/h = 20$\,Nm\,s/rad, twice the prior $k^d_0 = 10$\,Nm\,s/rad.
With $s = 1$, the exponential parameterization of Eq.~\ref{eq:learned_impedance} lets Learned-PD search $k^p \in [k^p_0 e^{-s},\, k^p_0 e^{s}] = [11.8,\, 87.0]$\,Nm/rad and $k^d \in [3.7,\, 27.2]$\,Nm\,s/rad.
Its damping ceiling therefore lies above Bound~2, so the policy must learn to remain stable in swing, as the parameterization does not enforce it.

\paragraph{Operating point.}
The gain sweeps retrain Fixed-PD and Learned-PD with one base gain set to a fraction of its bound, isolating each bound in turn.
The damping sweep uses the nominal reference, whereas the stiffness sweep keeps $q^*$ and scales the contact force sixfold, which raises $\tau^*$ until Bound~1 reaches the nominal $k^p_0$.
This crossing occurs when the peak torque exhausts the offset authority, $\max_t|\tau^*(t)| = \alpha k^p_0$, and substituting $k^p_0 = I_{jj}\omega_n^2 = mL^2\omega_n^2/3$ into Eq.~\ref{eq:bound1_rod} cancels the mass and places the crossing at
\begin{equation}
  \frac{F^*_{\max}}{mg} = \frac{1}{2} + \frac{\alpha L \omega_n^2}{3\,g\cos q_c} = 6.05,
  \label{eq:crossing}
\end{equation}
which depends only on the controller bandwidth, the action range, the rod geometry and gravity.
At the sixfold force, the nominal $k^p_0 = 32$\,Nm/rad lies on Bound~1, which sets the operating point of the stiffness sweep and of the gain trajectory in Fig.~\ref{fig:action_space_bounds}.

\paragraph{Gain sweeps.}
Sweeping $k^p_0$ at the sixfold contact force tests Bound~1 in isolation.
Fixed-PD's force error falls from $37$\,N at $k^p_0 = 8$\,Nm/rad to about $1$\,N from $k^p_0 = 48$\,Nm/rad upward, bracketing the bound at $31.7$\,Nm/rad, while its angle error stays between $9$ and $13^\circ$, so that below the bound the controller holds the pose and forfeits the force.
Sweeping $k^d_0$ at the nominal reference tests Bound~2, and torque jerk rises from $1.2$\,kNm/s at $k^d_0 = 15$\,Nm\,s/rad to $15$\,kNm/s at $k^d_0 = 25$\,Nm\,s/rad, bracketing the bound at $20$\,Nm\,s/rad, with angle and force error rising sharply.
Learned-PD holds a force error of $0.7$--$1.1$\,N down to $k^p_0 = 16$\,Nm/rad and reaches $12$\,N only at $k^p_0 = 8$\,Nm/rad, where its ceiling $k^p_0 e^{s} = 22$\,Nm/rad falls below Bound~1.
Across the damping sweep, its jerk, angle and force error stay at $0.3$--$1.1$\,kNm/s, $7$--$10^\circ$ and under $1$\,N up to twice Bound~2.

\subsubsection{ImpedanceMimic}
\label{App:ImpedanceMimic}

\renewcommand{\algorithmiccomment}[1]{\hfill$\triangleright$~#1}
\newcommand{\algstage}[1]{\item[] \textbf{#1}}

\paragraph{Kinematics-to-GRF estimator.}
\label{App:estimator}
The estimator maps reference kinematics to per-foot GRF in two stages, a closed-form physics prior followed by a learned residual (Algorithm~\ref{alg:training}). It is fitted once per split on the training clips with a fixed seed, so that all runs of a split share it, and is frozen thereafter. The prior alone reaches a vertical-GRF correlation of $0.94$ on the training frames, and the full estimator reaches activity-mean vertical, anterior--posterior and medio-lateral correlations of $0.979$, $0.882$ and $0.974$ on seen and $0.943$, $0.816$ and $0.963$ on unseen subjects. Wherever the actor observes a contact flag, it is derived from the estimate as $\hat{F}_{z,f} > 0.1$~BW rather than from the plates.

\vspace{-3pt}%
\begin{algorithm}[h]%
\caption{ImpedanceMimic Training.}
\label{alg:training}
\small
\begin{algorithmic}[1]
\REQUIRE Reference motions $\mathcal{D} = \{(\vect{q}^*, \dot{\vect{q}}^*, \vect{F}^*, \vect{c}^*)\}$,
         subject-specific models $\{m_s\}_{s=1}^{S}$
\ENSURE Policy $\pi_\theta$, GRF estimator $\hat{\vect{F}}$

\algstage{Stage 1: Kinematics-only GRF estimator}
\STATE $\vect{F}^{\mathrm{prior}} \leftarrow \textsc{PhysicsPrior}(\vect{q}^*, \dot{\vect{q}}^*)$
  \COMMENT{center-of-mass dynamics}
\STATE Train ensemble $\{f_{\phi_i}\}_{i=1}^{K}$ to regress the residual $\vect{F}^* - \vect{F}^{\mathrm{prior}}$
\STATE $\hat{\vect{F}} \leftarrow \vect{F}^{\mathrm{prior}} + \frac{1}{K}\sum_{i} f_{\phi_i}$;
  \quad $\hat{c} \leftarrow \mathbf{1}[\hat{F}_z > \epsilon_c]$
  \COMMENT{frozen during RL}

\algstage{Stage 2: Impedance policy learning}
\STATE Scale nominal gains $\vect{k}^p_0, \vect{k}^d_0$ per subject by joint inertia
\STATE Initialize actor $\pi_\theta$ with auxiliary head $h_\theta$, and critic $V_\psi$
\REPEAT
  \FOR{$t = 1$ \TO $T$, in parallel over environments}
    \STATE $\vect{o}_t \leftarrow \textsc{Obs}\bigl(\vect{q}_t, \dot{\vect{q}}_t,\; \vect{q}^*_{t+\Delta},\; \hat{\vect{F}}_{t+\Delta}, \hat{c}_{t+\Delta}\bigr)$
      \COMMENT{no force-plate input}
    \STATE $\vect{a}_t \sim \pi_\theta(\cdot \mid \vect{o}_t)$
    \STATE $(\vect{k}^p_t, \vect{k}^d_t, \vect{q}^{\mathrm{cmd}}_t) \leftarrow \textsc{Decode}(\vect{a}_t;\, \vect{k}^p_0, \vect{k}^d_0, \vect{q}^*_{t+1})$
      \COMMENT{gain lag}
    \STATE Simulate with $\vect{\tau} = \mathrm{clip}\bigl(\vect{k}^p_t \circ (\vect{q}^{\mathrm{cmd}}_t - \vect{q}) + \vect{k}^d_t \circ (\dot{\vect{q}}^*_{t+1} - \dot{\vect{q}}),\, \vect{\tau}^{\lim}\bigr)$
    \STATE $r_t \leftarrow r(\vect{q}_{t+1}, \vect{F}_{t+1};\, \vect{q}^*_{t+1}, \vect{F}^*_{t+1})$
      \COMMENT{kinematic + GRF tracking}
    \STATE Reset terminated episodes via reference-state initialization
  \ENDFOR
  \STATE Estimate advantages $\hat{A}_t$ with GAE using privileged critic $V_\psi(\vect{o}_t, \vect{F}^*_t)$
  \STATE Update $\theta, \psi$ by minimizing
    $\mathcal{L}_{\mathrm{PPO}} + w_{\mathrm{aux}}\, \ell\bigl(h_\theta(\vect{o}_t), \vect{F}^*_{t+\Delta}\bigr)$
\UNTIL{converged}
\end{algorithmic}
\end{algorithm}
\vspace{-3pt}

\textit{Physics prior.}
The whole-body CoM of each reference frame is computed with the subject's model, smoothed per clip with the binomial kernel $[1, 4, 6, 4, 1]/16$ and differentiated twice by central differences. The total vertical force $\max\bigl((\ddot{\vect{c}} + g)/g \cdot \vect{e}_{\mathrm{up}},\, 0\bigr)$ is distributed across the feet with weights $w_f = \mathrm{ramp}(h_{\mathrm{top}} - h_f;\, h_{\mathrm{ramp}})\, \mathrm{ramp}(v_{\mathrm{top}} - v_f;\, v_{\mathrm{ramp}})$, where $h_f$ and $v_f$ are the lowest height and speed among the bodies of foot $f$ ($h_{\mathrm{top}} = 0.10$~m, $h_{\mathrm{ramp}} = 0.04$~m, $v_{\mathrm{top}} = 1.0$~m/s, $v_{\mathrm{ramp}} = 0.6$~m/s). These weights are scaled by the inverse horizontal CoM-to-heel distance and normalized across feet. The horizontal components scale the vertical force by the horizontal-to-vertical ratio of the vector from the lowest foot point to the CoM, with a vertical leg length of at least $0.3$~m.

\textit{Residual network.}
The per-frame features, expressed in the pelvis heading frame without the root translation, comprise the pelvis height, tilt and list, the 25 non-root coordinates and their velocities, the root angular velocity, the root and CoM linear velocities, the CoM height, the heights and speeds of the four foot bodies, the heel offsets from the CoM, and the prior with its contact weights. Features at twelve offsets spanning $\pm 40$ frames (Table~\ref{tab:hyperparams}), clamped to the clip, are concatenated and standardized. An ensemble of MLPs with zero-initialized output layers, which therefore start at the prior, regresses the residual $\vect{F}^* - \vect{F}^{\mathrm{prior}}$ in the heading frame with a Huber loss and AdamW, using the plate-valid frames of the training clips and their left--right mirror images, with one segment in twenty withheld for diagnostics. The ensemble mean is smoothed per clip with the binomial kernel and rotated back to the world frame, and its vertical components are clamped at zero.
\vspace{-5pt}%
\paragraph{Observations.}
\label{App:obs}

\begin{table}[h]
\centering
\small
\caption{\textbf{Actor observation (215 dimensions).} Lookahead offsets are in reference frames at 100~Hz.}
  \vspace{-0.4em}
\label{tab:obs}
\settowidth{\toycolA}{Kinematic lookahead}\settowidth{\toycolB}{\textbf{Dims}}%
\begin{tabular}{@{}l p{\dimexpr\linewidth-\toycolA-\toycolB-4\tabcolsep\relax} r@{}}
\toprule
\textbf{Component} & \textbf{Contents} & \textbf{Dims} \\
\midrule
Proprioception & pelvis height, the three pelvis rotations and the 25 remaining coordinates, actuated and passive (29), and all generalized velocities (31) & 60 \\
Contact state & per foot: vertical contact force (BW, clamped at 0) and heel height & 4 \\
\midrule
Kinematic lookahead & $\vect{q}^*_{t+k} - \vect{q}_t$ for $k \in \{1, 5, 25\}$, clamped at the clip end & 93 \\
GRF lookahead & $\hat{\vect{F}}_{t+k}$, six components, $k \in \{1, 5, 25\}$ & 18 \\
Foot lookahead & per foot at $k \in \{0, 1, 5, 25\}$: heel height, toe height, heel vertical velocity and acceleration, estimated contact flag & 40 \\
\bottomrule
\end{tabular}
\vspace{-15pt}%
\end{table}

Raw observations are clipped to $\pm 100$ and normalized by running mean and variance, with the result clipped to $\pm 5$. The privileged features, namely the measured GRF at the current frame (6 dimensions) and at the three lookahead offsets (18 dimensions), have a separate running normalizer, $\mathcal{N}_{\mathrm{priv}}$ in Eq.~\ref{eq:aux}. The critic receives the normalized observation concatenated with the normalized current-frame GRF, and the auxiliary head regresses the lookahead GRF with a weighted Huber loss ($w_{\mathrm{aux}} = 0.25$) in the PPO objective,
\begingroup\setlength{\abovedisplayskip}{7pt}
\begin{equation}
  \mathcal{L} = \mathcal{L}_{\mathrm{PPO}}
    + w_{\mathrm{aux}}\,
      \mathrm{Huber}\bigl(\smash{\hat{\vect{F}}^{\mathrm{aux}}}(\vect{o}_t),\;
      \mathcal{N}_{\mathrm{priv}}(\vect{F}^*_{t+\{1,5,25\}})\bigr).
  \label{eq:aux}
\end{equation}
\endgroup

\paragraph{Reward.}
\label{App:reward}
The kinematic term is $r^{\mathrm{kin}} = 0.65\, r_{\mathrm{pose}} + 0.10\, r_{\mathrm{vel}} + 0.10\, r_{\mathrm{root}} + 0.15\, r_{\mathrm{lp}}$, with
\begin{equation}
  r_{\mathrm{pose}} = e^{-0.04 \sum_{j} (q_j - q^*_j)^2/\sigma_j^2 - 2 \sum_{k} \Delta\theta_k^2},\quad r_{\mathrm{vel}} = e^{-0.02\,\|\Delta\dot{\vect{q}}\|^2},\quad r_{\mathrm{root}} = e^{-5\,\|\Delta\vect{v}_{\mathrm{pelvis}}\|^2},
\label{eq:rpose}
\end{equation}
where the joint sums run over the actuated joints, $\Delta\theta$ is the pelvis tilt, list and rotation error, $\vect{v}_{\mathrm{pelvis}}$ is the three-dimensional pelvis velocity in world axes, and $\sigma_j$ is the standard deviation of $q^*_j$ over the clip, clipped to $[0.05, 0.141]$~rad. The end-effector term $r^{\mathrm{ee}} = \exp\bigl(-40 \sum_{b} \|\Delta(\vect{p}_b - \vect{p}_{\mathrm{pelvis}})\|^2\bigr)$ runs over the heel and toe bodies of both feet, with positions relative to the pelvis in world axes, which makes it invariant to pelvis drift but sensitive to yaw. The CoP errors of Eq.~\ref{eq:rgrf} are measured relative to each foot's calcaneus in world axes and are further gated on a valid measured CoP and a loaded simulated foot ($F_{z,f} > 0.05$~BW). The Torque baseline uses the unnormalized kernel $\exp(-2\sum_j (q_j - q^*_j)^2 - 2\sum_k \Delta\theta_k^2)$ instead (Table~\ref{tab:torque_variants}).

\vspace{-6pt}%
\paragraph{Root-velocity low-pass term.}
The within-stride velocity oscillation dominates the instantaneous $r_{\mathrm{root}}$ (running: RMS $0.35$\,m/s, kernel $\approx 0.54$), and a steady speed deficit of $0.14$\,m/s ($6\,\%$) lowers it by only $0.013$ per step. The low-passed horizontal error $\bar{\vect{e}}_t = \bar{\vect{e}}_{t-1} + \beta\,(\vect{v}_{xy} - \vect{v}^*_{xy} - \bar{\vect{e}}_{t-1})$, with $\beta = 1 - e^{-\Delta t/\tau}$ and $\tau = 1$\,s, is reset at the start of every episode and rewarded as $r_{\mathrm{lp}} = \exp(-30\,\|\bar{\vect{e}}_t\|^2)$, its weight of $0.15$ taken from $r_{\mathrm{root}}$. The $1$\,s window attenuates the $1.3$\,Hz stride oscillation about eightfold while passing the bias. On seen subjects the term lowers the drift rate from $0.047$ to $0.040$\,m/s (Table~\ref{tab:ablation_seen}) and the pelvis-position error from $0.24$ to $0.16$\,m at equal survival, kinematics and force, with $8\,\%$ and $18\,\%$ higher torque jerk on flexion and jumping.

\begin{table}[p]%
\centering
\caption{\textbf{ImpedanceMimic hyperparameters.}}
\vspace{-0.5em}
\label{tab:hyperparams}
\scriptsize
\begin{tabular}{@{}lll@{}}
\toprule
\textbf{Category} & \textbf{Parameter} & \textbf{Value} \\
\midrule
\multirow{7}{*}{\textit{Simulation}}
& Sim timestep $h$ & 0.002~s \\
& Control rate & 100~Hz (5 substeps) \\
& Contact time constant $t_c$ & 0.04~s \\
& Friction $\mu$ & 1.0 \\
& Foot model & 8 spheres per foot (extends Falisse et al.) \\
& Solver iterations & 6 (6 line-search) \\
& Parallel worlds $N$ & 8{,}192, equal per training subject \\
\midrule
\multirow{10}{*}{\makecell[l]{\itshape Action space\\\itshape (impedance)}}& Gain log-scale $s_p$ ($k^p$) & 1.25 \\
& Gain log-scale $s_d$ ($k^d$) & 1.5 \\
& Gain lag $T_{\mathrm{act}}$ & 50~ms \\
& Position offset scale $\alpha$ & 0.5~rad \\
& Joint groups & hip / knee / ankle / subtalar / lumbar \\
& $k^p_{\mathrm{nom}}$ (reference subject) & 400 / 400 / 300 / 100 / 500~Nm/rad \\
& $k^d_{\mathrm{nom}}$ (reference subject) & 40 / 40 / 30 / 10 / 50~Nm\,s/rad \\
& Per-subject gain scaling & joint inertia, Eq.~\ref{eq:inertia_scaling} \\
& Gain cap & none \\
& $\tau^{\lim}$ & 275 / 300 / 150 / 50 / 160~Nm \\
\midrule
\multirow{5}{*}{\makecell[l]{\itshape Action space\\\itshape (muscle)}}& Muscles & 80 Hill-type MTUs + 3 lumbar torque motors \\
& Lumbar torque scale & 100~Nm \\
& Excitation & $u = (a + 1)/2$, Millard activation dynamics \\
& Extra observation & 80 activations \\
& PPO deviations & entropy coeff 0, learning rate $10^{-4} \to 3 \times 10^{-6}$ (linear) \\
\midrule
\multirow{3}{*}{\textit{Observation}}
& Kinematic lookahead offsets & $\{1, 5, 25\}$ frames \\
& Foot lookahead offsets & $\{0, 1, 5, 25\}$ frames \\
& GRF lookahead offsets & $\{1, 5, 25\}$ frames \\
\midrule
\multirow{7}{*}{\textit{GRF estimator}}
& Window offsets & $\{$-$40, $-$25, $-$15, $-$10, $-$5, 0, 1, 5, 10, 15, 25, 40\}$ \\
& Hidden size & 1024 \\
& Layers & 2 \\
& Ensemble size $K$ & 3 \\
& Loss & Huber ($\delta = 0.1$) \\
& Gradient steps (per member) & 4000 \\
& Optimizer & AdamW, lr $10^{-3}$ (cosine), batch 4096 \\
\midrule
\multirow{6}{*}{\textit{Reward weights}}
& $w_{\mathrm{kin}}$ & 1.0 \\
& $w_{\mathrm{grf}}$  & $0.25 \to 0.5$ (annealed) \\
& $w_{\mathrm{shear}}$  & 0.25 \\
& $w_{\mathrm{ee}}$ & 0.15 \\
& Kinematic sub-weights & pose 0.65, vel 0.10, root 0.10, low-passed root 0.15 \\
& root low-pass $\tau$, $\alpha_{\mathrm{lp}}$ & 1~s, 30 \\
\midrule
\multirow{10}{*}{\textit{Reward scales}}
& $\alpha_{\mathrm{pose}}$ & 0.04 \\
& $\alpha_{\mathrm{vel}}$ & 0.02 \\
& $\alpha_{\mathrm{root}}$ & 5.0 \\
& $\alpha_{\mathrm{grf}}$ & 4.0 \\
& $\alpha_{\mathrm{cop}}$ & 5.0 \\
& $\alpha_{\mathrm{shear}}$ & 5.0 \\
& $\alpha_{\mathrm{ee}}$ & 40.0 \\
& GRF relative error floor & 0.5 \\
& Pose amplitude floor & 0.05~rad \\
& Pose amplitude cap & 0.141~rad \\
\midrule
\multirow{5}{*}{\textit{Network}}
& Actor& $[512, 256, 128]$, SiLU \\
& Mean head & Linear$(128 \to 45)$ \\
& Aux head & Linear$(128 \to 18)$ \\
& Critic & $[512, 256, 128, 1]$, SiLU \\
& Action distribution & tanh-Normal, global $\log\sigma$ \\
\midrule
\multirow{13}{*}{\textit{PPO}}
& Total timesteps & $2 \times 10^9$ \\
& Unroll length $T$ & 20 \\
& Minibatches & 32 ($5{,}120$ samples each) \\
& Epochs per collect & 4 \\
& Discount $\gamma$ & 0.98 \\
& GAE $\lambda$ & 0.95 \\
& Clip $\epsilon$ & 0.3 \\
& Learning rate & $3 \times 10^{-4} \to 3 \times 10^{-6}$ (linear) \\
& Entropy coeff & 0.01 \\
& Value loss coeff & 0.5 \\
& Aux loss weight $w_{\mathrm{aux}}$ & 0.25 \\
& Grad norm clip & 0.5 \\
& Optimizer & Adam \\
\midrule
\multirow{5}{*}{\makecell[l]{\itshape RSI and\\\itshape termination}}
& Fall height & 0.4~m \\
& $\dot{q}$ limit & 100~rad/s \\
& RSI noise $\sigma_q$ & 0.05~rad \\
& RSI noise $\sigma_{\dot{q}}$ & 0.2~rad/s \\
& Start-frame sampling & uniform over the subject's frames \\
\bottomrule
\end{tabular}
\end{table}
\begin{figure}[!b]%
    \vspace{-13pt}%
    \centering
    \includegraphics[width=\linewidth]{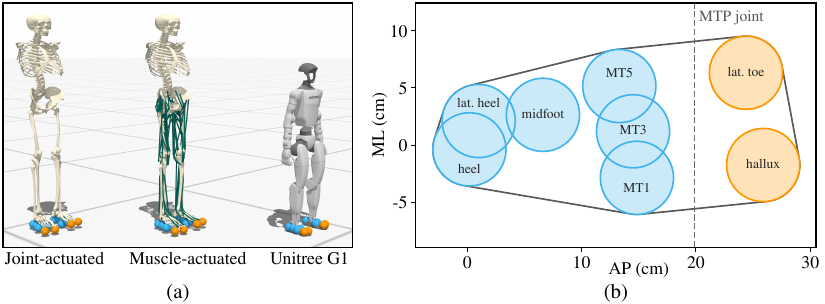}
    \vspace{-1.3em}%
    \caption{\textbf{Simulated bodies and contact model.} \textbf{(a)}~The joint-actuated human model, its 80-muscle counterpart and the Unitree G1, each supported by eight contact spheres per foot. \textbf{(b)}~Top view of the right foot, with six spheres on the calcaneus and two on the toes beyond the passive MTP joint.}
    \label{fig:contact_model}
\end{figure}
\vspace{-6pt}
\paragraph{Human model and contact.}
\label{App:contact}
Of the ten passive coordinates (App.~\ref{App:preprocessing}), eight are knee translations and rotations tied to the knee angle by equality constraints, and two are the metatarsophalangeal (MTP) joints, each with a linear spring--damper ($k = 25$~Nm/rad, $d = 2$~Nm\,s/rad) over $[-60^\circ, 0]$. In Eq.~\ref{eq:eom}, $\mat{M} \in \mathbb{R}^{n \times n}$, $\vect{h} \in \mathbb{R}^{n}$, $\vect{\tau} \in \mathbb{R}^{m}$, $\vect{\lambda} \in \mathbb{R}^{6 n_c}$ for the $n_c \le 16$ active contact wrenches and $\mat{J}_c \in \mathbb{R}^{6 n_c \times n}$, and the GRFs $\vect{F}_r, \vect{F}_l \in \mathbb{R}^{3}$ are the resultant forces on each foot. The constant selection matrix $\mat{S} \in \{0,1\}^{m \times n}$ has one unit entry per actuated joint and zero columns for the six pelvis and ten passive coordinates. MTP spring--damper torques enter $\vect{h}$, and knee coupling constraint forces are omitted. The eight contact spheres per foot (radius $32$~mm, Fig.~\ref{fig:contact_model}) extend the six-sphere foot of \citet{falisse2019rapid} with posterolateral-heel and third-metatarsal spheres. Contacts use sliding, torsional and rolling friction of $1.0$, $0.8$ and $0.005$ and a time constant $t_c = 0.04$~s, and a common $5$~mm vertical offset is fitted on standing frames.

\paragraph{Training and implementation.}
Worlds are distributed equally across the training subjects, and every reset draws a start frame uniformly from the frames of the world's subject. Gaussian noise of $0.05$~rad is added to all non-root coordinates and the pelvis orientation, and of $0.2$~rad/s to all velocities. The lagged gain state of Eq.~\ref{eq:gain_lag} persists across these training resets and is zeroed only at environment initialization and before every evaluation episode. Episodes terminate at the clip end, when the pelvis falls below $0.4$~m, on a non-finite state, or when $\max_j |\dot q_j|$ exceeds $100$~rad/s, and time-outs bootstrap the value. A single Adam optimizer updates actor, critic and auxiliary head on the clipped surrogate plus $0.5\cdot\tfrac12 (V - R)^2$, minus $0.01$ times the policy entropy, plus the auxiliary term of Eq.~\ref{eq:aux} (Huber, $\delta = 1$ in normalized units). Advantages are normalized per minibatch, gradient norms are clipped at $0.5$, and the actor's mean head is initialized uniformly in $\pm 3 \times 10^{-3}$ with zero log standard deviation. All worlds share one MuJoCo Warp data object with zero-copy GPU views of $\vect{q}$, $\dot{\vect{q}}$ and the contact wrenches, rollouts and PPO updates run on the GPU, and the physics substeps are captured as a CUDA graph. Per-foot forces and CoP follow from the heel and toe contact wrenches, $\vect{F}_f = \sum_b \vect{f}_b$ and $\vect{c}_f = (-\tau_y, \tau_x)/\max(F_{z,f}, 1~\mathrm{N})$, where $\vect{\tau} = \sum_b (\vect{\tau}_b + \vect{c}_{\mathrm{root}} \times \vect{f}_b)$ is the total moment shifted from the root subtree CoM, about which MuJoCo expresses the wrenches, to the world origin. Table~\ref{tab:compute} reports the resulting throughput.

\vspace{-5pt}\vspace{0pt}
\subsubsection{Baseline Implementation Details}
\label{App:baselines}
\vspace{-5pt}

\paragraph{Muscle.}
\label{App:muscle}
The Muscle baseline actuates each participant's scaled Rajagopal model~\citep{rajagopal2016full} through its 80 Hill-type muscle--tendon units, converted to MuJoCo's rigid-tendon muscle actuators with MyoConverter~\citep{wang2022myosim} and grafted onto the per-participant skeleton of the other action spaces, so that segment inertias, joint ranges and contact spheres are identical. Routed over 40 wrapping cylinders, the converted muscle paths reproduce the OpenSim moment arms with a median error of $0.2$\,mm across the 10{,}320 muscles of the 129 participants.
The policy outputs an excitation $u = (a+1)/2 \in [0, 1]$ for each muscle, where $a$ is the corresponding action, held constant over the five substeps of a control step.
MuJoCo converts excitations into activations through first-order activation dynamics~\citep{millard2013flexing} with time constants of $10$\,ms for activation and $40$\,ms for deactivation, scaled by the current activation.
The three lumbar joints remain driven by torque motors with an action scale of $100$\,Nm, the observation adds the 80 muscle activations, and the evaluated joint torques are the net actuator torques at the 15 actuated joints.

\begin{table*}[!b]%
\vspace{-15pt}%
\centering
\caption{\textbf{Muscle actuation at twice the training budget.} Mean$_{\pm\text{sd}}$ over 5 seeds after 4\,B steps, with the linear learning-rate decay stretched accordingly. Columns are RMSE ($\downarrow$) unless marked $\uparrow$.}
\label{tab:muscle_budget}
\vspace{-0.4em}%
\renewcommand{\arraystretch}{0.9}
\setlength{\tabcolsep}{2pt}
\newcommand{\gap}{\hspace{4pt}}
\newcommand{\sd}[1]{{\tiny\textcolor{black!55}{$\pm$#1}}}
\scriptsize
\resizebox{\textwidth}{!}{%
\begin{tabular}{@{}>{\fillL}l @{\gap} l @{\gap} l @{\gap} llll @{\gap} lp{0.75cm}p{0.75cm}p{0.75cm}ll @{\gap} ll>{\fillR}l @{}}
\toprule
 & & \multicolumn{1}{c}{\textbf{Stability}} & \multicolumn{4}{c}{\textbf{Kinematics}} & \multicolumn{6}{c}{\textbf{Ground Reaction Forces}} & \multicolumn{3}{c}{\textbf{Torques}$^\dagger$} \\
\cmidrule(lr){3-3}\cmidrule(lr){4-7}\cmidrule(lr){8-13}\cmidrule(lr){14-16}
\textbf{Budget} & \texttt{\textbf{Activity}}
  & \makecell{Ep.$\hi$ \\ (\%)}
  & \makecell{$q\lo$\\ \makebox[0pt][c]{($^\circ$)}}
  & \makecell{$\dot{q}\lo$\\ \makebox[0pt][c]{($^\circ$/s)}}
  & \makecell{$p_x\lo$\\ \makebox[0pt][c]{(m/s)}}
  & kin$_r\hi$
  & \makecell{$F_{\mathrm{Z}}\lo$\\ \makebox[0pt][c]{(BW)}}
  & $F^{\uparrow}_{\mathrm{Z}_r}$
  & $F^{\uparrow}_{\mathrm{ML}_r}$
  & $F^{\uparrow}_{\mathrm{AP}_r}$
  & \makecell{$\mathrm{CoP}^{\downarrow}_{\mathrm{ML}}$\\ {\tiny (mm)}}
  & \makecell{$\mathrm{CoP}^{\downarrow}_{\mathrm{AP}}$\\ {\tiny (mm)}}
  & \makecell{$\tau\lo$\\ \makebox[0pt][c]{\tiny (Nm/kg)}}
  & $\tau_r\hi$
  & \makecell{$\dot{\tau}\lo$\\ \makebox[0pt][c]{\tiny (Nm/kg/s)}} \\
\specialrule{\lightrulewidth}{\aboverulesep}{0pt}%
\rowcolor{papericeblue}
\multicolumn{16}{@{}>{\fillL}l}{\rule{0pt}{8pt}\textbf{Seen subjects, held-out clips}} \\[1pt]
\multirow{7}{*}{4\,B} & \act{walk} & 98.7\sd{.9} & 2.41\sd{.05} & 39.2\sd{.3} & .030\sd{.003} & .95\sd{.00} & .10\sd{.00} & .97\sd{.00} & .87\sd{.01} & .86\sd{.01} & 21\sd{1} & 50\sd{1} & .40\sd{.00} & .61\sd{.01} & 6.48\sd{.11} \\
 & \act{run} & 99.8\sd{.2} & 3.47\sd{.07} & 72.6\sd{1.3} & .071\sd{.015} & .94\sd{.00} & .14\sd{.01} & .98\sd{.00} & .41\sd{.04} & .80\sd{.02} & 20\sd{1} & 37\sd{2} & .77\sd{.01} & .72\sd{.01} & 13.5\sd{.2} \\
 & \act{stand} & 92.3\sd{1.3} & 1.97\sd{.15} & 8.54\sd{.97} & .013\sd{.002} & .44\sd{.02} & .07\sd{.00} & .57\sd{.02} & .70\sd{.04} & .19\sd{.03} & 16\sd{1} & 46\sd{2} & .77\sd{.04} & .09\sd{.01} & .70\sd{.07} \\
 & \act{lateral} & 91.8\sd{1.6} & 7.58\sd{.27} & 77.5\sd{1.9} & .112\sd{.006} & .73\sd{.01} & .24\sd{.01} & .89\sd{.01} & .77\sd{.02} & .62\sd{.02} & 30\sd{2} & 59\sd{4} & .55\sd{.01} & .49\sd{.01} & 11.6\sd{.3} \\
 & \act{flexion} & 67.0\sd{2.0} & 10.3\sd{.4} & 53.6\sd{5.3} & .066\sd{.002} & .63\sd{.01} & .20\sd{.00} & .62\sd{.02} & .56\sd{.03} & .52\sd{.01} & 26\sd{3} & 72\sd{2} & .45\sd{.02} & .40\sd{.02} & 4.22\sd{.08} \\
 & \act{jump} & 55.3\sd{1.0} & 11.0\sd{.6} & 83.0\sd{5.0} & .112\sd{.003} & .59\sd{.02} & .34\sd{.01} & .60\sd{.01} & .51\sd{.04} & .32\sd{.04} & 32\sd{6} & 79\sd{4} & .55\sd{.02} & .34\sd{.02} & 8.18\sd{.36} \\
\rowcolor{meangray}  & \textbf{Mean} & 84.2\sd{.9} & 6.12\sd{.16} & 55.7\sd{1.7} & .067\sd{.004} & .71\sd{.01} & .18\sd{.00} & .77\sd{.01} & .64\sd{.01} & .55\sd{.01} & 24\sd{1} & 57\sd{1} & .58\sd{.01} & .44\sd{.01} & 7.46\sd{.14} \\
\specialrule{\heavyrulewidth}{\aboverulesep}{0pt}
\rowcolor{papericeblue}
\multicolumn{16}{@{}>{\fillL}l}{\rule{0pt}{8pt}\textbf{Unseen subjects ($H$), zero-shot}} \\[1pt]
\multirow{7}{*}{4\,B} & \act{walk} & 91.3\sd{3.9} & 3.87\sd{.37} & 49.3\sd{2.0} & .058\sd{.006} & .90\sd{.02} & .14\sd{.01} & .93\sd{.01} & .79\sd{.03} & .81\sd{.02} & 23\sd{1} & 50\sd{3} & .42\sd{.01} & .57\sd{.02} & 6.82\sd{.13} \\
 & \act{run} & 98.8\sd{.4} & 3.75\sd{.15} & 78.7\sd{2.6} & .085\sd{.009} & .93\sd{.01} & .16\sd{.02} & .98\sd{.00} & .43\sd{.07} & .81\sd{.03} & 20\sd{0} & 38\sd{3} & .94\sd{.02} & .68\sd{.02} & 14.0\sd{.4} \\
 & \act{stand} & 94.9\sd{2.3} & 1.68\sd{.13} & 6.42\sd{1.03} & .009\sd{.002} & .47\sd{.02} & .07\sd{.01} & .55\sd{.04} & .78\sd{.05} & .26\sd{.13} & 19\sd{6} & 47\sd{12} & .87\sd{.08} & .11\sd{.01} & .64\sd{.05} \\
 & \act{lateral} & 83.0\sd{5.5} & 10.8\sd{.4} & 97.4\sd{3.9} & .127\sd{.013} & .59\sd{.02} & .31\sd{.02} & .83\sd{.02} & .69\sd{.03} & .51\sd{.05} & 33\sd{2} & 66\sd{5} & .64\sd{.03} & .43\sd{.02} & 10.5\sd{.4} \\
 & \act{flexion} & 54.4\sd{3.8} & 11.2\sd{.4} & 54.1\sd{2.1} & .079\sd{.006} & .59\sd{.01} & .22\sd{.00} & .58\sd{.01} & .53\sd{.02} & .59\sd{.03} & 22\sd{2} & 69\sd{2} & .47\sd{.01} & .35\sd{.01} & 3.98\sd{.31} \\
 & \act{jump} & 48.3\sd{4.2} & 10.4\sd{.6} & 76.6\sd{1.0} & .153\sd{.005} & .60\sd{.02} & .36\sd{.02} & .56\sd{.04} & .51\sd{.05} & .25\sd{.01} & 30\sd{4} & 83\sd{5} & .54\sd{.01} & .30\sd{.01} & 6.94\sd{.18} \\
\rowcolor{meangray}  & \textbf{Mean} & 78.5\sd{1.9} & 6.95\sd{.15} & 60.4\sd{1.1} & .085\sd{.004} & .68\sd{.01} & .21\sd{.01} & .74\sd{.01} & .62\sd{.02} & .54\sd{.02} & 24\sd{1} & 59\sd{2} & .65\sd{.02} & .41\sd{.00} & 7.15\sd{.16} \\
\bottomrule
\end{tabular}%
}
\vspace{0pt}
\parbox{\textwidth}{\scriptsize\raggedright
  $^\dagger$9 main joints (hip flexion / adduction, knee, ankle, both sides; lumbar extension), against inverse dynamics.
  Ep. = episode survival before termination (\%). $q$, $\dot q$ = joint-angle / velocity RMSE. $p_x$ = root forward-position drift rate. \{kin$_r$, $F_{\mathrm{Z}_r}$, $F_{\mathrm{AP}_r}$, $F_{\mathrm{ML}_r}$, $\tau_r$\} = Pearson correlation w.r.t. reference \{joint angles, vertical, AP, ML GRF, torques\}. CoP = center of pressure. $\dot\tau$ = torque jerk.}
\vspace{-0.8em}
\end{table*}
Data, reward, network and PPO settings follow \projname{}, except for a zero entropy cost and a learning rate of $10^{-4}$ instead of $3 \times 10^{-4}$, both found by a hyperparameter search, since the shared values stall training at a joint-angle error of $34^\circ$.
Doubling the training budget lowers the joint-angle error on seen subjects from $7.2^\circ$ to $6.1^\circ$ but leaves survival and the vertical-force correlation unchanged (Table~\ref{tab:muscle_budget}).
On seen walking and running, the activities on which MuscleMimic~\citep{li2026towards} trains its generalist policy, Muscle survives $98.6$ and $99.7\,\%$ of the episode at a joint-angle error of $2.7$ and $3.8^\circ$ (Table~\ref{tab:golden_full}).
MuscleMimic reports $6.6^\circ$ on its held-out motions after full training and $8.0$ to $10.7^\circ$ after the same 2\,B steps, for a full-body model with 416 muscles from toes to fingertips rather than our 80 lower-limb muscles.
Its lower survival in flexion and jumping is consistent with the rigid tendons of MuJoCo muscles, which store no elastic energy, and MuscleMimic jumps only after scaling the maximal muscle force fivefold and fine-tuning on that single motion.

\paragraph{GaitDynamics~\citep{tan2026gaitdynamics}.}
GaitDynamics is the only baseline that does not control a body. Its force-refinement model (\href{https://github.com/stanfordnmbl/GaitDynamics}{github.com/stanfordnmbl/GaitDynamics}), a six-layer transformer encoder ($d_\mathrm{model} = 192$, eight attention heads, ${\sim}2$\,M parameters), regresses bilateral GRF and CoP from a 150-frame (1.5\,s) window of reference kinematics, so only its force and CoP columns are defined. We retain the upstream architecture, input features, optimizer (Adan, learning rate $4 \times 10^{-4}$), exponential moving average of the weights and training budget (7{,}680 epochs), and retrain the model on our corpus with the subject and clip splits shared by every other method, in place of the upstream AddBiomechanics training set. The single substantive adaptation is to align the heading of every clip at inference. Upstream leaves clips whose pelvis heading spans more than $45^\circ$ unaligned, although the model is trained exclusively on aligned windows, and aligning them raises the vertical-force correlation on lateral clips of unseen subjects from $0.65$ to $0.89$.

\vspace{-6.5pt}%
\paragraph{PHC~\citep{luo2023perpetual}.}
PHC imitates motion capture on the SMPL humanoid with a fixed-gain PD controller. We re-implement it (\href{https://github.com/ZhengyiLuo/PHC}{github.com/ZhengyiLuo/PHC}) in MuJoCo Warp on the per-participant Rajagopal models, where it is trained and scored on the same data, physics and evaluation protocol as every other method (Tables~\ref{tab:existing}, \ref{tab:existing_full} and~\ref{tab:existing_transfer}). The 544-dimensional observation, the reward (exponential imitation kernels and an AMP discriminator), the networks of the upstream imitation configuration (actor, critic and discriminator with hidden layers of 1024 and 512 units), the PPO hyperparameters (App.~\ref{app:training}), implicit damping and the PMCP curriculum follow upstream. We drop the upstream joint-power penalty, which turns the per-step reward negative in running and makes early termination optimal. Training episodes terminate when any body deviates from the reference by more than $0.25$\,m, and every $245.76$\,M steps the curriculum evaluates the training clips at a $0.5$\,m failure threshold and resamples failed clips in proportion to their cumulative failures. The policy acts every 15 physics steps (33\,Hz), the rate closest to the upstream 30\,Hz, at which the rollout, discount horizon and AMP window stay near their upstream durations, and trains for $2 \times 10^9$ steps like every other method. A single primitive replaces the progressive multi-primitive composer of the original method, as in the single-primitive model released with the upstream code. Actions are absolute PD targets spanning $\pm 0.7$ of each joint range about its midpoint, with gains from our inertia-scaled priors, because the SMPL gains of PHC do not transfer to the Rajagopal skeleton.

We validate the port in PHC's original setting, the mean-shape SMPL humanoid imitating AMASS~\citep{mahmood2019amass} with the upstream gains, six-layer actor and critic (2048, 1536, 1024, 1024, 512 and 512 units), implicit damping, termination rule, joint-power penalty and 30\,Hz policy. To avoid downloading all of AMASS, the port trains on three of its subsets, CMU, ACCAD and BMLrub (4{,}021 motions after PHC's filter, about a third of its 11{,}313 training sequences), with a single primitive for $2 \times 10^9$ steps, and is scored with PHC's metrics on its 140 test motions, which come from the Transitions and SSM subsets (Table~\ref{tab:phc_amass}). The port imitates $92.2\,\%$ of the test motions, against $96.4\,\%$ for the multi-primitive PHC trained on all sequences.

For a comparison under a shared objective, we also retrain PHC with our full reward on every split, replacing only its imitation kernels and keeping the discriminator at its original weight, so each step earns $0.5\,r_{\mathrm{ours}} + 0.5\,r_{\mathrm{disc}}$ with all else as above. On seen subjects it holds the survival of the port ($93.5\,\%$), lowers its joint-angle error from $11.4$ to $8.7^\circ$ and raises its vertical-force correlation from $.61$ to $.71$, and Table~\ref{tab:existing} reports it next to the own-reward rows of Tables~\ref{tab:existing_full} and~\ref{tab:existing_transfer}.

\begin{table}[!h]%
\vspace{-5pt}%
\centering
\caption{\textbf{The PHC port approaches the reported PHC on AMASS.} SMPL humanoid with the upstream gains, network, damping and termination rule, scored with the upstream metrics on the 140 test motions of PHC. PHC trains on all 11{,}313 training sequences and the port on the 4{,}021 that pass its filter in CMU, ACCAD and BMLrub. Mean$_{\pm\text{sd}}$ over 3 seeds after $2 \times 10^9$ steps.}
\label{tab:phc_amass}
\vspace{-0.4em}
\renewcommand{\arraystretch}{0.9}
\setlength{\tabcolsep}{4pt}
\newcommand{\sd}[1]{{\tiny\textcolor{black!55}{$\pm$#1}}}
\scriptsize
\begin{tabular}{@{}lcc ccc ccc@{}}
\toprule
 & & & \multicolumn{3}{c}{\textbf{AMASS train}} & \multicolumn{3}{c}{\textbf{AMASS test}} \\
\cmidrule(lr){4-6}\cmidrule(lr){7-9}
 & Primitives & Motions & Succ.$\hi$ (\%) & G-MPJPE$\lo$ & MPJPE$\lo$ & Succ.$\hi$ (\%) & G-MPJPE$\lo$ & MPJPE$\lo$ \\
\midrule
PHC, reported~\citep{luo2023perpetual} & multiple & 11{,}313 & 98.9 & 37.5 & 26.9 & 96.4 & 47.4 & 30.9 \\
Our port & 1 & 4{,}021 & 96.2\sd{.1} & 49.3\sd{.3} & 39.0\sd{.1} & 92.2\sd{1.9} & 60.9\sd{.6} & 42.7\sd{.3} \\
\bottomrule
\end{tabular}
\par\vspace{1pt}
\parbox{\linewidth}{\scriptsize\raggedright Succ. = share of motions imitated without the mean body-position error exceeding $0.5$\,m. G-MPJPE / MPJPE = global / root-relative joint-position error (mm), over successfully imitated motions as in the upstream evaluation code.}
\end{table}

\vspace{-12.5pt}%
\paragraph{SimPoE~\citep{yuan2021simpoe}.}
SimPoE learns scales of the PD gains for every PD step together with residual PD targets (meta-PD control), which makes it the prior controller closest to ours. It has no public code, and we re-implement it from the paper, completing unspecified details from the controllers that precede it in the same code lineage, RFC~\citep{yuan2020residual} and UHC (\href{https://github.com/ZhengyiLuo/UHC}{github.com/ZhengyiLuo/UHC}). The policy acts every 15 physics steps (33\,Hz) and observes the joint angles, velocities and 3D joint positions in the heading frame together with their differences to the next reference frame, in place of SimPoE's kinematic estimate from video. Its action comprises residual targets on the next reference pose for the 15 actuated joints, one pair of gain scales $\lambda^p$, $\lambda^d$ for each of the 15 PD steps of a decision, shared by all joints as in UHC, and a residual wrench on the six root degrees of freedom, scaled by 500 as in the original. The scales multiply the nominal gains of our gain table, and the PD law runs as stable PD~\citep{tan2011stable} with damping toward zero velocity, as in RFC and UHC. The network (hidden layers of 2048 and 1024 units, fixed action noise of $0.05$), the PPO settings (discount $0.95$, learning rates of $5 \times 10^{-5}$ for the actor and $3 \times 10^{-4}$ for the critic, 10 updates per epoch) and the 200-step episodes, terminated when the root drops $0.5$\,m below the reference, follow the paper. SimPoE trains on our reward, including the force terms that supervise \projname{}, for $2 \times 10^9$ steps, twenty times its original budget, and is scored under our protocol (Tables~\ref{tab:existing}, \ref{tab:existing_full} and~\ref{tab:existing_transfer}). Its residual root wrench supports the body without passing through the contacts that the force metrics score. The SimPoE row of Table~\ref{tab:existing} keeps it, and the rows marked \emph{no root force} in Tables~\ref{tab:existing_full} and~\ref{tab:existing_transfer} remove it, as in SimPoE's own ablation.

\vspace{-6.5pt}%
\paragraph{Stable and implicit PD.}
Stable PD~\citep{tan2011stable} evaluates the PD law at the next state through the mass matrix, and implicit damping~\citep{todorov2012mujoco} integrates the damping force implicitly. With the setpoint $\bar{\vect{q}} = \vect{q}^{*}_{t+1} + \alpha\,\delta\vect{q}$ of Eq.~\ref{eq:pd_torque}, stable PD (S) and implicit damping (I) apply
\begin{equation}
  \hat{\vect{\tau}}_{\mathrm{S}} = \vect{k}^{p} \circ \bigl(\bar{\vect{q}} - \vect{q} - h\dot{\vect{q}}\bigr) + \vect{k}^{d} \circ \bigl(\dot{\vect{q}}^{*}_{t+1} - \dot{\vect{q}} - h\ddot{\vect{q}}\bigr),
  \quad
  \hat{\vect{\tau}}_{\mathrm{I}} = \vect{k}^{p} \circ \bigl(\bar{\vect{q}} - \vect{q}\bigr) + \vect{k}^{d} \circ \bigl(\dot{\vect{q}}^{*}_{t+1} - \dot{\vect{q}} - h\ddot{\vect{q}}\bigr),
  \label{eq:stable_implicit}
\end{equation}
where $\ddot{\vect{q}}$ is the acceleration of the physics step, solved from Eq.~\ref{eq:eom} with $\mat{M} + h\,\mat{S}^{\top}\mathrm{diag}(\vect{k}^{d})\,\mat{S}$ in place of $\mat{M}$. For stable PD we solve this system with the mass-matrix factorization of MuJoCo Warp at every physics step, and for implicit damping we write $\vect{k}^{d}$ before every control step into the joint damping, which the Euler integrator of MuJoCo treats implicitly. Each scheme remains stable at any damping gain and therefore lifts Bound~2, whereas Bound~1 still constrains fixed gains (App.~\ref{App:humanoid_bounds}). Fixed-PD and \projname{} are retrained under each scheme with data, reward, network and PPO settings unchanged (Tables~\ref{tab:existing}, \ref{tab:existing_full} and~\ref{tab:existing_transfer}). Under these schemes, Fixed-PD holds its nominal gains without the damping cap of $I_{jj}/h$ that explicit PD requires. The $2\times$ rows double the stiffness priors of either method and, under stable PD and implicit damping, also the damping priors, whereas under explicit PD \projname{} doubles its stiffness priors only, with the damping priors nominal, matched to the $2\times$ row of Fixed-PD, whose damping stays capped. PHC runs under implicit damping and SimPoE under stable PD, as in their original implementations, and both damp toward zero velocity instead of the reference velocity.

\begin{table}[p]%
\centering
\caption{%
  \textbf{Full metrics for seen subjects, held-out clips.}
  Columns are RMSE ($\downarrow$) unless marked $\uparrow$.
  Cells report mean$_{\pm\text{sd}}$ over 5 seeds.
  \textbf{Bold} / \underline{underline} = best / second best per activity.
  }
\vspace{-0.3em}
\label{tab:golden_full}
\renewcommand{\arraystretch}{0.86}
\setlength{\tabcolsep}{2pt}
\newcommand{\gap}{\hspace{4pt}}
\newcommand{\sd}[1]{{\tiny\textcolor{black!55}{$\pm$#1}}}
\newcommand{\gsd}[1]{{\tiny\textcolor{black!55}{\textit{$\pm$#1}}}}
\newcommand{\gd}[1]{\textit{#1}}
\scriptsize
\resizebox{\textwidth}{!}{%
\begin{tabular}{@{}>{\fillL}l @{\gap} l @{\gap} l @{\gap} llll @{\gap} lp{0.75cm}p{0.75cm}p{0.75cm}ll @{\gap} ll>{\fillR}l @{}}
\toprule
 & & \multicolumn{1}{c}{\textbf{Stability}} & \multicolumn{4}{c}{\textbf{Kinematics}} & \multicolumn{6}{c}{\textbf{Ground Reaction Forces}} & \multicolumn{3}{c}{\textbf{Torques}$^\dagger$} \\
\cmidrule(lr){3-3}\cmidrule(lr){4-7}\cmidrule(lr){8-13}\cmidrule(lr){14-16}
\texttt{\textbf{Activity}} & \textbf{Method}
  & \makecell{Ep.$\hi$ \\ (\%)}
  & \makecell{$q\lo$\\ \makebox[0pt][c]{($^\circ$)}}
  & \makecell{$\dot{q}\lo$\\ \makebox[0pt][c]{($^\circ$/s)}}
  & \makecell{$p_x\lo$\\ \makebox[0pt][c]{(m/s)}}
  & kin$_r\hi$
  & \makecell{$F_{\mathrm{Z}}\lo$\\ \makebox[0pt][c]{(BW)}}
  & $F^{\uparrow}_{\mathrm{Z}_r}$
  & $F^{\uparrow}_{\mathrm{ML}_r}$
  & $F^{\uparrow}_{\mathrm{AP}_r}$
  & \makecell{$\mathrm{CoP}^{\downarrow}_{\mathrm{ML}}$\\ {\tiny (mm)}}
  & \makecell{$\mathrm{CoP}^{\downarrow}_{\mathrm{AP}}$\\ {\tiny (mm)}}
  & \makecell{$\tau\lo$\\ \makebox[0pt][c]{\tiny (Nm/kg)}}
  & $\tau_r\hi$
  & \makecell{$\dot{\tau}\lo$\\ \makebox[0pt][c]{\tiny (Nm/kg/s)}} \\
\midrule
\multirow{5}{*}{\act{walk}} & \gd{GaitDynamics}$^\ast$ & -- & -- & -- & -- & -- & \gd{.04}\gsd{.00} & \gd{1.00}\gsd{.00} & \gd{.99}\gsd{.00} & \gd{.99}\gsd{.00} & \gd{7}\gsd{0} & \gd{14}\gsd{0} & -- & -- & -- \\
 & Fixed-PD & \underline{99.9}\sd{.1} & 2.48\sd{.04} & 57.2\sd{1.9} & \underline{.022}\sd{.002} & \underline{.97}\sd{.00} & .13\sd{.00} & .95\sd{.00} & .77\sd{.01} & .72\sd{.00} & \underline{20}\sd{1} & 45\sd{1} & .55\sd{.01} & .50\sd{.01} & 26.3\sd{2.0} \\
 & Torque & 99.7\sd{.2} & \underline{2.33}\sd{.03} & 55.7\sd{2.5} & \underline{.022}\sd{.001} & .93\sd{.00} & \textbf{.09}\sd{.00} & \textbf{.98}\sd{.00} & \textbf{.89}\sd{.00} & .85\sd{.00} & \textbf{19}\sd{0} & \underline{41}\sd{1} & .42\sd{.00} & \underline{.61}\sd{.01} & 9.06\sd{.22} \\
 & Muscle & 98.6\sd{.4} & 2.72\sd{.08} & \underline{41.1}\sd{.6} & .032\sd{.001} & .93\sd{.00} & .10\sd{.00} & .97\sd{.00} & .86\sd{.01} & \textbf{.86}\sd{.01} & \underline{20}\sd{1} & 51\sd{2} & \underline{.41}\sd{.00} & .60\sd{.01} & \textbf{6.12}\sd{.13} \\
 & ImpedanceMimic & \textbf{100}\sd{0} & \textbf{1.66}\sd{.03} & \textbf{28.8}\sd{.4} & \textbf{.013}\sd{.001} & \textbf{.98}\sd{.00} & \textbf{.09}\sd{.00} & \textbf{.98}\sd{.00} & \textbf{.89}\sd{.00} & \textbf{.86}\sd{.00} & \underline{20}\sd{0} & \textbf{40}\sd{1} & \textbf{.38}\sd{.00} & \textbf{.65}\sd{.00} & \underline{6.38}\sd{.08} \\
\addlinespace[4pt]
\multirow{5}{*}{\act{run}} & \gd{GaitDynamics}$^\ast$ & -- & -- & -- & -- & -- & \gd{.04}\gsd{.01} & \gd{1.00}\gsd{.00} & \gd{.96}\gsd{.00} & \gd{.99}\gsd{.00} & \gd{5}\gsd{0} & \gd{11}\gsd{0} & -- & -- & -- \\
 & Fixed-PD & 99.7\sd{.2} & \underline{3.36}\sd{.02} & 109\sd{4} & \underline{.064}\sd{.005} & \underline{.96}\sd{.00} & .21\sd{.00} & .96\sd{.00} & .32\sd{.01} & .52\sd{.01} & 22\sd{1} & 43\sd{1} & 1.25\sd{.01} & .47\sd{.01} & 110\sd{2} \\
 & Torque & \underline{99.8}\sd{.3} & 3.51\sd{.02} & 114\sd{3} & .075\sd{.005} & .93\sd{.00} & .23\sd{.00} & .96\sd{.00} & .40\sd{.03} & .58\sd{.01} & \underline{21}\sd{0} & 49\sd{1} & .86\sd{.00} & .66\sd{.00} & 19.8\sd{.6} \\
 & Muscle & 99.7\sd{.2} & 3.84\sd{.06} & \underline{75.5}\sd{.3} & \underline{.064}\sd{.002} & .93\sd{.00} & \underline{.15}\sd{.00} & \underline{.98}\sd{.00} & \underline{.42}\sd{.07} & \underline{.80}\sd{.01} & \textbf{20}\sd{0} & \textbf{37}\sd{2} & \textbf{.78}\sd{.01} & \textbf{.72}\sd{.01} & \textbf{13.0}\sd{.1} \\
 & ImpedanceMimic & \textbf{100}\sd{0} & \textbf{2.24}\sd{.02} & \textbf{59.7}\sd{1.3} & \textbf{.043}\sd{.003} & \textbf{.98}\sd{.00} & \textbf{.12}\sd{.00} & \textbf{.99}\sd{.00} & \textbf{.58}\sd{.03} & \textbf{.86}\sd{.00} & \underline{21}\sd{0} & \underline{42}\sd{0} & \underline{.80}\sd{.00} & \textbf{.72}\sd{.01} & \underline{19.7}\sd{1.9} \\
\addlinespace[4pt]
\multirow{5}{*}{\act{stand}} & \gd{GaitDynamics}$^\ast$ & -- & -- & -- & -- & -- & \gd{.03}\gsd{.01} & \gd{.75}\gsd{.06} & \gd{.88}\gsd{.02} & \gd{.52}\gsd{.06} & \gd{4}\gsd{1} & \gd{8}\gsd{0} & -- & -- & -- \\
 & Fixed-PD & 97.8\sd{.1} & \underline{1.34}\sd{.01} & 10.0\sd{.4} & .005\sd{.001} & \underline{.63}\sd{.01} & .10\sd{.00} & .36\sd{.02} & .48\sd{.06} & .20\sd{.01} & 14\sd{1} & 50\sd{1} & \underline{.56}\sd{.01} & \underline{.13}\sd{.01} & 6.40\sd{.53} \\
 & Torque & \underline{99.6}\sd{.3} & 2.33\sd{.03} & 14.3\sd{1.2} & \underline{.004}\sd{.001} & .37\sd{.02} & \textbf{.05}\sd{.00} & \textbf{.64}\sd{.04} & .55\sd{.04} & \textbf{.26}\sd{.04} & \textbf{13}\sd{1} & 52\sd{4} & \textbf{.35}\sd{.01} & .11\sd{.03} & 1.64\sd{.21} \\
 & Muscle & 93.2\sd{.9} & 2.14\sd{.17} & \underline{8.14}\sd{.49} & .012\sd{.001} & .47\sd{.02} & .07\sd{.00} & .52\sd{.03} & \underline{.60}\sd{.11} & \underline{.21}\sd{.02} & 15\sd{2} & \underline{47}\sd{3} & .78\sd{.08} & .08\sd{.01} & \textbf{.56}\sd{.05} \\
 & ImpedanceMimic & \textbf{99.7}\sd{.2} & \textbf{1.07}\sd{.00} & \textbf{4.28}\sd{.13} & \textbf{.003}\sd{.001} & \textbf{.66}\sd{.01} & \underline{.06}\sd{.00} & \underline{.56}\sd{.02} & \textbf{.80}\sd{.03} & .20\sd{.03} & \textbf{13}\sd{0} & \textbf{32}\sd{1} & .84\sd{.01} & \textbf{.15}\sd{.00} & \underline{.92}\sd{.08} \\
\addlinespace[4pt]
\multirow{5}{*}{\act{lateral}} & \gd{GaitDynamics}$^\ast$ & -- & -- & -- & -- & -- & \gd{.21}\gsd{.01} & \gd{.90}\gsd{.01} & \gd{.86}\gsd{.02} & \gd{.86}\gsd{.01} & \gd{29}\gsd{1} & \gd{31}\gsd{1} & -- & -- & -- \\
 & Fixed-PD & 95.9\sd{.5} & \underline{3.77}\sd{.09} & \underline{71.7}\sd{.8} & \underline{.087}\sd{.003} & \underline{.91}\sd{.00} & .30\sd{.01} & .85\sd{.01} & .72\sd{.00} & .61\sd{.01} & 30\sd{1} & \textbf{53}\sd{3} & .71\sd{.01} & .50\sd{.01} & 43.3\sd{1.7} \\
 & Torque & \underline{97.6}\sd{1.3} & 6.49\sd{.22} & 149\sd{7} & .102\sd{.008} & .77\sd{.01} & .25\sd{.01} & \underline{.89}\sd{.01} & .74\sd{.01} & \underline{.65}\sd{.01} & \textbf{28}\sd{1} & \textbf{53}\sd{2} & .62\sd{.01} & \underline{.51}\sd{.01} & 28.5\sd{.3} \\
 & Muscle & 92.4\sd{1.2} & 8.91\sd{.42} & 82.7\sd{2.0} & .114\sd{.006} & .67\sd{.02} & \underline{.24}\sd{.01} & \underline{.89}\sd{.01} & \underline{.76}\sd{.02} & .58\sd{.02} & \underline{29}\sd{1} & 61\sd{4} & \textbf{.56}\sd{.02} & .47\sd{.02} & \textbf{10.6}\sd{.4} \\
 & ImpedanceMimic & \textbf{99.9}\sd{.1} & \textbf{3.11}\sd{.07} & \textbf{59.6}\sd{1.4} & \textbf{.084}\sd{.004} & \textbf{.94}\sd{.00} & \textbf{.23}\sd{.00} & \textbf{.91}\sd{.01} & \textbf{.80}\sd{.01} & \textbf{.71}\sd{.02} & 31\sd{2} & \textbf{53}\sd{4} & \textbf{.56}\sd{.02} & \textbf{.57}\sd{.02} & \underline{21.2}\sd{1.6} \\
\addlinespace[4pt]
\multirow{5}{*}{\act{flexion}} & \gd{GaitDynamics}$^\ast$ & -- & -- & -- & -- & -- & \gd{.05}\gsd{.00} & \gd{.92}\gsd{.01} & \gd{.95}\gsd{.00} & \gd{.83}\gsd{.01} & \gd{9}\gsd{0} & \gd{20}\gsd{0} & -- & -- & -- \\
 & Fixed-PD & 91.3\sd{.7} & \underline{2.76}\sd{.05} & \underline{42.7}\sd{1.1} & .034\sd{.002} & \underline{.92}\sd{.00} & .21\sd{.00} & .63\sd{.01} & .48\sd{.01} & .47\sd{.01} & 24\sd{1} & 63\sd{1} & .49\sd{.01} & .45\sd{.01} & 19.2\sd{1.4} \\
 & Torque & \underline{96.8}\sd{.8} & 4.62\sd{.09} & 112\sd{4} & \underline{.032}\sd{.003} & .81\sd{.01} & \textbf{.15}\sd{.00} & \textbf{.73}\sd{.01} & \underline{.58}\sd{.01} & \textbf{.54}\sd{.01} & \textbf{23}\sd{1} & \underline{61}\sd{1} & \textbf{.42}\sd{.00} & \underline{.48}\sd{.01} & 14.1\sd{.6} \\
 & Muscle & 68.2\sd{.9} & 12.3\sd{.5} & 52.0\sd{1.9} & .071\sd{.001} & .58\sd{.02} & .20\sd{.01} & .62\sd{.01} & \textbf{.59}\sd{.02} & \underline{.52}\sd{.01} & 24\sd{2} & 73\sd{2} & .46\sd{.01} & .38\sd{.02} & \textbf{3.66}\sd{.15} \\
 & ImpedanceMimic & \textbf{97.7}\sd{1.3} & \textbf{2.40}\sd{.04} & \textbf{42.3}\sd{1.1} & \textbf{.031}\sd{.003} & \textbf{.93}\sd{.00} & \underline{.18}\sd{.00} & \underline{.70}\sd{.01} & \underline{.58}\sd{.01} & \underline{.52}\sd{.01} & \textbf{23}\sd{0} & \textbf{60}\sd{1} & \underline{.44}\sd{.00} & \textbf{.49}\sd{.01} & \underline{11.4}\sd{.5} \\
\addlinespace[4pt]
\multirow{5}{*}{\act{jump}} & \gd{GaitDynamics}$^\ast$ & -- & -- & -- & -- & -- & \gd{.13}\gsd{.01} & \gd{.90}\gsd{.02} & \gd{.86}\gsd{.01} & \gd{.75}\gsd{.01} & \gd{34}\gsd{2} & \gd{51}\gsd{1} & -- & -- & -- \\
 & Fixed-PD & 73.6\sd{1.5} & \underline{3.76}\sd{.13} & \underline{77.9}\sd{2.3} & \underline{.072}\sd{.005} & \underline{.89}\sd{.00} & .37\sd{.01} & .64\sd{.01} & .43\sd{.01} & .31\sd{.02} & \underline{28}\sd{3} & \underline{72}\sd{2} & .89\sd{.02} & .29\sd{.01} & 67.1\sd{2.0} \\
 & Torque & \underline{78.9}\sd{1.0} & 7.61\sd{.47} & 171\sd{14} & .075\sd{.005} & .72\sd{.01} & \textbf{.28}\sd{.01} & \textbf{.74}\sd{.01} & \textbf{.55}\sd{.02} & \underline{.38}\sd{.01} & 32\sd{7} & 73\sd{9} & \underline{.57}\sd{.02} & \textbf{.41}\sd{.01} & \underline{25.8}\sd{1.2} \\
 & Muscle & 56.4\sd{2.1} & 13.2\sd{.7} & 87.9\sd{2.5} & .104\sd{.003} & .53\sd{.02} & \underline{.33}\sd{.01} & .62\sd{.01} & .52\sd{.04} & .31\sd{.03} & 35\sd{4} & 83\sd{3} & \textbf{.56}\sd{.02} & .34\sd{.02} & \textbf{7.40}\sd{.62} \\
 & ImpedanceMimic & \textbf{83.8}\sd{2.1} & \textbf{3.22}\sd{.16} & \textbf{69.7}\sd{2.8} & \textbf{.063}\sd{.003} & \textbf{.92}\sd{.00} & \underline{.33}\sd{.01} & \underline{.69}\sd{.01} & \textbf{.55}\sd{.03} & \textbf{.40}\sd{.02} & \textbf{27}\sd{2} & \textbf{68}\sd{3} & .69\sd{.02} & \underline{.38}\sd{.01} & 36.0\sd{2.7} \\
\specialrule{\cmidrulewidth}{\aboverulesep}{0pt}
\rowcolor{meangray}  & \padT \gd{GaitDynamics}$^\ast$ & -- & -- & -- & -- & -- & \gd{.08}\gsd{.00} & \gd{.91}\gsd{.01} & \gd{.92}\gsd{.01} & \gd{.82}\gsd{.01} & \gd{15}\gsd{1} & \gd{23}\gsd{0} & -- & -- & -- \\
\rowcolor{meangray}  & Fixed-PD & 93.0\sd{.2} & \underline{2.91}\sd{.03} & 61.4\sd{.9} & \underline{.047}\sd{.002} & \underline{.88}\sd{.00} & .22\sd{.00} & .73\sd{.00} & .53\sd{.01} & .47\sd{.01} & \textbf{23}\sd{1} & \underline{54}\sd{1} & .74\sd{.00} & .39\sd{.00} & 45.3\sd{.6} \\
\rowcolor{meangray}  & Torque & \underline{95.4}\sd{.4} & 4.48\sd{.09} & 103\sd{4} & .052\sd{.002} & .76\sd{.00} & \textbf{.17}\sd{.00} & \textbf{.82}\sd{.01} & \underline{.62}\sd{.01} & .54\sd{.01} & \textbf{23}\sd{1} & 55\sd{2} & \textbf{.54}\sd{.00} & \underline{.46}\sd{.00} & 16.5\sd{.4} \\
\rowcolor{meangray}  & Muscle & 84.8\sd{.5} & 7.18\sd{.26} & \underline{57.9}\sd{.8} & .066\sd{.001} & .69\sd{.01} & .18\sd{.00} & .77\sd{.00} & \underline{.62}\sd{.02} & \underline{.55}\sd{.01} & 24\sd{1} & 59\sd{0} & \underline{.59}\sd{.01} & .43\sd{.01} & \textbf{6.89}\sd{.15} \\
\rowcolor{meangray} \multirow{-6}{*}{\textbf{Mean}} & \padB ImpedanceMimic & \textbf{96.9}\sd{.5} & \textbf{2.28}\sd{.04} & \textbf{44.1}\sd{.2} & \textbf{.040}\sd{.001} & \textbf{.90}\sd{.00} & \textbf{.17}\sd{.00} & \underline{.80}\sd{.00} & \textbf{.70}\sd{.01} & \textbf{.59}\sd{.01} & \textbf{23}\sd{1} & \textbf{49}\sd{1} & .62\sd{.00} & \textbf{.49}\sd{.00} & \underline{15.9}\sd{.4} \\

\specialrule{\heavyrulewidth}{0pt}{\belowbottomsep}
\end{tabular}%
}
\vspace{0pt}
\parbox{\textwidth}{\scriptsize\raggedright
$^\ast$GaitDynamics is shown in italics for reference and excluded from the ranking, since it is not a controller and serves as a reference for GRF.
  $^\dagger$9 main joints (hip flexion / adduction, knee, ankle, both sides; lumbar extension), against inverse dynamics.
  Ep. = episode survival before termination (\%). $q$, $\dot q$ = joint-angle / velocity RMSE. $p_x$ = root forward-position drift rate. \{kin$_r$, $F_{\mathrm{Z}_r}$, $F_{\mathrm{AP}_r}$, $F_{\mathrm{ML}_r}$, $\tau_r$\} = Pearson correlation w.r.t. reference \{joint angles, vertical, AP, ML GRF, torques\}. CoP = center of pressure. $\dot\tau$ = torque jerk.}
\vspace{-1em}
\end{table}
\begin{table}[p]%
\centering
\caption{%
  \textbf{Full metrics for unseen subjects, zero-shot.}
  Columns are RMSE ($\downarrow$) unless marked $\uparrow$.
  Cells report mean$_{\pm\text{sd}}$ over 5 seeds.
  \textbf{Bold} / \underline{underline} = best / second best per activity.
  }
\vspace{-0.3em}
\label{tab:golden_unseen}
\renewcommand{\arraystretch}{0.86}
\setlength{\tabcolsep}{2pt}
\newcommand{\gap}{\hspace{4pt}}
\newcommand{\sd}[1]{{\tiny\textcolor{black!55}{$\pm$#1}}}
\newcommand{\gsd}[1]{{\tiny\textcolor{black!55}{\textit{$\pm$#1}}}}
\newcommand{\gd}[1]{\textit{#1}}
\scriptsize
\resizebox{\textwidth}{!}{%
\begin{tabular}{@{}>{\fillL}l @{\gap} l @{\gap} l @{\gap} llll @{\gap} lp{0.75cm}p{0.75cm}p{0.75cm}ll @{\gap} ll>{\fillR}l @{}}
\toprule
 & & \multicolumn{1}{c}{\textbf{Stability}} & \multicolumn{4}{c}{\textbf{Kinematics}} & \multicolumn{6}{c}{\textbf{Ground Reaction Forces}} & \multicolumn{3}{c}{\textbf{Torques}$^\dagger$} \\
\cmidrule(lr){3-3}\cmidrule(lr){4-7}\cmidrule(lr){8-13}\cmidrule(lr){14-16}
\texttt{\textbf{Activity}} & \textbf{Method}
  & \makecell{Ep.$\hi$ \\ (\%)}
  & \makecell{$q\lo$\\ \makebox[0pt][c]{($^\circ$)}}
  & \makecell{$\dot{q}\lo$\\ \makebox[0pt][c]{($^\circ$/s)}}
  & \makecell{$p_x\lo$\\ \makebox[0pt][c]{(m/s)}}
  & kin$_r\hi$
  & \makecell{$F_{\mathrm{Z}}\lo$\\ \makebox[0pt][c]{(BW)}}
  & $F^{\uparrow}_{\mathrm{Z}_r}$
  & $F^{\uparrow}_{\mathrm{ML}_r}$
  & $F^{\uparrow}_{\mathrm{AP}_r}$
  & \makecell{\makebox[0pt][c]{$\mathrm{CoP}^{\downarrow}_{\mathrm{ML}}$}\\ \makebox[0pt][c]{\tiny (mm)}}
  & \makecell{\makebox[0pt][c]{$\mathrm{CoP}^{\downarrow}_{\mathrm{AP}}$}\\ \makebox[0pt][c]{\tiny (mm)}}
  & \makecell{$\tau\lo$\\ \makebox[0pt][c]{\tiny (Nm/kg)}}
  & $\tau_r\hi$
  & \makecell{$\dot{\tau}\lo$\\ \makebox[0pt][c]{\tiny (Nm/kg/s)}} \\
\midrule
\multirow{5}{*}{\act{walk}} & \gd{GaitDynamics}$^\ast$ & -- & -- & -- & -- & -- & \gd{.06}\gsd{.00} & \gd{.99}\gsd{.00} & \gd{.98}\gsd{.00} & \gd{.97}\gsd{.00} & \gd{11}\gsd{0} & \gd{18}\gsd{0} & -- & -- & -- \\
 & Fixed-PD & \underline{99.8}\sd{.2} & 2.75\sd{.05} & 60.5\sd{.4} & .027\sd{.003} & \underline{.96}\sd{.00} & .16\sd{.00} & .94\sd{.00} & .71\sd{.01} & .72\sd{.01} & 24\sd{1} & 41\sd{1} & .55\sd{.01} & .50\sd{.01} & 28.0\sd{.5} \\
 & Torque & \underline{99.8}\sd{.3} & \underline{2.69}\sd{.04} & 64.1\sd{3.5} & \underline{.025}\sd{.002} & .93\sd{.00} & \textbf{.09}\sd{.00} & \textbf{.98}\sd{.00} & \textbf{.88}\sd{.00} & \textbf{.86}\sd{.00} & \textbf{22}\sd{1} & \textbf{37}\sd{1} & \underline{.41}\sd{.00} & \underline{.62}\sd{.01} & 9.54\sd{.31} \\
 & Muscle & 90.9\sd{7.9} & 3.98\sd{.24} & \underline{50.4}\sd{1.6} & .044\sd{.005} & .90\sd{.01} & .13\sd{.01} & .95\sd{.01} & .83\sd{.02} & .83\sd{.01} & \textbf{22}\sd{1} & 48\sd{1} & \underline{.41}\sd{.00} & .60\sd{.01} & \textbf{6.30}\sd{.07} \\
 & ImpedanceMimic & \textbf{100}\sd{0} & \textbf{1.95}\sd{.03} & \textbf{32.7}\sd{.3} & \textbf{.016}\sd{.001} & \textbf{.98}\sd{.00} & \underline{.11}\sd{.00} & \underline{.97}\sd{.00} & \textbf{.88}\sd{.01} & \underline{.85}\sd{.01} & 23\sd{1} & \underline{39}\sd{1} & \textbf{.38}\sd{.01} & \textbf{.65}\sd{.01} & \underline{6.78}\sd{.11} \\
\addlinespace[4pt]
\multirow{5}{*}{\act{run}} & \gd{GaitDynamics}$^\ast$ & -- & -- & -- & -- & -- & \gd{.07}\gsd{.00} & \gd{1.00}\gsd{.00} & \gd{.96}\gsd{.00} & \gd{.98}\gsd{.00} & \gd{7}\gsd{0} & \gd{15}\gsd{1} & -- & -- & -- \\
 & Fixed-PD & \underline{98.6}\sd{.1} & \underline{3.31}\sd{.07} & 112\sd{5} & \underline{.058}\sd{.007} & \underline{.96}\sd{.00} & .21\sd{.01} & .96\sd{.00} & .20\sd{.03} & .54\sd{.04} & 22\sd{1} & \underline{41}\sd{1} & 1.28\sd{.02} & .48\sd{.00} & 96.8\sd{3.3} \\
 & Torque & 94.2\sd{3.7} & 3.61\sd{.09} & 121\sd{2} & .073\sd{.010} & .92\sd{.01} & .20\sd{.01} & .96\sd{.00} & .30\sd{.03} & .64\sd{.01} & \textbf{21}\sd{0} & 43\sd{1} & .97\sd{.01} & \underline{.67}\sd{.00} & \underline{20.1}\sd{.4} \\
 & Muscle & 97.0\sd{2.6} & 4.19\sd{.14} & \underline{83.5}\sd{1.4} & .077\sd{.004} & .91\sd{.01} & \underline{.16}\sd{.01} & \textbf{.98}\sd{.00} & \underline{.44}\sd{.06} & \underline{.82}\sd{.03} & \textbf{21}\sd{1} & \textbf{37}\sd{3} & \textbf{.95}\sd{.02} & \underline{.67}\sd{.02} & \textbf{13.0}\sd{.3} \\
 & ImpedanceMimic & \textbf{100}\sd{0} & \textbf{2.24}\sd{.04} & \textbf{63.2}\sd{1.6} & \textbf{.052}\sd{.006} & \textbf{.98}\sd{.00} & \textbf{.14}\sd{.01} & \textbf{.98}\sd{.00} & \textbf{.47}\sd{.03} & \textbf{.89}\sd{.01} & \textbf{21}\sd{1} & \underline{41}\sd{1} & \textbf{.95}\sd{.01} & \textbf{.69}\sd{.01} & 20.6\sd{2.0} \\
\addlinespace[4pt]
\multirow{5}{*}{\act{stand}} & \gd{GaitDynamics}$^\ast$ & -- & -- & -- & -- & -- & \gd{.04}\gsd{.00} & \gd{.65}\gsd{.04} & \gd{.95}\gsd{.01} & \gd{.36}\gsd{.05} & \gd{6}\gsd{0} & \gd{15}\gsd{1} & -- & -- & -- \\
 & Fixed-PD & 99.1\sd{.2} & \underline{1.35}\sd{.03} & \underline{6.88}\sd{.86} & \underline{.004}\sd{.001} & \textbf{.66}\sd{.02} & .08\sd{.01} & .49\sd{.05} & .54\sd{.08} & .16\sd{.03} & \underline{14}\sd{1} & \underline{41}\sd{1} & \underline{.54}\sd{.01} & \underline{.17}\sd{.01} & 4.30\sd{.97} \\
 & Torque & \textbf{99.8}\sd{.1} & 2.24\sd{.08} & 16.2\sd{1.4} & \underline{.004}\sd{.001} & .37\sd{.02} & \textbf{.05}\sd{.00} & \textbf{.64}\sd{.02} & .49\sd{.16} & .17\sd{.10} & \underline{14}\sd{1} & 48\sd{5} & \textbf{.35}\sd{.02} & .12\sd{.02} & 2.16\sd{.23} \\
 & Muscle & 92.0\sd{2.8} & 2.30\sd{.26} & 8.30\sd{1.19} & .009\sd{.002} & .42\sd{.05} & .07\sd{.01} & .52\sd{.06} & \underline{.75}\sd{.07} & \textbf{.18}\sd{.06} & 15\sd{1} & 45\sd{2} & .88\sd{.06} & .10\sd{.01} & \textbf{.64}\sd{.05} \\
 & ImpedanceMimic & \underline{99.6}\sd{.2} & \textbf{.98}\sd{.02} & \textbf{3.70}\sd{.16} & \textbf{.003}\sd{.000} & \textbf{.66}\sd{.01} & \underline{.06}\sd{.00} & \underline{.54}\sd{.03} & \textbf{.88}\sd{.03} & \textbf{.18}\sd{.04} & \textbf{13}\sd{0} & \textbf{30}\sd{3} & .89\sd{.01} & \textbf{.18}\sd{.01} & \underline{1.14}\sd{.15} \\
\addlinespace[4pt]
\multirow{5}{*}{\act{lateral}} & \gd{GaitDynamics}$^\ast$ & -- & -- & -- & -- & -- & \gd{.23}\gsd{.01} & \gd{.89}\gsd{.02} & \gd{.86}\gsd{.01} & \gd{.84}\gsd{.02} & \gd{33}\gsd{2} & \gd{41}\gsd{3} & -- & -- & -- \\
 & Fixed-PD & \underline{93.7}\sd{.7} & \underline{4.02}\sd{.08} & \underline{84.2}\sd{2.9} & \textbf{.114}\sd{.005} & \underline{.90}\sd{.00} & .38\sd{.02} & .78\sd{.02} & .61\sd{.02} & \underline{.56}\sd{.01} & 36\sd{6} & \textbf{64}\sd{7} & .93\sd{.03} & \underline{.42}\sd{.02} & 69.6\sd{4.8} \\
 & Torque & 71.4\sd{5.2} & 8.77\sd{.25} & 232\sd{20} & .154\sd{.008} & .68\sd{.02} & .37\sd{.01} & .79\sd{.01} & .59\sd{.03} & .54\sd{.02} & 36\sd{5} & 71\sd{8} & .90\sd{.03} & .38\sd{.02} & 52.9\sd{2.9} \\
 & Muscle & 79.3\sd{7.7} & 12.9\sd{.7} & 104\sd{4} & .133\sd{.008} & .53\sd{.03} & \textbf{.31}\sd{.01} & \underline{.83}\sd{.02} & \underline{.67}\sd{.03} & .44\sd{.04} & \textbf{31}\sd{1} & 67\sd{3} & \textbf{.67}\sd{.01} & .41\sd{.01} & \textbf{9.54}\sd{.21} \\
 & ImpedanceMimic & \textbf{99.9}\sd{.2} & \textbf{3.62}\sd{.07} & \textbf{73.8}\sd{2.7} & \underline{.121}\sd{.010} & \textbf{.91}\sd{.00} & \textbf{.31}\sd{.00} & \textbf{.84}\sd{.01} & \textbf{.69}\sd{.01} & \textbf{.60}\sd{.01} & \underline{35}\sd{2} & \textbf{64}\sd{8} & \underline{.76}\sd{.02} & \textbf{.49}\sd{.01} & \underline{34.2}\sd{3.7} \\
\addlinespace[4pt]
\multirow{5}{*}{\act{flexion}} & \gd{GaitDynamics}$^\ast$ & -- & -- & -- & -- & -- & \gd{.07}\gsd{.00} & \gd{.91}\gsd{.01} & \gd{.91}\gsd{.01} & \gd{.75}\gsd{.02} & \gd{15}\gsd{0} & \gd{33}\gsd{1} & -- & -- & -- \\
 & Fixed-PD & \underline{86.6}\sd{5.0} & \underline{3.19}\sd{.09} & 63.1\sd{4.0} & \underline{.070}\sd{.009} & \underline{.86}\sd{.02} & .31\sd{.02} & .52\sd{.02} & .31\sd{.01} & .39\sd{.02} & 24\sd{2} & 71\sd{4} & .77\sd{.06} & .33\sd{.01} & 53.3\sd{9.5} \\
 & Torque & 64.3\sd{8.2} & 5.77\sd{.43} & 163\sd{15} & .075\sd{.010} & .71\sd{.02} & \textbf{.21}\sd{.02} & \underline{.62}\sd{.03} & .45\sd{.05} & \underline{.54}\sd{.04} & 27\sd{6} & \textbf{67}\sd{3} & .54\sd{.02} & \underline{.37}\sd{.02} & 24.2\sd{1.6} \\
 & Muscle & 49.1\sd{7.2} & 14.4\sd{.6} & \underline{60.9}\sd{3.1} & .085\sd{.009} & .49\sd{.02} & \underline{.23}\sd{.00} & .58\sd{.04} & \textbf{.54}\sd{.02} & \textbf{.57}\sd{.03} & \underline{23}\sd{3} & 74\sd{2} & \textbf{.50}\sd{.01} & .32\sd{.01} & \textbf{3.50}\sd{.19} \\
 & ImpedanceMimic & \textbf{99.7}\sd{.3} & \textbf{2.53}\sd{.07} & \textbf{45.2}\sd{1.7} & \textbf{.058}\sd{.007} & \textbf{.90}\sd{.01} & \underline{.23}\sd{.01} & \textbf{.64}\sd{.01} & \underline{.47}\sd{.02} & .51\sd{.02} & \textbf{22}\sd{1} & \textbf{67}\sd{1} & \underline{.53}\sd{.01} & \textbf{.40}\sd{.01} & \underline{14.8}\sd{.7} \\
\addlinespace[4pt]
\multirow{5}{*}{\act{jump}} & \gd{GaitDynamics}$^\ast$ & -- & -- & -- & -- & -- & \gd{.16}\gsd{.01} & \gd{.88}\gsd{.01} & \gd{.85}\gsd{.01} & \gd{.67}\gsd{.02} & \gd{37}\gsd{3} & \gd{54}\gsd{1} & -- & -- & -- \\
 & Fixed-PD & \underline{71.5}\sd{3.3} & \underline{4.18}\sd{.22} & 90.4\sd{2.1} & .118\sd{.006} & \underline{.87}\sd{.01} & .48\sd{.02} & .48\sd{.05} & .34\sd{.05} & .20\sd{.03} & 37\sd{16} & 96\sd{38} & 1.06\sd{.03} & .21\sd{.02} & 86.6\sd{4.6} \\
 & Torque & 45.7\sd{10.6} & 6.94\sd{.72} & 204\sd{25} & \underline{.108}\sd{.012} & .74\sd{.03} & \textbf{.31}\sd{.03} & \textbf{.72}\sd{.03} & .47\sd{.06} & \textbf{.37}\sd{.04} & \textbf{27}\sd{3} & \textbf{72}\sd{3} & \underline{.65}\sd{.04} & \textbf{.37}\sd{.02} & \underline{33.5}\sd{5.0} \\
 & Muscle & 43.7\sd{7.8} & 14.2\sd{.8} & \underline{87.8}\sd{2.7} & .141\sd{.011} & .50\sd{.02} & \underline{.35}\sd{.01} & .57\sd{.03} & \textbf{.51}\sd{.03} & .23\sd{.02} & \underline{29}\sd{2} & 81\sd{5} & \textbf{.58}\sd{.01} & .27\sd{.02} & \textbf{6.62}\sd{.36} \\
 & ImpedanceMimic & \textbf{89.3}\sd{2.6} & \textbf{3.41}\sd{.09} & \textbf{74.9}\sd{3.8} & \textbf{.088}\sd{.007} & \textbf{.90}\sd{.01} & .39\sd{.01} & \underline{.62}\sd{.02} & \underline{.49}\sd{.02} & \underline{.34}\sd{.03} & \underline{29}\sd{1} & \underline{76}\sd{4} & .78\sd{.02} & \underline{.30}\sd{.01} & 41.7\sd{2.1} \\
\specialrule{\cmidrulewidth}{\aboverulesep}{0pt}
\rowcolor{meangray}  & \padT \gd{GaitDynamics}$^\ast$ & -- & -- & -- & -- & -- & \gd{.10}\gsd{.00} & \gd{.89}\gsd{.01} & \gd{.92}\gsd{.00} & \gd{.76}\gsd{.01} & \gd{18}\gsd{1} & \gd{29}\gsd{1} & -- & -- & -- \\
\rowcolor{meangray}  & Fixed-PD & \underline{91.5}\sd{1.2} & \underline{3.13}\sd{.05} & 69.6\sd{1.1} & \underline{.065}\sd{.003} & \underline{.87}\sd{.01} & .27\sd{.01} & .70\sd{.02} & .45\sd{.01} & .43\sd{.01} & 26\sd{2} & 59\sd{6} & .86\sd{.01} & .35\sd{.01} & 56.4\sd{2.0} \\
\rowcolor{meangray}  & Torque & 79.2\sd{4.4} & 5.01\sd{.10} & 133\sd{6} & .073\sd{.005} & .73\sd{.00} & \textbf{.21}\sd{.01} & \textbf{.79}\sd{.01} & .53\sd{.03} & \underline{.52}\sd{.02} & \textbf{24}\sd{2} & \underline{56}\sd{2} & \textbf{.63}\sd{.01} & \underline{.42}\sd{.01} & 23.7\sd{.7} \\
\rowcolor{meangray}  & Muscle & 75.3\sd{5.1} & 8.65\sd{.26} & \underline{65.8}\sd{1.6} & .081\sd{.002} & .62\sd{.01} & \textbf{.21}\sd{.00} & .74\sd{.02} & \underline{.62}\sd{.02} & .51\sd{.01} & \textbf{24}\sd{1} & 59\sd{1} & \underline{.66}\sd{.01} & .40\sd{.00} & \textbf{6.60}\sd{.06} \\
\rowcolor{meangray} \multirow{-6}{*}{\textbf{Mean}} & \padB ImpedanceMimic & \textbf{98.1}\sd{.4} & \textbf{2.45}\sd{.02} & \textbf{48.9}\sd{.9} & \textbf{.057}\sd{.002} & \textbf{.89}\sd{.00} & \textbf{.21}\sd{.00} & \underline{.76}\sd{.01} & \textbf{.65}\sd{.01} & \textbf{.56}\sd{.01} & \textbf{24}\sd{0} & \textbf{53}\sd{1} & .72\sd{.01} & \textbf{.45}\sd{.00} & \underline{19.9}\sd{1.0} \\
\specialrule{\heavyrulewidth}{0pt}{\belowbottomsep}
\end{tabular}%
}
\vspace{0pt}
\parbox{\textwidth}{\scriptsize\raggedright
$^\ast$GaitDynamics is shown in italics for reference and excluded from the ranking, since it is not a controller and serves as a reference for GRF.
  $^\dagger$9 main joints (hip flexion / adduction, knee, ankle, both sides; lumbar extension), against inverse dynamics.
  Ep. = episode survival before termination (\%). $q$, $\dot q$ = joint-angle / velocity RMSE. $p_x$ = root forward-position drift rate. \{kin$_r$, $F_{\mathrm{Z}_r}$, $F_{\mathrm{AP}_r}$, $F_{\mathrm{ML}_r}$, $\tau_r$\} = Pearson correlation w.r.t. reference \{joint angles, vertical, AP, ML GRF, torques\}. CoP = center of pressure. $\dot\tau$ = torque jerk.}
\vspace{-1em}
\end{table}
\begin{figure}[!t]%
    \centering
    \includegraphics[width=\linewidth, trim=0 0 0 6pt, clip]{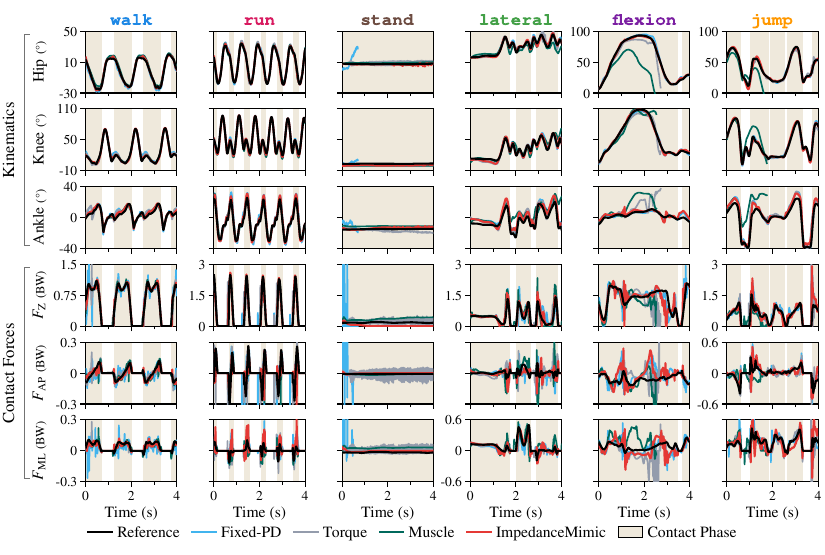}%
    \par\vspace{-6pt}%
    \caption{\textbf{Reference and simulated trajectories of the four controllers.} Right hip, knee and ankle angles and the three GRF components of the right foot on one held-out clip per activity. A trace that ends early marks a terminated episode, and \projname{} alone completes all six clips.}
    \label{fig:baseline_waveforms}
    \vspace{-14pt}%
\end{figure}

\vspace{-0.9em}
\subsection{Results}
\label{App:results}
\vspace{-0.7em}

\subsubsection{A Single Policy for Force-Accurate Motion Imitation}
\label{App:seen}
\vspace{-4pt}%

On held-out clips of the 129 training subjects (seen split, App.~\ref{App:splits}), \projname{} attains the highest survival and the lowest joint-angle error in every activity and leads on nearly every force and center-of-pressure metric (Table~\ref{tab:golden_full}, Fig.~\ref{fig:baseline_waveforms}).
Fixed-PD tracks the kinematics but loses the force where the motion does not determine it, its vertical-force correlation falling from $0.95$--$0.96$ in walking and running to $0.36$--$0.64$ in standing, flexion and jumping.
Torque recovers the vertical force but not the pose, with about twice the joint-angle error of \projname{}.
\projname{} trails Torque in vertical-force correlation only in standing, flexion and jumping, by at most $0.08$, and leads on both horizontal force correlations, the anterior--posterior center of pressure and the torque correlation, at a third of Fixed-PD's torque jerk and on par with Torque.
Muscle produces the smoothest torques but the lowest survival in flexion and jumping.
GaitDynamics, which predicts forces without a body, reaches the highest vertical-force correlation.

\vspace{-6pt}\vspace{0pt}%
\subsubsection{Zero-Shot Transfer to Unseen Subjects}
\label{App:unseen}
\vspace{-6pt}

On the 22 held-out subjects of the unseen split, each simulated with its own scaled model and inertia-scaled nominal gains (App.~\ref{App:preprocessing}, Eq.~\ref{eq:inertia_scaling}), \projname{} transfers without loss of survival and with little loss of fidelity, its survival even exceeding that of the seen-split policy on held-out clips and its joint-angle error rising by only $0.17^\circ$ (Tables~\ref{tab:golden_full} and~\ref{tab:golden_unseen}).
The baselines lose 1.5 to 16.2 points of survival, mostly in the non-cyclic activities, as every controller still survives at least $90.9\,\%$ of walking and running episodes.
In lateral cutting, flexion and jumping, Torque and Muscle, the two action spaces without a feedback law about the reference, survive $44$--$79\,\%$ of episodes, against $89$--$100\,\%$ for \projname{}.
Force accuracy transfers with a small loss, as the vertical-force correlation of \projname{} decreases by $0.04$ and remains second only to Torque, while \projname{} stays first on both horizontal force correlations, the anterior--posterior center of pressure and the torque correlation.
Fixed-PD retains its survival but not its force accuracy, reproducing on new bodies the trade-off it exhibits on seen subjects.

\begin{table}[!b]%
\vspace{-12pt}%
\centering
\caption{\textbf{Zero-shot transfer to the Unitree G1.} Unseen-split policies are evaluated without retraining on the 22 held-out subjects' clips retargeted to the robot. Cells report mean$_{\pm\text{sd}}$ over 5 seeds. \textbf{Bold} / \underline{underline} = best / second best within each activity.}
\vspace{-0.4em}
\label{tab:embodiments}
\renewcommand{\arraystretch}{0.80}
\setlength{\tabcolsep}{2pt}
\newcommand{\gap}{\hspace{4pt}}
\newcommand{\sd}[1]{{\tiny\textcolor{black!55}{$\pm$#1}}}
\scriptsize
\resizebox{\textwidth}{!}{%
\begin{tabular}{@{}>{\fillL}l @{\gap} l @{\gap} lllll @{\gap} lp{0.75cm}p{0.75cm}p{0.75cm}p{0.75cm}p{0.75cm} @{\gap} >{\fillR}l @{}}
\toprule
 & & \multicolumn{1}{c}{\textbf{Stability}} & \multicolumn{4}{c}{\textbf{Kinematics}} & \multicolumn{6}{c}{\textbf{Ground Reaction Forces}} & \multicolumn{1}{c}{\textbf{Torques}$^\dagger$} \\
\cmidrule(lr){3-3}\cmidrule(lr){4-7}\cmidrule(lr){8-13}\cmidrule(lr){14-14}
\texttt{\textbf{Activity}} & \textbf{Method}
  & \makecell{Ep.$\hi$ \\ (\%)}
  & \makecell{$q\lo$\\ \makebox[0pt][c]{($^\circ$)}}
  & \makecell{$\dot{q}\lo$\\ \makebox[0pt][c]{($^\circ$/s)}}
  & \makecell{$p_x\lo$\\ \makebox[0pt][c]{(m/s)}}
  & kin$_r\hi$
  & \makecell{$F_{\mathrm{Z}}\lo$\\ \makebox[0pt][c]{(BW)}}
  & $F^{\uparrow}_{\mathrm{Z}_r}$
  & $F^{\uparrow}_{\mathrm{ML}_r}$
  & $F^{\uparrow}_{\mathrm{AP}_r}$
  & \makecell{\makebox[0pt][c]{$\mathrm{CoP}^{\downarrow}_{\mathrm{ML}}$}\\ \makebox[0pt][c]{\tiny (mm)}}
  & \makecell{\makebox[0pt][c]{$\mathrm{CoP}^{\downarrow}_{\mathrm{AP}}$}\\ \makebox[0pt][c]{\tiny (mm)}}
  & \makecell{$\dot{\tau}\lo$\\ \makebox[0pt][c]{\tiny (Nm/kg/s)}} \\
\midrule
\multirow{3}{*}{\act{walk}} & Fixed-PD & \underline{76.7}\sd{4.0} & \underline{4.70}\sd{.10} & \underline{107}\sd{2} & \underline{.078}\sd{.018} & \underline{.87}\sd{.01} & \underline{.62}\sd{.02} & .41\sd{.03} & \underline{.39}\sd{.03} & .21\sd{.03} & \underline{24}\sd{3} & \underline{58}\sd{1} & \underline{32.8}\sd{.9} \\
 & Torque & 34.5\sd{3.9} & 14.2\sd{.4} & 393\sd{30} & .186\sd{.016} & .46\sd{.02} & .67\sd{.03} & \underline{.44}\sd{.03} & .33\sd{.04} & \underline{.23}\sd{.02} & 28\sd{2} & \textbf{57}\sd{1} & 43.9\sd{3.1} \\
 & ImpedanceMimic & \textbf{95.5}\sd{6.6} & \textbf{3.85}\sd{.16} & \textbf{83.7}\sd{6.3} & \textbf{.035}\sd{.004} & \textbf{.91}\sd{.01} & \textbf{.46}\sd{.02} & \textbf{.60}\sd{.03} & \textbf{.68}\sd{.02} & \textbf{.43}\sd{.05} & \textbf{22}\sd{1} & \underline{58}\sd{3} & \textbf{18.2}\sd{2.3} \\
\addlinespace[3pt]
\multirow{3}{*}{\act{run}} & Fixed-PD & \underline{43.0}\sd{13.8} & \underline{5.83}\sd{.20} & \underline{120}\sd{2} & \underline{.109}\sd{.016} & \underline{.89}\sd{.00} & \underline{.60}\sd{.01} & \underline{.66}\sd{.02} & .10\sd{.06} & \underline{.28}\sd{.03} & \textbf{28}\sd{2} & \textbf{50}\sd{1} & \underline{39.8}\sd{1.8} \\
 & Torque & 20.4\sd{2.8} & 17.5\sd{1.0} & 427\sd{41} & .178\sd{.023} & .46\sd{.02} & .68\sd{.03} & .57\sd{.02} & \underline{.13}\sd{.06} & .16\sd{.03} & 36\sd{3} & 51\sd{1} & 42.5\sd{2.6} \\
 & ImpedanceMimic & \textbf{71.0}\sd{17.1} & \textbf{5.00}\sd{.50} & \textbf{119}\sd{14} & \textbf{.093}\sd{.032} & \textbf{.91}\sd{.03} & \textbf{.55}\sd{.03} & \textbf{.72}\sd{.03} & \textbf{.25}\sd{.15} & \textbf{.38}\sd{.02} & \textbf{28}\sd{4} & \textbf{50}\sd{2} & \textbf{30.4}\sd{4.2} \\
\addlinespace[3pt]
\multirow{3}{*}{\act{stand}} & Fixed-PD & 41.4\sd{8.5} & \underline{3.65}\sd{.21} & \underline{97.5}\sd{11.0} & .139\sd{.036} & \underline{.14}\sd{.04} & \underline{.41}\sd{.04} & \negs{}.02\sd{.02} & .12\sd{.08} & \textbf{.03}\sd{.02} & 23\sd{2} & \underline{50}\sd{1} & \underline{17.3}\sd{2.2} \\
 & Torque & \underline{86.5}\sd{8.5} & 8.15\sd{1.61} & 289\sd{89} & \underline{.036}\sd{.017} & .07\sd{.02} & .57\sd{.14} & \textbf{.04}\sd{.01} & \underline{.18}\sd{.06} & \textbf{.03}\sd{.02} & \underline{21}\sd{3} & 56\sd{3} & 35.8\sd{17.2} \\
 & ImpedanceMimic & \textbf{97.7}\sd{3.6} & \textbf{2.77}\sd{.15} & \textbf{27.7}\sd{11.0} & \textbf{.031}\sd{.014} & \textbf{.39}\sd{.07} & \textbf{.18}\sd{.05} & \underline{.00}\sd{.05} & \textbf{.65}\sd{.16} & \negs{}.04\sd{.11} & \textbf{17}\sd{1} & \textbf{35}\sd{4} & \textbf{4.38}\sd{1.79} \\
\addlinespace[3pt]
\multirow{3}{*}{\act{lateral}} & Fixed-PD & \underline{85.2}\sd{1.7} & \underline{5.03}\sd{.16} & \underline{113}\sd{2} & \underline{.090}\sd{.010} & \underline{.85}\sd{.01} & \underline{.55}\sd{.01} & \underline{.52}\sd{.01} & \underline{.45}\sd{.02} & \underline{.31}\sd{.01} & \underline{37}\sd{1} & 54\sd{1} & \underline{37.5}\sd{1.5} \\
 & Torque & 62.0\sd{5.3} & 15.5\sd{.9} & 428\sd{22} & .096\sd{.009} & .37\sd{.02} & .66\sd{.02} & .40\sd{.03} & .34\sd{.03} & .18\sd{.03} & 41\sd{1} & \textbf{50}\sd{1} & 41.7\sd{2.1} \\
 & ImpedanceMimic & \textbf{93.7}\sd{1.7} & \textbf{4.46}\sd{.14} & \textbf{100}\sd{6} & \textbf{.074}\sd{.008} & \textbf{.88}\sd{.01} & \textbf{.47}\sd{.02} & \textbf{.62}\sd{.03} & \textbf{.59}\sd{.03} & \textbf{.38}\sd{.04} & \textbf{33}\sd{1} & \underline{52}\sd{1} & \textbf{24.4}\sd{1.3} \\
\addlinespace[3pt]
\multirow{3}{*}{\act{flexion}} & Fixed-PD & \underline{50.4}\sd{5.5} & \underline{4.33}\sd{.17} & \underline{92.2}\sd{4.6} & \underline{.069}\sd{.010} & \underline{.78}\sd{.01} & \underline{.42}\sd{.01} & .21\sd{.02} & \underline{.19}\sd{.04} & \underline{.32}\sd{.02} & \underline{27}\sd{2} & \underline{57}\sd{2} & \underline{24.7}\sd{.8} \\
 & Torque & 29.2\sd{2.9} & 11.6\sd{1.0} & 383\sd{54} & .094\sd{.021} & .48\sd{.05} & .54\sd{.08} & \underline{.22}\sd{.06} & .16\sd{.03} & .26\sd{.04} & \underline{27}\sd{2} & \underline{57}\sd{1} & 40.9\sd{7.3} \\
 & ImpedanceMimic & \textbf{73.3}\sd{5.1} & \textbf{3.73}\sd{.08} & \textbf{80.9}\sd{5.9} & \textbf{.063}\sd{.008} & \textbf{.83}\sd{.02} & \textbf{.33}\sd{.01} & \textbf{.34}\sd{.02} & \textbf{.33}\sd{.04} & \textbf{.33}\sd{.02} & \textbf{25}\sd{1} & \textbf{56}\sd{1} & \textbf{18.9}\sd{2.1} \\
\addlinespace[3pt]
\multirow{3}{*}{\act{jump}} & Fixed-PD & \underline{44.7}\sd{5.5} & \underline{4.57}\sd{.07} & \underline{110}\sd{6} & .160\sd{.007} & \underline{.81}\sd{.02} & \underline{.53}\sd{.02} & \underline{.21}\sd{.02} & \underline{.25}\sd{.05} & .03\sd{.02} & \textbf{28}\sd{2} & \underline{56}\sd{1} & \underline{29.0}\sd{1.3} \\
 & Torque & 31.4\sd{3.8} & 15.7\sd{.7} & 451\sd{51} & \textbf{.130}\sd{.026} & .45\sd{.03} & .61\sd{.05} & \underline{.21}\sd{.02} & .24\sd{.04} & \underline{.06}\sd{.02} & 31\sd{4} & 59\sd{8} & 39.9\sd{6.5} \\
 & ImpedanceMimic & \textbf{61.4}\sd{5.7} & \textbf{4.42}\sd{.14} & \textbf{103}\sd{5} & \underline{.138}\sd{.013} & \textbf{.85}\sd{.01} & \textbf{.48}\sd{.01} & \textbf{.27}\sd{.03} & \textbf{.38}\sd{.07} & \textbf{.11}\sd{.04} & \textbf{28}\sd{1} & \textbf{54}\sd{1} & \textbf{22.8}\sd{1.1} \\
\specialrule{\cmidrulewidth}{\aboverulesep}{0pt}
\rowcolor{meangray}  & \padT Fixed-PD & \underline{56.9}\sd{3.6} & \underline{4.68}\sd{.11} & \underline{106}\sd{3} & \underline{.108}\sd{.007} & \underline{.72}\sd{.01} & \underline{.52}\sd{.01} & \underline{.33}\sd{.01} & \underline{.25}\sd{.02} & \underline{.20}\sd{.01} & \underline{28}\sd{1} & \underline{54}\sd{0} & \underline{30.2}\sd{.8} \\
\rowcolor{meangray}  & Torque & 44.0\sd{2.4} & 13.8\sd{.6} & 395\sd{32} & .120\sd{.013} & .38\sd{.02} & .62\sd{.05} & .32\sd{.02} & .23\sd{.03} & .15\sd{.01} & 31\sd{2} & 55\sd{2} & 40.8\sd{6.0} \\
\rowcolor{meangray} \multirow{-3}{*}{\textbf{Mean}} & \padB ImpedanceMimic & \textbf{82.1}\sd{4.3} & \textbf{4.04}\sd{.14} & \textbf{85.7}\sd{4.5} & \textbf{.072}\sd{.008} & \textbf{.79}\sd{.01} & \textbf{.41}\sd{.01} & \textbf{.42}\sd{.02} & \textbf{.48}\sd{.05} & \textbf{.27}\sd{.02} & \textbf{25}\sd{1} & \textbf{51}\sd{1} & \textbf{19.8}\sd{.6} \\
\specialrule{\heavyrulewidth}{0pt}{\belowbottomsep}
\end{tabular}%
}
\vspace{0pt}
\parbox{\textwidth}{\scriptsize\raggedright
  $^\dagger$Torque jerk over the 9 main joints (hip flexion / adduction, knee, ankle, both sides; lumbar extension). Forces and CoP are scored against the physics prior.
  Ep. = episode survival before termination (\%). $q$, $\dot q$ = joint-angle / velocity RMSE. $p_x$ = root forward-position drift rate. \{kin$_r$, $F_{\mathrm{Z}_r}$, $F_{\mathrm{AP}_r}$, $F_{\mathrm{ML}_r}$\} = Pearson correlation w.r.t. reference \{joint angles, vertical, AP, ML GRF\}. CoP = center of pressure.}
\vspace{-1em}
\end{table}
\vspace{-6pt}\vspace{0pt}%
\subsubsection{Zero-Shot Transfer to a Humanoid Robot}
\label{App:embodiments}
\vspace{-6pt}
\paragraph{Robot model and retargeting.}
The unseen-split policies (107 training subjects, five seeds) are evaluated without retraining on the Unitree G1.
The MuJoCo Menagerie model is rebuilt in the Rajagopal topology, with 16 bodies, the 31 joints matched by name and axis and eight contact spheres per foot on the robot sole, and the joints the robot lacks are locked with their action channels zero-filled.
The clips of the 22 held-out subjects are retargeted to the robot by a joint map with a per-frame foot-height correction.
Since the robot has no force plates, the force target is the physics prior on the retargeted kinematics, which the policy also observes, without the learned residual.
Actuator torque limits are enforced, and the gain priors follow from the robot's joint-space inertia (Eq.~\ref{eq:inertia_scaling}).

\vspace{-4pt}%
\paragraph{Transfer.}
\projname{} transfers best on every activity-mean metric among the action spaces of Table~\ref{tab:embodiments}, losing 16 points of survival from the held-out humans to the robot against about 35 for Fixed-PD and Torque.
The gap is widest in quiet standing, where fixed nominal gains, although derived from the robot's own inertia, do not stabilize the robot under explicit PD ($41.4$ against $97.7\,\%$), while under stable PD and implicit damping they stand at $89.9$ and $85.0\,\%$ (activity means in Table~\ref{tab:embodiments_integrators}).
Torque survives standing but tracks poorly throughout, with about three times the joint-angle error of either PD action space.
\projname{} also produces the smoothest torques of the three controllers, with a torque jerk unchanged from the human subjects.
Running is its hardest activity to track, consistent with the robot's ankle torque limit of $1.5$\,Nm/kg, which covers every walking frame of the scaled human references but is exceeded on a fifth of the running frames.
Force correlations are lower than on humans for every action space, as the target is a synthetic prior rather than a measurement.

\begin{table}[H]%
\vspace{-5pt}%
\centering
\caption{\textbf{Zero-shot transfer to the Unitree G1 across integration schemes and gain levels.} Unseen-split policies are evaluated as in Table~\ref{tab:embodiments}. Activity mean$_{\pm\text{sd}}$ over 5 seeds.}
\vspace{-0.4em}
\label{tab:embodiments_integrators}
\scriptsize
\setlength{\tabcolsep}{0pt}%
\setlength{\aboverulesep}{0pt}
\newcommand{\sd}[1]{{\tiny\textcolor{black!55}{$\pm$#1}}}
\begin{tabular}{@{}l<{\hspace{20.00pt}}l>{\hspace{30.08pt}}c>{\hspace{30.08pt}}c>{\hspace{30.08pt}}c>{\hspace{30.08pt}}c>{\hspace{30.08pt}}c@{}}
\toprule
\textbf{PD} & \textbf{Method} & \textbf{\boldmath Ep.$\hi$ (\%)} & \textbf{\boldmath $q\lo$ ($^\circ$)} & {\boldmath$F^{\uparrow}_{\mathrm{Z}_r}$} & {\boldmath$F^{\uparrow}_{\mathrm{ML}_r}$} & \textbf{\boldmath $\dot{\tau}\lo$ {\tiny (Nm/kg/s)}} \\%
\midrule
 & Fixed-PD & 56.9\sd{3.6} & 4.68\sd{.11} & .33\sd{.01} & .25\sd{.02} & 30.2\sd{.8} \\
 & Fixed-PD, $2\times$ & 63.3\sd{5.5} & 4.58\sd{.15} & .32\sd{.01} & .21\sd{.02} & 54.8\sd{2.9} \\
\rowcolor{imptint}\cellcolor{white} & \textcolor[RGB]{229,57,53}{\textbf{\projname{}}} & 82.1\sd{4.3} & 4.04\sd{.14} & .42\sd{.02} & .48\sd{.05} & 19.8\sd{.6} \\
\rowcolor{imptint}\cellcolor{white}\multirow{-4}{*}{Explicit} & \textcolor[RGB]{229,57,53}{\textbf{\projname{}, $2\times$}} & 83.5\sd{3.6} & 3.98\sd{.12} & .40\sd{.02} & .41\sd{.02} & 31.4\sd{3.1} \\
\midrule
 & Fixed-PD & 74.5\sd{2.6} & 3.82\sd{.09} & .45\sd{.01} & .46\sd{.03} & 13.4\sd{.5} \\
 & Fixed-PD, $2\times$ & 71.1\sd{4.9} & 3.46\sd{.05} & .40\sd{.01} & .44\sd{.04} & 22.1\sd{.8} \\
\rowcolor{imptint}\cellcolor{white} & \textcolor[RGB]{229,57,53}{\textbf{\projname{}}} & 75.9\sd{5.0} & 3.65\sd{.08} & .43\sd{.02} & .55\sd{.03} & 15.5\sd{.8} \\
\rowcolor{imptint}\cellcolor{white}\multirow{-4}{*}{Stable} & \textcolor[RGB]{229,57,53}{\textbf{\projname{}, $2\times$}} & 79.7\sd{3.5} & 3.44\sd{.09} & .39\sd{.01} & .48\sd{.03} & 24.4\sd{1.1} \\
\midrule
 & Fixed-PD & 60.6\sd{4.0} & 4.33\sd{.09} & .39\sd{.01} & .37\sd{.02} & 19.0\sd{.5} \\
 & Fixed-PD, $2\times$ & 65.8\sd{2.1} & 4.01\sd{.09} & .32\sd{.01} & .28\sd{.03} & 33.6\sd{1.5} \\
\rowcolor{imptint}\cellcolor{white} & \textcolor[RGB]{229,57,53}{\textbf{\projname{}}} & 75.5\sd{4.3} & 4.09\sd{.17} & .39\sd{.01} & .44\sd{.02} & 23.6\sd{2.4} \\
\rowcolor{imptint}\cellcolor{white}\multirow{-4}{*}{Implicit} & \textcolor[RGB]{229,57,53}{\textbf{\projname{}, $2\times$}} & 77.4\sd{1.9} & 3.84\sd{.03} & .34\sd{.00} & .35\sd{.04} & 37.9\sd{2.4} \\
\bottomrule
\end{tabular}
\end{table}

\vspace{-19pt}\vspace{0pt}%
\subsubsection{Cross-Activity Generalization}
\label{App:loao}
\vspace{-4pt}

Trained on five activities and evaluated zero-shot on the sixth (App.~\ref{App:splits}),%
\begin{table}[!b]%
\vspace{-9pt}%
\centering
\caption{%
  \textbf{Full metrics for withheld activities, zero-shot.}
  Each activity is scored by the fold trained without it (App.~\ref{App:splits}).
  Columns are RMSE ($\downarrow$) unless marked $\uparrow$.
  Cells report mean$_{\pm\text{sd}}$ over 5 seeds.
  \textbf{Bold} / \underline{underline} = best / second best per activity.
  }
\vspace{-0.5em}
\label{tab:loao_full}
\renewcommand{\arraystretch}{0.85}
\setlength{\tabcolsep}{2.0pt}
\newcommand{\gap}{\hspace{4pt}}\newcommand{\gapin}{\hspace{2.5pt}}%
\newcommand{\w}{0.85cm}
\newcommand{\wc}{0.88cm}
\newcommand{\sd}[1]{{\tiny\textcolor{black!55}{$\pm$#1}}}
\newcommand{\gsd}[1]{{\tiny\textcolor{black!55}{\textit{$\pm$#1}}}}
\newcommand{\gd}[1]{\textit{#1}}
\scriptsize
\resizebox{\textwidth}{!}{%
\begin{tabular}{@{}>{\fillL}l @{\gap} >{\gapin}l @{\gap} >{\gapin}l @{\gap} >{\gapin}llll @{\gap} >{\gapin}lp{0.75cm}p{0.75cm}p{0.75cm}ll @{\gap} >{\gapin}ll>{\fillR}l @{}}
\toprule
 & & \multicolumn{1}{c}{\textbf{Stability}} & \multicolumn{4}{c}{\textbf{Kinematics}} & \multicolumn{6}{c}{\textbf{Ground Reaction Forces}} & \multicolumn{3}{c}{\textbf{Torques}$^\dagger$} \\
\cmidrule(lr){3-3}\cmidrule(lr){4-7}\cmidrule(lr){8-13}\cmidrule(lr){14-16}
\makecell[l]{\texttt{\textbf{Withheld}} \\ \texttt{\textbf{Activity}}} & \textbf{Method}
  & \makecell{Ep.$\hi$ \\ (\%)}
  & \makecell{$q\lo$\\ \makebox[0pt][c]{($^\circ$)}}
  & \makecell{$\dot{q}\lo$\\ \makebox[0pt][c]{($^\circ$/s)}}
  & \makecell{$p_x\lo$\\ \makebox[0pt][c]{(m/s)}}
  & kin$_r\hi$
  & \makecell{$F_{\mathrm{Z}}\lo$\\ \makebox[0pt][c]{(BW)}}
  & $F^{\uparrow}_{\mathrm{Z}_r}$
  & $F^{\uparrow}_{\mathrm{ML}_r}$
  & $F^{\uparrow}_{\mathrm{AP}_r}$
  & \makecell{\makebox[0pt][c]{$\mathrm{CoP}^{\downarrow}_{\mathrm{ML}}$}\\ \makebox[0pt][c]{\tiny (mm)}}
  & \makecell{\makebox[0pt][c]{$\mathrm{CoP}^{\downarrow}_{\mathrm{AP}}$}\\ \makebox[0pt][c]{\tiny (mm)}}
  & \makecell{$\tau\lo$\\ \makebox[0pt][c]{\tiny (Nm/kg)}}
  & $\tau_r\hi$
  & \makecell{$\dot{\tau}\lo$\\ \makebox[0pt][c]{\tiny (Nm/kg/s)}} \\
\midrule
\multirow{5}{*}{\act{walk}} & \gd{GaitDynamics}$^\ast$ & -- & -- & -- & -- & -- & \gd{.12}\gsd{.00} & \gd{.96}\gsd{.00} & \gd{.93}\gsd{.01} & \gd{.91}\gsd{.01} & \gd{15}\gsd{1} & \gd{40}\gsd{2} & -- & -- & -- \\
 & Fixed-PD & 40.0\sd{5.0} & \underline{4.52}\sd{.14} & 115\sd{5} & .322\sd{.034} & \underline{.85}\sd{.01} & .61\sd{.00} & .43\sd{.01} & .13\sd{.02} & .11\sd{.01} & 76\sd{22} & 137\sd{18} & 1.40\sd{.03} & .14\sd{.01} & 158\sd{6} \\
 & Torque & \underline{79.5}\sd{7.7} & 6.53\sd{.44} & 172\sd{14} & \underline{.251}\sd{.064} & .76\sd{.02} & \textbf{.32}\sd{.03} & \textbf{.76}\sd{.04} & \textbf{.51}\sd{.03} & \underline{.41}\sd{.02} & \textbf{27}\sd{3} & \textbf{81}\sd{6} & \underline{.68}\sd{.04} & \textbf{.34}\sd{.04} & \underline{27.6}\sd{2.7} \\
 & Muscle & 32.8\sd{7.8} & 12.2\sd{.8} & \underline{93.3}\sd{4.0} & \textbf{.214}\sd{.055} & .54\sd{.03} & \underline{.33}\sd{.02} & \underline{.74}\sd{.02} & \underline{.50}\sd{.04} & .34\sd{.04} & \underline{28}\sd{4} & \underline{94}\sd{8} & \textbf{.61}\sd{.02} & .28\sd{.03} & \textbf{9.06}\sd{.38} \\
 & ImpedanceMimic & \textbf{87.1}\sd{4.2} & \textbf{3.43}\sd{.20} & \textbf{78.0}\sd{5.3} & .263\sd{.050} & \textbf{.91}\sd{.01} & .41\sd{.04} & .66\sd{.04} & .49\sd{.04} & \textbf{.42}\sd{.05} & 38\sd{5} & 95\sd{9} & .83\sd{.07} & \underline{.30}\sd{.04} & 31.5\sd{5.6} \\
\addlinespace[4pt]
\multirow{5}{*}{\act{run}} & \gd{GaitDynamics}$^\ast$ & -- & -- & -- & -- & -- & \gd{.11}\gsd{.01} & \gd{.99}\gsd{.00} & \gd{.97}\gsd{.00} & \gd{.92}\gsd{.01} & \gd{13}\gsd{1} & \gd{26}\gsd{2} & -- & -- & -- \\
 & Fixed-PD & 73.4\sd{4.5} & \underline{5.10}\sd{.24} & 114\sd{4} & \underline{.251}\sd{.019} & \underline{.88}\sd{.01} & .49\sd{.02} & .75\sd{.02} & .30\sd{.09} & .28\sd{.06} & 25\sd{6} & 66\sd{2} & 1.18\sd{.04} & .35\sd{.03} & 92.7\sd{4.5} \\
 & Torque & 73.8\sd{3.8} & 9.29\sd{.30} & 214\sd{12} & .257\sd{.016} & .70\sd{.01} & \underline{.38}\sd{.01} & \underline{.85}\sd{.01} & .25\sd{.10} & .34\sd{.03} & 28\sd{4} & 67\sd{4} & .92\sd{.01} & .44\sd{.02} & 27.9\sd{1.8} \\
 & Muscle & \underline{80.6}\sd{4.5} & 8.39\sd{.49} & \underline{112}\sd{4} & .313\sd{.076} & .72\sd{.01} & .39\sd{.02} & \underline{.85}\sd{.02} & \textbf{.69}\sd{.04} & \underline{.50}\sd{.01} & \textbf{21}\sd{1} & \textbf{56}\sd{6} & \underline{.84}\sd{.02} & \underline{.46}\sd{.04} & \textbf{11.1}\sd{.3} \\
 & ImpedanceMimic & \textbf{95.6}\sd{4.1} & \textbf{3.55}\sd{.14} & \textbf{68.8}\sd{4.0} & \textbf{.161}\sd{.008} & \textbf{.95}\sd{.01} & \textbf{.28}\sd{.02} & \textbf{.92}\sd{.02} & \underline{.50}\sd{.11} & \textbf{.57}\sd{.05} & \underline{22}\sd{1} & \underline{61}\sd{6} & \textbf{.83}\sd{.02} & \textbf{.55}\sd{.02} & \underline{22.2}\sd{1.5} \\
\addlinespace[4pt]
\multirow{5}{*}{\act{stand}} & \gd{GaitDynamics}$^\ast$ & -- & -- & -- & -- & -- & \gd{.07}\gsd{.00} & \gd{.85}\gsd{.01} & \gd{.95}\gsd{.00} & \gd{.38}\gsd{.05} & \gd{15}\gsd{1} & \gd{25}\gsd{1} & -- & -- & -- \\
 & Fixed-PD & 68.6\sd{3.2} & \underline{3.08}\sd{.10} & 74.4\sd{4.8} & .080\sd{.010} & \underline{.65}\sd{.03} & .39\sd{.03} & .39\sd{.01} & .25\sd{.02} & .07\sd{.01} & 29\sd{14} & 85\sd{10} & .93\sd{.06} & .13\sd{.01} & 103\sd{9} \\
 & Torque & \textbf{93.2}\sd{1.3} & 4.92\sd{.62} & 183\sd{37} & .267\sd{.098} & .61\sd{.05} & .27\sd{.04} & .49\sd{.06} & .36\sd{.08} & .10\sd{.02} & \underline{23}\sd{5} & 72\sd{7} & .58\sd{.06} & \underline{.20}\sd{.02} & 34.1\sd{7.3} \\
 & Muscle & 65.0\sd{2.8} & 7.45\sd{.93} & \underline{40.2}\sd{4.9} & \underline{.067}\sd{.006} & .49\sd{.04} & \underline{.19}\sd{.01} & \underline{.53}\sd{.02} & \textbf{.69}\sd{.02} & \underline{.19}\sd{.04} & 24\sd{9} & \underline{71}\sd{9} & \textbf{.36}\sd{.01} & .16\sd{.01} & \textbf{3.80}\sd{.53} \\
 & ImpedanceMimic & \underline{91.9}\sd{1.2} & \textbf{2.34}\sd{.07} & \textbf{36.9}\sd{2.5} & \textbf{.039}\sd{.006} & \textbf{.81}\sd{.01} & \textbf{.18}\sd{.01} & \textbf{.64}\sd{.02} & \underline{.68}\sd{.02} & \textbf{.22}\sd{.02} & \textbf{17}\sd{1} & \textbf{56}\sd{3} & \underline{.45}\sd{.02} & \textbf{.26}\sd{.01} & \underline{14.0}\sd{1.3} \\
\addlinespace[4pt]
\multirow{5}{*}{\act{lateral}} & \gd{GaitDynamics}$^\ast$ & -- & -- & -- & -- & -- & \gd{.26}\gsd{.01} & \gd{.86}\gsd{.02} & \gd{.79}\gsd{.01} & \gd{.68}\gsd{.01} & \gd{32}\gsd{2} & \gd{41}\gsd{3} & -- & -- & -- \\
 & Fixed-PD & \underline{93.7}\sd{.5} & \underline{4.10}\sd{.09} & \underline{75.8}\sd{1.0} & \underline{.094}\sd{.003} & \underline{.90}\sd{.00} & .36\sd{.01} & .77\sd{.01} & .65\sd{.02} & .46\sd{.02} & \underline{31}\sd{4} & 62\sd{7} & .81\sd{.01} & .41\sd{.01} & 51.2\sd{1.6} \\
 & Torque & 86.9\sd{2.3} & 7.23\sd{.20} & 185\sd{9} & .105\sd{.004} & .74\sd{.00} & \underline{.30}\sd{.01} & \underline{.82}\sd{.01} & .69\sd{.01} & \underline{.51}\sd{.02} & \textbf{29}\sd{3} & \textbf{58}\sd{2} & .68\sd{.02} & \underline{.44}\sd{.01} & 33.3\sd{1.8} \\
 & Muscle & 83.3\sd{4.4} & 10.1\sd{.2} & 84.3\sd{1.5} & .116\sd{.004} & .62\sd{.01} & \textbf{.28}\sd{.01} & \textbf{.84}\sd{.01} & \underline{.71}\sd{.02} & .37\sd{.04} & 32\sd{3} & 63\sd{4} & \textbf{.59}\sd{.02} & .42\sd{.02} & \textbf{10.2}\sd{.3} \\
 & ImpedanceMimic & \textbf{98.9}\sd{.5} & \textbf{3.41}\sd{.07} & \textbf{64.9}\sd{2.4} & \textbf{.082}\sd{.002} & \textbf{.93}\sd{.00} & \underline{.30}\sd{.01} & \underline{.82}\sd{.01} & \textbf{.73}\sd{.00} & \textbf{.56}\sd{.02} & 32\sd{7} & \underline{59}\sd{7} & \underline{.65}\sd{.02} & \textbf{.48}\sd{.01} & \underline{26.0}\sd{2.2} \\
\addlinespace[4pt]
\multirow{5}{*}{\act{flexion}} & \gd{GaitDynamics}$^\ast$ & -- & -- & -- & -- & -- & \gd{.11}\gsd{.00} & \gd{.86}\gsd{.01} & \gd{.88}\gsd{.02} & \gd{.69}\gsd{.01} & \gd{24}\gsd{3} & \gd{41}\gsd{2} & -- & -- & -- \\
 & Fixed-PD & \underline{71.7}\sd{1.3} & \underline{3.52}\sd{.10} & \textbf{41.2}\sd{2.4} & \textbf{.046}\sd{.002} & \underline{.85}\sd{.00} & .23\sd{.01} & .61\sd{.02} & .39\sd{.02} & .44\sd{.02} & 25\sd{1} & \underline{70}\sd{1} & .58\sd{.01} & \underline{.37}\sd{.01} & 28.5\sd{2.1} \\
 & Torque & 67.9\sd{3.5} & 8.71\sd{.38} & 152\sd{13} & .060\sd{.003} & .64\sd{.01} & \textbf{.21}\sd{.00} & \underline{.64}\sd{.01} & .47\sd{.04} & \underline{.48}\sd{.01} & 26\sd{1} & 72\sd{2} & \underline{.54}\sd{.02} & .36\sd{.02} & 21.7\sd{2.1} \\
 & Muscle & 64.0\sd{.7} & 11.7\sd{.7} & 53.0\sd{1.5} & .072\sd{.004} & .58\sd{.01} & \textbf{.21}\sd{.00} & .60\sd{.01} & \textbf{.57}\sd{.01} & \textbf{.49}\sd{.02} & \textbf{24}\sd{0} & 75\sd{1} & \textbf{.48}\sd{.01} & .34\sd{.01} & \textbf{3.90}\sd{.20} \\
 & ImpedanceMimic & \textbf{80.5}\sd{3.0} & \textbf{3.06}\sd{.07} & \underline{47.6}\sd{7.3} & \underline{.047}\sd{.003} & \textbf{.88}\sd{.00} & .22\sd{.01} & \textbf{.65}\sd{.01} & \underline{.51}\sd{.01} & .47\sd{.02} & \textbf{24}\sd{1} & \textbf{67}\sd{2} & .55\sd{.04} & \textbf{.38}\sd{.02} & \underline{18.5}\sd{4.4} \\
\addlinespace[4pt]
\multirow{5}{*}{\act{jump}} & \gd{GaitDynamics}$^\ast$ & -- & -- & -- & -- & -- & \gd{.16}\gsd{.00} & \gd{.87}\gsd{.01} & \gd{.84}\gsd{.01} & \gd{.63}\gsd{.01} & \gd{33}\gsd{2} & \gd{57}\gsd{1} & -- & -- & -- \\
 & Fixed-PD & 71.2\sd{1.3} & \underline{3.98}\sd{.09} & \underline{80.2}\sd{1.1} & \underline{.076}\sd{.004} & \underline{.88}\sd{.00} & .38\sd{.00} & .58\sd{.00} & .40\sd{.02} & .29\sd{.01} & \textbf{25}\sd{1} & \underline{71}\sd{2} & .90\sd{.01} & .27\sd{.00} & 65.6\sd{1.7} \\
 & Torque & \underline{74.5}\sd{1.2} & 8.72\sd{.53} & 218\sd{11} & .086\sd{.009} & .69\sd{.01} & \textbf{.30}\sd{.01} & \textbf{.67}\sd{.01} & \textbf{.51}\sd{.01} & \underline{.33}\sd{.00} & 39\sd{8} & 76\sd{13} & \underline{.60}\sd{.01} & \textbf{.39}\sd{.01} & \underline{26.0}\sd{1.3} \\
 & Muscle & 57.5\sd{2.1} & 13.3\sd{1.1} & 86.6\sd{8.3} & .107\sd{.007} & .52\sd{.03} & \underline{.34}\sd{.01} & .53\sd{.03} & .49\sd{.04} & .26\sd{.01} & 32\sd{5} & 81\sd{6} & \textbf{.55}\sd{.02} & .30\sd{.02} & \textbf{6.94}\sd{.43} \\
 & ImpedanceMimic & \textbf{83.8}\sd{.6} & \textbf{3.48}\sd{.12} & \textbf{77.5}\sd{4.9} & \textbf{.067}\sd{.005} & \textbf{.91}\sd{.01} & \underline{.34}\sd{.01} & \underline{.65}\sd{.01} & \textbf{.51}\sd{.01} & \textbf{.35}\sd{.01} & \textbf{25}\sd{0} & \textbf{65}\sd{1} & .70\sd{.03} & \underline{.36}\sd{.01} & 32.5\sd{3.2} \\
\specialrule{\cmidrulewidth}{\aboverulesep}{0pt}
\rowcolor{meangray}  & \padT \gd{GaitDynamics}$^\ast$ & -- & -- & -- & -- & -- & \gd{.14}\gsd{.00} & \gd{.90}\gsd{.01} & \gd{.89}\gsd{.01} & \gd{.70}\gsd{.01} & \gd{22}\gsd{1} & \gd{38}\gsd{1} & -- & -- & -- \\
\rowcolor{meangray}  & Fixed-PD & 69.8\sd{1.7} & \underline{4.05}\sd{.06} & 83.5\sd{1.4} & \underline{.145}\sd{.008} & \underline{.84}\sd{.01} & .41\sd{.01} & .59\sd{.01} & .35\sd{.02} & .27\sd{.01} & 35\sd{3} & 82\sd{1} & .97\sd{.01} & .28\sd{.01} & 83.2\sd{1.8} \\
\rowcolor{meangray}  & Torque & \underline{79.3}\sd{1.2} & 7.57\sd{.25} & 187\sd{10} & .171\sd{.026} & .69\sd{.01} & .30\sd{.01} & \underline{.71}\sd{.01} & .47\sd{.03} & \underline{.36}\sd{.01} & 29\sd{2} & \underline{71}\sd{2} & \underline{.67}\sd{.01} & \underline{.36}\sd{.01} & 28.5\sd{1.3} \\
\rowcolor{meangray}  & Muscle & 63.9\sd{2.1} & 10.5\sd{.2} & \underline{78.2}\sd{1.3} & .148\sd{.019} & .58\sd{.01} & \textbf{.29}\sd{.00} & .68\sd{.00} & \textbf{.61}\sd{.01} & \underline{.36}\sd{.01} & \underline{27}\sd{1} & 73\sd{2} & \textbf{.57}\sd{.01} & .33\sd{.01} & \textbf{7.51}\sd{.22} \\
\rowcolor{meangray} \multirow{-6}{*}{\textbf{Mean}} & \padB ImpedanceMimic & \textbf{89.6}\sd{.6} & \textbf{3.21}\sd{.03} & \textbf{62.3}\sd{1.5} & \textbf{.110}\sd{.007} & \textbf{.89}\sd{.00} & \textbf{.29}\sd{.00} & \textbf{.72}\sd{.01} & \underline{.57}\sd{.02} & \textbf{.43}\sd{.01} & \textbf{26}\sd{1} & \textbf{67}\sd{1} & \underline{.67}\sd{.01} & \textbf{.39}\sd{.01} & \underline{24.1}\sd{1.1} \\
\specialrule{\heavyrulewidth}{0pt}{\belowbottomsep}
\end{tabular}%
}
\vspace{0pt}
\parbox{\textwidth}{\scriptsize\raggedright
$^\ast$GaitDynamics is shown in italics for reference and excluded from the ranking, since it is not a controller and serves as a reference for GRF.
  $^\dagger$9 main joints (hip flexion / adduction, knee, ankle, both sides; lumbar extension), against inverse dynamics.
  Ep. = episode survival before termination (\%). $q$, $\dot q$ = joint-angle / velocity RMSE. $p_x$ = root forward-position drift rate. \{kin$_r$, $F_{\mathrm{Z}_r}$, $F_{\mathrm{AP}_r}$, $F_{\mathrm{ML}_r}$, $\tau_r$\} = Pearson correlation w.r.t. reference \{joint angles, vertical, AP, ML GRF, torques\}. CoP = center of pressure. $\dot\tau$ = torque jerk.}
\vspace{-7pt}%
\end{table}%
\projname{} generalizes best to the withheld activity, with the lowest joint-angle error in every fold and the highest survival in every fold except standing, where Torque leads by 1.3 points (Table~\ref{tab:loao_full}).
Relative to training on all six activities, its mean survival decreases by 7 points, against 16 to 23 points for the baselines.
Withheld walking is the hardest fold for Fixed-PD and Muscle, which survive $33$--$40\,\%$ of its episodes, whereas \projname{} survives $87.1\,\%$ of them but loses force accuracy, its vertical-force correlation falling to $0.66$ against $0.98$ when walking is trained (Table~\ref{tab:golden_full}).
Over the six folds, \projname{} nevertheless retains the highest vertical-force correlation among the controllers and leads the anterior--posterior force correlation and both center-of-pressure errors, whereas Muscle attains the best medio-lateral force correlation but the second-lowest survival.

\vspace{-8pt plus 1fil}\vspace{0pt}%
\subsubsection{Action-Space Bounds on the Human Model}
\label{App:humanoid_bounds}
\vspace{-4pt}

To test whether the single-joint bounds carry over to the human model, we evaluate both (Eqs.~\ref{eq:bound1} and~\ref{eq:bound2}) frame by frame on the held-out clips of the seen split, for the learned and for the nominal fixed gains.
Bound~1 counts the contact frames on which the realized stiffness $k^p_t$ exceeds the demand $|\tau^*_t|/\alpha$, with $\tau^*_t$ the inverse-dynamics torque of the reference clip, and Bound~2 the swing frames on which $k^d_t < 2I_{jj}/h$, with $I_{jj}$ the diagonal joint-space inertia at the rest pose.
Since $k^d_0 \propto I_{jj}$, Bound~2 is a subject-independent multiple of the nominal damping, far above it at the hip and knee but at $0.77\times$ nominal at the ankle (Table~\ref{tab:humanoid_bounds}).

\begin{figure}[t]%
    \centering
    \includegraphics[width=\linewidth]{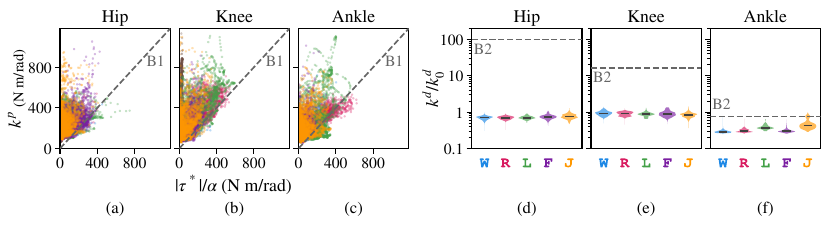}
    \par\vspace{-5pt}%
    \caption{\textbf{Action-space bounds on the joint-actuated human model.} \textbf{(a--c)}~Contact stiffness against the Bound-1 demand $|\tau^*|/\alpha$. \textbf{(d--f)}~Swing damping against Bound~2 on a log scale.}
    \label{fig:humanoid_bounds}
    \vspace{-17.5pt}%
\end{figure}
\vspace{-4pt plus 1fil}%
\paragraph{Bound satisfaction.}
The single-joint analysis predicts where the bounds bind on the human model, Bound~2 only at the ankle in swing, the one joint whose nominal damping exceeds it, and Bound~1 in the contact phases of running and lateral cutting, where the torque demand exceeds the nominal stiffness (Table~\ref{tab:humanoid_bounds}, Fig.~\ref{fig:humanoid_bounds}).
The nominal ankle damping violates Bound~2 on every swing frame, whereas the learned policy holds it at a median $0.29$--$0.39\times$ nominal and satisfies it on at least $96\,\%$ of swing frames.
Bound~1 binds where the median per-cycle demand $\max_t |\tau^*_t|/(\alpha k^p_0)$ reaches unity, at the walking ankle ($1.0$), the running knee ($1.5$) and the running ankle ($1.7$), and there the learned stiffness meets the demand on more than $90\,\%$ of contact frames, whereas the nominal stiffness misses up to half in running.
Both fall short at the lateral-cutting ankle, on about a fifth of its contact frames.
As the policy raises $k^p$ when the demand arrives rather than throughout contact, the strict per-cycle form $\min_t k^p_t \ge \max_t |\tau^*_t|/\alpha$ holds for at most $2\,\%$ of the cycles where Bound~1 binds.

\vspace{0pt plus 1fil}%
\begin{table}[h]%
\vspace{-7pt}
\centering
\caption{\textbf{Action-space bounds on the joint-actuated human model.} Shares (\%) of contact frames satisfying Bound~1 (B1) and of swing frames satisfying Bound~2 (B2) on seen held-out clips. Each cell reads learned gains (mean over five seeds) / nominal gains. Standing has no swing phase.}
\label{tab:humanoid_bounds}
\scriptsize
\setlength{\tabcolsep}{2.5pt}
\begin{tabular*}{\textwidth}{@{\extracolsep{\fill}}lccccccccccccc@{}}
\toprule
 & \textbf{Bound 2} & \multicolumn{2}{c}{\act{walk}} & \multicolumn{2}{c}{\act{run}} & \multicolumn{2}{c}{\act{stand}} & \multicolumn{2}{c}{\act{lateral}} & \multicolumn{2}{c}{\act{flexion}} & \multicolumn{2}{c}{\act{jump}} \\
\cmidrule(lr){3-4}\cmidrule(lr){5-6}\cmidrule(lr){7-8}\cmidrule(lr){9-10}\cmidrule(lr){11-12}\cmidrule(lr){13-14}
\textbf{Joint} & ($\times k^d_0$) & B1 \% & B2 \% & B1 \% & B2 \% & B1 \% & B2 \% & B1 \% & B2 \% & B1 \% & B2 \% & B1 \% & B2 \% \\
\midrule
hip & 100.27 & 98 / 100 & 100 / 100 & 99 / 99 & 100 / 100 & 100 / 100 & -- & 95 / 97 & 100 / 100 & 96 / 98 & 100 / 100 & 98 / 99 & 100 / 100 \\
knee & 16.43 & 98 / 100 & 100 / 100 & 97 / 72 & 100 / 100 & 100 / 100 & -- & 97 / 89 & 100 / 100 & 97 / 100 & 100 / 100 & 98 / 99 & 100 / 100 \\
ankle & 0.77 & 96 / 93 & 100 / 0 & 92 / 53 & 100 / 0 & 100 / 100 & -- & 79 / 76 & 99 / 0 & 100 / 100 & 100 / 0 & 94 / 94 & 96 / 0 \\
\bottomrule
\end{tabular*}
\vspace{-8pt}
\end{table}

\vspace{-5pt plus 1fil}%
\paragraph{Fixed-gain interventions.}
The explicit Fixed-PD variants of Table~\ref{tab:existing_full} turn the bounds into fixed-gain interventions.
Capping only the ankle and subtalar damping at $I_{jj}/h$, removing the Bound-2 violation, cuts torque jerk to a third and raises survival and vertical-force correlation, the latter most in standing.
Doubling every stiffness prior on top of the cap closes the force gap in running, where Bound~1 binds, but not in jumping, and nearly doubles torque jerk again, as the stiffer spring is underdamped in contact at the swing damping ceiling.
No fixed pair of gains therefore satisfies both phases (Remark~\ref{cor:sensitivity}), whereas \projname{} does so at a third of the capped variant's torque jerk.

\pagebreak%
\vspace{-6.5pt}\paragraph{Gain schedule.}
\label{App:controller}
Frozen as a subject-mean, phase-indexed schedule that drives a retrained position network, the learned gains keep part of their fidelity advantage, with joint-angle error, vertical-force correlation and torque jerk between those of Fixed-PD and \projname{} (Table~\ref{tab:existing_full}).
The schedule nevertheless survives less often than Fixed-PD ($88.5$ against $93.0\,\%$), most clearly in standing, flexion and jumping. The learned gains thus depend on the state and not on the phase alone.

\vspace{-0.85em}
\subsubsection{Learned Impedance as an Encoder of Motor Control}
\label{App:learned_impedance}
\vspace{-0.6em}

\begin{figure}[t]
    \centering
    \includegraphics[width=\linewidth]{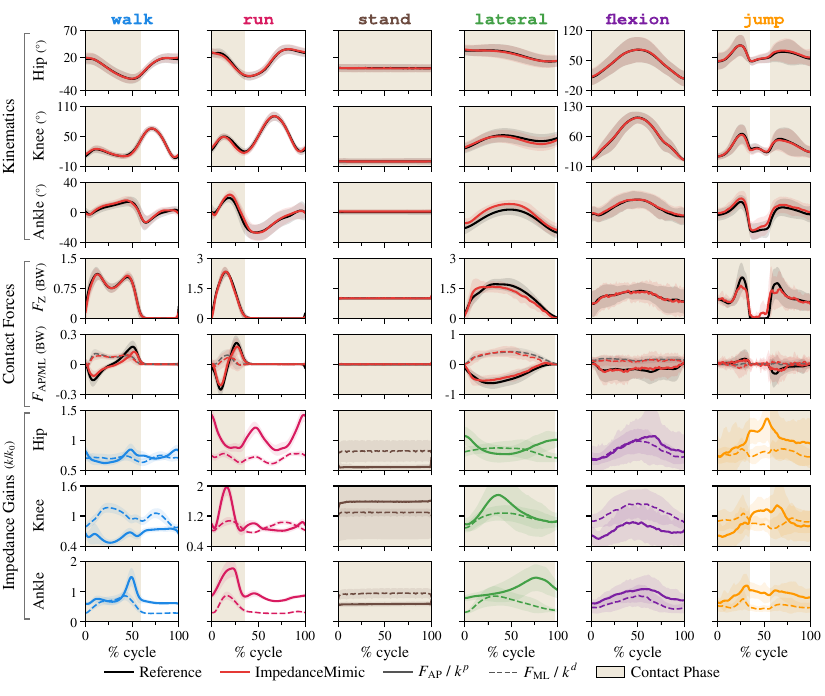}
    \vspace{-1.6em}%
    \caption{\textbf{Reference and \projname{} trajectories across the six activities.} Cycle-normalized hip, knee and ankle angles, ground reaction forces and impedance gains relative to nominal. Solid lines denote $F_{\mathrm{AP}}$ and $k^p$, dashed lines $F_{\mathrm{ML}}$ and $k^d$. Bands are mean $\pm$ 1 sd over held-out cycles, shading marks contact, and panels without tick labels share the \act{walk} y-axis.}%
    \vspace{-0.6em}%
    \label{fig:waveforms}
    \vspace{-10.5pt}%
\end{figure}

To characterize what the learned impedance encodes, we segment the held-out rollouts into movement cycles, resample each to 101 samples and express its gains as multiples of the subject's nominal $k_0$.
Walking and running cycles span consecutive ipsilateral heel strikes, standing uses quiet windows and lateral cutting its contact phases, flexion cycles run from descent onset to the return, and jump cycles are centered on flight (take-off at $35\,\%$, landing at $55\,\%$).
On the held-out clips of the 129 training subjects, the five seen-split policies yield 16{,}867--16{,}874 cycles and the unseen-split policy (seed 0) 13{,}282 from the 22 held-out subjects.
Fig.~\ref{fig:waveforms} averages the joint angles, GRF and gains over these cycles, and Table~\ref{tab:gain_units} reduces the gains to phase means and peaks.
Table~\ref{tab:gain_units}, Fig.~\ref{fig:pca_scree} and all standard deviations average the five seen-split seeds, and all others refer to seed 0.

\begin{table}[!b]%
\vspace{-0.4em}\vspace{-7.5pt}
\centering
\caption{\textbf{Normalized learned joint gains per activity.} Contact and swing columns are cycle-median phase means. The $k^p/k^p_0$ peak is the maximum of the median waveform, the $k^p/m$ peak the median of per-cycle maxima, with body mass $m$. Means over 5 seeds. Standing has no swing phase.}
\label{tab:gain_units}
\vspace{-0.3em}%
\scriptsize
\setlength{\tabcolsep}{3pt}
\renewcommand{\arraystretch}{0.80}%
\begin{tabular*}{\textwidth}{@{\extracolsep{\fill}}ll ccc cc cc cc@{}}
\toprule
 & & \multicolumn{3}{c}{\textbf{\boldmath$k^p/k^p_0$}} & \multicolumn{2}{c}{\textbf{\boldmath$k^d/k^d_0$}} & \multicolumn{2}{c}{\textbf{\boldmath$k^p/m$}} & \multicolumn{2}{c}{\textbf{\boldmath$k^d/m$}} \\
 & & & & & & & \multicolumn{2}{c}{(Nm/rad/kg)} & \multicolumn{2}{c}{(Nm\,s/rad/kg)} \\
\cmidrule(lr){3-5}\cmidrule(lr){6-7}\cmidrule(lr){8-9}\cmidrule(lr){10-11}
\texttt{\textbf{Activity}} & \textbf{Joint} & contact & swing & peak & contact & swing & contact & peak & contact & swing \\
\midrule
\multirow{3}{*}{\act{walk}} & hip & 0.70 & 0.73 & 0.83 & 0.71 & 0.69 & 2.9 & 3.6 & 0.29 & 0.28 \\
 & knee & 0.57 & 0.73 & 0.75 & 1.02 & 0.91 & 2.3 & 3.3 & 0.41 & 0.37 \\
 & ankle & 0.87 & 0.67 & 1.54 & 0.66 & 0.29 & 2.6 & 5.1 & 0.20 & 0.09 \\
\addlinespace[2pt]
\multirow{3}{*}{\act{run}} & hip & 1.00 & 1.04 & 1.44 & 0.70 & 0.69 & 3.8 & 5.6 & 0.27 & 0.27 \\
 & knee & 1.29 & 0.92 & 1.94 & 0.98 & 0.90 & 4.9 & 7.6 & 0.37 & 0.34 \\
 & ankle & 1.43 & 0.82 & 1.88 & 0.63 & 0.30 & 4.3 & 5.9 & 0.19 & 0.09 \\
\addlinespace[2pt]
\multirow{3}{*}{\act{stand}} & hip & 0.56 & -- & 0.64 & 0.87 & -- & 2.0 & 2.4 & 0.30 & -- \\
 & knee & 1.07 & -- & 1.13 & 1.06 & -- & 4.1 & 4.8 & 0.41 & -- \\
 & ankle & 0.59 & -- & 0.65 & 0.93 & -- & 1.7 & 1.9 & 0.26 & -- \\
\addlinespace[2pt]
\multirow{3}{*}{\act{lateral}} & hip & 0.86 & 0.92 & 1.08 & 0.79 & 0.69 & 3.8 & 5.1 & 0.35 & 0.31 \\
 & knee & 1.16 & 1.04 & 1.49 & 0.92 & 0.84 & 5.5 & 7.9 & 0.45 & 0.42 \\
 & ankle & 1.11 & 1.04 & 1.47 & 0.61 & 0.38 & 3.4 & 5.0 & 0.20 & 0.12 \\
\addlinespace[2pt]
\multirow{3}{*}{\act{flexion}} & hip & 0.90 & 0.80 & 1.03 & 0.85 & 0.71 & 3.6 & 5.7 & 0.34 & 0.30 \\
 & knee & 0.72 & 0.73 & 0.82 & 1.05 & 0.84 & 3.1 & 5.7 & 0.45 & 0.35 \\
 & ankle & 0.91 & 0.66 & 1.05 & 0.64 & 0.31 & 2.8 & 4.6 & 0.20 & 0.09 \\
\addlinespace[2pt]
\multirow{3}{*}{\act{jump}} & hip & 0.78 & 1.07 & 1.16 & 0.77 & 0.78 & 3.5 & 5.9 & 0.34 & 0.32 \\
 & knee & 0.76 & 0.97 & 1.10 & 0.94 & 0.82 & 3.6 & 7.0 & 0.45 & 0.38 \\
 & ankle & 0.81 & 0.92 & 1.23 & 0.53 & 0.39 & 2.7 & 4.9 & 0.20 & 0.13 \\
\bottomrule
\end{tabular*}
\vspace{-1.9em}
\end{table}

\vspace{-7pt}
\paragraph{Per-activity schedules.}
Each activity has a distinct gain schedule (Table~\ref{tab:gain_units}).
Standing holds hip and ankle stiffness below nominal and knee stiffness slightly above it.
Walking modulates the ankle most, stiffening it from $0.6\times$ nominal at heel strike to its push-off peak ($49\,\%$ of the cycle) and releasing its damping in swing, and running stiffens the knee and ankle during impact absorption (first $20\,\%$ of the cycle), with the same swing release.
Flexion ramps stiffness with depth while raising ankle damping, jumping stiffens at push-off and landing and otherwise stays below nominal, and lateral cutting stiffens the hip at foot strike and the knee and ankle through mid- and late contact.

\vspace{-0.2em}
\paragraph{A low-dimensional code of the activity.}
Two principal components carry most of the variance of the gain cycles (Fig.~\ref{fig:pca_scree}a), and a nested decomposition over all ${\sim}16{,}870$ cycles attributes nearly all of the first to activity rather than subject or speed (Fig.~\ref{fig:pca_scree}b).
In the 10-PC space, activity explains $\eta^2 = 0.52 \pm 0.03$ of the variance, well above Cohen's threshold of $0.14$ for a large effect.
A logistic regression on 10~PCs with leave-subject-out cross-validation, trained on a balanced subsample of 34 cycles per activity, recovers the activity of a single cycle (Fig.~\ref{fig:pca_scree}c), confusing mainly flexion with jumping and lateral cutting, which share its deep knee flexion and high ground forces.
Subject identity within walking remains recoverable at $70.4 \pm 1.0\,\%$, $27$ times the chance level of $2.6\,\%$.
Morphology is carried chiefly by the inertia scaling of the nominal gains, by which the features are normalized, and the modulation about them encodes the activity far more strongly than the subject.

\begin{figure*}[t]%
\centering
\newcommand{\sd}[1]{{\tiny\textcolor{black!55}{$\pm$#1}}}
\begin{minipage}[t]{0.285\textwidth}\vspace{0pt}
  \centering
  \includegraphics{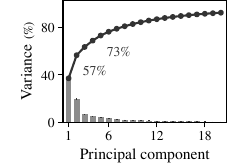}
\end{minipage}\hfill
\begin{minipage}[t]{0.345\textwidth}\vspace{0pt}
  \centering
  {\scriptsize\setlength{\tabcolsep}{1.5pt}%
  \begin{tabular}{@{}ccccc@{}}
  \toprule
  \textbf{PC} & \textbf{Activity} & \textbf{Dataset} & \textbf{Subject} & \textbf{Speed} \\
  \midrule
  1 & 89.3\sd{.5} & 0.6\sd{.1} & 4.0\sd{.3} & 2.3\sd{.1} \\
  2 & 61.1\sd{3.0} & 8.3\sd{1.0} & 17.9\sd{1.7} & 0.8\sd{.3} \\
  3 & 21.9\sd{3.5} & 36.2\sd{5.8} & 16.2\sd{2.4} & 0.2\sd{.3} \\
  4 & 16.3\sd{7.3} & 8.6\sd{5.5} & 27.5\sd{3.8} & 10.0\sd{5.3} \\
  5 & 24.8\sd{12.5} & 5.4\sd{3.5} & 20.1\sd{7.1} & 2.5\sd{4.0} \\
  \bottomrule

  \end{tabular}}\\[2pt]
  {\scriptsize Mean $\pm$ sd over five seeds.}%
\end{minipage}\hfill
\begin{minipage}[t]{0.345\textwidth}\vspace{0pt}
  \centering
  {\scriptsize\setlength{\tabcolsep}{3.5pt}%
  \begin{tabular}{@{}lccc@{}}
  \toprule
  \texttt{\textbf{Activity}} & \textbf{Recall} & \textbf{Precision} & \textbf{F1} \\
  \midrule
  \act{walk} & 98.8\sd{1.6} & 97.1\sd{2.0} & 98.0\sd{1.6} \\
  \act{run} & 99.4\sd{1.3} & 97.1\sd{2.0} & 98.3\sd{1.6} \\
  \act{stand} & 100.0\sd{.0} & 99.4\sd{1.3} & 99.7\sd{.6} \\
  \act{lateral} & 92.3\sd{5.7} & 92.6\sd{8.3} & 92.4\sd{6.9} \\
  \act{flexion} & 78.8\sd{9.2} & 82.9\sd{9.9} & 80.7\sd{9.2} \\
  \act{jump} & 85.3\sd{8.1} & 85.3\sd{6.3} & 85.2\sd{6.8} \\
  \midrule
  \textbf{Macro} & 92.5\sd{3.3} & 92.4\sd{3.5} & 92.4\sd{3.4} \\
  \bottomrule
  \end{tabular}}
\end{minipage}\\[0.3em]
\makebox[0.285\textwidth]{\hspace*{0.38in}\footnotesize (a)}\hfill\makebox[0.345\textwidth]{\footnotesize (b)}\hfill\makebox[0.345\textwidth]{\footnotesize (c)}
\vspace{-0.1em}
\caption{\textbf{A low-dimensional code of the activity.} Principal components (PCs) of the standardized hip, knee and ankle gain cycles. \textbf{(a)}~Variance explained. \textbf{(b)}~Variance of each PC attributed to each factor by nested ANOVA (\%). \textbf{(c)}~Leave-subject-out activity classification from 10~PCs (\%).}
\label{fig:pca_scree}
\vspace{-14pt}%
\end{figure*}

\vspace{-4pt}
\paragraph{Unseen subjects.}
The unseen-split cycles are standardized and projected with the scaler and 10-component basis of each seen-split seed (Fig.~\ref{fig:pca_scree}a).
The nearest seen-activity centroid recovers $83$--$100\,\%$ of the unseen walking, running and standing cycles but only $40$--$75\,\%$ of the lateral, flexion and jumping cycles (means over five bases).
A logistic regression trained on all seen cycles classifies $99.3 \pm 0.1\,\%$ of the unseen cycles correctly, with a macro recall of $79.0 \pm 4.5\,\%$, and balanced training on 20 draws of 34 seen cycles per activity raises it to $81.3 \pm 2.5\,\%$, chiefly through jumping.
The gap to the $92.5\,\%$ reached on held-out clips of seen subjects thus arises almost entirely from the three activities with the fewest training cycles (48 lateral, 121 flexion and 38 jumping).

\vspace{-4pt}
\paragraph{Load scaling and anticipatory damping.}
Stiffness scales with load, the contact-mean ankle $k^p/k^p_0$ increasing linearly with the cycle's peak vertical GRF across all six activities ($r = 0.93$, Fig.~\ref{fig:encoder}c).
Damping anticipates unloading, the mean ankle $k^d/k^d_0$ waveform peaking $196 \pm 8$, $136 \pm 3$ and $128 \pm 10$\,ms before toe-off or take-off in walking, running and jumping, and every individual walking cycle peaks before toe-off (11{,}533 cycles in each seed).
By toe-off the ankle damping has fallen to $0.34\times$ nominal in walking and $0.38\times$ in running, and since the realized gains include the $50$\,ms lag of Eq.~\ref{eq:gain_lag}, the commanded lead is about $50$\,ms longer.
Within contact, ankle damping co-varies with stiffness, as Remark~\ref{cor:sensitivity} prescribes (median $r = 0.88$ in running, $0.39$ in walking, $0.45$ in flexion and $0.30$ in jumping), and in swing it remains below Bound~2 on $99.99\,\%$ of 764{,}695 frames.

\vspace{-4pt}%
\paragraph{Withheld activities.}
The timing of the damping release transfers to a withheld activity, whereas the magnitude of the gain modulation does not.
On the walking clips it never trained on, the leave-walking-out policy (seed 0) yields 28{,}653 cycles from 21 subjects of \citet{moore2015elaborate} and \citet{wang2023wearable}.
Against the seen-split policy (seed 0, 11{,}533 cycles), its median ankle $k^d/k^d_0$ waveform correlates at $r = 0.99$, peaks at the same $42\,\%$ of the cycle, precedes toe-off in $98\,\%$ of cycles against $100\,\%$ and settles at $0.30\times$ nominal in swing against $0.29\times$, but reaches only $66\,\%$ of the seen-split peak ($0.57\times$ against $0.87\times$ nominal).
Its $k^p/k^p_0$ waveform correlates at $r = 0.85$ and rises from $0.73\times$ at heel strike to $1.14\times$ at push-off, against $0.60\times$ to $1.51\times$, or $46\,\%$ of the seen-split push-off rise.
Within walking, contact stiffness correlates with peak GRF at $r = 0.11$, against $0.40$.

\vspace{-4pt}%
\paragraph{Comparison with human impedance.}
\label{App:identifiability}
Over the contact window of \citet{rouse2014estimation}, learned ankle stiffness rises from $2.3$ to $3.0$\,Nm/rad/kg, against $1.5$ to $6.5$\,Nm/rad/kg measured.
Quiet standing holds $0.4$ of the toppling stiffness ($mgh = 8.4$\,Nm/rad/kg for the 61 standing subjects, CoM $0.86$\,m above the ankle), against about $0.9$ measured in humans~\citep{loram2002direct}, and the running peak exceeds the walking one, as for a leg spring that stiffens with speed~\citep{mcmahon1990mechanics, farley1996leg}.
This comparison holds only where the spring term carries the torque, and decomposing $\tau = k^p(q^* - q) + k^p\alpha\,\delta q + k^d(\dot q^* - \dot q)$ over 11{,}533 walking cycles shows that the feedforward lever $k^p\alpha\,\delta q$ supplies $86\,\%$ of the mid-contact torque and $109\,\%$ at push-off (offset at $70\,\%$ of its authority), so the push-off peak reflects the lever, not a stiffness about the reference.
Damping leads unloading as in human data~\citep{lee2015timevarying} but not in magnitude, which Bound~2 caps. A model with one torque per joint has no antagonist muscles, and kinematics and GRF cannot identify the co-activation by which humans raise joint stiffness~\citep{hogan1984adaptive}.

\vspace{-0.3em}
\subsubsection{Physical Consistency: GaitDynamics vs.\ ImpedanceMimic}
\label{App:physics_realisticness}
\vspace{-0.2em}

\begin{figure}[!b]%
  \centering
  \includegraphics[width=\linewidth]{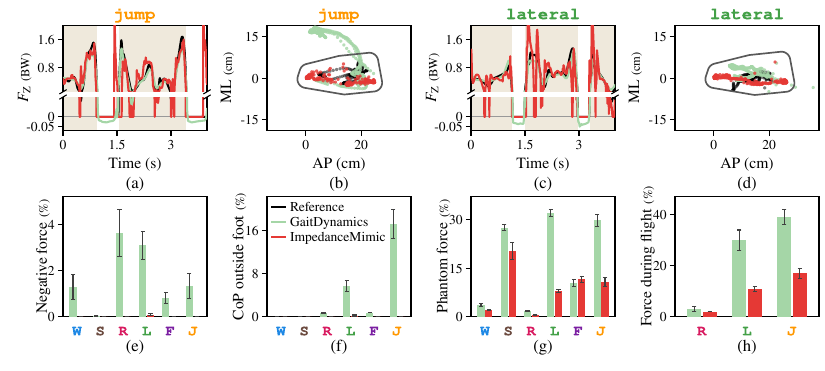}
  \par\vspace{-8pt}%
  \caption{\textbf{Physical consistency of GaitDynamics and \projname{} forces.} Unseen subjects. \textbf{(a--d)}~One foot's vertical force (a, c) and center of pressure in its outline (b, d), seed 0, reference contact shaded, zero band magnified. \textbf{(e--h)}~Violation rates on 1{,}365 clips, mean $\pm$ sd of five seeds.}
  \label{fig:plausibility}
\vspace{-0.7em}%
\end{figure}

GaitDynamics regresses the force-plate distribution of its training clips from joint-angle windows, without a body, a contact model or momentum, whereas the forces of \projname{} arise from the contact model of a physical simulation.
The regressor attains the higher vertical-force correlation on every activity of the unseen split (Table~\ref{tab:golden}), yet agreement with the plates does not establish that a force is physically realizable.
We therefore score both methods with identical per-frame metrics on the unseen split, where neither has seen the subject, using the predictions of GaitDynamics and the rollout contact forces of \projname{} (mean $\pm$ sd over five seeds).
Figure~\ref{fig:plausibility}a--d shows held-out examples, with the center of pressure drawn in the contact-model foot outline (Fig.~\ref{fig:contact_model}).

\vspace{-0.5em}%
\paragraph{Structural constraints.}
Two constraints hold for any physical contact.
Contact is unilateral, so the vertical force cannot pull on the ground, which GaitDynamics violates on $1.2 \pm 0.2\,\%$ of foot-frames ($3.6 \pm 1.0\,\%$ in running), whereas \projname{} does so only in numerical contact transients, on at most $0.09 \pm 0.04\,\%$ of frames (lateral, Fig.~\ref{fig:plausibility}e).
The center of pressure must lie beneath the foot, which GaitDynamics violates on $5.7 \pm 1.0\,\%$ of loaded lateral and $17.2 \pm 2.7\,\%$ of loaded jump frames, against $0.2 \pm 0.1\,\%$ of lateral frames and none elsewhere for \projname{} (Fig.~\ref{fig:plausibility}f).

\vspace{-0.5em}%
\paragraph{Contact timing.}
Two metrics capture when contact occurs: \emph{phantom force}, a force above $0.05$\,BW on a foot the plates show unloaded, and \emph{force during flight}, a total force above $0.1$\,BW while both feet are measured airborne (scored on activities with more than 100 flight frames).
Away from steady gait, GaitDynamics produces phantom force on $28$--$32\,\%$ of unloaded foot-frames in standing, lateral and jumping, against $8$--$20\,\%$ for \projname{} (Fig.~\ref{fig:plausibility}g), whereas flexion shows comparable rates ($10.5\,\%$ and $11.6\,\%$).
During flight, the regressor loads the ground on $30 \pm 4\,\%$ of lateral and $39 \pm 3\,\%$ of jump frames, against $11 \pm 1\,\%$ and $17 \pm 2\,\%$ for \projname{} (Fig.~\ref{fig:plausibility}h).
On walking and running both methods are nearly consistent, with phantom force below $4\,\%$ and $2$--$3\,\%$ of running flight frames loaded.
\projname{}'s residual errors concern the timing of contact rather than its physical validity, with missed contact (below $0.02$\,BW on a foot the plates load above $0.2$\,BW) on up to $13\,\%$ of loaded foot-frames in jumping, where the regressor misses almost none.

\vspace{-8pt}%
\subsection{Training Details and Compute}
\label{app:training}
\vspace{-2.5pt}%

\paragraph{Training.}
Fixed-PD, Torque, Muscle and \projname{} train with PPO for $2 \times 10^9$ steps on 8{,}192 parallel environments in MJWarp~\citep{mjwarp2025} with the hyperparameters of Table~\ref{tab:hyperparams} (Muscle exceptions in App.~\ref{App:muscle}), and PHC~\citep{luo2023perpetual} keeps its own hyperparameters, actor and AMP discriminator for the same budget (App.~\ref{App:baselines}).
Each controller is trained on the eight benchmark splits (seen, unseen and six withheld activities) with five seeds, for 40 runs per method, and the GaitDynamics~\citep{tan2026gaitdynamics} transformer (2\,M parameters) trains for 7{,}680 epochs with Adan in about 15\,min per run, three runs sharing one H200 GPU.

\vspace{-6.5pt}%
\paragraph{Compute.}
Reinforcement-learning runs train three at a time per GPU under MPS, and Table~\ref{tab:compute} reports their cost on an NVIDIA RTX PRO 6000.
\projname{} keeps the throughput of Fixed-PD and Torque despite three times as many actions, whereas Muscle and PHC run about $30$ and $60\,\%$ slower per process.
The benchmark totals 240 runs and ${\sim}550$\,\mbox{GPU-h}, and the appendix experiments 508 runs and ${\sim}870$\,\mbox{GPU-h}.
MuscleMimic~\citep{li2026towards}, by comparison, reports 13\,k steps/s on an H100 at the same environment count, simulator and control rate, with 416 Hill-type muscles against our 80 and a 50\,M-parameter policy against our $0.28$\,M.

\vspace{2.5pt}%
\begin{table}[H]
\vspace{-0.8em}
\centering
\caption{\textbf{Network size and training cost per method.} 2\,B steps per run, 3 runs packed per GPU.}
\vspace{-0.4em}
\label{tab:compute}
\footnotesize
\renewcommand{\arraystretch}{0.9}
\setlength{\tabcolsep}{3pt}
\begin{tabular*}{\textwidth}{@{\extracolsep{\fill}}lcccccc@{}}
\toprule
\textbf{Method} & \textbf{Actions} & \textbf{Policy parameters} & \textbf{Steps/s per process} & \textbf{Hours per run} & \textbf{Runs} & \textbf{GPU-h} \\
\midrule
Fixed-PD & 15 & 0.28\,M & 139\,k & 4.1 & 40 & 55 \\
Torque & 15 & 0.28\,M & 141\,k & 4.0 & 40 & 54 \\
Muscle & 83 & 0.33\,M & 103\,k & 5.5 & 40 & 73 \\
PHC & 15 & 1.09\,M & 53\,k & 10.6 & 40 & 141 \\
SimPoE & 51 & 2.46\,M & 42\,k & 13.3 & 40 & 177 \\
ImpedanceMimic & 45 & 0.28\,M & 143\,k & 4.0 & 40 & 53 \\
\bottomrule
\end{tabular*}
\vspace{-1.5em}
\end{table}

\begin{figure}[H]%
\vspace{-7.5pt}%
\centering
\includegraphics[width=\textwidth]{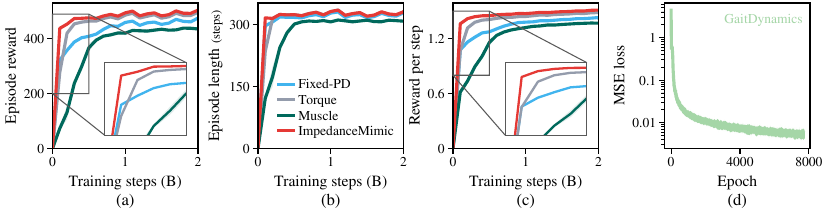}
\vspace{-1.5em}
\caption{\textbf{Training curves.} \textbf{(a)}~Episode reward, \textbf{(b)}~length and \textbf{(c)}~reward per step. (a,~c) magnify the first 0.5\,B steps. PHC and SimPoE are not included, as PHC trains on its own reward and SimPoE on 200-step episodes with a residual root wrench. \textbf{(d)}~GaitDynamics MSE loss.}
\label{fig:training_curves}
\vspace{-13pt}%
\end{figure}

\vspace{-6.5pt}%
\paragraph{Training curves.}
On the seen split (Fig.~\ref{fig:training_curves}, mean $\pm$ sd of five seeds), \projname{} converges fastest, near its plateau by $0.5 \times 10^9$ steps.
Torque starts lower but closes most of the gap, Fixed-PD finishes lowest of the three, and all three still improve at $2 \times 10^9$ steps.
Muscle plateaus lowest, near $10^9$ steps.
PHC and SimPoE act at 33\,Hz, and each of their steps simulates 30\,ms instead of 10\,ms.

\vspace{-10pt}\vspace{0pt}%
\subsection{Ablations, Sensitivity and Robustness}
\label{App:ablation}
\vspace{-6pt}%

\paragraph{Existing controllers.}\label{App:existing}
PHC, whose reward tracks body positions and orientations and carries no force term, reaches $11.4$ to $15.7^\circ$ of joint-angle error, against at most $4.16^\circ$ for the PD rows, and its vertical-force correlation is the lowest of all rows on every benchmark, except for uncapped Fixed-PD on seen subjects and SimPoE without its wrench on withheld activities.
Its port approaches the reported PHC on AMASS (Table~\ref{tab:phc_amass}), which places this gap in the setting rather than in the implementation.
Its error is largest in hip flexion, knee and lumbar extension ($13.1$ to $13.6^\circ$, against at most $2.4^\circ$ for \projname{}), whereas its pelvis drifts least of all methods, tied only with SimPoE (Table~\ref{tab:existing_full}), consistent with a reward and termination on body positions rather than the joint angles scored here.
With our full reward in place of its own, its joint-angle error falls to $8.7$, $10.8$ and $13.3^\circ$ and its vertical-force correlation rises by $.07$ to $.10$, still below every \projname{} row (Tables~\ref{tab:existing_full} and~\ref{tab:existing_transfer}).
SimPoE survives $96.4$, $96.2$ and $93.0\,\%$ of episodes on the three benchmarks, more often than \projname{} under stable PD, chiefly in flexion, and on withheld activities also in walking and standing (Tables~\ref{tab:existing_full} and~\ref{tab:existing_transfer}).
Its joint-angle error is larger in every activity of every benchmark except standing on seen subjects, its vertical-force correlation is lower on every benchmark, and in standing its mediolateral-force correlation reaches only $0.52$ against $0.82$, consistent with a residual wrench that balances the body instead of its feet.
Without the wrench, SimPoE's survival falls to $84.8$, $79.6$ and $55.8\,\%$, most in flexion and jumping, down to $27.1\,\%$ in withheld walking, and its joint-angle error rises to $3.53$, $4.38$ and $5.69^\circ$.
The wrench carries no net body weight, its signed vertical mean staying within $1.5\,\%$\,BW, yet supplies $51$ to $86\,\%$ of the mediolateral and $45$ to $69\,\%$ of the anterior--posterior force, and remains active in $96$ to $100\,\%$ of airborne frames, when no contact can act (Table~\ref{tab:simpoe_wrench}).
SimPoE therefore balances through the pelvis, whereas every other method must balance through its feet.
Retraining Fixed-PD and \projname{} under stable PD and implicit damping separates the integration scheme from the learned gains (Tables~\ref{tab:existing}, \ref{tab:existing_full} and~\ref{tab:existing_transfer}).
With the same gain priors, \projname{} leads Fixed-PD in joint-angle error and{\parfillskip=0pt\par}%

\begin{table}[t]%
\centering
\caption{\textbf{Residual root wrench of SimPoE.} Mean$_{\pm\text{sd}}$ over 5 seeds. Root forces (\%\,BW) along and moments (Nm/kg) about the vertical (Z), mediolateral (ML) and anteroposterior (AP) pelvis axes, as frame means. \emph{Root share} = proportion (\%) of the force on the body applied by the root. \emph{No contact} = share of airborne frames with a root force above $10\,\%$\,BW or moment above $1$\,Nm/kg.}
\vspace{-0.4em}
\label{tab:simpoe_wrench}
\renewcommand{\arraystretch}{0.82}%
\setlength{\tabcolsep}{2pt}
\newcommand{\sd}[1]{{\tiny\textcolor{black!55}{$\pm$#1}}}
\scriptsize
\begin{tabular}{@{}>{\fillL}p{1.40in} ccc p{4pt} ccc p{4pt} ccc p{4pt} >{\fillR}c@{}}
\toprule
 & \multicolumn{3}{c}{Force (\%\,BW)} & & \multicolumn{3}{c}{Moment about (Nm/kg)} & & \multicolumn{3}{c}{Root share (\%)} & & No contact \\
\cmidrule(lr){2-4}\cmidrule(lr){6-8}\cmidrule(lr){10-12}\cmidrule(lr){14-14}
\texttt{\textbf{Activity}} & Z & ML & AP & & Z & ML & AP & & Z & ML & AP & & frames (\%) \\
\specialrule{\lightrulewidth}{\aboverulesep}{0pt}
\rowcolor{papericeblue}
\multicolumn{14}{@{}>{\fillL}l}{\padT\textbf{Seen subjects, held-out clips}} \\[1pt]
\act{walk} & 10.3\sd{1.0} & 16.7\sd{1.1} & 12.0\sd{1.1} &  & .40\sd{.03} & .55\sd{.03} & .64\sd{.02} &  & 8\sd{1} & 64\sd{1} & 57\sd{2} &  & -- \\
\act{run} & 12.5\sd{1.3} & 19.1\sd{1.0} & 22.4\sd{2.0} &  & .66\sd{.04} & .97\sd{.08} & .96\sd{.05} &  & 11\sd{1} & 76\sd{1} & 69\sd{2} &  & 99\sd{0} \\
\act{stand} & 1.3\sd{.2} & 7.8\sd{.9} & 1.9\sd{.2} &  & .16\sd{.04} & .17\sd{.01} & .59\sd{.03} &  & 1\sd{0} & 86\sd{1} & 46\sd{3} &  & -- \\
\act{lateral} & 13.1\sd{1.0} & 13.0\sd{.4} & 12.6\sd{.6} &  & .56\sd{.02} & .87\sd{.06} & .92\sd{.03} &  & 12\sd{1} & 51\sd{1} & 51\sd{1} &  & 98\sd{1} \\
\act{flexion} & 7.2\sd{.4} & 15.9\sd{1.3} & 8.6\sd{.3} &  & .43\sd{.01} & .46\sd{.02} & .64\sd{.02} &  & 6\sd{0} & 71\sd{2} & 51\sd{1} &  & -- \\
\act{jump} & 9.7\sd{.4} & 13.0\sd{.3} & 10.9\sd{.5} &  & .42\sd{.02} & .64\sd{.01} & .85\sd{.04} &  & 9\sd{0} & 63\sd{1} & 56\sd{1} &  & 97\sd{2} \\
\rowcolor{meangray} \textbf{Mean} & 9.0\sd{.5} & 14.2\sd{.5} & 11.4\sd{.7} &  & .44\sd{.01} & .61\sd{.02} & .77\sd{.01} &  & 8\sd{0} & 68\sd{1} & 55\sd{1} &  & -- \\
\specialrule{\lightrulewidth}{\aboverulesep}{0pt}
\rowcolor{papericeblue}
\multicolumn{14}{@{}>{\fillL}l}{\padT\textbf{Unseen subjects, zero-shot}} \\[1pt]
\act{walk} & 9.2\sd{.9} & 15.1\sd{.5} & 12.0\sd{.9} &  & .43\sd{.03} & .55\sd{.03} & .66\sd{.02} &  & 7\sd{1} & 61\sd{1} & 54\sd{2} &  & -- \\
\act{run} & 15.5\sd{1.4} & 21.2\sd{1.9} & 24.6\sd{3.2} &  & .71\sd{.05} & 1.16\sd{.05} & 1.11\sd{.14} &  & 13\sd{1} & 79\sd{2} & 66\sd{3} &  & 100\sd{0} \\
\act{stand} & 1.2\sd{.0} & 7.7\sd{2.7} & 1.6\sd{.2} &  & .14\sd{.03} & .17\sd{.01} & .53\sd{.02} &  & 1\sd{0} & 84\sd{4} & 45\sd{3} &  & -- \\
\act{lateral} & 14.1\sd{.9} & 13.3\sd{.9} & 12.4\sd{.5} &  & .59\sd{.03} & .98\sd{.03} & .97\sd{.02} &  & 13\sd{1} & 51\sd{1} & 53\sd{2} &  & 99\sd{1} \\
\act{flexion} & 6.6\sd{.4} & 19.2\sd{3.6} & 10.2\sd{.8} &  & .37\sd{.01} & .50\sd{.02} & .71\sd{.06} &  & 6\sd{0} & 72\sd{4} & 54\sd{1} &  & -- \\
\act{jump} & 9.3\sd{.3} & 13.5\sd{.8} & 11.3\sd{.2} &  & .45\sd{.03} & .68\sd{.02} & .83\sd{.04} &  & 9\sd{0} & 64\sd{2} & 55\sd{1} &  & 97\sd{1} \\
\rowcolor{meangray} \textbf{Mean} & 9.3\sd{.3} & 15.0\sd{1.2} & 12.0\sd{.7} &  & .45\sd{.01} & .67\sd{.02} & .80\sd{.02} &  & 8\sd{0} & 69\sd{2} & 54\sd{1} &  & -- \\
\specialrule{\lightrulewidth}{\aboverulesep}{0pt}
\rowcolor{papericeblue}
\multicolumn{14}{@{}>{\fillL}l}{\padT\textbf{Withheld activity, zero-shot}} \\[1pt]
\act{walk} & 11.5\sd{1.1} & 16.4\sd{1.6} & 16.4\sd{1.8} &  & .44\sd{.04} & .78\sd{.05} & 1.16\sd{.16} &  & 10\sd{1} & 66\sd{3} & 54\sd{2} &  & -- \\
\act{run} & 17.5\sd{1.9} & 18.3\sd{1.4} & 17.2\sd{1.1} &  & .56\sd{.02} & 1.00\sd{.09} & 1.22\sd{.10} &  & 14\sd{1} & 72\sd{1} & 59\sd{2} &  & 100\sd{0} \\
\act{stand} & 4.7\sd{.3} & 7.8\sd{.7} & 4.7\sd{.4} &  & .31\sd{.03} & .47\sd{.03} & .48\sd{.06} &  & 4\sd{0} & 63\sd{3} & 54\sd{1} &  & -- \\
\act{lateral} & 12.7\sd{.9} & 12.7\sd{.8} & 12.2\sd{.1} &  & .50\sd{.03} & .81\sd{.02} & .90\sd{.02} &  & 12\sd{1} & 51\sd{1} & 52\sd{1} &  & 96\sd{1} \\
\act{flexion} & 7.5\sd{.5} & 13.0\sd{1.9} & 9.1\sd{1.0} &  & .40\sd{.02} & .54\sd{.05} & .74\sd{.05} &  & 7\sd{0} & 65\sd{3} & 51\sd{2} &  & -- \\
\act{jump} & 9.3\sd{.8} & 14.0\sd{.4} & 11.4\sd{.1} &  & .42\sd{.02} & .61\sd{.06} & .84\sd{.03} &  & 9\sd{1} & 66\sd{2} & 57\sd{1} &  & 98\sd{0} \\
\rowcolor{meangray} \textbf{Mean} & 10.6\sd{.4} & 13.7\sd{.4} & 11.8\sd{.6} &  & .44\sd{.01} & .70\sd{.03} & .89\sd{.02} &  & 9\sd{0} & 64\sd{0} & 55\sd{1} &  & -- \\
\bottomrule
\end{tabular}
\vspace{-12pt}%
\end{table}

\vspace{-4pt plus 1fil}%
\begin{table}[H]%
\centering
\caption{%
  \textbf{Existing controllers and PD integration on seen subjects, held-out clips, per activity.} Mean$_{\pm\text{sd}}$ over 5 seeds.
  (S) = stable PD, (I) = implicit damping, unmarked = explicit PD (App.~\ref{App:baselines}).
  Columns are RMSE ($\downarrow$) unless marked $\uparrow$.
  \textbf{Bold} / \underline{underline} = best / second best per activity.}
\vspace{-0.2em}%
\label{tab:existing_full}
\renewcommand{\arraystretch}{0.72}
\setlength{\tabcolsep}{2pt}
\newcommand{\gap}{\hspace{4pt}}
\newcommand{\sd}[1]{{\tiny\textcolor{black!55}{$\pm$#1}}}
\scriptsize
\resizebox{\textwidth}{!}{%
\begin{tabular}{@{}>{\fillL}l @{\gap} l @{\gap} lllll @{\gap} lp{0.75cm}p{0.75cm}p{0.75cm}ll @{\gap} ll>{\fillR}l @{}}
\toprule
 & & \multicolumn{1}{c}{\textbf{Stability}} & \multicolumn{4}{c}{\textbf{Kinematics}} & \multicolumn{6}{c}{\textbf{Ground Reaction Forces}} & \multicolumn{3}{c}{\textbf{Torques}$^\dagger$} \\
\cmidrule(lr){3-3}\cmidrule(lr){4-7}\cmidrule(lr){8-13}\cmidrule(lr){14-16}
\texttt{\textbf{Activity}} & \textbf{Method}
  & \makecell{Ep.$\hi$ \\ (\%)}
  & \makecell{$q\lo$\\ \makebox[0pt][c]{($^\circ$)}}
  & \makecell{$\dot{q}\lo$\\ \makebox[0pt][c]{($^\circ$/s)}}
  & \makecell{$p_x\lo$\\ \makebox[0pt][c]{(m/s)}}
  & kin$_r\hi$
  & \makecell{$F_{\mathrm{Z}}\lo$\\ \makebox[0pt][c]{(BW)}}
  & $F^{\uparrow}_{\mathrm{Z}_r}$
  & $F^{\uparrow}_{\mathrm{ML}_r}$
  & $F^{\uparrow}_{\mathrm{AP}_r}$
  & \makecell{$\mathrm{CoP}^{\downarrow}_{\mathrm{ML}}$\\ {\tiny (mm)}}
  & \makecell{$\mathrm{CoP}^{\downarrow}_{\mathrm{AP}}$\\ {\tiny (mm)}}
  & \makecell{$\tau\lo$\\ \makebox[0pt][c]{\tiny (Nm/kg)}}
  & $\tau_r\hi$
  & \makecell{$\dot{\tau}\lo$\\ \makebox[0pt][c]{\tiny (Nm/kg/s)}} \\
\midrule
\multirow{18}{*}{\act{walk}} & Fixed-PD & 99.9\sd{.1} & 2.48\sd{.04} & 57.2\sd{1.9} & .022\sd{.002} & .97\sd{.00} & .13\sd{.00} & .95\sd{.00} & .77\sd{.01} & .72\sd{.00} & \textbf{20}\sd{1} & 45\sd{1} & .55\sd{.01} & .50\sd{.01} & 26.3\sd{2.0} \\
 & Fixed-PD, uncapped \smash[t]{$k^d$} & 99.0\sd{.2} & 3.89\sd{.03} & 112\sd{0} & .071\sd{.006} & .91\sd{.00} & .32\sd{.00} & .82\sd{.00} & .45\sd{.01} & .22\sd{.02} & 28\sd{1} & 56\sd{2} & 1.16\sd{.01} & .24\sd{.00} & 145\sd{2} \\
 & Fixed-PD, \smash[t]{$2\times k^p$} & 99.9\sd{.1} & 2.19\sd{.02} & 88.2\sd{15.1} & .020\sd{.001} & .97\sd{.00} & .15\sd{.01} & .94\sd{.00} & .67\sd{.02} & .64\sd{.01} & 23\sd{0} & 44\sd{1} & .84\sd{.02} & .38\sd{.00} & 97.5\sd{3.2} \\
\bottomrule
\end{tabular}%
}
\end{table}
\begin{table}[H]%
\addtocounter{table}{-1}%
\centering
\caption{\textbf{Existing controllers and PD integration, per activity (continued).}}
\vspace{-0.2em}%
\renewcommand{\arraystretch}{0.72}
\setlength{\tabcolsep}{2pt}
\newcommand{\gap}{\hspace{4pt}}
\newcommand{\sd}[1]{{\tiny\textcolor{black!55}{$\pm$#1}}}
\scriptsize
\resizebox{\textwidth}{!}{%
\begin{tabular}{@{}>{\fillL}l @{\gap} l @{\gap} lllll @{\gap} lp{0.75cm}p{0.75cm}p{0.75cm}ll @{\gap} ll>{\fillR}l @{}}
\toprule
 & & \multicolumn{1}{c}{\textbf{Stability}} & \multicolumn{4}{c}{\textbf{Kinematics}} & \multicolumn{6}{c}{\textbf{Ground Reaction Forces}} & \multicolumn{3}{c}{\textbf{Torques}$^\dagger$} \\
\cmidrule(lr){3-3}\cmidrule(lr){4-7}\cmidrule(lr){8-13}\cmidrule(lr){14-16}
\texttt{\textbf{Activity}} & \textbf{Method}
  & \makecell{Ep.$\hi$ \\ (\%)}
  & \makecell{$q\lo$\\ \makebox[0pt][c]{($^\circ$)}}
  & \makecell{$\dot{q}\lo$\\ \makebox[0pt][c]{($^\circ$/s)}}
  & \makecell{$p_x\lo$\\ \makebox[0pt][c]{(m/s)}}
  & kin$_r\hi$
  & \makecell{$F_{\mathrm{Z}}\lo$\\ \makebox[0pt][c]{(BW)}}
  & $F^{\uparrow}_{\mathrm{Z}_r}$
  & $F^{\uparrow}_{\mathrm{ML}_r}$
  & $F^{\uparrow}_{\mathrm{AP}_r}$
  & \makecell{$\mathrm{CoP}^{\downarrow}_{\mathrm{ML}}$\\ {\tiny (mm)}}
  & \makecell{$\mathrm{CoP}^{\downarrow}_{\mathrm{AP}}$\\ {\tiny (mm)}}
  & \makecell{$\tau\lo$\\ \makebox[0pt][c]{\tiny (Nm/kg)}}
  & $\tau_r\hi$
  & \makecell{$\dot{\tau}\lo$\\ \makebox[0pt][c]{\tiny (Nm/kg/s)}} \\
\midrule
\multirow{18}{*}{\act{run}} & Fixed-PD & 99.7\sd{.2} & 3.36\sd{.02} & 109\sd{4} & .064\sd{.005} & .96\sd{.00} & .21\sd{.00} & .96\sd{.00} & .32\sd{.01} & .52\sd{.01} & 22\sd{1} & 43\sd{1} & 1.25\sd{.01} & .47\sd{.01} & 110\sd{2} \\
 & Fixed-PD, uncapped \smash[t]{$k^d$} & 99.4\sd{.3} & 4.80\sd{.04} & 161\sd{1} & .103\sd{.007} & .90\sd{.00} & .30\sd{.00} & .93\sd{.00} & .19\sd{.01} & .29\sd{.01} & 24\sd{1} & 48\sd{4} & 1.57\sd{.00} & .37\sd{.00} & 192\sd{1} \\
 & Fixed-PD, \smash[t]{$2\times k^p$} & \textbf{100}\sd{0} & 2.45\sd{.03} & 107\sd{2} & .060\sd{.004} & .97\sd{.00} & .15\sd{.00} & .98\sd{.00} & .46\sd{.03} & .81\sd{.01} & \textbf{21}\sd{0} & 45\sd{0} & 1.29\sd{.02} & .48\sd{.00} & 125\sd{4} \\
\end{tabular}%
}
\vspace{-2pt}%
\parbox{\textwidth}{\scriptsize\raggedright
  $^\ast$SimPoE is shown in italics for reference and excluded from the ranking, since its residual wrench acts directly on the unactuated pelvis.\\
  \emph{Uncapped $k^d$} = Fixed-PD without its damping cap. $2\times k^p$ = stiffness priors doubled, $2\times$ (\emph{gains}) = stiffness and damping priors doubled. \emph{Gain schedule} = subject-mean learned gains by activity and phase, with a retrained position network.
  $^\dagger$9 main joints (hip flexion / adduction, knee, ankle, both sides; lumbar extension), against inverse dynamics.
  Columns as in Table~\ref{tab:golden_full}.}
\vspace{-1em}
\end{table}

\begin{table}[H]
\centering
\caption{%
  \textbf{Existing controllers and PD integration on unseen subjects and withheld activities.} Unseen subjects zero-shot and each withheld activity from the fold trained without it (App.~\ref{App:splits}), mean$_{\pm\text{sd}}$ over 5 seeds.
  Rows, columns, markers and \textbf{bold} / \underline{underline} as in Table~\ref{tab:existing_full}, per benchmark.}
\vspace{-0.2em}%
\label{tab:existing_transfer}
\renewcommand{\arraystretch}{0.80}
\setlength{\tabcolsep}{2pt}
\newcommand{\gap}{\hspace{4pt}}
\newcommand{\sd}[1]{{\tiny\textcolor{black!55}{$\pm$#1}}}
\scriptsize
\resizebox{\textwidth}{!}{%
\begin{tabular}{@{}>{\fillL}l @{\gap} l @{\gap} llllllll<{\hspace{2pt}}|>{\hspace{2pt}}lllllll>{\fillR}l @{}}
\toprule
 & & \multicolumn{8}{c<{\hspace{2pt}}}{\textbf{Unseen subjects, zero-shot}} & \multicolumn{8}{>{\hspace{2pt}}c}{\textbf{Withheld activity}} \\
\cmidrule(lr){3-10}\cmidrule(lr){11-18}
\texttt{\textbf{Activity}} & \textbf{Method}
  & \makecell{Ep.$\hi$ \\ (\%)} & \makecell{$q\lo$\\ \makebox[0pt][c]{($^\circ$)}} & \makecell{$p_x\lo$\\ \makebox[0pt][c]{(m/s)}} & $F^{\uparrow}_{\mathrm{Z}_r}$ & $F^{\uparrow}_{\mathrm{ML}_r}$ & $F^{\uparrow}_{\mathrm{AP}_r}$ & \makecell{$\tau\lo{}^{\dagger}$\\ \makebox[0pt][c]{\tiny (Nm/kg)}} & \multicolumn{1}{l<{\hspace{2pt}}}{\makecell{$\dot{\tau}\lo{}^{\dagger}$\\ \makebox[0pt][c]{\tiny (Nm/kg/s)}}}
  & \makecell{Ep.$\hi$ \\ (\%)} & \makecell{$q\lo$\\ \makebox[0pt][c]{($^\circ$)}} & \makecell{$p_x\lo$\\ \makebox[0pt][c]{(m/s)}} & $F^{\uparrow}_{\mathrm{Z}_r}$ & $F^{\uparrow}_{\mathrm{ML}_r}$ & $F^{\uparrow}_{\mathrm{AP}_r}$ & \makecell{$\tau\lo{}^{\dagger}$\\ \makebox[0pt][c]{\tiny (Nm/kg)}} & \makecell{$\dot{\tau}\lo{}^{\dagger}$\\ \makebox[0pt][c]{\tiny (Nm/kg/s)}} \\
\specialrule{\lightrulewidth}{\aboverulesep}{0pt}
\multirow{16}{*}{\act{walk}} & \padT Fixed-PD & 99.8\sd{.2} & 2.75\sd{.05} & .027\sd{.003} & .94\sd{.00} & .71\sd{.01} & .72\sd{.01} & .55\sd{.01} & 28.0\sd{.5} & 40.0\sd{5.0} & 4.52\sd{.14} & .322\sd{.034} & .43\sd{.01} & .13\sd{.02} & .11\sd{.01} & 1.40\sd{.03} & 158\sd{6} \\
 & Fixed-PD, \smash[t]{$2\times k^p$} & \textbf{100}\sd{0} & 2.42\sd{.03} & .022\sd{.002} & .93\sd{.00} & .65\sd{.01} & .65\sd{.01} & .85\sd{.01} & 99.1\sd{2.0} & 67.5\sd{4.1} & 3.93\sd{.11} & .452\sd{.046} & .42\sd{.02} & .16\sd{.01} & .11\sd{.01} & 1.56\sd{.01} & 160\sd{5} \\
\end{tabular}%
}
\vspace{-2pt}%
\parbox{\textwidth}{\scriptsize\raggedright
  $^\ast$SimPoE is shown in italics for reference and excluded from the ranking, since its residual wrench acts directly on the unactuated pelvis.\\
  $^\dagger$9 main joints (hip flexion / adduction, knee, ankle, both sides; lumbar extension), against inverse dynamics. Columns as in Table~\ref{tab:golden_full}.}%
\vspace{-1em}
\end{table}
\vspace{-7pt}%
\noindent every force correlation on every benchmark under any scheme, and survives more often in every matched comparison but one ($96.3$ against $96.6\,\%$ on seen subjects at doubled stable priors).
The same priors give it the lowest joint-angle error of all rows on every benchmark.

\begin{table}[H]%
\centering
\caption{%
  \textbf{Torque variants on seen subjects, held-out clips.} Mean$_{\pm\text{sd}}$ over 5 seeds.
  \emph{Fixed scale} replaces the inertia-scaled output scale $\alpha k^p_0$ by a body-independent constant, the \emph{kernel} rows exchange the pose-reward kernels of Torque and \projname{}, \emph{PD torque limits} gives Torque the torque limits of the PD action spaces, and \emph{500\,Hz policy} acts at every 2\,ms physics step.
  Columns are RMSE ($\downarrow$) unless marked $\uparrow$.
  \textbf{Bold} / \underline{underline} = best / second best per activity.}
\vspace{-0.6em}%
\label{tab:torque_variants}
\renewcommand{\arraystretch}{0.80}
\setlength{\tabcolsep}{2pt}
\newcommand{\gap}{\hspace{4pt}}
\newcommand{\sd}[1]{{\tiny\textcolor{black!55}{$\pm$#1}}}
\scriptsize
\resizebox{\textwidth}{!}{%
\begin{tabular}{@{}>{\fillL}l @{\gap} l @{\gap} lllll @{\gap} lp{0.75cm}p{0.75cm}p{0.75cm}ll @{\gap} ll>{\fillR}l @{}}
\toprule
 & & \multicolumn{1}{c}{\textbf{Stability}} & \multicolumn{4}{c}{\textbf{Kinematics}} & \multicolumn{6}{c}{\textbf{Ground Reaction Forces}} & \multicolumn{3}{c}{\textbf{Torques}$^\dagger$} \\
\cmidrule(lr){3-3}\cmidrule(lr){4-7}\cmidrule(lr){8-13}\cmidrule(lr){14-16}
\texttt{\textbf{Activity}} & \textbf{Method}
  & \makecell{Ep.$\hi$ \\ (\%)}
  & \makecell{$q\lo$\\ \makebox[0pt][c]{($^\circ$)}}
  & \makecell{$\dot{q}\lo$\\ \makebox[0pt][c]{($^\circ$/s)}}
  & \makecell{$p_x\lo$\\ \makebox[0pt][c]{(m/s)}}
  & kin$_r\hi$
  & \makecell{$F_{\mathrm{Z}}\lo$\\ \makebox[0pt][c]{(BW)}}
  & $F^{\uparrow}_{\mathrm{Z}_r}$
  & $F^{\uparrow}_{\mathrm{ML}_r}$
  & $F^{\uparrow}_{\mathrm{AP}_r}$
  & \makecell{\makebox[0pt][c]{$\mathrm{CoP}^{\downarrow}_{\mathrm{ML}}$}\\ \makebox[0pt][c]{\tiny (mm)}}
  & \makecell{\makebox[0pt][c]{$\mathrm{CoP}^{\downarrow}_{\mathrm{AP}}$}\\ \makebox[0pt][c]{\tiny (mm)}}
  & \makecell{$\tau\lo$\\ \makebox[0pt][c]{\tiny (Nm/kg)}}
  & $\tau_r\hi$
  & \makecell{$\dot{\tau}\lo$\\ \makebox[0pt][c]{\tiny (Nm/kg/s)}} \\
\midrule
\multirow{7}{*}{\act{walk}} & ImpedanceMimic & \textbf{100}\sd{0} & \textbf{1.66}\sd{.03} & \textbf{28.8}\sd{.4} & \textbf{.013}\sd{.001} & \textbf{.98}\sd{.00} & \underline{.09}\sd{.00} & \textbf{.98}\sd{.00} & \underline{.89}\sd{.00} & \underline{.86}\sd{.00} & \underline{20}\sd{0} & \underline{40}\sd{1} & \textbf{.38}\sd{.00} & \textbf{.65}\sd{.00} & \underline{6.38}\sd{.08} \\
 & ImpedanceMimic, Torque's kernel & \textbf{100}\sd{0} & \underline{1.87}\sd{.02} & \underline{29.8}\sd{.5} & \textbf{.013}\sd{.001} & \underline{.96}\sd{.00} & \textbf{.08}\sd{.00} & \textbf{.98}\sd{.00} & \textbf{.90}\sd{.00} & \textbf{.87}\sd{.00} & \underline{20}\sd{0} & \textbf{39}\sd{1} & \textbf{.38}\sd{.00} & \textbf{.65}\sd{.00} & \textbf{6.14}\sd{.05} \\
 & Torque & 99.7\sd{.2} & 2.33\sd{.03} & 55.7\sd{2.5} & .022\sd{.001} & .93\sd{.00} & \underline{.09}\sd{.00} & \textbf{.98}\sd{.00} & \underline{.89}\sd{.00} & .85\sd{.00} & \textbf{19}\sd{0} & 41\sd{1} & .42\sd{.00} & .61\sd{.01} & 9.06\sd{.22} \\
 & Torque, fixed scale & 98.6\sd{.4} & 2.69\sd{.03} & 57.4\sd{1.2} & .031\sd{.002} & .93\sd{.00} & .10\sd{.00} & \textbf{.98}\sd{.00} & .88\sd{.00} & .85\sd{.01} & \underline{20}\sd{0} & \underline{40}\sd{0} & .39\sd{.00} & .63\sd{.00} & 7.44\sd{.11} \\
 & Torque, ImpedanceMimic's kernel & 94.6\sd{1.5} & 44.8\sd{.5} & 146\sd{25} & .015\sd{.004} & .20\sd{.08} & .17\sd{.02} & .93\sd{.02} & .75\sd{.04} & .77\sd{.04} & 25\sd{4} & 59\sd{8} & .94\sd{.05} & .26\sd{.03} & 10.9\sd{.5} \\
 & Torque, PD torque limits & 80.0\sd{15.9} & 52.9\sd{3.8} & 306\sd{67} & .019\sd{.014} & .02\sd{.01} & 1.36\sd{.21} & .05\sd{.03} & .05\sd{.03} & .12\sd{.09} & 30\sd{7} & 97\sd{18} & 2.64\sd{.09} & .05\sd{.03} & 66.1\sd{4.6} \\
 & Torque, 500\,Hz policy & 96.1\sd{2.1} & 3.34\sd{.17} & 69.7\sd{2.8} & .024\sd{.001} & .90\sd{.01} & .23\sd{.01} & .88\sd{.01} & .74\sd{.01} & .62\sd{.01} & 22\sd{2} & 56\sd{3} & .47\sd{.00} & .53\sd{.01} & 10.2\sd{.3} \\
\addlinespace[3pt]
\multirow{7}{*}{\act{run}} & ImpedanceMimic & \textbf{100}\sd{0} & \textbf{2.24}\sd{.02} & \textbf{59.7}\sd{1.3} & .043\sd{.003} & \textbf{.98}\sd{.00} & \textbf{.12}\sd{.00} & \textbf{.99}\sd{.00} & \underline{.58}\sd{.03} & \underline{.86}\sd{.00} & \textbf{21}\sd{0} & 42\sd{0} & \textbf{.80}\sd{.00} & \textbf{.72}\sd{.01} & 19.7\sd{1.9} \\
 & ImpedanceMimic, Torque's kernel & \textbf{100}\sd{0} & \underline{2.38}\sd{.03} & \underline{60.1}\sd{1.3} & .043\sd{.005} & \underline{.97}\sd{.00} & \textbf{.12}\sd{.00} & \textbf{.99}\sd{.00} & \textbf{.63}\sd{.03} & \textbf{.87}\sd{.00} & \textbf{21}\sd{0} & \underline{41}\sd{1} & \underline{.81}\sd{.01} & \underline{.70}\sd{.01} & 19.4\sd{1.3} \\
 & Torque & 99.8\sd{.3} & 3.51\sd{.02} & 114\sd{3} & .075\sd{.005} & .93\sd{.00} & .23\sd{.00} & .96\sd{.00} & .40\sd{.03} & .58\sd{.01} & \textbf{21}\sd{0} & 49\sd{1} & .86\sd{.00} & .66\sd{.00} & 19.8\sd{.6} \\
 & Torque, fixed scale & 99.5\sd{.3} & 4.09\sd{.05} & 119\sd{3} & .159\sd{.019} & .92\sd{.00} & .27\sd{.01} & .94\sd{.00} & .37\sd{.03} & .53\sd{.02} & \textbf{21}\sd{0} & 48\sd{2} & .84\sd{.01} & .65\sd{.01} & \underline{18.1}\sd{.3} \\
 & Torque, ImpedanceMimic's kernel & 97.2\sd{.8} & 42.6\sd{2.3} & 284\sd{14} & \textbf{.030}\sd{.004} & .26\sd{.07} & .43\sd{.03} & .86\sd{.02} & .46\sd{.04} & .68\sd{.04} & 25\sd{2} & \textbf{40}\sd{2} & 1.08\sd{.05} & .44\sd{.08} & 21.1\sd{.4} \\
 & Torque, PD torque limits & 70.5\sd{25.0} & 53.2\sd{.5} & 413\sd{68} & \underline{.041}\sd{.026} & \negs{}.03\sd{.01} & 1.38\sd{.11} & .21\sd{.04} & .08\sd{.03} & .09\sd{.07} & 53\sd{52} & 125\sd{71} & 2.81\sd{.05} & .11\sd{.02} & 67.2\sd{7.3} \\
 & Torque, 500\,Hz policy & 99.4\sd{.4} & 3.71\sd{.04} & 113\sd{2} & .079\sd{.006} & .93\sd{.00} & .28\sd{.00} & .93\sd{.00} & .43\sd{.02} & .51\sd{.02} & 23\sd{2} & 55\sd{2} & .86\sd{.00} & .66\sd{.01} & \textbf{16.9}\sd{.4} \\
\addlinespace[3pt]
\multirow{7}{*}{\act{stand}} & ImpedanceMimic & \underline{99.7}\sd{.2} & \textbf{1.07}\sd{.00} & \textbf{4.28}\sd{.13} & \underline{.003}\sd{.001} & \textbf{.66}\sd{.01} & .06\sd{.00} & .56\sd{.02} & \underline{.80}\sd{.03} & .20\sd{.03} & \underline{13}\sd{0} & \textbf{32}\sd{1} & .84\sd{.01} & \underline{.15}\sd{.00} & \underline{.92}\sd{.08} \\
 & ImpedanceMimic, Torque's kernel & \textbf{99.8}\sd{.1} & \underline{2.04}\sd{.04} & \underline{4.50}\sd{.19} & \textbf{.002}\sd{.001} & \underline{.51}\sd{.01} & \textbf{.04}\sd{.00} & \textbf{.76}\sd{.02} & \textbf{.86}\sd{.02} & \textbf{.41}\sd{.03} & \textbf{12}\sd{1} & \underline{38}\sd{7} & .59\sd{.03} & \textbf{.18}\sd{.01} & \textbf{.70}\sd{.10} \\
 & Torque & 99.6\sd{.3} & 2.33\sd{.03} & 14.3\sd{1.2} & .004\sd{.001} & .37\sd{.02} & \underline{.05}\sd{.00} & \underline{.64}\sd{.04} & .55\sd{.04} & \underline{.26}\sd{.04} & \underline{13}\sd{1} & 52\sd{4} & \underline{.35}\sd{.01} & .11\sd{.03} & 1.64\sd{.21} \\
 & Torque, fixed scale & 98.2\sd{.9} & 2.43\sd{.12} & 12.6\sd{1.4} & \underline{.003}\sd{.000} & .36\sd{.01} & \underline{.05}\sd{.00} & .55\sd{.03} & .48\sd{.07} & .22\sd{.02} & 14\sd{0} & 49\sd{3} & \textbf{.32}\sd{.01} & .11\sd{.01} & 1.26\sd{.26} \\
 & Torque, ImpedanceMimic's kernel & 89.3\sd{9.4} & 51.2\sd{8.1} & 95.9\sd{13.5} & .014\sd{.004} & .00\sd{.01} & .34\sd{.05} & .07\sd{.02} & .25\sd{.04} & .05\sd{.02} & 20\sd{10} & 85\sd{7} & 1.03\sd{.15} & .01\sd{.01} & 3.78\sd{2.00} \\
 & Torque, PD torque limits & 92.7\sd{12.0} & 60.5\sd{7.3} & 247\sd{101} & .011\sd{.009} & .00\sd{.00} & 1.41\sd{.20} & .00\sd{.01} & \negs{}.01\sd{.04} & .00\sd{.00} & 23\sd{8} & 73\sd{25} & 2.63\sd{.21} & .00\sd{.00} & 60.5\sd{5.0} \\
 & Torque, 500\,Hz policy & 98.7\sd{.7} & 2.05\sd{.07} & 52.1\sd{4.1} & .010\sd{.001} & .30\sd{.02} & .41\sd{.03} & .11\sd{.01} & .20\sd{.07} & .05\sd{.02} & 23\sd{1} & 72\sd{6} & .38\sd{.01} & .05\sd{.01} & 12.8\sd{1.1} \\
\addlinespace[3pt]
\multirow{7}{*}{\act{lateral}} & ImpedanceMimic & \textbf{99.9}\sd{.1} & \textbf{3.11}\sd{.07} & \textbf{59.6}\sd{1.4} & .084\sd{.004} & \textbf{.94}\sd{.00} & \underline{.23}\sd{.00} & \underline{.91}\sd{.01} & \underline{.80}\sd{.01} & \underline{.71}\sd{.02} & 31\sd{2} & 53\sd{4} & \underline{.56}\sd{.02} & \underline{.57}\sd{.02} & \textbf{21.2}\sd{1.6} \\
 & ImpedanceMimic, Torque's kernel & \textbf{99.9}\sd{.1} & \underline{3.74}\sd{.06} & \underline{63.5}\sd{.9} & \underline{.082}\sd{.004} & \underline{.90}\sd{.00} & \textbf{.22}\sd{.01} & \textbf{.92}\sd{.01} & \textbf{.81}\sd{.01} & \textbf{.74}\sd{.01} & 29\sd{1} & \textbf{51}\sd{2} & \textbf{.55}\sd{.01} & \textbf{.58}\sd{.01} & \textbf{21.2}\sd{1.2} \\
 & Torque & 97.6\sd{1.3} & 6.49\sd{.22} & 149\sd{7} & .102\sd{.008} & .77\sd{.01} & .25\sd{.01} & .89\sd{.01} & .74\sd{.01} & .65\sd{.01} & \textbf{28}\sd{1} & 53\sd{2} & .62\sd{.01} & .51\sd{.01} & 28.5\sd{.3} \\
 & Torque, fixed scale & 93.2\sd{1.8} & 7.75\sd{.30} & 151\sd{5} & .114\sd{.003} & .73\sd{.01} & .27\sd{.01} & .87\sd{.01} & .74\sd{.03} & .59\sd{.02} & \textbf{28}\sd{1} & \underline{52}\sd{1} & \underline{.56}\sd{.01} & .50\sd{.01} & 22.8\sd{.9} \\
 & Torque, ImpedanceMimic's kernel & 89.0\sd{4.2} & 42.9\sd{1.9} & 236\sd{18} & .083\sd{.011} & .13\sd{.02} & .40\sd{.02} & .77\sd{.03} & .61\sd{.04} & .47\sd{.04} & 34\sd{3} & 69\sd{8} & .93\sd{.04} & .33\sd{.04} & 22.0\sd{.9} \\
 & Torque, PD torque limits & 90.1\sd{4.6} & 49.6\sd{.5} & 371\sd{54} & \textbf{.078}\sd{.014} & .02\sd{.03} & 1.37\sd{.08} & .13\sd{.02} & .09\sd{.02} & .12\sd{.03} & 97\sd{78} & 118\sd{19} & 2.66\sd{.05} & .05\sd{.00} & 73.8\sd{3.6} \\
 & Torque, 500\,Hz policy & 94.5\sd{2.1} & 5.48\sd{.18} & 128\sd{2} & .126\sd{.004} & .83\sd{.00} & .32\sd{.01} & .83\sd{.01} & .71\sd{.01} & .58\sd{.01} & 37\sd{11} & 71\sd{12} & .61\sd{.01} & .50\sd{.01} & 24.5\sd{.9} \\
\addlinespace[3pt]
\multirow{7}{*}{\act{flexion}} & ImpedanceMimic & \underline{97.7}\sd{1.3} & \textbf{2.40}\sd{.04} & \underline{42.3}\sd{1.1} & \underline{.031}\sd{.003} & \textbf{.93}\sd{.00} & .18\sd{.00} & .70\sd{.01} & \underline{.58}\sd{.01} & .52\sd{.01} & \textbf{23}\sd{0} & \underline{60}\sd{1} & .44\sd{.00} & \underline{.49}\sd{.01} & 11.4\sd{.5} \\
 & ImpedanceMimic, Torque's kernel & \textbf{99.9}\sd{.1} & \underline{3.18}\sd{.06} & \textbf{40.0}\sd{1.8} & \textbf{.019}\sd{.002} & \underline{.89}\sd{.01} & \underline{.16}\sd{.00} & \textbf{.75}\sd{.01} & \textbf{.65}\sd{.01} & \textbf{.59}\sd{.01} & \textbf{23}\sd{1} & \textbf{58}\sd{1} & \textbf{.39}\sd{.01} & \textbf{.52}\sd{.01} & \textbf{7.86}\sd{.53} \\
 & Torque & 96.8\sd{.8} & 4.62\sd{.09} & 112\sd{4} & .032\sd{.003} & .81\sd{.01} & \textbf{.15}\sd{.00} & \underline{.73}\sd{.01} & \underline{.58}\sd{.01} & \underline{.54}\sd{.01} & \textbf{23}\sd{1} & 61\sd{1} & .42\sd{.00} & .48\sd{.01} & 14.1\sd{.6} \\
 & Torque, fixed scale & 80.8\sd{.3} & 9.07\sd{.27} & 112\sd{8} & .040\sd{.001} & .68\sd{.00} & .17\sd{.00} & .68\sd{.01} & .52\sd{.02} & \underline{.54}\sd{.02} & 24\sd{1} & 62\sd{1} & \underline{.40}\sd{.01} & .45\sd{.01} & 9.84\sd{.42} \\
 & Torque, ImpedanceMimic's kernel & 72.8\sd{4.4} & 44.1\sd{2.5} & 135\sd{10} & .063\sd{.006} & .20\sd{.05} & .32\sd{.03} & .45\sd{.06} & .45\sd{.03} & .40\sd{.03} & 24\sd{2} & 81\sd{3} & .94\sd{.02} & .07\sd{.01} & \underline{8.64}\sd{1.21} \\
 & Torque, PD torque limits & 90.6\sd{8.6} & 52.7\sd{1.2} & 288\sd{68} & .033\sd{.013} & .04\sd{.02} & 1.38\sd{.11} & .02\sd{.01} & .02\sd{.02} & .02\sd{.01} & 30\sd{3} & 85\sd{10} & 2.66\sd{.14} & \negs{}.01\sd{.02} & 65.2\sd{4.0} \\
 & Torque, 500\,Hz policy & 81.3\sd{1.1} & 6.04\sd{.25} & 97.8\sd{1.4} & .045\sd{.004} & .77\sd{.01} & .27\sd{.01} & .53\sd{.01} & .40\sd{.01} & .41\sd{.02} & 25\sd{0} & 70\sd{1} & .47\sd{.01} & .42\sd{.01} & 14.2\sd{.3} \\
\addlinespace[3pt]
\multirow{7}{*}{\act{jump}} & ImpedanceMimic & \underline{83.8}\sd{2.1} & \textbf{3.22}\sd{.16} & \textbf{69.7}\sd{2.8} & .063\sd{.003} & \textbf{.92}\sd{.00} & .33\sd{.01} & .69\sd{.01} & \underline{.55}\sd{.03} & \underline{.40}\sd{.02} & \underline{27}\sd{2} & \underline{68}\sd{3} & .69\sd{.02} & .38\sd{.01} & 36.0\sd{2.7} \\
 & ImpedanceMimic, Torque's kernel & \textbf{86.1}\sd{.9} & \underline{3.96}\sd{.06} & \underline{70.9}\sd{2.6} & \underline{.056}\sd{.004} & \underline{.87}\sd{.00} & \underline{.30}\sd{.01} & \underline{.73}\sd{.01} & \textbf{.62}\sd{.02} & \textbf{.44}\sd{.02} & \textbf{26}\sd{1} & \textbf{64}\sd{2} & .64\sd{.02} & .40\sd{.01} & 31.8\sd{2.1} \\
 & Torque & 78.9\sd{1.0} & 7.61\sd{.47} & 171\sd{14} & .075\sd{.005} & .72\sd{.01} & \textbf{.28}\sd{.01} & \textbf{.74}\sd{.01} & \underline{.55}\sd{.02} & .38\sd{.01} & 32\sd{7} & 73\sd{9} & .57\sd{.02} & \textbf{.41}\sd{.01} & 25.8\sd{1.2} \\
 & Torque, fixed scale & 66.2\sd{1.5} & 11.7\sd{.2} & 173\sd{7} & .088\sd{.006} & .62\sd{.01} & \underline{.30}\sd{.01} & .68\sd{.02} & .49\sd{.02} & .35\sd{.02} & 30\sd{4} & \underline{68}\sd{3} & \textbf{.52}\sd{.01} & \textbf{.41}\sd{.01} & \underline{18.6}\sd{1.0} \\
 & Torque, ImpedanceMimic's kernel & 54.2\sd{1.7} & 44.0\sd{1.9} & 217\sd{30} & .091\sd{.004} & .13\sd{.03} & .43\sd{.03} & .52\sd{.07} & .43\sd{.03} & .24\sd{.03} & 38\sd{18} & 97\sd{15} & .92\sd{.03} & .20\sd{.02} & \textbf{15.7}\sd{.8} \\
 & Torque, PD torque limits & 70.8\sd{6.1} & 51.7\sd{1.9} & 345\sd{57} & \textbf{.054}\sd{.011} & .00\sd{.01} & 1.39\sd{.11} & .11\sd{.04} & .02\sd{.02} & .04\sd{.02} & 92\sd{60} & 167\sd{74} & 2.61\sd{.09} & .05\sd{.02} & 69.0\sd{4.4} \\
 & Torque, 500\,Hz policy & 81.5\sd{1.8} & 6.66\sd{.47} & 132\sd{7} & .084\sd{.007} & .78\sd{.01} & .40\sd{.01} & .61\sd{.01} & .50\sd{.03} & .31\sd{.02} & 29\sd{3} & 76\sd{2} & \underline{.56}\sd{.01} & \textbf{.41}\sd{.01} & 21.2\sd{.5} \\
\specialrule{\cmidrulewidth}{\aboverulesep}{0pt}
\rowcolor{meangray}  & \padT ImpedanceMimic & \underline{96.9}\sd{.5} & \textbf{2.28}\sd{.04} & \textbf{44.1}\sd{.2} & .040\sd{.001} & \textbf{.90}\sd{.00} & \underline{.17}\sd{.00} & .80\sd{.00} & \underline{.70}\sd{.01} & \underline{.59}\sd{.01} & \underline{23}\sd{1} & \underline{49}\sd{1} & .62\sd{.00} & \underline{.49}\sd{.00} & 15.9\sd{.4} \\
\rowcolor{meangray}  & ImpedanceMimic, Torque's kernel & \textbf{97.6}\sd{.1} & \underline{2.86}\sd{.01} & \underline{44.8}\sd{.9} & \textbf{.036}\sd{.002} & \underline{.85}\sd{.00} & \textbf{.15}\sd{.00} & \textbf{.85}\sd{.00} & \textbf{.75}\sd{.01} & \textbf{.65}\sd{.01} & \textbf{22}\sd{0} & \textbf{48}\sd{2} & .56\sd{.01} & \textbf{.51}\sd{.00} & 14.5\sd{.7} \\
\rowcolor{meangray}  & Torque & 95.4\sd{.4} & 4.48\sd{.09} & 103\sd{4} & .052\sd{.002} & .76\sd{.00} & \underline{.17}\sd{.00} & \underline{.82}\sd{.01} & .62\sd{.01} & .54\sd{.01} & \underline{23}\sd{1} & 55\sd{2} & \underline{.54}\sd{.00} & .46\sd{.00} & 16.5\sd{.4} \\
\rowcolor{meangray}  & Torque, fixed scale & 89.4\sd{.3} & 6.29\sd{.10} & 104\sd{3} & .072\sd{.002} & .71\sd{.00} & .19\sd{.00} & .78\sd{.01} & .58\sd{.01} & .51\sd{.01} & \underline{23}\sd{1} & 53\sd{1} & \textbf{.50}\sd{.00} & .46\sd{.00} & \textbf{13.0}\sd{.4} \\
\rowcolor{meangray}  & Torque, ImpedanceMimic's kernel & 82.8\sd{1.5} & 44.9\sd{.6} & 186\sd{14} & .049\sd{.003} & .15\sd{.02} & .35\sd{.02} & .60\sd{.03} & .49\sd{.02} & .43\sd{.01} & 28\sd{5} & 72\sd{3} & .98\sd{.04} & .22\sd{.02} & \underline{13.7}\sd{.8} \\
\rowcolor{meangray}  & Torque, PD torque limits & 82.5\sd{10.5} & 53.4\sd{2.4} & 328\sd{69} & \underline{.039}\sd{.013} & .01\sd{.01} & 1.38\sd{.13} & .09\sd{.02} & .04\sd{.01} & .06\sd{.03} & 54\sd{16} & 111\sd{22} & 2.67\sd{.10} & .04\sd{.01} & 67.0\sd{3.9} \\
\rowcolor{meangray} \multirow{-7}{*}{\textbf{Mean}} & \padB Torque, 500\,Hz policy & 91.9\sd{.6} & 4.55\sd{.11} & 98.8\sd{.5} & .061\sd{.001} & .75\sd{.00} & .32\sd{.01} & .65\sd{.00} & .50\sd{.01} & .41\sd{.00} & 27\sd{2} & 67\sd{2} & .56\sd{.00} & .43\sd{.00} & 16.6\sd{.1} \\
\specialrule{\heavyrulewidth}{0pt}{\belowbottomsep}
\end{tabular}%
}
\vspace{-2pt}%
\parbox{\textwidth}{\scriptsize\raggedright
  $^\dagger$9 main joints (hip flexion / adduction, knee, ankle, both sides; lumbar extension), against inverse dynamics.
  Columns as in Table~\ref{tab:golden_full}.}
\vspace{-1em}
\end{table}
\vspace{-21pt}%
\paragraph{Torque variants.}
Each row changes one training choice of Torque or \projname{} (Table~\ref{tab:torque_variants}).
Torque fails to learn with the amplitude-normalized pose kernel of \projname{} ($44.9^\circ$ of joint-angle error) or with the torque limits of the PD action spaces ($53.4^\circ$), which makes its own kernel and inertia-scaled output range prerequisites for learning rather than tuned advantages.
Both failures are consistent with an action that starts from zero torque rather than, like a PD policy, at the reference pose, and with exploration noise that scales with the output range.
A body-independent output scale lowers survival by $6.0$ points, because the inertia prior adapts the torque authority to each body as it adapts the PD gains.
Trained for the same simulated time ($10^{10}$ steps of 2\,ms), with the discount, rollout length and lookahead offsets rescaled accordingly, the 500\,Hz policy leaves the joint-angle error unchanged ($4.55^\circ$ against $4.48^\circ$) while lowering the force correlations.
With the kernel of Torque, \projname{} exceeds Torque in survival, joint-angle error and every force correlation, the vertical one $0.85$ against $0.82$.
Torque's slim lead in vertical-force correlation (Table~\ref{tab:golden}) therefore reflects its kernel, which raises the joint-angle error of \projname{} from $2.28$ to $2.86^\circ$ in exchange for force correlation, and the advantage of \projname{} lies in its action space.

{\setlength{\intextsep}{2pt}%
\vspace{2pt}%
\begin{table}[H]
\centering
\caption{\textbf{Robustness to test-time perturbations.} Seen subjects, held-out clips, activity mean$_{\pm\text{sd}}$ over 5 seeds. Perturbed rows show the change from the unperturbed row, without retraining.}
\label{tab:robustness}
\vspace{-0.2em}%
\renewcommand{\arraystretch}{0.85}
\newcommand{\sd}[1]{{\tiny\textcolor{black!55}{$\pm$#1}}}
\newcommand{\secrule}{\arrayrulecolor{black!25}\midrule[0.3pt]\arrayrulecolor{black}}
\scriptsize
\setlength{\tabcolsep}{3pt}
\resizebox{\textwidth}{!}{%
\begin{tabular}{@{}l cccc cccc cccc cccc@{}}
\toprule
 & \multicolumn{4}{c}{\textbf{\projname{}}} & \multicolumn{4}{c}{\textbf{Fixed-PD}} & \multicolumn{4}{c}{\textbf{Torque}} & \multicolumn{4}{c}{\textbf{Muscle}} \\
\cmidrule(lr){2-5}\cmidrule(lr){6-9}\cmidrule(lr){10-13}\cmidrule(lr){14-17}
\textbf{Condition} & Ep.$\hi$ & $q\lo$ & $F^{\uparrow}_{\mathrm{Z}_r}$ & $\dot{\tau}\lo{}^{\dagger}$ & Ep.$\hi$ & $q\lo$ & $F^{\uparrow}_{\mathrm{Z}_r}$ & $\dot{\tau}\lo{}^{\dagger}$ & Ep.$\hi$ & $q\lo$ & $F^{\uparrow}_{\mathrm{Z}_r}$ & $\dot{\tau}\lo{}^{\dagger}$ & Ep.$\hi$ & $q\lo$ & $F^{\uparrow}_{\mathrm{Z}_r}$ & $\dot{\tau}\lo{}^{\dagger}$ \\
\midrule
Unperturbed & 96.9\sd{.5} & 2.28\sd{.04} & .80\sd{.00} & 15.9\sd{.4} & 93.0\sd{.2} & 2.91\sd{.03} & .73\sd{.00} & 45.3\sd{.6} & 95.4\sd{.4} & 4.48\sd{.09} & .82\sd{.01} & 16.5\sd{.4} & 84.8\sd{.5} & 7.18\sd{.26} & .77\sd{.00} & 6.89\sd{.15} \\
\secrule
pelvis pushes $\le 0.05$ BW & $-$0.2 & $+$0.02 & \cellcolor{dworse!9}$-$.01 & $+$0.2 & $-$0.1 & 0.00 & \cellcolor{dworse!10}$-$.01 & 0.0 & $+$0.2 & $-$0.03 & $-$.01 & $+$0.1 & $-$0.4 & $+$0.01 & \cellcolor{dworse!9}$-$.01 & $+$0.01 \\
pelvis pushes $\le 0.10$ BW & $-$0.2 & $+$0.03 & \cellcolor{dworse!21}$-$.03 & $+$0.3 & \cellcolor{dworse!6}$-$0.4 & $+$0.03 & \cellcolor{dworse!10}$-$.01 & $+$0.2 & 0.0 & $+$0.07 & \cellcolor{dworse!21}$-$.03 & $+$0.3 & \cellcolor{dworse!11}$-$1.3 & $+$0.09 & \cellcolor{dworse!22}$-$.03 & $+$0.05 \\
pelvis pushes $\le 0.20$ BW & $-$0.3 & \cellcolor{dworse!22}$+$0.09 & \cellcolor{dworse!30}$-$.06 & \cellcolor{dworse!28}$+$1.0 & \cellcolor{dworse!6}$-$0.7 & \cellcolor{dworse!19}$+$0.09 & \cellcolor{dworse!26}$-$.04 & \cellcolor{dworse!19}$+$1.4 & \cellcolor{dworse!6}$-$0.6 & \cellcolor{dworse!27}$+$0.26 & \cellcolor{dworse!29}$-$.06 & \cellcolor{dworse!31}$+$1.4 & \cellcolor{dworse!28}$-$5.7 & \cellcolor{dworse!32}$+$0.68 & \cellcolor{dworse!30}$-$.06 & \cellcolor{dworse!18}$+$0.20 \\
\secrule
contact stiffness $\times 0.5$ & $-$0.2 & 0.00 & \cellcolor{dbetter!9}$+$.01 & $-$0.2 & \cellcolor{dworse!6}$-$0.5 & $+$0.01 & \cellcolor{dbetter!10}$+$.01 & $-$0.5 & $-$0.1 & $+$0.03 & .00 & \cellcolor{dbetter!19}$-$0.5 & $-$0.4 & $-$0.02 & .00 & $-$0.10 \\
contact stiffness $\times 2$ & $-$0.1 & $+$0.02 & .00 & $+$0.2 & 0.0 & $+$0.02 & \cellcolor{dworse!10}$-$.01 & $+$0.6 & 0.0 & $+$0.02 & .00 & \cellcolor{dworse!21}$+$0.6 & $+$0.1 & $+$0.02 & \cellcolor{dworse!9}$-$.01 & $+$0.09 \\
friction $\times 0.6$ & $-$0.5 & \cellcolor{dworse!25}$+$0.12 & \cellcolor{dworse!21}$-$.03 & \cellcolor{dworse!43}$+$3.6 & \cellcolor{dworse!9}$-$1.2 & \cellcolor{dworse!18}$+$0.08 & \cellcolor{dworse!29}$-$.05 & \cellcolor{dworse!32}$+$4.0 & \cellcolor{dworse!7}$-$1.0 & \cellcolor{dworse!28}$+$0.30 & \cellcolor{dworse!16}$-$.02 & \cellcolor{dworse!32}$+$1.5 & \cellcolor{dworse!6}$-$0.8 & \cellcolor{dworse!23}$+$0.31 & \cellcolor{dworse!22}$-$.03 & \cellcolor{dworse!37}$+$0.96 \\
observation delay $20$\,ms & \cellcolor{dworse!21}$-$3.4 & \cellcolor{dworse!53}$+$1.24 & \cellcolor{dworse!31}$-$.07 & \cellcolor{dworse!59}$+$14.5 & \cellcolor{dworse!46}$-$27.4 & \cellcolor{dworse!54}$+$1.76 & \cellcolor{dworse!46}$-$.22 & \cellcolor{dworse!60}$+$61 & \cellcolor{dworse!60}$-$92.1 & \cellcolor{dworse!60}$+$25.0 & \cellcolor{dworse!55}$-$.52 & \cellcolor{dworse!60}$+$44.3 & \cellcolor{dworse!54}$-$48.9 & \cellcolor{dworse!60}$+$11.8 & \cellcolor{dworse!50}$-$.33 & \cellcolor{dworse!60}$+$15.5 \\
\secrule
contact-sphere radius $\times 0.5$ & $-$0.3 & $+$0.01 & .00 & $+$0.3 & \cellcolor{dworse!6}$-$0.4 & $+$0.01 & .00 & $-$0.1 & $-$0.4 & $+$0.05 & .00 & $+$0.2 & $-$0.3 & $-$0.04 & .00 & 0.00 \\
contact-sphere radius $\times 1.5$ & \cellcolor{dworse!19}$-$3.0 & \cellcolor{dworse!45}$+$0.62 & \cellcolor{dworse!44}$-$.21 & \cellcolor{dworse!60}$+$21.3 & \cellcolor{dworse!32}$-$8.8 & \cellcolor{dworse!42}$+$0.63 & \cellcolor{dworse!43}$-$.17 & \cellcolor{dworse!54}$+$26.4 & \cellcolor{dworse!30}$-$7.5 & \cellcolor{dworse!50}$+$1.87 & \cellcolor{dworse!41}$-$.16 & \cellcolor{dworse!55}$+$11.2 & \cellcolor{dworse!48}$-$30.6 & \cellcolor{dworse!53}$+$4.1 & \cellcolor{dworse!45}$-$.22 & \cellcolor{dworse!60}$+$7.7 \\
contact-sphere radius $\times 2$ & \cellcolor{dworse!44}$-$25.4 & \cellcolor{dworse!57}$+$1.70 & \cellcolor{dworse!50}$-$.33 & \cellcolor{dworse!60}$+$45.7 & \cellcolor{dworse!49}$-$36.8 & \cellcolor{dworse!52}$+$1.43 & \cellcolor{dworse!49}$-$.29 & \cellcolor{dworse!60}$+$57 & \cellcolor{dworse!42}$-$20.1 & \cellcolor{dworse!60}$+$4.31 & \cellcolor{dworse!46}$-$.24 & \cellcolor{dworse!60}$+$18.2 & \cellcolor{dworse!52}$-$43.9 & \cellcolor{dworse!59}$+$6.5 & \cellcolor{dworse!49}$-$.31 & \cellcolor{dworse!60}$+$10.8 \\
coarse 4-sphere foot & $-$0.3 & $+$0.04 & \cellcolor{dworse!9}$-$.01 & $+$0.2 & \cellcolor{dworse!6}$-$0.3 & \cellcolor{dworse!10}$+$0.04 & \cellcolor{dworse!10}$-$.01 & $+$0.1 & 0.0 & \cellcolor{dworse!17}$+$0.11 & \cellcolor{dworse!16}$-$.02 & \cellcolor{dworse!19}$+$0.5 & \cellcolor{dworse!9}$-$1.1 & $+$0.09 & \cellcolor{dworse!9}$-$.01 & $-$0.04 \\
\bottomrule
\end{tabular}%
}
\vspace{-2pt}%
\parbox{\textwidth}{\scriptsize\raggedright
  $^\dagger$9 main joints (hip flexion / adduction, knee, ankle, both sides; lumbar extension), against inverse dynamics. Columns as in Table~\ref{tab:golden_full}. Shading = relative change on a log scale, 1\,\% \tintbar{dworse} ${\geq}$100\,\% if worse, green if better, none within one seed sd.}
\par\vspace{-3.8pt}%
\end{table}}

\paragraph{Robustness.}
\label{App:robustness}
We re-score the seen-split policies of \projname{}, Fixed-PD, Torque and Muscle under eleven test-time perturbations (Table~\ref{tab:robustness}), where \projname{} keeps the lowest joint-angle error.
Pushes of magnitude uniform in $[0, b]$ body weights act on the pelvis in a random direction for $0.2$\,s, the $20$\,ms delay applies to every simulator-derived observation except Muscle's activations, and the contact spheres are rescaled at fixed sole height or reduced to four.
Pushes and changes in contact compliance, friction or foot discretization cost no controller more than $1.3$ points of survival or $0.06$ in $F_{\mathrm{Z}_r}$, except Muscle under the strongest pushes, whereas observation delay separates the action spaces by where their feedback is closed (Sec.~\ref{sec:benchmark}).
Enlarged contact spheres degrade every controller, and since smaller or fewer spheres leave every controller unchanged, this failure points to rolling geometry, as a larger radius moves the contact point further per degree of foot rotation and reshapes the heel-strike and push-off rocker on which every policy was trained.

\begin{table}[H]%
\centering
\caption{\textbf{Observation delay across integration schemes and gain levels.} The seen-split policies of Fixed-PD and \projname{} under each integration scheme are re-scored without retraining under the $20$\,ms observation delay of Table~\ref{tab:robustness}. Activity mean$_{\pm\text{sd}}$ over 5 seeds.}
\vspace{-0.4em}
\label{tab:robustness_integrators}
\scriptsize
\setlength{\tabcolsep}{0pt}%
\setlength{\aboverulesep}{0pt}
\newcommand{\sd}[1]{{\tiny\textcolor{black!55}{$\pm$#1}}}
\begin{tabular}{@{}l<{\hspace{20.00pt}}l>{\hspace{37.60pt}}c>{\hspace{11.00pt}}c>{\hspace{37.60pt}}c>{\hspace{11.00pt}}c>{\hspace{37.60pt}}c>{\hspace{11.00pt}}c@{}}
\toprule
& & \multicolumn{2}{>{\hspace{37.60pt}}c}{\textbf{\boldmath Ep.$\hi$ (\%)}} & \multicolumn{2}{>{\hspace{37.60pt}}c}{\textbf{\boldmath $q\lo$ ($^\circ$)}} & \multicolumn{2}{>{\hspace{37.60pt}}c}{\boldmath$F^{\uparrow}_{\mathrm{Z}_r}$} \\
\cmidrule(l{34.85pt}r{0pt}){3-4}\cmidrule(l{34.85pt}r{0pt}){5-6}\cmidrule(l{34.85pt}r{0pt}){7-8}
\textbf{PD} & \textbf{Method} & $0$\,ms & $20$\,ms & $0$\,ms & $20$\,ms & $0$\,ms & $20$\,ms \\
\midrule
 & Fixed-PD & 93.0\sd{.2} & 65.6\sd{1.3} & 2.91\sd{.03} & 4.67\sd{.07} & .73\sd{.00} & .51\sd{.01} \\
 & Fixed-PD, $2\times$ & 96.2\sd{.2} & 60.1\sd{1.0} & 2.53\sd{.02} & 5.82\sd{.08} & .72\sd{.00} & .42\sd{.01} \\
\rowcolor{imptint}\cellcolor{white} & \textcolor[RGB]{229,57,53}{\textbf{\projname{}}} & 96.9\sd{.5} & 93.5\sd{.3} & 2.28\sd{.04} & 3.52\sd{.02} & .80\sd{.00} & .73\sd{.00} \\
\rowcolor{imptint}\cellcolor{white}\multirow{-4}{*}{Explicit} & \textcolor[RGB]{229,57,53}{\textbf{\projname{}, $2\times$}} & 96.9\sd{.2} & 86.6\sd{1.8} & 2.30\sd{.02} & 4.12\sd{.13} & .79\sd{.00} & .65\sd{.01} \\
\midrule
 & Fixed-PD & 93.5\sd{.4} & 89.1\sd{.8} & 2.44\sd{.02} & 3.33\sd{.05} & .75\sd{.01} & .69\sd{.00} \\
 & Fixed-PD, $2\times$ & 96.6\sd{.5} & 90.7\sd{.7} & 2.10\sd{.02} & 3.22\sd{.05} & .75\sd{.01} & .63\sd{.01} \\
\rowcolor{imptint}\cellcolor{white} & \textcolor[RGB]{229,57,53}{\textbf{\projname{}}} & 93.8\sd{.3} & 91.6\sd{.2} & 2.22\sd{.03} & 3.09\sd{.04} & .79\sd{.00} & .74\sd{.00} \\
\rowcolor{imptint}\cellcolor{white}\multirow{-4}{*}{Stable} & \textcolor[RGB]{229,57,53}{\textbf{\projname{}, $2\times$}} & 96.3\sd{.5} & 93.5\sd{1.0} & 2.07\sd{.02} & 3.05\sd{.08} & .77\sd{.00} & .70\sd{.01} \\
\midrule
 & Fixed-PD & 95.7\sd{.2} & 84.4\sd{.7} & 2.52\sd{.03} & 3.71\sd{.06} & .72\sd{.00} & .61\sd{.01} \\
 & Fixed-PD, $2\times$ & 95.8\sd{.4} & 72.3\sd{1.2} & 2.21\sd{.02} & 4.09\sd{.10} & .67\sd{.00} & .44\sd{.01} \\
\rowcolor{imptint}\cellcolor{white} & \textcolor[RGB]{229,57,53}{\textbf{\projname{}}} & 97.2\sd{.3} & 91.4\sd{.2} & 2.22\sd{.02} & 3.30\sd{.02} & .77\sd{.00} & .68\sd{.00} \\
\rowcolor{imptint}\cellcolor{white}\multirow{-4}{*}{Implicit} & \textcolor[RGB]{229,57,53}{\textbf{\projname{}, $2\times$}} & 96.8\sd{.2} & 82.0\sd{.8} & 2.16\sd{.03} & 3.60\sd{.06} & .72\sd{.00} & .56\sd{.00} \\
\bottomrule
\end{tabular}
\end{table}

\vspace{-13pt}%
\paragraph{Gain-range sensitivity.}
The log-ranges $s_p$ and $s_d$ are narrowed from their defaults of $1.25$ and $1.5$ and the activation lag $T_{\mathrm{act}}$ is varied around $50$\,ms, one at a time (Table~\ref{tab:sensitivity}).
The values bracket both bounds, as the ankle can drop below Bound~2 only for $s_d > 0.27$ and running can reach Bound~1 only for $s_p > 0.35$.
Survival stays within $2.4$ points of the default throughout, and narrowing the stiffness range costs little, whereas torque jerk remains five times the default even at $s_d = 0.50$, so smooth tracking needs a damping range that reaches well below Bound~2 rather than merely across it.
A $200$\,ms lag also triples torque jerk, as the gains then trail the contact events.

{\setlength{\intextsep}{5pt}%
\vspace{7.9pt}%
\begin{table}[H]
\centering
\caption{\textbf{Sensitivity to the gain ranges and gain lag.} Seen subjects, held-out clips, activity mean$_{\pm\text{sd}}$ over 5 seeds. Other rows vary one parameter and show the change from the first row.}
\label{tab:sensitivity}
\vspace{-0.4em}%
\scriptsize%
\renewcommand{\arraystretch}{0.85}
\setlength{\tabcolsep}{4pt}
\newcommand{\sd}[1]{{\tiny\textcolor{black!55}{$\pm$#1}}}
\newcommand{\secrule}{\arrayrulecolor{black!25}\midrule[0.3pt]\arrayrulecolor{black}}
\begin{tabular*}{\textwidth}{@{\extracolsep{\fill}}lccccc@{}}
\toprule
\textbf{Configuration} & \textbf{\boldmath Ep.$\hi$} & {\boldmath$q\lo$} & {\boldmath$F^{\uparrow}_{\mathrm{Z}_r}$} & {\boldmath$\dot{\tau}\lo{}^{\dagger}$} & {\boldmath$p_x\lo$} \\
\midrule
\projname{} ($s_p = 1.25$, $s_d = 1.5$, $T_{\mathrm{act}} = 50$\,ms) & 96.9\sd{.5} & 2.28\sd{.04} & .80\sd{.00} & 15.9\sd{.4} & .040\sd{.001} \\
\secrule
$s_p = 0.10$ & \cellcolor{dworse!17}$-$2.4 & \cellcolor{dworse!34}$+$0.24 & \cellcolor{dworse!21}$-$.03 & \cellcolor{dworse!32}$+$1.5 & \cellcolor{dworse!38}$+$.006 \\
$s_p = 0.35$ & \cellcolor{dworse!16}$-$2.3 & \cellcolor{dworse!30}$+$0.17 & \cellcolor{dworse!17}$-$.02 & \cellcolor{dworse!28}$+$1.0 & \cellcolor{dworse!36}$+$.005 \\
$s_p = 0.75$ & \cellcolor{dworse!10}$-$1.4 & \cellcolor{dworse!21}$+$0.08 & \cellcolor{dworse!9}$-$.01 & $-$0.1 & $+$.001 \\
\secrule
$s_d = 0.10$ & \cellcolor{dworse!16}$-$2.2 & \cellcolor{dworse!49}$+$0.90 & \cellcolor{dworse!39}$-$.13 & \cellcolor{dworse!60}$+$86 & \cellcolor{dworse!47}$+$.013 \\
$s_d = 0.25$ & \cellcolor{dworse!10}$-$1.4 & \cellcolor{dworse!46}$+$0.72 & \cellcolor{dworse!34}$-$.09 & \cellcolor{dworse!60}$+$74.7 & \cellcolor{dworse!43}$+$.009 \\
$s_d = 0.50$ & \cellcolor{dworse!9}$-$1.2 & \cellcolor{dworse!42}$+$0.48 & \cellcolor{dworse!30}$-$.06 & \cellcolor{dworse!60}$+$61.9 & \cellcolor{dworse!36}$+$.005 \\
\secrule
$T_{\mathrm{act}} = 0$ (no lag) & $+$0.1 & \cellcolor{dworse!15}$+$0.05 & .00 & \cellcolor{dworse!25}$+$0.8 & .000 \\
$T_{\mathrm{act}} = 100$\,ms & \cellcolor{dworse!6}$-$0.8 & \cellcolor{dworse!25}$+$0.12 & \cellcolor{dworse!9}$-$.01 & \cellcolor{dworse!52}$+$8.0 & \cellcolor{dworse!25}$+$.002 \\
$T_{\mathrm{act}} = 200$\,ms & \cellcolor{dworse!17}$-$2.4 & \cellcolor{dworse!36}$+$0.29 & \cellcolor{dworse!21}$-$.03 & \cellcolor{dworse!60}$+$33.2 & \cellcolor{dworse!33}$+$.004 \\
\bottomrule
\end{tabular*}
\vspace{0pt}
\parbox{\textwidth}{\scriptsize\raggedright
  $^\dagger$9 main joints (hip flexion / adduction, knee, ankle, both sides; lumbar extension), against inverse dynamics. Columns as in Table~\ref{tab:golden_full}. Shading = relative change on a log scale, 1\,\% \tintbar{dworse} ${\geq}$100\,\% if worse, green if better, none within one seed sd.}
\par\vspace{-1.9pt}%
\end{table}}

\paragraph{Reward factorial.}
Without the GRF reward (Table~\ref{tab:reward_factorial}, Fig.~\ref{fig:benchmark}a--c), every action space lowers its joint-angle error by up to $3.1^\circ$ with survival unchanged or slightly higher, yet its forces move further from the reference, the vertical-force correlation falling by $0.12$ to $0.16$ and the horizontal ones by up to $0.25$, while torque jerk rises, nearly doubling for Torque.
\projname{} gains the least from the reward and, without it, overtakes Torque in vertical-force correlation, so it recovers the most force from kinematic tracking alone, whereas Torque depends most on direct force supervision.

{\setlength{\intextsep}{3pt}%
\begin{table}[H]
\centering
\caption{%
  \textbf{Training without the GRF reward.} Seen subjects, held-out clips, mean$_{\pm\text{sd}}$ over 5 seeds.
  Columns are RMSE ($\downarrow$) unless marked $\uparrow$.
  \textbf{Bold} / \underline{underline} = best / second best per activity.}
\vspace{-0.4em}
\label{tab:reward_factorial}
\renewcommand{\arraystretch}{0.80}
\setlength{\tabcolsep}{2pt}
\newcommand{\gap}{\hspace{4pt}}
\newcommand{\sd}[1]{{\tiny\textcolor{black!55}{$\pm$#1}}}
\scriptsize
\resizebox{\textwidth}{!}{%
\begin{tabular}{@{}>{\fillL}l @{\gap} l @{\gap} lllll @{\gap} lp{0.75cm}p{0.75cm}p{0.75cm}ll @{\gap} ll>{\fillR}l @{}}
\toprule
 & & \multicolumn{1}{c}{\textbf{Stability}} & \multicolumn{4}{c}{\textbf{Kinematics}} & \multicolumn{6}{c}{\textbf{Ground Reaction Forces}} & \multicolumn{3}{c}{\textbf{Torques}$^\dagger$} \\
\cmidrule(lr){3-3}\cmidrule(lr){4-7}\cmidrule(lr){8-13}\cmidrule(lr){14-16}
\texttt{\textbf{Activity}} & \textbf{Method}
  & \makecell{Ep.$\hi$ \\ (\%)}
  & \makecell{$q\lo$\\ \makebox[0pt][c]{($^\circ$)}}
  & \makecell{$\dot{q}\lo$\\ \makebox[0pt][c]{($^\circ$/s)}}
  & \makecell{$p_x\lo$\\ \makebox[0pt][c]{(m/s)}}
  & kin$_r\hi$
  & \makecell{$F_{\mathrm{Z}}\lo$\\ \makebox[0pt][c]{(BW)}}
  & $F^{\uparrow}_{\mathrm{Z}_r}$
  & $F^{\uparrow}_{\mathrm{ML}_r}$
  & $F^{\uparrow}_{\mathrm{AP}_r}$
  & \makecell{$\mathrm{CoP}^{\downarrow}_{\mathrm{ML}}$\\ {\tiny (mm)}}
  & \makecell{$\mathrm{CoP}^{\downarrow}_{\mathrm{AP}}$\\ {\tiny (mm)}}
  & \makecell{$\tau\lo$\\ \makebox[0pt][c]{\tiny (Nm/kg)}}
  & $\tau_r\hi$
  & \makecell{$\dot{\tau}\lo$\\ \makebox[0pt][c]{\tiny (Nm/kg/s)}} \\
\midrule
\multirow{4}{*}{\act{walk}} & Fixed-PD & 99.7\sd{.1} & \underline{1.81}\sd{.03} & 63.8\sd{.7} & .011\sd{.001} & \underline{.98}\sd{.00} & .38\sd{.01} & .67\sd{.01} & .29\sd{.01} & .21\sd{.01} & 23\sd{2} & 52\sd{2} & .76\sd{.01} & .34\sd{.00} & 49.9\sd{1.2} \\
 & Torque & \underline{99.9}\sd{.2} & 1.92\sd{.03} & 58.6\sd{1.6} & \underline{.009}\sd{.001} & .95\sd{.00} & .29\sd{.00} & .79\sd{.01} & .57\sd{.02} & .45\sd{.02} & \textbf{21}\sd{1} & \underline{48}\sd{1} & .52\sd{.00} & .47\sd{.01} & 18.8\sd{.9} \\
 & Muscle & 99.0\sd{.2} & 2.00\sd{.01} & \underline{36.6}\sd{.3} & .022\sd{.002} & .96\sd{.00} & \textbf{.23}\sd{.01} & \textbf{.86}\sd{.01} & \textbf{.71}\sd{.03} & \textbf{.58}\sd{.01} & 22\sd{1} & 56\sd{3} & \textbf{.44}\sd{.01} & \textbf{.57}\sd{.01} & \textbf{8.30}\sd{.25} \\
 & ImpedanceMimic & \textbf{100}\sd{0} & \textbf{1.25}\sd{.01} & \textbf{28.1}\sd{.3} & \textbf{.008}\sd{.000} & \textbf{.99}\sd{.00} & \underline{.27}\sd{.00} & \underline{.81}\sd{.01} & \underline{.66}\sd{.01} & \underline{.50}\sd{.01} & \textbf{21}\sd{0} & \textbf{45}\sd{1} & \underline{.49}\sd{.00} & \underline{.48}\sd{.00} & \underline{11.1}\sd{.3} \\
\addlinespace[3pt]
\multirow{4}{*}{\act{run}} & Fixed-PD & 99.7\sd{.2} & \underline{2.87}\sd{.04} & 107\sd{1} & .025\sd{.001} & \underline{.97}\sd{.00} & .38\sd{.02} & .88\sd{.01} & .15\sd{.01} & .14\sd{.01} & 23\sd{0} & 49\sd{1} & 1.36\sd{.02} & .38\sd{.01} & 117\sd{4} \\
 & Torque & \textbf{100}\sd{0} & 2.89\sd{.03} & 123\sd{4} & \underline{.022}\sd{.002} & .95\sd{.00} & .35\sd{.01} & .90\sd{.00} & .27\sd{.01} & .29\sd{.02} & \underline{21}\sd{0} & 53\sd{1} & .96\sd{.01} & .53\sd{.01} & 35.0\sd{1.2} \\
 & Muscle & 99.8\sd{.2} & 3.20\sd{.04} & \underline{67.8}\sd{.6} & .038\sd{.003} & .96\sd{.00} & \textbf{.25}\sd{.00} & \textbf{.95}\sd{.00} & \underline{.33}\sd{.05} & \textbf{.55}\sd{.04} & \textbf{20}\sd{1} & \textbf{46}\sd{3} & \textbf{.86}\sd{.01} & \textbf{.64}\sd{.01} & \textbf{14.1}\sd{.3} \\
 & ImpedanceMimic & \textbf{100}\sd{0} & \textbf{2.01}\sd{.04} & \textbf{57.6}\sd{1.5} & \textbf{.016}\sd{.000} & \textbf{.98}\sd{.00} & \underline{.28}\sd{.02} & \underline{.94}\sd{.01} & \textbf{.37}\sd{.01} & \underline{.44}\sd{.02} & \underline{21}\sd{0} & \underline{48}\sd{0} & \underline{.90}\sd{.01} & \underline{.62}\sd{.01} & \underline{21.8}\sd{.6} \\
\addlinespace[3pt]
\multirow{4}{*}{\act{stand}} & Fixed-PD & 97.9\sd{.5} & 1.14\sd{.02} & 10.5\sd{.9} & .006\sd{.001} & \underline{.75}\sd{.01} & .12\sd{.00} & .28\sd{.02} & .18\sd{.03} & \textbf{.01}\sd{.02} & 17\sd{0} & 38\sd{1} & \underline{.65}\sd{.01} & .13\sd{.01} & 8.44\sd{.82} \\
 & Torque & \underline{99.5}\sd{.3} & 1.64\sd{.02} & 9.96\sd{.36} & \textbf{.003}\sd{.000} & .62\sd{.01} & \textbf{.07}\sd{.00} & \textbf{.41}\sd{.03} & \underline{.33}\sd{.04} & .00\sd{.01} & \textbf{13}\sd{0} & \textbf{23}\sd{1} & \textbf{.48}\sd{.01} & \textbf{.16}\sd{.01} & 1.98\sd{.04} \\
 & Muscle & 97.4\sd{.1} & \underline{.99}\sd{.03} & \underline{5.82}\sd{.40} & .007\sd{.000} & .67\sd{.02} & .13\sd{.01} & .29\sd{.05} & .26\sd{.13} & \textbf{.01}\sd{.02} & 17\sd{1} & 42\sd{3} & .90\sd{.05} & .09\sd{.01} & \textbf{1.34}\sd{.39} \\
 & ImpedanceMimic & \textbf{99.8}\sd{.0} & \textbf{.71}\sd{.01} & \textbf{4.18}\sd{.13} & \underline{.004}\sd{.001} & \textbf{.82}\sd{.00} & \underline{.09}\sd{.00} & \underline{.34}\sd{.01} & \textbf{.37}\sd{.05} & .00\sd{.01} & \underline{16}\sd{0} & \underline{31}\sd{1} & .87\sd{.01} & \textbf{.16}\sd{.01} & \underline{1.40}\sd{.07} \\
\addlinespace[3pt]
\multirow{4}{*}{\act{lateral}} & Fixed-PD & 96.3\sd{.4} & \underline{3.33}\sd{.09} & 78.3\sd{1.4} & \underline{.080}\sd{.003} & \underline{.94}\sd{.00} & .44\sd{.01} & .71\sd{.01} & .60\sd{.01} & .39\sd{.02} & 33\sd{4} & 72\sd{18} & .88\sd{.02} & .38\sd{.01} & 52.6\sd{.4} \\
 & Torque & \underline{98.2}\sd{.2} & 5.43\sd{.23} & 189\sd{8} & .086\sd{.005} & .82\sd{.00} & .42\sd{.01} & .72\sd{.02} & .56\sd{.01} & .40\sd{.01} & 33\sd{3} & 67\sd{4} & .83\sd{.01} & .35\sd{.01} & 56.6\sd{.4} \\
 & Muscle & 92.6\sd{1.0} & 6.15\sd{.03} & \underline{71.4}\sd{1.1} & .102\sd{.008} & .79\sd{.01} & \textbf{.39}\sd{.01} & \underline{.73}\sd{.02} & \underline{.64}\sd{.01} & \underline{.42}\sd{.02} & \textbf{30}\sd{1} & \underline{65}\sd{2} & \textbf{.65}\sd{.01} & \underline{.39}\sd{.01} & \textbf{13.0}\sd{.2} \\
 & ImpedanceMimic & \textbf{99.8}\sd{.1} & \textbf{2.79}\sd{.07} & \textbf{68.6}\sd{1.3} & \textbf{.070}\sd{.003} & \textbf{.95}\sd{.00} & \textbf{.39}\sd{.01} & \textbf{.77}\sd{.01} & \textbf{.66}\sd{.01} & \textbf{.47}\sd{.02} & \underline{31}\sd{4} & \textbf{62}\sd{4} & \underline{.77}\sd{.01} & \textbf{.44}\sd{.01} & \underline{33.2}\sd{.9} \\
\addlinespace[3pt]
\multirow{4}{*}{\act{flexion}} & Fixed-PD & 95.9\sd{.6} & \underline{2.13}\sd{.02} & 49.7\sd{1.7} & .037\sd{.004} & \underline{.95}\sd{.00} & .29\sd{.01} & .52\sd{.01} & .25\sd{.03} & .33\sd{.01} & \underline{24}\sd{1} & 65\sd{1} & .63\sd{.01} & .38\sd{.01} & 34.2\sd{1.3} \\
 & Torque & \underline{98.1}\sd{.4} & 3.43\sd{.18} & 102\sd{9} & \textbf{.030}\sd{.004} & .88\sd{.01} & \underline{.24}\sd{.01} & \underline{.57}\sd{.01} & .32\sd{.02} & \underline{.36}\sd{.01} & 25\sd{2} & \underline{64}\sd{5} & \underline{.52}\sd{.01} & \underline{.40}\sd{.01} & 24.8\sd{1.3} \\
 & Muscle & 66.4\sd{1.2} & 5.96\sd{.20} & \textbf{33.7}\sd{.3} & .064\sd{.004} & .79\sd{.00} & .25\sd{.01} & .51\sd{.03} & \textbf{.40}\sd{.03} & \textbf{.41}\sd{.02} & 29\sd{5} & 75\sd{1} & \textbf{.48}\sd{.01} & .36\sd{.01} & \textbf{5.24}\sd{.27} \\
 & ImpedanceMimic & \textbf{99.8}\sd{.2} & \textbf{1.76}\sd{.05} & \underline{36.7}\sd{.7} & \underline{.031}\sd{.003} & \textbf{.96}\sd{.00} & \textbf{.23}\sd{.00} & \textbf{.62}\sd{.01} & \underline{.37}\sd{.02} & \underline{.36}\sd{.01} & \textbf{23}\sd{0} & \textbf{63}\sd{1} & \underline{.52}\sd{.01} & \textbf{.45}\sd{.00} & \underline{14.5}\sd{.5} \\
\addlinespace[3pt]
\multirow{4}{*}{\act{jump}} & Fixed-PD & 76.9\sd{.9} & \underline{3.14}\sd{.05} & 79.2\sd{1.0} & \underline{.067}\sd{.004} & \underline{.93}\sd{.00} & .44\sd{.01} & .52\sd{.01} & .25\sd{.01} & .23\sd{.01} & \textbf{27}\sd{1} & \underline{72}\sd{1} & .93\sd{.01} & .27\sd{.01} & 70.1\sd{1.2} \\
 & Torque & \underline{81.4}\sd{.9} & 5.85\sd{.21} & 187\sd{4} & .069\sd{.005} & .80\sd{.01} & \textbf{.38}\sd{.00} & \textbf{.60}\sd{.01} & .34\sd{.02} & \underline{.26}\sd{.02} & 32\sd{6} & 74\sd{10} & \underline{.69}\sd{.01} & \underline{.34}\sd{.01} & 42.1\sd{.8} \\
 & Muscle & 57.5\sd{1.8} & 6.38\sd{.31} & \textbf{64.6}\sd{.9} & .114\sd{.009} & .76\sd{.01} & \underline{.40}\sd{.01} & .46\sd{.01} & \underline{.36}\sd{.03} & .22\sd{.03} & 36\sd{8} & 88\sd{10} & \textbf{.59}\sd{.00} & .30\sd{.01} & \textbf{9.74}\sd{.42} \\
 & ImpedanceMimic & \textbf{83.3}\sd{1.2} & \textbf{2.66}\sd{.07} & \underline{67.4}\sd{2.5} & \textbf{.059}\sd{.004} & \textbf{.94}\sd{.00} & \underline{.40}\sd{.01} & \underline{.58}\sd{.01} & \textbf{.39}\sd{.02} & \textbf{.32}\sd{.01} & \underline{28}\sd{1} & \textbf{68}\sd{1} & .73\sd{.02} & \textbf{.35}\sd{.01} & \underline{36.9}\sd{2.4} \\
\specialrule{\cmidrulewidth}{\aboverulesep}{0pt}
\rowcolor{meangray}  & \padT Fixed-PD & 94.4\sd{.3} & \underline{2.40}\sd{.02} & 64.7\sd{.5} & .038\sd{.001} & \underline{.92}\sd{.00} & .34\sd{.01} & .59\sd{.01} & .29\sd{.01} & .22\sd{.01} & 25\sd{1} & 58\sd{3} & .87\sd{.01} & .31\sd{.00} & 55.3\sd{.9} \\
\rowcolor{meangray}  & Torque & \underline{96.2}\sd{.1} & 3.53\sd{.03} & 112\sd{2} & \underline{.036}\sd{.001} & .84\sd{.00} & .29\sd{.00} & \underline{.66}\sd{.01} & .40\sd{.00} & .29\sd{.01} & \underline{24}\sd{1} & \underline{55}\sd{2} & \underline{.67}\sd{.00} & .37\sd{.00} & 29.9\sd{.3} \\
\rowcolor{meangray}  & Muscle & 85.5\sd{.6} & 4.11\sd{.06} & \underline{46.6}\sd{.4} & .058\sd{.002} & .82\sd{.00} & \textbf{.28}\sd{.01} & .63\sd{.01} & \underline{.45}\sd{.02} & \textbf{.36}\sd{.01} & 26\sd{2} & 62\sd{3} & \textbf{.65}\sd{.01} & \underline{.39}\sd{.00} & \textbf{8.61}\sd{.12} \\
\rowcolor{meangray} \multirow{-4}{*}{\textbf{Mean}} & \padB ImpedanceMimic & \textbf{97.1}\sd{.2} & \textbf{1.86}\sd{.02} & \textbf{43.8}\sd{.8} & \textbf{.031}\sd{.001} & \textbf{.94}\sd{.00} & \textbf{.28}\sd{.00} & \textbf{.68}\sd{.00} & \textbf{.47}\sd{.01} & \underline{.35}\sd{.01} & \textbf{23}\sd{1} & \textbf{53}\sd{0} & .72\sd{.00} & \textbf{.42}\sd{.00} & \underline{19.8}\sd{.4} \\
\specialrule{\heavyrulewidth}{0pt}{\belowbottomsep}
\end{tabular}%
}
\vspace{0pt}
\parbox{\textwidth}{\scriptsize\raggedright
  $^\dagger$9 main joints (hip flexion / adduction, knee, ankle, both sides; lumbar extension), against inverse dynamics.
  Ep. = episode survival before termination (\%). $q$, $\dot q$ = joint-angle / velocity RMSE. $p_x$ = root forward-position drift rate. \{kin$_r$, $F_{\mathrm{Z}_r}$, $F_{\mathrm{AP}_r}$, $F_{\mathrm{ML}_r}$, $\tau_r$\} = Pearson correlation w.r.t. reference \{joint angles, vertical, AP, ML GRF, torques\}. CoP = center of pressure. $\dot\tau$ = torque jerk.}
\vspace{-1em}
\end{table}}
\paragraph{Ablations.}
Each ablation removes one element of \projname{} and retrains it on the seen split (Table~\ref{tab:ablation_seen}, Fig.~\ref{fig:benchmark}d), three of them and the oracle also on unseen subjects (Table~\ref{tab:ablation_unseen}), where removing the inertia scaling nearly triples torque jerk and removing the force estimate lowers the mean $F_{\mathrm{Z}_r}$ by $0.04$.
The end-effector term secures survival and limits drift, the root-velocity term suppresses drift at some cost in jerk, and amplitude normalization favors kinematics over force.
Since only removing the end-effector term lowers survival, the elements shape the quality of the motion rather than the ability to complete it.

\vfill%
{\setlength{\intextsep}{3pt}%
\begin{table}[H]
\centering
\caption{%
  \textbf{Ablations of \projname{}.} Unseen subjects, zero-shot, mean$_{\pm\text{sd}}$ over 5 seeds. Rows marked ``$-$'' remove one element, \emph{Oracle GRF obs} observes the force-plate GRF, and \textbf{bold} / \underline{underline} mark the best / second best per activity.}
\vspace{-0.4em}
\label{tab:ablation_unseen}
\renewcommand{\arraystretch}{0.80}
\setlength{\tabcolsep}{2pt}
\newcommand{\gap}{\hspace{4pt}}
\newcommand{\sd}[1]{{\tiny\textcolor{black!55}{$\pm$#1}}}
\scriptsize
\resizebox{\textwidth}{!}{%
\begin{tabular}{@{}>{\fillL}l @{\gap} l @{\gap} lllll @{\gap} lp{0.75cm}p{0.75cm}p{0.75cm}ll @{\gap} ll>{\fillR}l @{}}
\toprule
 & & \multicolumn{1}{c}{\textbf{Stability}} & \multicolumn{4}{c}{\textbf{Kinematics}} & \multicolumn{6}{c}{\textbf{Ground Reaction Forces}} & \multicolumn{3}{c}{\textbf{Torques}$^\dagger$} \\
\cmidrule(lr){3-3}\cmidrule(lr){4-7}\cmidrule(lr){8-13}\cmidrule(lr){14-16}
\texttt{\textbf{Activity}} & \textbf{Method}
  & \makecell{Ep.$\hi$ \\ (\%)}
  & \makecell{$q\lo$\\ \makebox[0pt][c]{($^\circ$)}}
  & \makecell{$\dot{q}\lo$\\ \makebox[0pt][c]{($^\circ$/s)}}
  & \makecell{$p_x\lo$\\ \makebox[0pt][c]{(m/s)}}
  & kin$_r\hi$
  & \makecell{$F_{\mathrm{Z}}\lo$\\ \makebox[0pt][c]{(BW)}}
  & $F^{\uparrow}_{\mathrm{Z}_r}$
  & $F^{\uparrow}_{\mathrm{ML}_r}$
  & $F^{\uparrow}_{\mathrm{AP}_r}$
  & \makecell{$\mathrm{CoP}^{\downarrow}_{\mathrm{ML}}$\\ {\tiny (mm)}}
  & \makecell{$\mathrm{CoP}^{\downarrow}_{\mathrm{AP}}$\\ {\tiny (mm)}}
  & \makecell{$\tau\lo$\\ \makebox[0pt][c]{\tiny (Nm/kg)}}
  & $\tau_r\hi$
  & \makecell{$\dot{\tau}\lo$\\ \makebox[0pt][c]{\tiny (Nm/kg/s)}} \\
\midrule
\multirow{5}{*}{\act{walk}} & ImpedanceMimic & \textbf{100}\sd{0} & \textbf{1.95}\sd{.03} & \textbf{32.7}\sd{.3} & \textbf{.016}\sd{.001} & \textbf{.98}\sd{.00} & \textbf{.11}\sd{.00} & \textbf{.97}\sd{.00} & \textbf{.88}\sd{.01} & \textbf{.85}\sd{.01} & \textbf{23}\sd{1} & 39\sd{1} & \textbf{.38}\sd{.01} & \textbf{.65}\sd{.01} & 6.78\sd{.11} \\
 & \textit{Oracle GRF obs} & \textit{100}\sd{0} & \textit{1.97}\sd{.02} & \textit{32.8}\sd{.4} & \textit{.017}\sd{.001} & \textit{.98}\sd{.00} & \textit{.11}\sd{.00} & \textit{.97}\sd{.00} & \textit{.89}\sd{.00} & \textit{.86}\sd{.01} & \textit{22}\sd{1} & \textit{39}\sd{1} & \textit{.38}\sd{.00} & \textit{.66}\sd{.01} & \textit{6.68}\sd{.16} \\
 & - Inertia scaling & \textbf{100}\sd{0} & 2.26\sd{.06} & 74.4\sd{4.7} & .019\sd{.002} & .97\sd{.00} & .15\sd{.01} & .95\sd{.00} & .81\sd{.02} & .74\sd{.02} & \textbf{23}\sd{0} & \textbf{38}\sd{1} & .62\sd{.02} & .50\sd{.01} & 39.7\sd{4.0} \\
 & - Estimator obs & \textbf{100}\sd{0} & \textbf{1.95}\sd{.01} & \underline{32.8}\sd{.6} & .017\sd{.002} & \textbf{.98}\sd{.00} & .12\sd{.00} & .96\sd{.00} & \underline{.87}\sd{.00} & .84\sd{.01} & \textbf{23}\sd{1} & 39\sd{0} & \textbf{.38}\sd{.00} & \textbf{.65}\sd{.01} & \textbf{6.46}\sd{.11} \\
 & - Learned residual & \textbf{100}\sd{0} & 1.98\sd{.02} & 33.1\sd{.5} & \textbf{.016}\sd{.002} & \textbf{.98}\sd{.00} & \textbf{.11}\sd{.00} & \textbf{.97}\sd{.00} & \underline{.87}\sd{.01} & \textbf{.85}\sd{.01} & \textbf{23}\sd{0} & \textbf{38}\sd{0} & \textbf{.38}\sd{.00} & \textbf{.65}\sd{.01} & \underline{6.62}\sd{.15} \\
\addlinespace[3pt]
\multirow{5}{*}{\act{run}} & ImpedanceMimic & \textbf{100}\sd{0} & \textbf{2.24}\sd{.04} & \underline{63.2}\sd{1.6} & .052\sd{.006} & \textbf{.98}\sd{.00} & \textbf{.14}\sd{.01} & \textbf{.98}\sd{.00} & \textbf{.47}\sd{.03} & \textbf{.89}\sd{.01} & \textbf{21}\sd{1} & \underline{41}\sd{1} & \textbf{.95}\sd{.01} & \textbf{.69}\sd{.01} & \textbf{20.6}\sd{2.0} \\
 & \textit{Oracle GRF obs} & \textit{99.9}\sd{.2} & \textit{2.28}\sd{.04} & \textit{64.1}\sd{1.5} & \textit{.048}\sd{.011} & \textit{.98}\sd{.00} & \textit{.13}\sd{.00} & \textit{.99}\sd{.00} & \textit{.49}\sd{.06} & \textit{.90}\sd{.01} & \textit{21}\sd{0} & \textit{41}\sd{2} & \textit{.94}\sd{.01} & \textit{.69}\sd{.01} & \textit{22.4}\sd{2.3} \\
 & - Inertia scaling & \textbf{100}\sd{0} & 2.51\sd{.06} & 151\sd{2} & .052\sd{.005} & .97\sd{.00} & .19\sd{.00} & .97\sd{.00} & .28\sd{.06} & .81\sd{.02} & \textbf{21}\sd{0} & 42\sd{2} & 1.47\sd{.01} & .44\sd{.01} & 144\sd{4} \\
 & - Estimator obs & 99.8\sd{.5} & 2.27\sd{.03} & \textbf{62.8}\sd{.8} & \textbf{.047}\sd{.014} & \textbf{.98}\sd{.00} & \underline{.15}\sd{.01} & \textbf{.98}\sd{.00} & \textbf{.47}\sd{.03} & \textbf{.89}\sd{.01} & \textbf{21}\sd{1} & \textbf{40}\sd{2} & \textbf{.95}\sd{.01} & \textbf{.69}\sd{.01} & \underline{21.1}\sd{.4} \\
 & - Learned residual & 99.8\sd{.3} & \underline{2.26}\sd{.03} & 64.5\sd{1.9} & \underline{.048}\sd{.002} & \textbf{.98}\sd{.00} & \underline{.15}\sd{.00} & \textbf{.98}\sd{.00} & .44\sd{.07} & .88\sd{.01} & \textbf{21}\sd{0} & \underline{41}\sd{2} & \textbf{.95}\sd{.01} & \textbf{.69}\sd{.01} & 22.1\sd{3.1} \\
\addlinespace[3pt]
\multirow{5}{*}{\act{stand}} & ImpedanceMimic & \textbf{99.6}\sd{.2} & .98\sd{.02} & \textbf{3.70}\sd{.16} & \textbf{.003}\sd{.000} & .66\sd{.01} & \textbf{.06}\sd{.00} & \textbf{.54}\sd{.03} & \textbf{.88}\sd{.03} & .18\sd{.04} & \textbf{13}\sd{0} & \textbf{30}\sd{3} & \textbf{.89}\sd{.01} & .18\sd{.01} & 1.14\sd{.15} \\
 & \textit{Oracle GRF obs} & \textit{99.7}\sd{.1} & \textit{1.00}\sd{.03} & \textit{3.74}\sd{.17} & \textit{.003}\sd{.000} & \textit{.68}\sd{.01} & \textit{.06}\sd{.00} & \textit{.58}\sd{.01} & \textit{.82}\sd{.03} & \textit{.28}\sd{.05} & \textit{14}\sd{1} & \textit{29}\sd{1} & \textit{.88}\sd{.02} & \textit{.21}\sd{.01} & \textit{1.20}\sd{.10} \\
 & - Inertia scaling & 99.5\sd{.2} & 1.02\sd{.02} & 5.44\sd{.34} & \textbf{.003}\sd{.001} & .68\sd{.00} & .07\sd{.00} & \textbf{.54}\sd{.01} & \underline{.84}\sd{.04} & .18\sd{.02} & 14\sd{1} & 33\sd{3} & .95\sd{.02} & .18\sd{.01} & 2.58\sd{.19} \\
 & - Estimator obs & \textbf{99.6}\sd{.1} & \textbf{.93}\sd{.01} & 3.98\sd{.25} & \textbf{.003}\sd{.001} & \textbf{.70}\sd{.01} & \textbf{.06}\sd{.00} & .41\sd{.05} & .83\sd{.04} & \underline{.21}\sd{.07} & \textbf{13}\sd{0} & 33\sd{2} & \underline{.92}\sd{.02} & \textbf{.19}\sd{.01} & \underline{1.12}\sd{.13} \\
 & - Learned residual & \textbf{99.6}\sd{.1} & \underline{.96}\sd{.02} & \underline{3.72}\sd{.24} & \textbf{.003}\sd{.000} & \underline{.69}\sd{.02} & \textbf{.06}\sd{.00} & .45\sd{.04} & .79\sd{.06} & \textbf{.22}\sd{.04} & \textbf{13}\sd{0} & \underline{32}\sd{3} & .93\sd{.02} & \textbf{.19}\sd{.02} & \textbf{1.08}\sd{.08} \\
\addlinespace[3pt]
\multirow{5}{*}{\act{lateral}} & ImpedanceMimic & \textbf{99.9}\sd{.2} & \textbf{3.62}\sd{.07} & \textbf{73.8}\sd{2.7} & \underline{.121}\sd{.010} & .91\sd{.00} & \textbf{.31}\sd{.00} & \textbf{.84}\sd{.01} & \textbf{.69}\sd{.01} & \textbf{.60}\sd{.01} & 35\sd{2} & 64\sd{8} & \textbf{.76}\sd{.02} & \textbf{.49}\sd{.01} & 34.2\sd{3.7} \\
 & \textit{Oracle GRF obs} & \textit{98.8}\sd{1.3} & \textit{3.60}\sd{.09} & \textit{72.7}\sd{2.0} & \textit{.120}\sd{.003} & \textit{.92}\sd{.01} & \textit{.31}\sd{.01} & \textit{.85}\sd{.01} & \textit{.71}\sd{.01} & \textit{.63}\sd{.03} & \textit{37}\sd{6} & \textit{65}\sd{9} & \textit{.75}\sd{.02} & \textit{.50}\sd{.01} & \textit{32.7}\sd{2.6} \\
 & - Inertia scaling & \underline{99.7}\sd{.3} & \underline{3.71}\sd{.06} & 81.0\sd{1.8} & \textbf{.100}\sd{.005} & \textbf{.92}\sd{.00} & \underline{.32}\sd{.01} & \textbf{.84}\sd{.01} & \textbf{.69}\sd{.01} & \underline{.58}\sd{.03} & \textbf{32}\sd{4} & \textbf{58}\sd{3} & \underline{.78}\sd{.02} & \underline{.47}\sd{.01} & 37.1\sd{1.3} \\
 & - Estimator obs & 99.5\sd{.7} & 3.82\sd{.07} & \underline{75.1}\sd{1.9} & .130\sd{.006} & \textbf{.92}\sd{.01} & .36\sd{.01} & .79\sd{.02} & .66\sd{.02} & .53\sd{.01} & 42\sd{17} & 70\sd{13} & .80\sd{.03} & .46\sd{.02} & \textbf{33.5}\sd{2.3} \\
 & - Learned residual & \underline{99.7}\sd{.6} & 3.75\sd{.17} & \underline{75.1}\sd{1.0} & .126\sd{.010} & \textbf{.92}\sd{.01} & .34\sd{.01} & .81\sd{.02} & .68\sd{.04} & .57\sd{.02} & \underline{33}\sd{3} & \underline{61}\sd{3} & \underline{.78}\sd{.02} & \underline{.47}\sd{.01} & \underline{33.7}\sd{2.1} \\
\addlinespace[3pt]
\multirow{5}{*}{\act{flexion}} & ImpedanceMimic & \textbf{99.7}\sd{.3} & \textbf{2.53}\sd{.07} & \textbf{45.2}\sd{1.7} & \textbf{.058}\sd{.007} & \textbf{.90}\sd{.01} & \textbf{.23}\sd{.01} & \textbf{.64}\sd{.01} & \textbf{.47}\sd{.02} & .51\sd{.02} & \textbf{22}\sd{1} & \underline{67}\sd{1} & \textbf{.53}\sd{.01} & \textbf{.40}\sd{.01} & \textbf{14.8}\sd{.7} \\
 & \textit{Oracle GRF obs} & \textit{99.4}\sd{.7} & \textit{2.52}\sd{.03} & \textit{45.4}\sd{2.5} & \textit{.057}\sd{.004} & \textit{.90}\sd{.01} & \textit{.23}\sd{.01} & \textit{.65}\sd{.01} & \textit{.49}\sd{.03} & \textit{.53}\sd{.02} & \textit{25}\sd{7} & \textit{68}\sd{5} & \textit{.52}\sd{.00} & \textit{.42}\sd{.01} & \textit{13.5}\sd{.3} \\
 & - Inertia scaling & 97.4\sd{1.2} & 2.90\sd{.04} & 72.2\sd{5.5} & .112\sd{.007} & .86\sd{.01} & .30\sd{.02} & .57\sd{.02} & \underline{.43}\sd{.02} & .41\sd{.02} & \underline{23}\sd{1} & \underline{67}\sd{2} & .73\sd{.07} & .35\sd{.02} & 35.2\sd{10.1} \\
 & - Estimator obs & 98.9\sd{1.1} & \underline{2.56}\sd{.07} & 49.6\sd{2.2} & \underline{.065}\sd{.005} & \textbf{.90}\sd{.01} & \underline{.25}\sd{.00} & \underline{.61}\sd{.01} & .40\sd{.03} & \textbf{.53}\sd{.01} & 25\sd{5} & 69\sd{4} & \underline{.55}\sd{.01} & .39\sd{.01} & \underline{16.4}\sd{.8} \\
 & - Learned residual & \underline{99.0}\sd{.4} & 2.57\sd{.05} & \underline{48.9}\sd{1.9} & .066\sd{.003} & \textbf{.90}\sd{.01} & \underline{.25}\sd{.01} & \underline{.61}\sd{.01} & .41\sd{.02} & \underline{.52}\sd{.02} & \underline{23}\sd{1} & \textbf{66}\sd{1} & \underline{.55}\sd{.01} & \textbf{.40}\sd{.01} & 16.8\sd{.9} \\
\addlinespace[3pt]
\multirow{5}{*}{\act{jump}} & ImpedanceMimic & \underline{89.3}\sd{2.6} & \textbf{3.41}\sd{.09} & \textbf{74.9}\sd{3.8} & .088\sd{.007} & \underline{.90}\sd{.01} & \textbf{.39}\sd{.01} & \textbf{.62}\sd{.02} & \textbf{.49}\sd{.02} & \textbf{.34}\sd{.03} & \underline{29}\sd{1} & 76\sd{4} & \textbf{.78}\sd{.02} & \textbf{.30}\sd{.01} & \textbf{41.7}\sd{2.1} \\
 & \textit{Oracle GRF obs} & \textit{89.1}\sd{1.9} & \textit{3.44}\sd{.15} & \textit{77.4}\sd{1.9} & \textit{.082}\sd{.005} & \textit{.91}\sd{.01} & \textit{.39}\sd{.01} & \textit{.63}\sd{.01} & \textit{.51}\sd{.04} & \textit{.35}\sd{.02} & \textit{29}\sd{2} & \textit{76}\sd{3} & \textit{.79}\sd{.02} & \textit{.30}\sd{.01} & \textit{43.3}\sd{2.9} \\
 & - Inertia scaling & 88.5\sd{2.5} & 3.71\sd{.11} & 92.1\sd{4.7} & \textbf{.083}\sd{.011} & \underline{.90}\sd{.00} & \underline{.41}\sd{.01} & \underline{.61}\sd{.01} & .46\sd{.01} & \underline{.32}\sd{.01} & \underline{29}\sd{1} & \textbf{71}\sd{2} & .94\sd{.05} & .26\sd{.01} & 64.0\sd{7.7} \\
 & - Estimator obs & \textbf{90.4}\sd{1.5} & \underline{3.51}\sd{.14} & \underline{77.6}\sd{2.6} & \underline{.084}\sd{.008} & \textbf{.91}\sd{.01} & .42\sd{.01} & .58\sd{.02} & \underline{.48}\sd{.02} & \underline{.32}\sd{.03} & \textbf{28}\sd{0} & 75\sd{4} & \underline{.81}\sd{.02} & \underline{.29}\sd{.01} & \underline{44.2}\sd{1.9} \\
 & - Learned residual & 88.3\sd{2.1} & 3.54\sd{.10} & 77.9\sd{2.0} & .093\sd{.013} & \underline{.90}\sd{.00} & .42\sd{.00} & .58\sd{.01} & .46\sd{.02} & \underline{.32}\sd{.01} & \underline{29}\sd{1} & \underline{74}\sd{3} & .82\sd{.03} & \underline{.29}\sd{.01} & 44.6\sd{3.5} \\
\specialrule{\cmidrulewidth}{\aboverulesep}{0pt}
\rowcolor{meangray}  & \padT ImpedanceMimic & \textbf{98.1}\sd{.4} & \textbf{2.45}\sd{.02} & \textbf{48.9}\sd{.9} & \textbf{.057}\sd{.002} & \underline{.89}\sd{.00} & \textbf{.21}\sd{.00} & \textbf{.76}\sd{.01} & \textbf{.65}\sd{.01} & \textbf{.56}\sd{.01} & \underline{24}\sd{0} & 53\sd{1} & \textbf{.72}\sd{.01} & \textbf{.45}\sd{.00} & \textbf{19.9}\sd{1.0} \\
\rowcolor{meangray}  & \textit{Oracle GRF obs} & \textit{97.8}\sd{.5} & \textit{2.47}\sd{.04} & \textit{49.4}\sd{.8} & \textit{.055}\sd{.001} & \textit{.89}\sd{.00} & \textit{.20}\sd{.00} & \textit{.78}\sd{.00} & \textit{.65}\sd{.02} & \textit{.59}\sd{.01} & \textit{25}\sd{1} & \textit{53}\sd{1} & \textit{.71}\sd{.01} & \textit{.46}\sd{.01} & \textit{20.0}\sd{.6} \\
\rowcolor{meangray}  & - Inertia scaling & 97.5\sd{.4} & 2.68\sd{.03} & 79.3\sd{1.0} & .062\sd{.002} & .88\sd{.00} & .24\sd{.00} & \underline{.75}\sd{.00} & .59\sd{.01} & .51\sd{.01} & \underline{24}\sd{1} & \textbf{52}\sd{1} & .91\sd{.01} & .37\sd{.00} & 53.7\sd{1.7} \\
\rowcolor{meangray}  & - Estimator obs & \underline{98.0}\sd{.4} & \underline{2.51}\sd{.04} & \underline{50.3}\sd{.8} & \underline{.058}\sd{.003} & \textbf{.90}\sd{.00} & .23\sd{.00} & .72\sd{.01} & \underline{.62}\sd{.01} & .55\sd{.01} & 25\sd{3} & 54\sd{2} & .74\sd{.01} & .44\sd{.01} & \underline{20.5}\sd{.6} \\
\rowcolor{meangray} \multirow{-5}{*}{\textbf{Mean}} & \padB - Learned residual & 97.7\sd{.4} & \underline{2.51}\sd{.02} & 50.5\sd{.8} & .059\sd{.002} & \underline{.89}\sd{.00} & \underline{.22}\sd{.00} & .73\sd{.01} & .61\sd{.02} & \textbf{.56}\sd{.01} & \textbf{23}\sd{0} & \textbf{52}\sd{1} & \underline{.73}\sd{.01} & \textbf{.45}\sd{.01} & 20.8\sd{1.4} \\
\specialrule{\heavyrulewidth}{0pt}{\belowbottomsep}
\end{tabular}%
}
\vspace{0pt}
\parbox{\textwidth}{\scriptsize\raggedright
  $^\dagger$9 main joints (hip flexion / adduction, knee, ankle, both sides; lumbar extension), against inverse dynamics.
  Ep. = episode survival before termination (\%). $q$, $\dot q$ = joint-angle / velocity RMSE. $p_x$ = root forward-position drift rate. \{kin$_r$, $F_{\mathrm{Z}_r}$, $F_{\mathrm{AP}_r}$, $F_{\mathrm{ML}_r}$, $\tau_r$\} = Pearson correlation w.r.t. reference \{joint angles, vertical, AP, ML GRF, torques\}. CoP = center of pressure. $\dot\tau$ = torque jerk.}
\vspace{-1em}
\end{table}}

\begin{table}[H]
\centering
\caption{%
  \textbf{Ablations of \projname{}.} Seen subjects, held-out clips, mean$_{\pm\text{sd}}$ over 5 seeds. Rows marked ``$-$'' remove one element, and \emph{Oracle GRF obs} observes the ground-truth force-plate GRF.
  Columns are RMSE ($\downarrow$) unless marked $\uparrow$.
  \textbf{Bold} / \underline{underline} = best / second best per activity.}
\vspace{-0.4em}
\label{tab:ablation_seen}
\renewcommand{\arraystretch}{0.80}
\setlength{\tabcolsep}{2pt}
\newcommand{\gap}{\hspace{4pt}}
\newcommand{\sd}[1]{{\tiny\textcolor{black!55}{$\pm$#1}}}
\scriptsize
\resizebox{\textwidth}{!}{%
\begin{tabular}{@{}>{\fillL}l @{\gap} l @{\gap} lllll @{\gap} lp{0.75cm}p{0.75cm}p{0.75cm}ll @{\gap} ll>{\fillR}l @{}}
\toprule
 & & \multicolumn{1}{c}{\textbf{Stability}} & \multicolumn{4}{c}{\textbf{Kinematics}} & \multicolumn{6}{c}{\textbf{Ground Reaction Forces}} & \multicolumn{3}{c}{\textbf{Torques}$^\dagger$} \\
\cmidrule(lr){3-3}\cmidrule(lr){4-7}\cmidrule(lr){8-13}\cmidrule(lr){14-16}
\texttt{\textbf{Activity}} & \textbf{Method}
  & \makecell{Ep.$\hi$ \\ (\%)}
  & \makecell{$q\lo$\\ \makebox[0pt][c]{($^\circ$)}}
  & \makecell{$\dot{q}\lo$\\ \makebox[0pt][c]{($^\circ$/s)}}
  & \makecell{$p_x\lo$\\ \makebox[0pt][c]{(m/s)}}
  & kin$_r\hi$
  & \makecell{$F_{\mathrm{Z}}\lo$\\ \makebox[0pt][c]{(BW)}}
  & $F^{\uparrow}_{\mathrm{Z}_r}$
  & $F^{\uparrow}_{\mathrm{ML}_r}$
  & $F^{\uparrow}_{\mathrm{AP}_r}$
  & \makecell{\makebox[0pt][c]{$\mathrm{CoP}^{\downarrow}_{\mathrm{ML}}$}\\ \makebox[0pt][c]{\tiny (mm)}}
  & \makecell{\makebox[0pt][c]{$\mathrm{CoP}^{\downarrow}_{\mathrm{AP}}$}\\ \makebox[0pt][c]{\tiny (mm)}}
  & \makecell{$\tau\lo$\\ \makebox[0pt][c]{\tiny (Nm/kg)}}
  & $\tau_r\hi$
  & \makecell{$\dot{\tau}\lo$\\ \makebox[0pt][c]{\tiny (Nm/kg/s)}} \\
\midrule
\multirow{14}{*}{\act{walk}} & ImpedanceMimic & \textbf{100}\sd{0} & 1.66\sd{.03} & 28.8\sd{.4} & \underline{.013}\sd{.001} & \textbf{.98}\sd{.00} & \underline{.09}\sd{.00} & \textbf{.98}\sd{.00} & .89\sd{.00} & .86\sd{.00} & \underline{20}\sd{0} & 40\sd{1} & \underline{.38}\sd{.00} & \underline{.65}\sd{.00} & 6.38\sd{.08} \\
 & \textit{Oracle GRF obs} & \textit{100}\sd{0} & \textit{1.67}\sd{.03} & \textit{28.9}\sd{.2} & \textit{.014}\sd{.001} & \textit{.98}\sd{.00} & \textit{.09}\sd{.00} & \textit{.98}\sd{.00} & \textit{.90}\sd{.01} & \textit{.87}\sd{.00} & \textit{20}\sd{0} & \textit{40}\sd{0} & \textit{.38}\sd{.00} & \textit{.65}\sd{.01} & \textit{6.16}\sd{.05} \\
\end{tabular}%
}
\vspace{0pt}
\parbox{\textwidth}{\scriptsize\raggedright
  $^\dagger$9 main joints (hip flexion / adduction, knee, ankle, both sides; lumbar extension), against inverse dynamics.
  Ep. = episode survival before termination (\%). $q$, $\dot q$ = joint-angle / velocity RMSE. $p_x$ = root forward-position drift rate. \{kin$_r$, $F_{\mathrm{Z}_r}$, $F_{\mathrm{AP}_r}$, $F_{\mathrm{ML}_r}$, $\tau_r$\} = Pearson correlation w.r.t. reference \{joint angles, vertical, AP, ML GRF, torques\}. CoP = center of pressure. $\dot\tau$ = torque jerk.}
\vspace{-1em}
\end{table}

\end{document}

%% file: math_commands.tex
\usepackage{amsmath,amsfonts,bm}

\def\eqref#1{equation~\ref{#1}}

\def\1{\bm{1}}

\DeclareMathAlphabet{\mathsfit}{\encodingdefault}{\sfdefault}{m}{sl}
\SetMathAlphabet{\mathsfit}{bold}{\encodingdefault}{\sfdefault}{bx}{n}